\documentclass[sigconf]{acmart}

\AtBeginDocument{%
  \providecommand\BibTeX{{%
    \normalfont B\kern-0.5em{\scshape i\kern-0.25em b}\kern-0.8em\TeX}}}

\copyrightyear{2020}
\acmYear{2020}
\setcopyright{acmcopyright}\acmConference[MM '20]{Proceedings of the 28th ACM International Conference on Multimedia}{October 12--16, 2020}{Seattle, WA, USA}
\acmBooktitle{Proceedings of the 28th ACM International Conference on Multimedia (MM '20), October 12--16, 2020, Seattle, WA, USA}
\acmPrice{15.00}
\acmDOI{10.1145/3394171.3413559}
\acmISBN{978-1-4503-7988-5/20/10}

\begin{document}
\fancyhead{}
\title{Joint Self-Attention and Scale-Aggregation for Self-Calibrated Deraining Network}

\author{Cong Wang}
\authornote{Both authors contributed equally to this research.}
\affiliation{%
	\institution{Dalian University of Technology}}
%\streetaddress{8600 Datapoint Drive}
\email{supercong94@gmail.com}

\author{Yutong Wu}
\authornotemark[1]
\affiliation{\institution{Dalian University of Technology}}
\email{ytongwu@mail.dlut.edu.cn}

\author{Zhixun Su}
\authornote{Corresponding author.}
\affiliation{%
  \institution{Dalian University of Technology}
  \institution{Key Laboratory for Computational Mathematics and Data Intelligence of Liaoning Province}
  %\city{Ganjingzi Qu}
  %\state{Dalian Shi}
  %\country{China}
}
\email{zxsu@dlut.edu.cn}

\author{Junyang Chen}
\affiliation{%
  \institution{University of Macau}
  %\streetaddress{8600 Datapoint Drive}
  %\city{San Antonio}
  %\state{Texas}
  %\postcode{78229}
}
\email{yb77403@umac.mo}

\begin{abstract}
In the field of multimedia, single image deraining is a basic pre-processing work, which can greatly improve the visual effect of subsequent high-level tasks in rainy conditions.
In this paper, we propose an effective algorithm, called JDNet, to solve the single image deraining problem and conduct the segmentation and detection task for applications.
Specifically, considering the important information on multi-scale features, we propose a Scale-Aggregation module to learn the features with different scales.
Simultaneously, Self-Attention module is introduced to match or outperform their convolutional counterparts, which allows the feature aggregation to adapt to each channel.
Furthermore, to improve the basic convolutional feature transformation process of Convolutional Neural Networks (CNNs), Self-Calibrated convolution is applied to build long-range spatial and inter-channel dependencies around each spatial location that explicitly expand fields-of-view of each convolutional layer through internal communications and hence enriches the output features.
By designing the Scale-Aggregation and Self-Attention modules with Self-Calibrated convolution skillfully, the proposed model has better deraining results both on real-world and synthetic datasets.
Extensive experiments are conducted to demonstrate the superiority of our method compared with state-of-the-art methods. The source code will be available at \url{https://supercong94.wixsite.com/supercong94}.

\end{abstract}




\maketitle

\begin{figure}[!t]
	\begin{center}
		\begin{tabular}{cccc}
			\includegraphics[width = 0.3\linewidth]{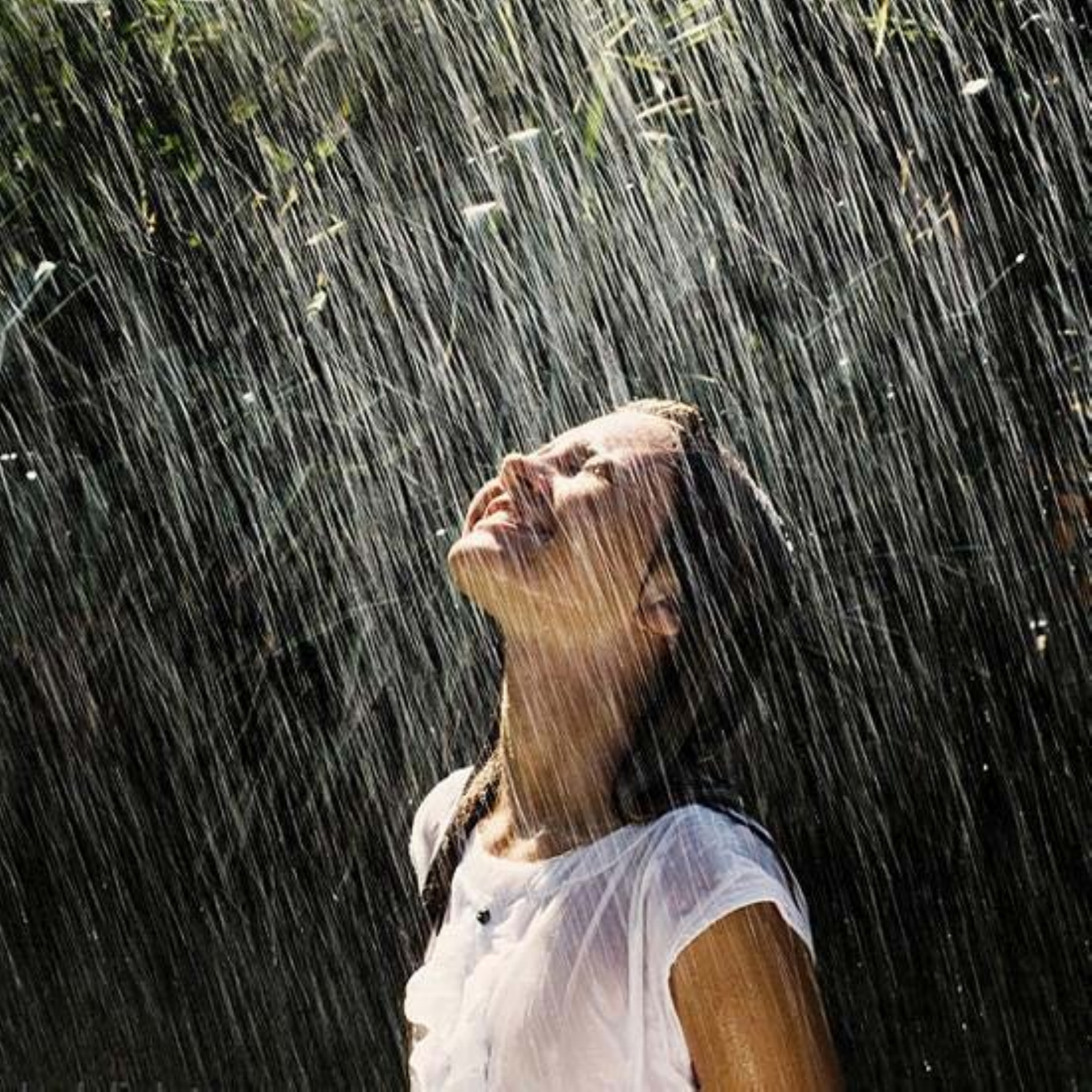}&\hspace{-2mm}
			\includegraphics[width = 0.3\linewidth]{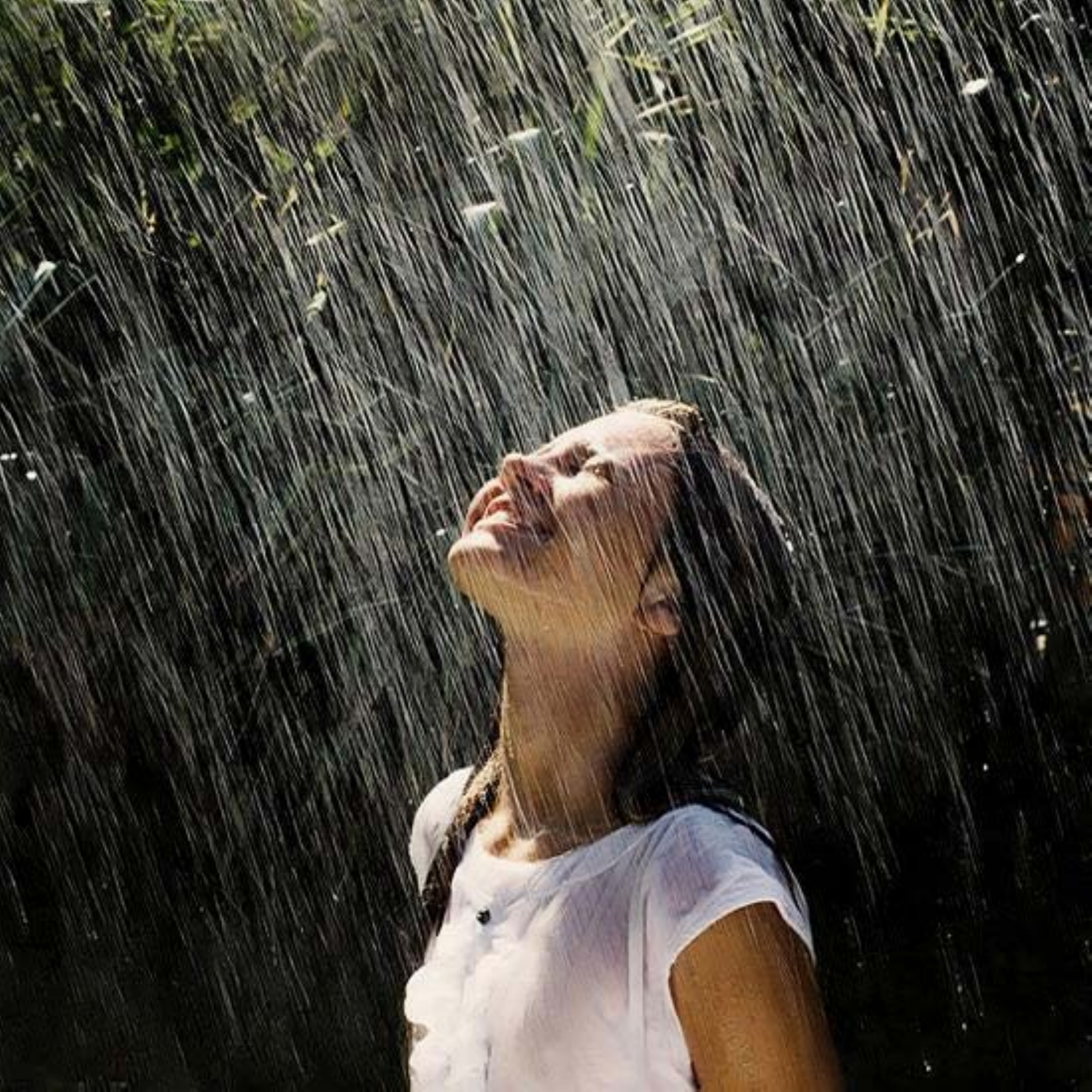}&\hspace{-2mm}
			\includegraphics[width = 0.3\linewidth]{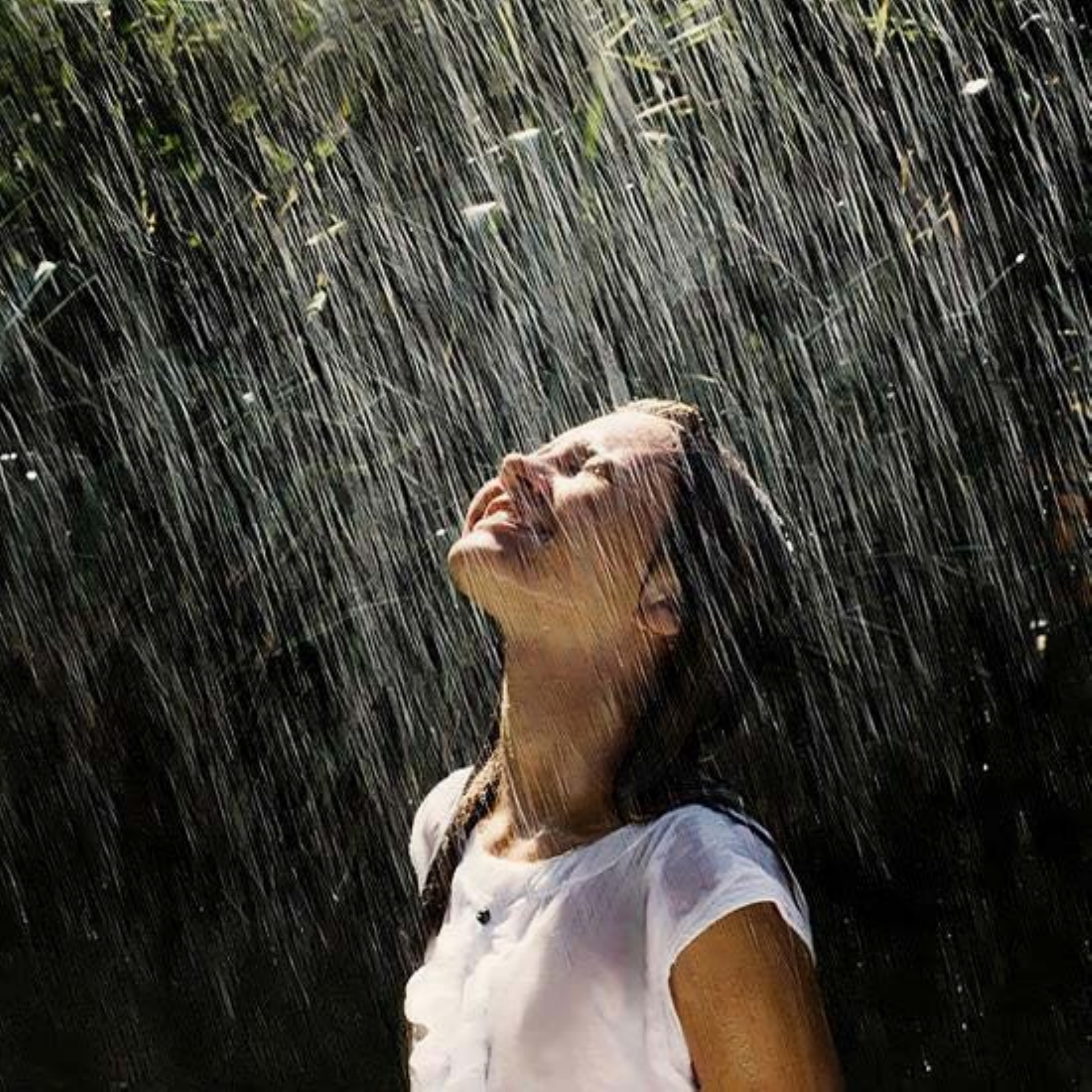}&\hspace{-2mm}
			\\
			(a) Input&\hspace{-2mm} (b) REHEN~\cite{derain_acmmm19_rehen} &\hspace{-2mm} (c) PreNet~\cite{derain_prenet_Ren_2019_CVPR}
			\\
			\includegraphics[width = 0.3\linewidth]{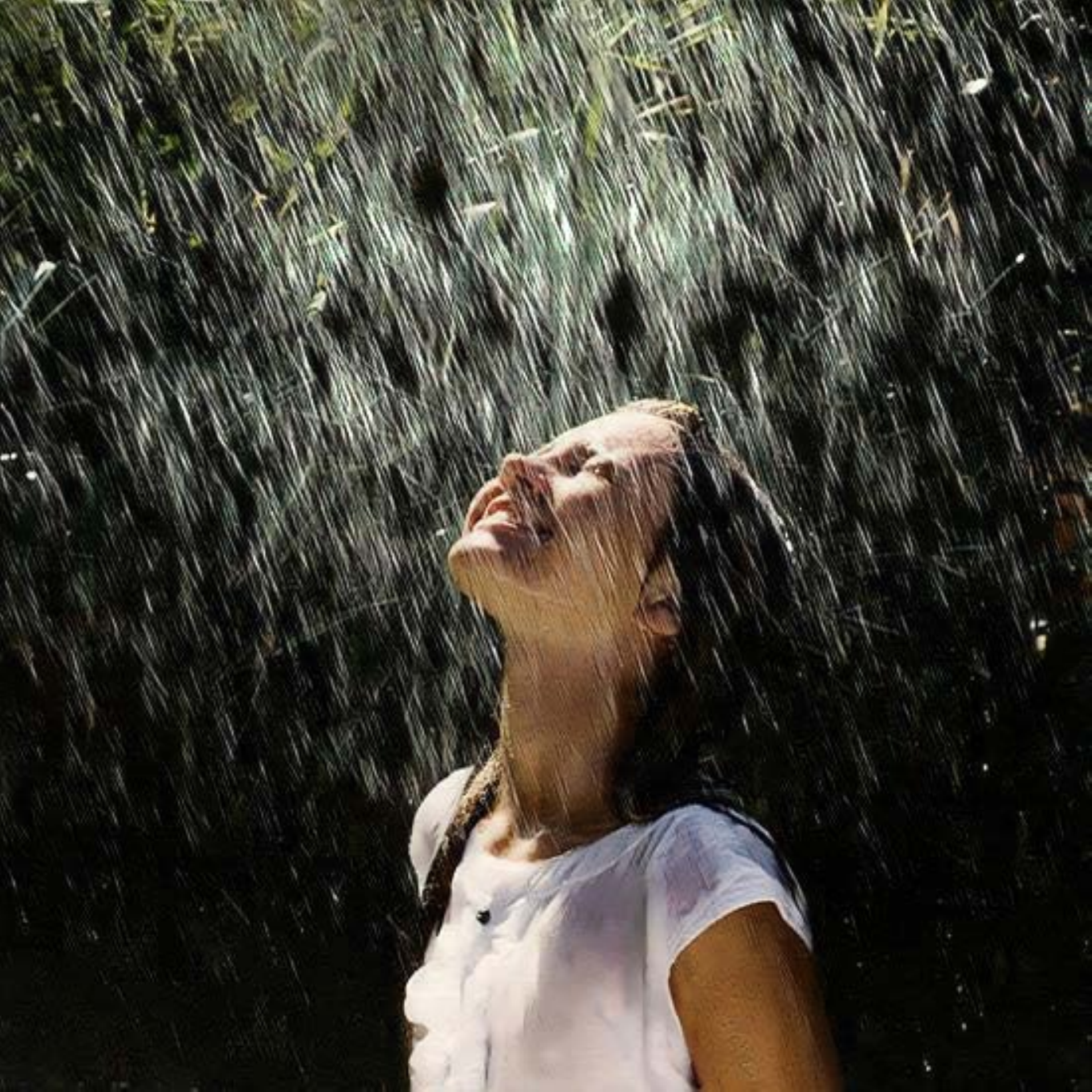}&\hspace{-2mm}
			\includegraphics[width = 0.3\linewidth]{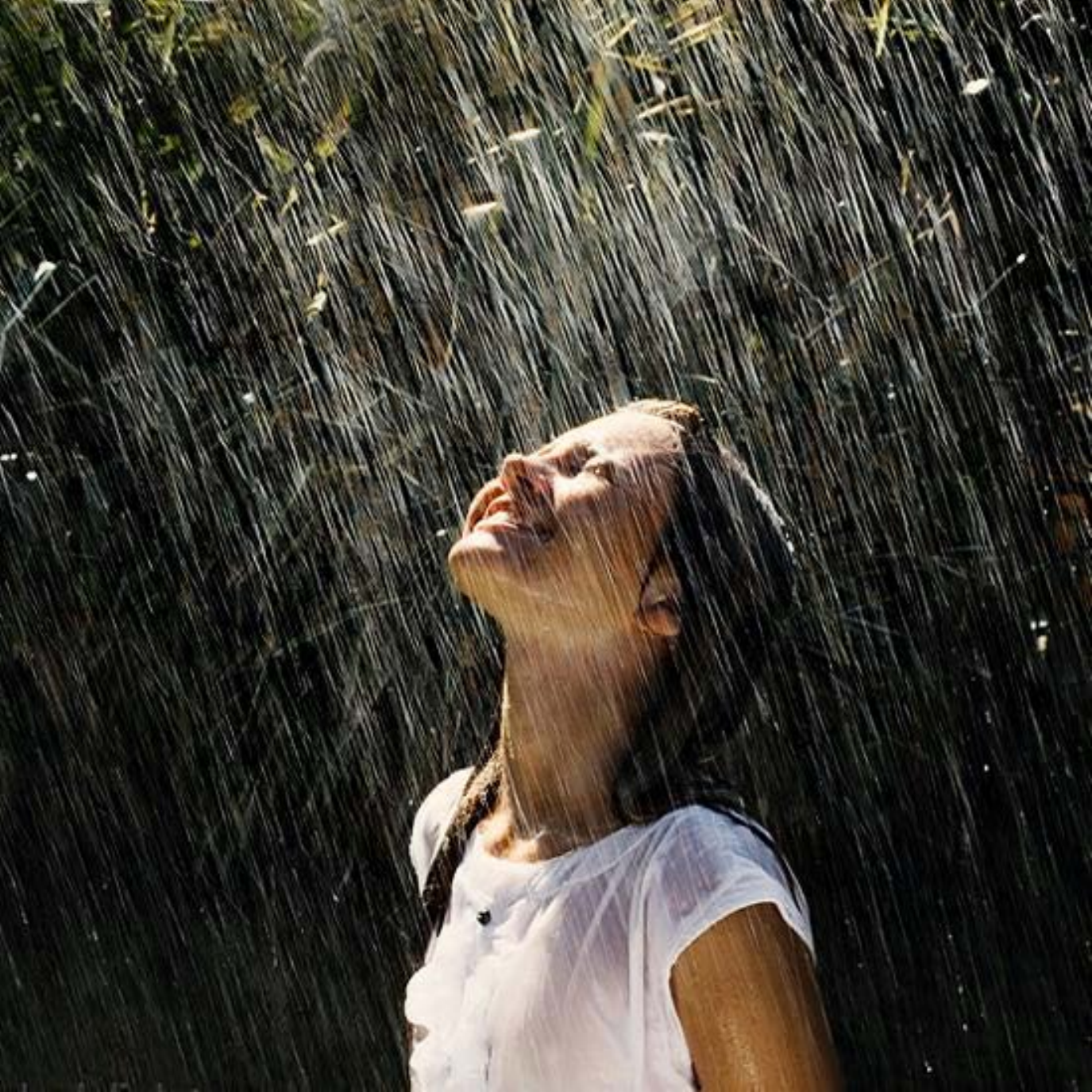}&\hspace{-2mm}
			\includegraphics[width = 0.3\linewidth]{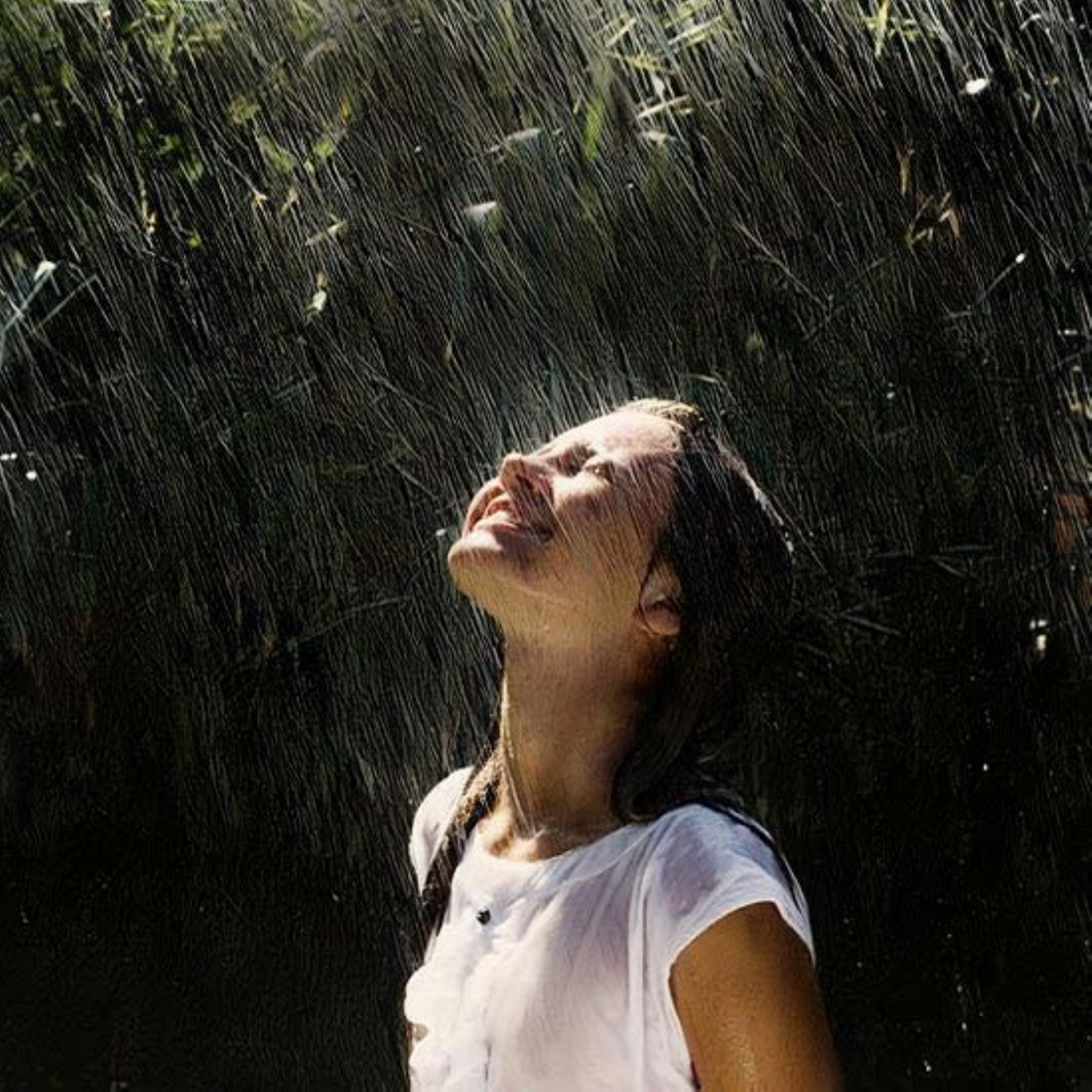}
			\\(d) SpaNet~\cite{derain_2019_CVPR_spa}&\hspace{-2mm} (e) SSIR~\cite{derain-cvpr19-semi}&\hspace{-2mm} (f) JDNet (Ours) \\
		\end{tabular}
	\end{center}
	\caption{An example from real-world datasets.}
	\label{fig:an example with other methods}
\end{figure}

\section{Introduction}

As we all know, plenty of influence factors will greatly reduce the image quality, while affects the effect of subsequent image processing, such as Object Detection, Semantic Segmentation and Optical Character Recognition, etc.
Rain is one of the common influence factors.
How to restore a rain-free image from a given rainy image is a challenging problem.
In this paper, we strive to solve the task by exploring a series of inner properties in the convolution layer.
%In order to describe the importance of deraining in general computer vision processing, Figure 1 shows the comparison results of rainy images and deraining images in a high-level task.

We formulate the process of image deraining:
\begin{equation}
B=O-R,
\label{eq:deraining}
\end{equation}
where \begin{math}O\end{math} represents the input rainy image, \begin{math}R\end{math} is the rain streaks layer, and \begin{math}B\end{math} is the deraining image, which is generally called the background layer.
%Deraining can be roughly divided into the video-based deraining methods and single image deraining methods.
%Different from the video-based deraining methods which use coherent inter-frame information and temporal content, the single image deraining methods have no additional information.
The restoration from a single rainy image has always been regarded as an ill-posed problem that there is numerous solution for a given rainy image.
How to regulate the solution space to get stable and sole rain-free image becomes a core problem for the image deraining task.

In the early period, the deraining task was generally regarded as an optimization problem based on the prior information of images.
\cite{yang-pami} classifies mainly into two categories in prior-based methods, 1) employing Sparse Coding and 2) establishing Gaussian mixture models.
The Sparse Coding-based methods convert the processing of the image into the processing of the signal and decompose the image into a low-frequency component and a high-frequency component.
Based on this frequency decomposition process, \citeauthor{derain_id_kang} \cite{derain_id_kang} remove the rain component in rainy images using dictionary learning, \citeauthor{derain_dsc_luo} \cite{derain_dsc_luo} learn the dictionary of rain streak and background layers via Discriminative Sparse Coding, while others~\cite{derain_lp_li} apply Gaussian mixture models to model rain and background layers.
These prior-based methods achieve better deraining performance to some extent under certain conditions by the given assumptions, but they either smooth out the edge details as rain streaks, or cannot handle large and dense rainy images.
Moreover, the prior-based methods can generally be regarded as optimizing the cost function, which has a high time consumption.

With the rapid development of deep learning in recent years, many Convolutional Neural Networks (CNNs)-based methods for single image deraining have been proposed, gradually replaced by prior-based methods, such as \cite{derain_zhang_did,derain_ddn_fu,derain_jorder_yang,derain_rescan_li,derain_depth_hu,derain_acmmm19_rehen}.
These methods employ deep networks to automatically extract features of layers, enabling them to model more complex mappings from rainy images to clean images, such as learning the binary rain streaks map \cite{derain_jorder_yang}.
Many subsequent methods enhance the deraining effects from the aspects of network structure complexity and image priors.
For example, \cite{derain_rescan_li} and \cite{derain_depth_hu} combine multi-stage recurrent network and depth information with single image deraining, respectively.
However, most of them rely on the traditional convolution pattern which achieves the network step by step by stacking several convolution layers that attach stationary weights to specific locations so that these layers do not learn features from different locations, leading to information drop-out.
Furthermore, these models only consider the network design to apply to the deraining task while ignoring the inner structure of CNNs that lead to these methods is not robust to real-world rainy conditions.
As shown in Fig.~\ref{fig:an example with other methods}, other state-of-the-art methods fail to restore rain-free images, while our approach is able to remove most rain streaks and gets a clearer rain-free image.

To solve the above problems, we propose an effective deraining algorithm, called JDNet, from three aspects:
1) Pairwise Self-Attention module, 2) Scale-Aggregation module and 3) Self-Calibrated convolution.
First of all, considering that the aggregation of information from a neighborhood cannot adapt to its content which will lose much important information in the traditional convolutional layer, we introduce a Pairwise Self-Attention module by paying attention to learn different locations at convolution.
The introduced Pairwise Self-Attention module does not attach stationary weights to specific locations and is invariant to permutation and cardinality.
In particular, weight computations of the Self-Attention module do not collapse the channel dimension and allows the feature aggregation to adapt to each channel.
Secondly, we design a Scale-Aggregation module to learn the features from different scales.
The proposed Scale-Aggregation module not only converts the convolution layer into deeper features, but also can maintain the original features from shallower ones.
By designing the inner fusion between shallower and deeper layers, the proposed module can learn adaptively the features which part is more effective for rain removal.
Thirdly, to improve the basic convolutional feature transformation process of CNNs, we bring in Self-Calibrated convolution to build long-range spatial and inter-channel dependencies around each spatial location that explicitly expand fields-of-view of each convolutional layer through internal communications and hence enriches the output features, which can help CNNs generate more discriminative representations by explicitly incorporating richer information.
By devising jointly the there parts in the convolution layer, the proposed method has better deraining performance that can be able to remove heavy rain streaks and preserve better details.

\section{Related Work}
\subsection{Single Image Deraining}

In order to generate accurate deraining results, researchers have made many trials, which can be simply divided into image prior-based methods and CNNs-based methods.

These image prior-based methods need to represent the characteristics of the rain-free image while maintaining consistency with the input image content.
\citeauthor{derain_id_kang}~\cite{derain_id_kang} employ the bilateral filter to decompose high-frequency information and low-frequency information from a rainy image.
Sparse coding and dictionary learning are used to remove rain components in a rainy image.
This work successfully eliminates the sparse light rain streaks.
However, due to the extreme dependence on the preprocessing of the bilateral filter, it will produce blurred background details.
In order not to confuse the rain line layer and the background layer, \citeauthor{derain_dsc_luo}~\cite{derain_dsc_luo} introduce the mutual exclusivity property into discriminative sparse coding, and finally obtain deraining results which retain the clean texture details.
In addition, \cite{derain_lp_li} proposes Gaussian mixture models to model the rain layers and the background layers.
These priors have a good effect on the removal of rain streaks with multiple directions and scales.
Moreover, \citeauthor{derain_zhu_bilayer}~\cite{derain_zhu_bilayer} construct an iterative process to remove more rain.
Although many prior-based methods have tried and improved for single image deraining, there are two common limits in these methods.
On the one hand, the removal of large and dense rain streaks is not enough. On the other hand, the test time is too long.

Deep learning has an epoch-making significance in the field of image processing, which can simultaneously improve the speed of operations and the quality of completions.
The CNNs-based methods generally use artificial means to generate a large number of paired datasets for training.
\citeauthor{derain_jorder_yang}~\cite{derain_jorder_yang} jointly perform a recurrent network of detection and rain removal, where the binary map is used in the detection process.
In recent years, researchers have improved the network from different perspectives. In terms of models, \citeauthor{derain_ddn_fu}~\cite{derain_ddn_fu} and \citeauthor{derain_rescan_li}~\cite{derain_rescan_li} introduce ResNet and SE-block in their deraining networks, respectively. In terms of processes, \citeauthor{derain_prenet_Ren_2019_CVPR}~\cite{derain_prenet_Ren_2019_CVPR} choose to implement simple networks in multiple stages instead of designing complex models. In particular, \citeauthor{derain_2019_CVPR_spa}~\cite{derain_2019_CVPR_spa} design a deraining network based on the spatial attention module to pay special attention to the area where the rain is located.
\begin{figure*}[!t]
	\begin{center}
		\begin{tabular}{cccc}
			\includegraphics[width = 0.98\linewidth]{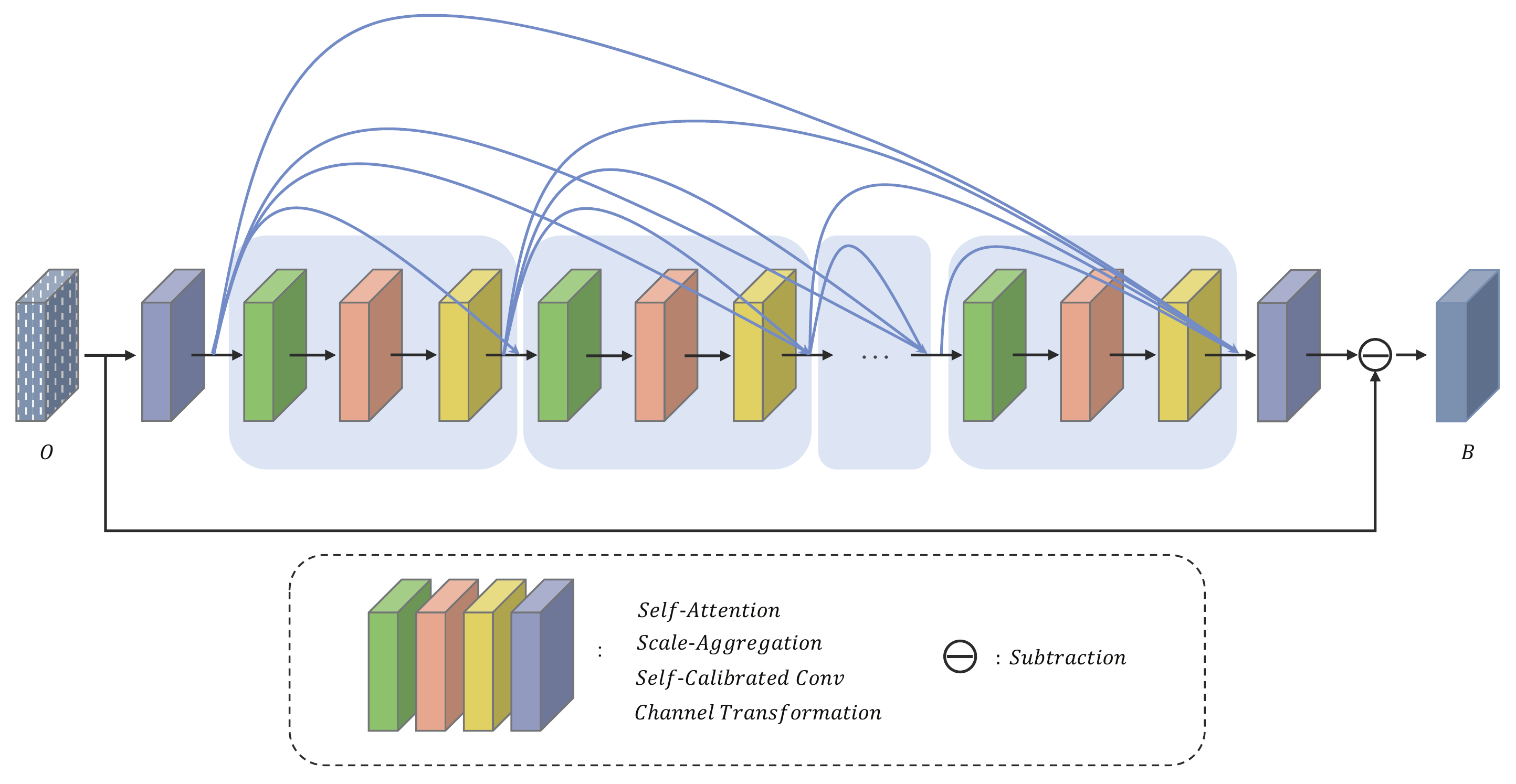}
		\end{tabular}
	\end{center}
	\caption{The architecture of Joint Network for deraining (JDNet).
		Each joint unit is composed of a Self-Attention module, Scale-Aggregation module and Self-Calibrated convolution.
		We use dense connections to inner-connect several joint units.
		And at the first layer and last layer, we use 3$\times$3 convolution following a LeakyReLU to change the channel dimensions.}
	\label{fig: overall}
\end{figure*}
\subsection{Attention Mechanisms}

%As we all know, the design of the network's architecture is crucial to the feature extraction of the final task. In order to make the extracted representation of the network more effective, common methods are to manually design a powerful network structure, such as VGG~\cite{vgg}, ResNet~\cite{He_2016_CVPR}, DenseNet~\cite{densenet} and their various variants~\cite{res1,res2,res3,res4}. These structures are proposed from object recognition, and can be used for computer vision and other applications.

In recent years, some methods have tried to add non-local modules~\cite{local1,local2} or attention mechanisms~\cite{att1,att2,att3,att4} to complex networks, which will establish the dependency of spatial location or channels or both.
Since~\cite{self_att1}, Self-Attention module has been widely paid attention while it becomes a research hotspot. They propose Google Neural Machine Translation, which ignores the distance between words and directly calculates the dependency relationship. Later, Self-Attention is also used in computer vision tasks as a complement to convolution.
In general channel-wise attention methods~\cite{att1,att2,self_att2}, attention weights reweight the activation in different channels. In particular, other methods~\cite{att3,self_att3,self_att4} perform weighting operations in different contents and channels.
Moreover, \cite{scnet} discusses the Self-Calibrated convolution that considers more efficiently exploiting the convolutional filters in convolutional layers and designing powerful feature transformations to generate more expressive feature representations.

\section{Proposed Methods (JDNet)}
\subsection{Overview}
In this section, we detail the proposed joint network for single image deraining, called JDNet.
To take full advantage of features with different levels, dense connections are utilized to connect several joint units that are composed of Self-Attention module, Scale-Aggregation module and Self-Calibrated convolution.

Specifically, an input rainy image first passes through a convolution layer following an activation function to transform the channel dimension from image to feature.
And then the transformed features are input into several joint units further extract the rain streaks information.
At last, we obtain the rain streaks by a convolution layer following an activation function to transform the features to image.
The detailed structure is shown in Fig.~\ref{fig: overall}.

\subsection{Self-Attention Module}

\begin{figure}[!h]
	\begin{center}
		\begin{tabular}{cccc}
			\includegraphics[width = 1\linewidth]{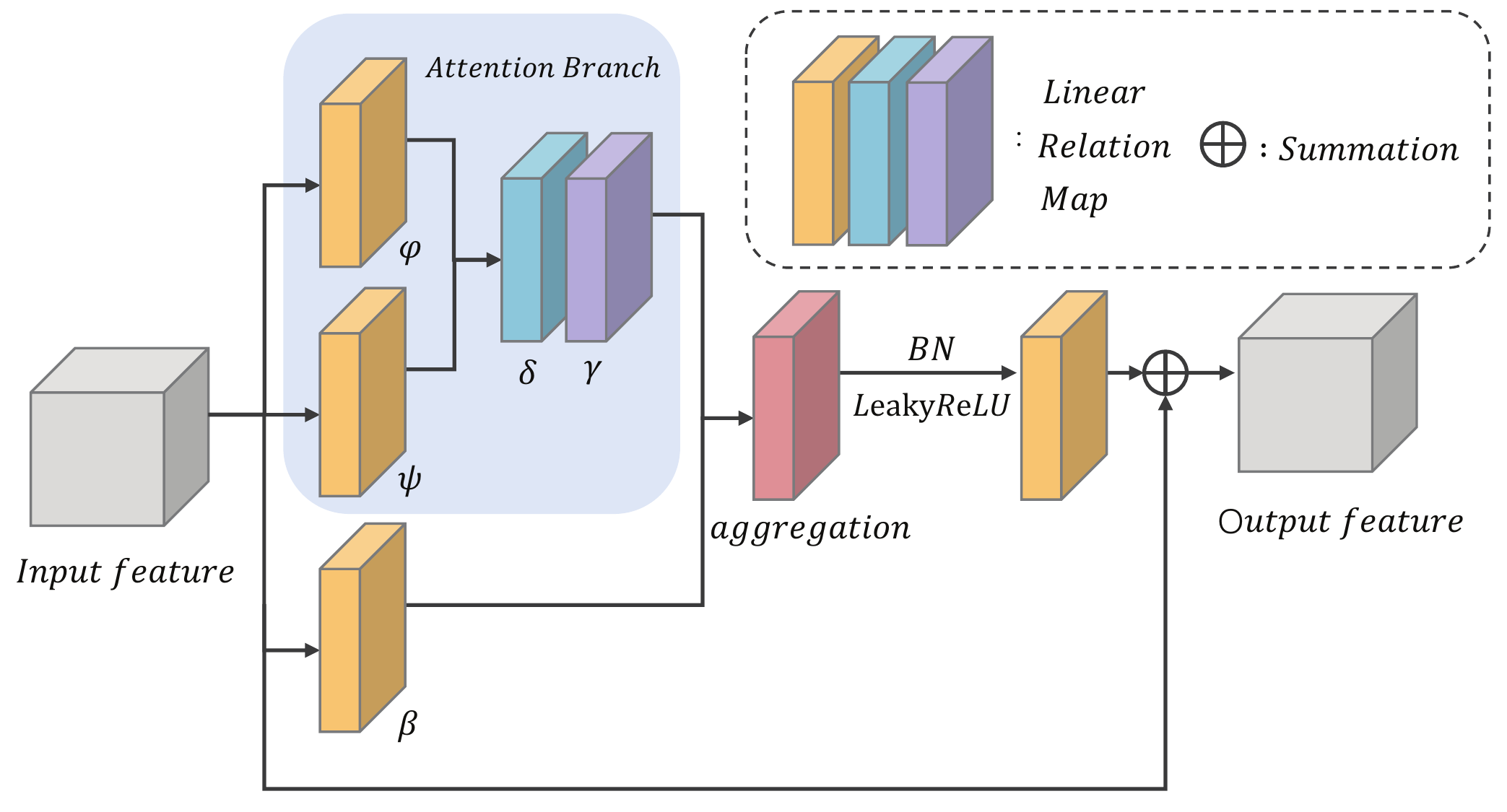}
		\end{tabular}
	\end{center}
	\caption{The architecture of Pairwise Self-Attention Module.}
	\label{fig: Pairwise Self-attention}
\end{figure}

In CNNs for image processing, each layer simultaneously realizes two functions, i.e., feature aggregation and feature transformation. The former integrates the features of all positions extracted by the kernel, and the latter performs transformation through linear mapping and nonlinear scalar functions.
Since these two functions can be decoupled, if the feature transformation is simply set as an element-level operation composed of linear mapping and nonlinear scalar functions, then only the design of feature aggregation will be considered next.
In this paper, we introduce the Pairwise Self-Attention module~\cite{self-att} to establish feature aggregation.
Consistent with general Self-Attention modules, the final result is expressed as a weighted sum of adaptive weights and features:
\begin{equation}y_i=\sum_{j\in\mathcal{R}(i)}\alpha(x_{i},x_{j})\odot\beta(x_j),\label{eq:self-attention}
\end{equation}
where $x_i$ and $x_j$ are feature maps with indexes $i$ and $j$, respectively. $\odot$ is the Hadamard product called aggregation with the local footprint $\mathcal{R}(i)$, which is a set of indexes that can be aggregated with $x_i$. Please note that the number of parameters in the Pairwise Self-Attention module will not be affected by the size of footprint. In order to utilize more surrounding pixels, we set the size of footprint to $7\times7$.
After this aggregation, the result $y_i$ can be obtained.

The vector $\beta(x_j)$ generated by the function $\beta(\cdot)$ will be aggregated with the adaptive vector $\alpha(x_i,x_j)$ introduced later. Compared with ordinary weights, adaptive vector $\alpha(x_i,x_j)$ has strong content adaptability. It can be decomposed as follows:
\begin{equation}\alpha(x_{i},x_{j})=\gamma(\delta(x_i,x_j)),\end{equation}
where $\delta(\cdot)$ and $\gamma(\cdot)$ respectively represent a relation function and a hybrid map composed of linear and nonlinear functions. Here, we use LeakyReLU as the previously mentioned nonlinear function.
Based on the relation $\delta(\cdot)$, the function $\gamma(\cdot)$ is used to obtain a vector result, which can be combined with $\beta(x_j)$ in Eq.~\ref{eq:self-attention}. In general, matching the output dimension of $\gamma(\cdot)$ with the dimension of $\beta(\cdot)$ is not a necessary thing because attention weights can be shared among a group of channels.

We choose the subtraction as the relation function, which can be formulated:
\begin{equation}\delta(x_i,x_j)=\varphi(x_i)-\psi(x_j),\end{equation}
where $\varphi(\cdot)$ and $\psi(\cdot)$ are convolution operations while have matching output dimensions. $\delta(\cdot)$ calculates spatial attention for each channel instead of sharing between channels.

This Pairwise Self-Attention module we introduced is shown in Fig~\ref{fig: Pairwise Self-attention}. In order to perform a more efficient process, these two branches through which the input feature passes reduce the dimensionality of channels appropriately.

The first branch is called the attention branch, which further extracts features through two convolutional layers of $\varphi(\cdot)$ and $\psi(\cdot)$, and then the relation $\delta(\cdot)$ and the map $\gamma(\cdot)$ are used successively to obtain the final attention weight $\alpha$.
The second branch employs the convolution operation $\beta(\cdot)$ to get features of reducing the channel dimension. After that, we use the Hadamard product to aggregate the results of the two branches and input them into Batch Normalization, LeakyReLU function and convolution operation. The last convolution operation expands the channel dimension of feature result back to the input channel and adds it to the original input feature.

Conventional convolution uses fixed kernels for feature aggregation. The kernel weights do not change with contents of input images, but change across the channels. The above Self-Attention module uses novel vector attention, which can generate content adaptation ability while maintaining the channel adaptation ability. This makes our deraining model have strong adaptability, which can effectively remove rain streaks when the distribution of rain is different from training datasets.

\subsection{Scale-Aggregation Module}

\begin{figure}[!h]
	\begin{center}
		\begin{tabular}{cccc}
			\includegraphics[width = 1\linewidth]{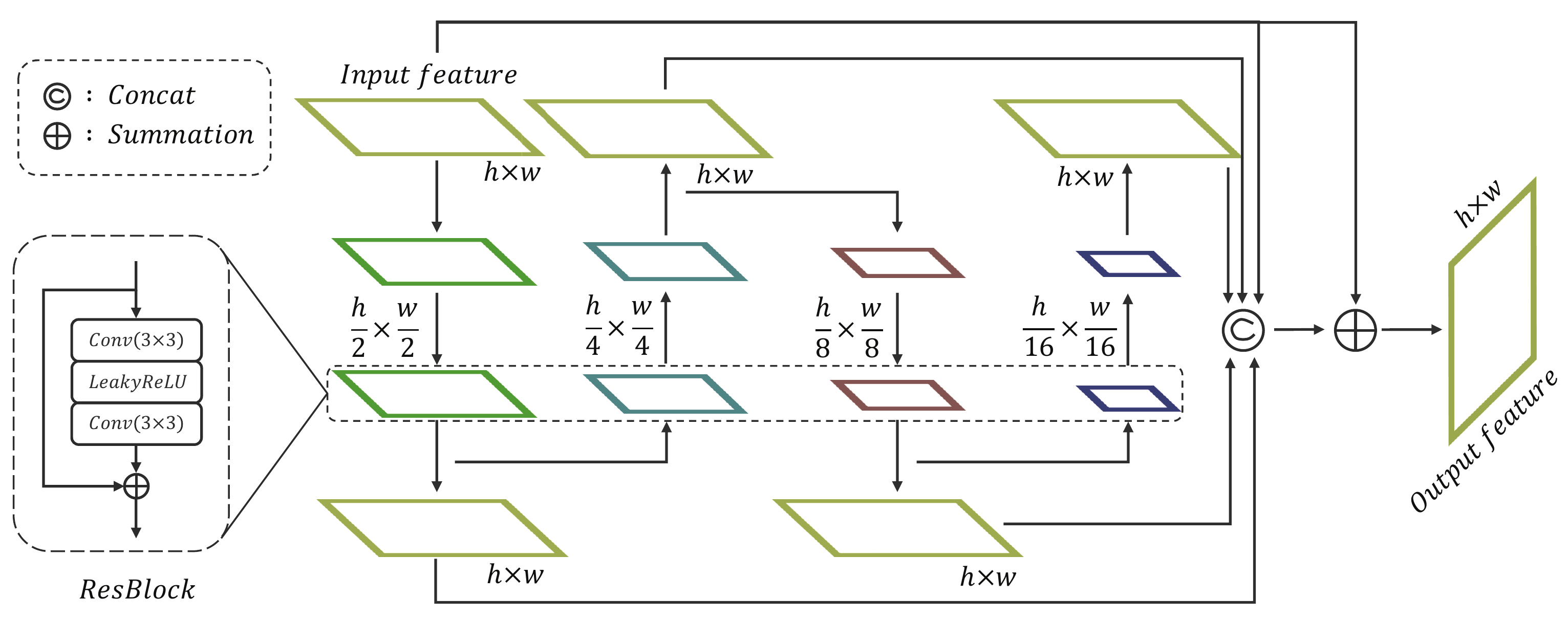}
		\end{tabular}
	\end{center}
	\caption{Scale-Aggregation Module.}
	\label{fig: Scale-Agg}
\end{figure}

Extracting features of different scales is a means to improve the performance of vision tasks.
In this paper, we design a Scale-Aggregation module to learn the features from different scales. The proposed Scale-Aggregation module not only converts the convolution into deeper features, but also can maintain the original features from shallower ones.
By designing the inner fusion between shallower and deeper layers, the proposed module can learn adaptively the features which part is more effective for rain removal.
The structure of the Scale-Aggregation module which we propose is shown in Fig~\ref{fig: Scale-Agg}.
At the beginning of this module, for a given input $X$, the feature map with a downsampling rate of 2 can be obtained by the following two formulas:
\begin{equation}X^{1}=LeakyReLU(Conv_3^{2}(X)),\end{equation}
\begin{equation}X^{1}=ResBlock_1(X^{1}),\end{equation}
where $Conv_i^{j}(\cdot)$ represents $i{\times}i$ convolution operation with stride j, $LeakyReLU(\cdot)$ represents an activation function with the parameter of 0.2, $ResBlock(\cdot)$ consists of an activation function between two $3\times3$ convolution layers. Let $X^{0}=X$, the general formulas of the above process are:
\begin{equation}X^{i}=LeakyReLU(Conv_3^{2}(X^{i-1})),\end{equation}
\begin{equation}X^{i}=ResBlock_i(X^{i}),\end{equation}
where $i=1,2,\dots,n$. Eventually, we upsample each $X^i$ to the scale of original input and concatenate these results with the input features following a $1\times1$ convolution:
\begin{equation}X^{out}=Conv_1^{1}\{Concat[X^{0},Up(X^{1}),Up(X^{2}),\dots,Up(X^{n})]\},\end{equation}
where $Up(\cdot)$ is an interpolation operation, $Concat(\cdot)$ represents the concatenation operation. As a result, we get the feature map restored to the original shape containing feature information of different scales.

\subsection{Self-Calibrated Convolution}

\begin{figure}[!h]
	\begin{center}
		\begin{tabular}{cccc}
			\includegraphics[width = 1\linewidth]{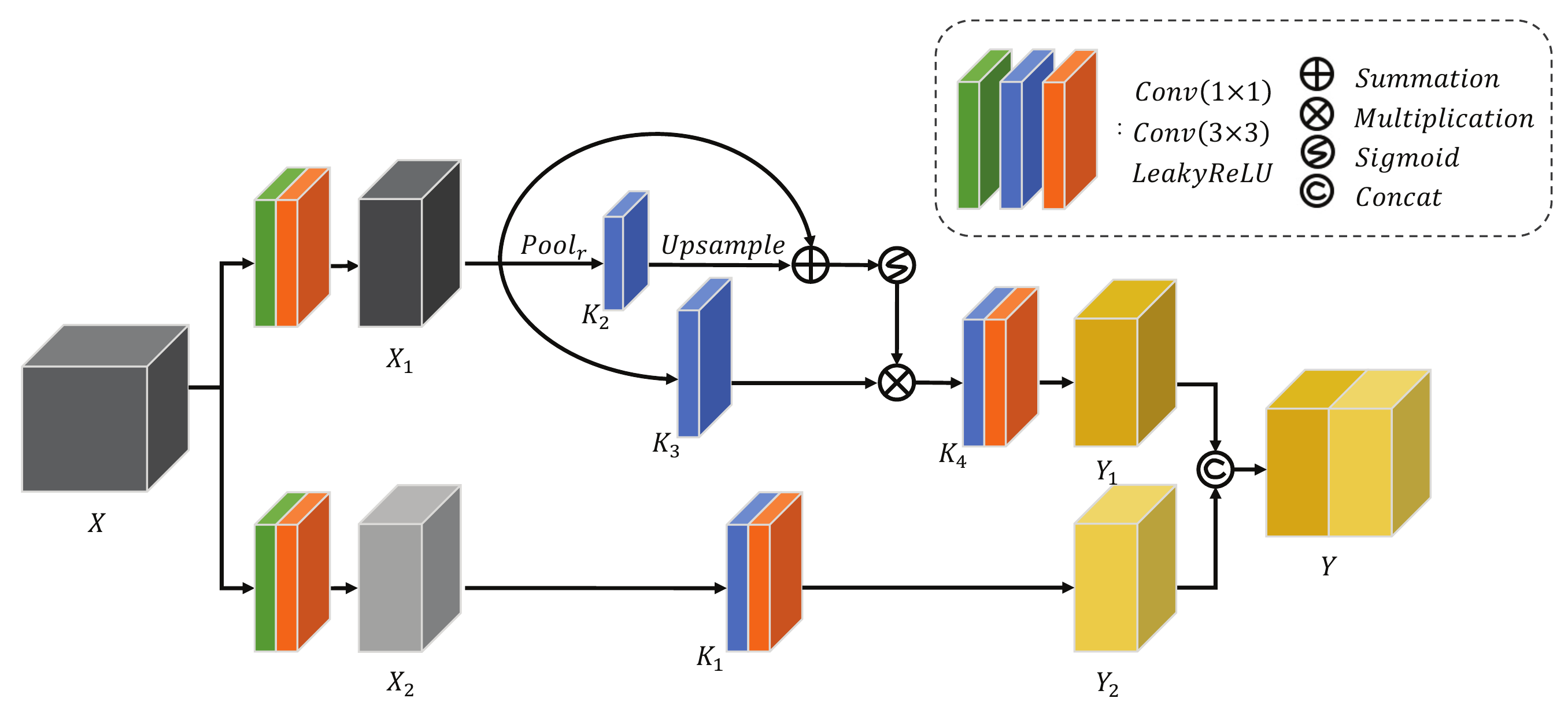}
		\end{tabular}
	\end{center}
	\caption{Self-Calibrated Convolution.}
	\label{fig: Self-Calibrated Convolution}
\end{figure}

The structure of deep CNNs is becoming more and more complicated, which can enhance the learning ability of the network.
The novel convolution called Self-Calibrated convolution~\cite{scnet} that we introduce considers improving the feature transformation process in convolution, as Fig~\ref{fig: Self-Calibrated Convolution}.

A given group of filter sets $K$ with the shape $(C,C,k_h,k_w)$ is divided into four parts, i.e., $[K_1,K_2,K_3,K_4]$, where $k_h$ and $k_w$ are the spatial height and width, respectively. Each part, whose shape is $(C/2,C/2,k_h,k_w)$, is responsible for performing different functions.
After splitting filters, the input $X$ with the channel $C$ is split into $X_1$ and $X_2$ through $1\times1$ convolution with the channel $C/2$.

In Self-Calibrated convolution, we perform feature transform at two scales: the original scale and the smaller scale after downsampling. For a given $X$, we adopt average pooling to reduce the scale:
\begin{equation}T_1=AvgPool_r(X_1),\end{equation}
where $r$ is the downsampling rate and stride of the pooling process.
Benefiting from the downsampling operation, the receptive field at each spatial location can be effectively expanded. Next, $T_1$ can be used as an input to the filter $K_2$ following the upsample operation which restores the feature back to the original scale, resulting in:
\begin{equation}X_1^{'}=Up(\mathcal{F}_2(T_1))=Up(T_1*K_2),\end{equation}
where $\mathcal{F}_2(T_1)=T_1*K_2$ is a simplified form of convolution. Then, the calibrated operation can be formulated as:
\begin{equation}Y_1^{'}=\mathcal{F}_3(X_1)\odot{Sigmoid(X_1+X_1^{'})},\end{equation}
where $\mathcal{F}_3(X_1)=X_1*K_3$, $Sigmoid(\cdot)$ is an activation function. The final result of the calibrated branch is calculated:
\begin{equation}Y_1=\mathcal{F}_4(Y_1^{'}),\end{equation}
where $\mathcal{F}_4(Y_1^{'})=Y_1^{'}*K_4$.

The other half of result can be obtained from another branch that does not require scale transformation. The formula is as follows:
\begin{equation}Y_2=\mathcal{F}_1(X_2)=X_2*K_1.\end{equation}

Finally, we concatenate $Y_1$ and $Y_2$ in order to get the final result $Y$.
Reviewing the entire Self-Calibrated convolution, it enables each spatial position to adaptively encode the context from a long-range region, which is also a huge difference between it and traditional convolution.

\begin{table*}[!t]
	\caption{Quantitative experiments evaluated on three synthetic datasets.
		The best results are highlighted in boldface.}
	\centering
	\scalebox{0.95}{
		\begin{tabular}{ccccccccccccccccc}
			\toprule
			& \multicolumn{2}{c}{DDN}                   & \multicolumn{2}{c}{RESCAN}    & \multicolumn{2}{c}{REHEN} & \multicolumn{2}{c}{PreNet}  & \multicolumn{2}{c}{SpaNet}       & \multicolumn{2}{c}{SSIR}       & \multicolumn{2}{c}{Ours}
			\\
			\midrule
			Dataset       & PSNR& SSIM & PSNR&SSIM & PSNR&SSIM & PSNR & SSIM& PSNR & SSIM & PSNR & SSIM & PSNR & SSIM    \\
			\midrule
			Rain100H   &22.26&0.69    &25.92&0.84  & 27.52 & 0.86& 27.89&0.89  &  26.54 &0.90 &22.47 &0.71 & \textbf{30.02}&\textbf{0.92} \\
			\midrule
			Rain100L   &34.85&0.95    &36.12&0.97 & 37.91 & 0.98&36.69&0.98   &36.20 &0.98 &32.37 &0.92 & \textbf{38.65}&\textbf{0.99} \\
			\midrule
			Rain12     &28.66&0.91     &33.75&0.95   &35.84&0.96  &34.77&0.96   &33.59&0.96   &24.14&0.78  & \textbf{37.02}&\textbf{0.97} \\
			\bottomrule
	\end{tabular}}
	\label{tab: the results in synthetic datasets}
\end{table*}

\begin{figure*}[!t]
	\begin{center}
		\begin{tabular}{ccccccccc}
			\includegraphics[width = 0.106\linewidth]{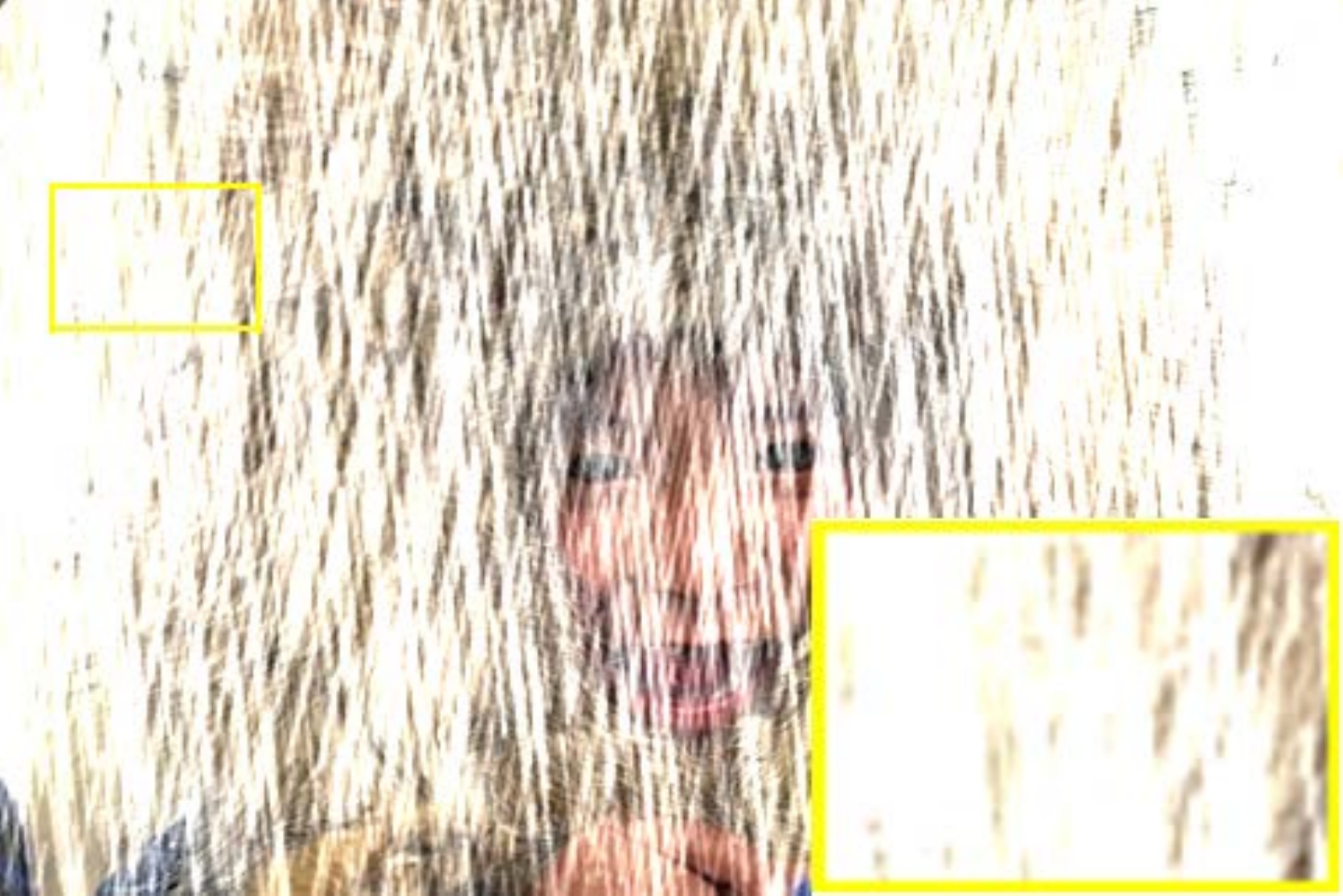}&\hspace{-4mm}
			\includegraphics[width = 0.106\linewidth]{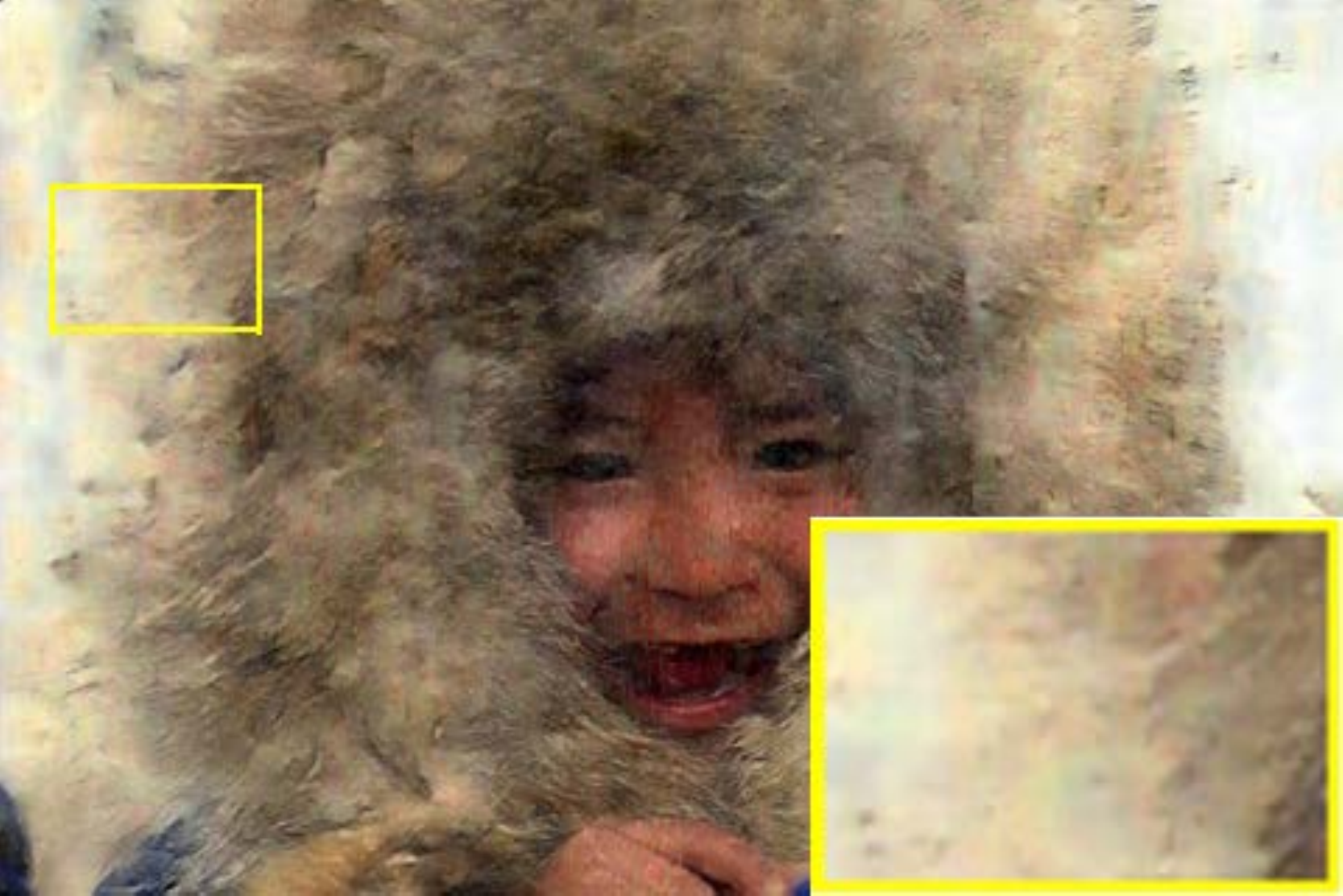}&\hspace{-4mm}
			\includegraphics[width = 0.106\linewidth]{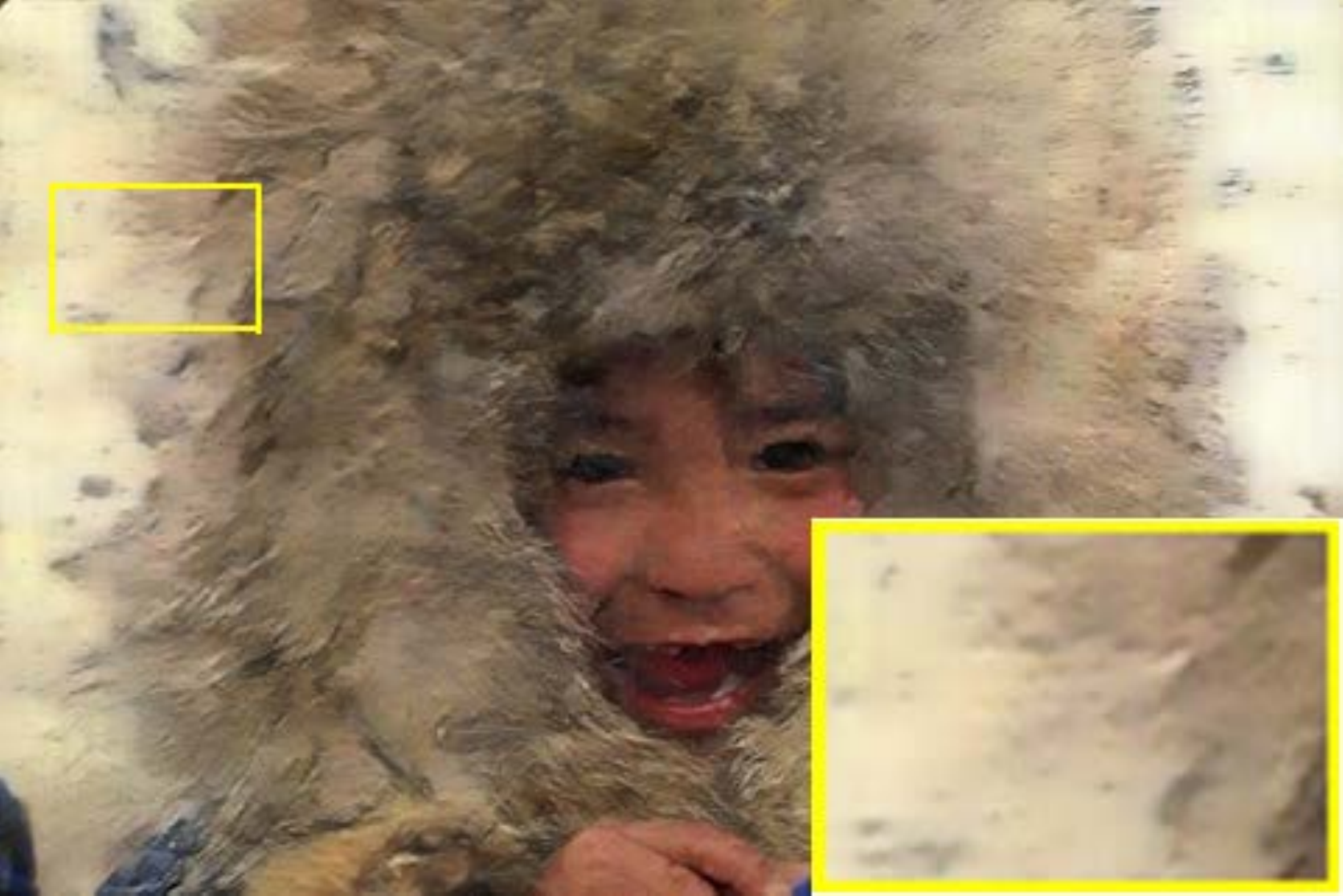}&\hspace{-4mm}
			\includegraphics[width = 0.106\linewidth]{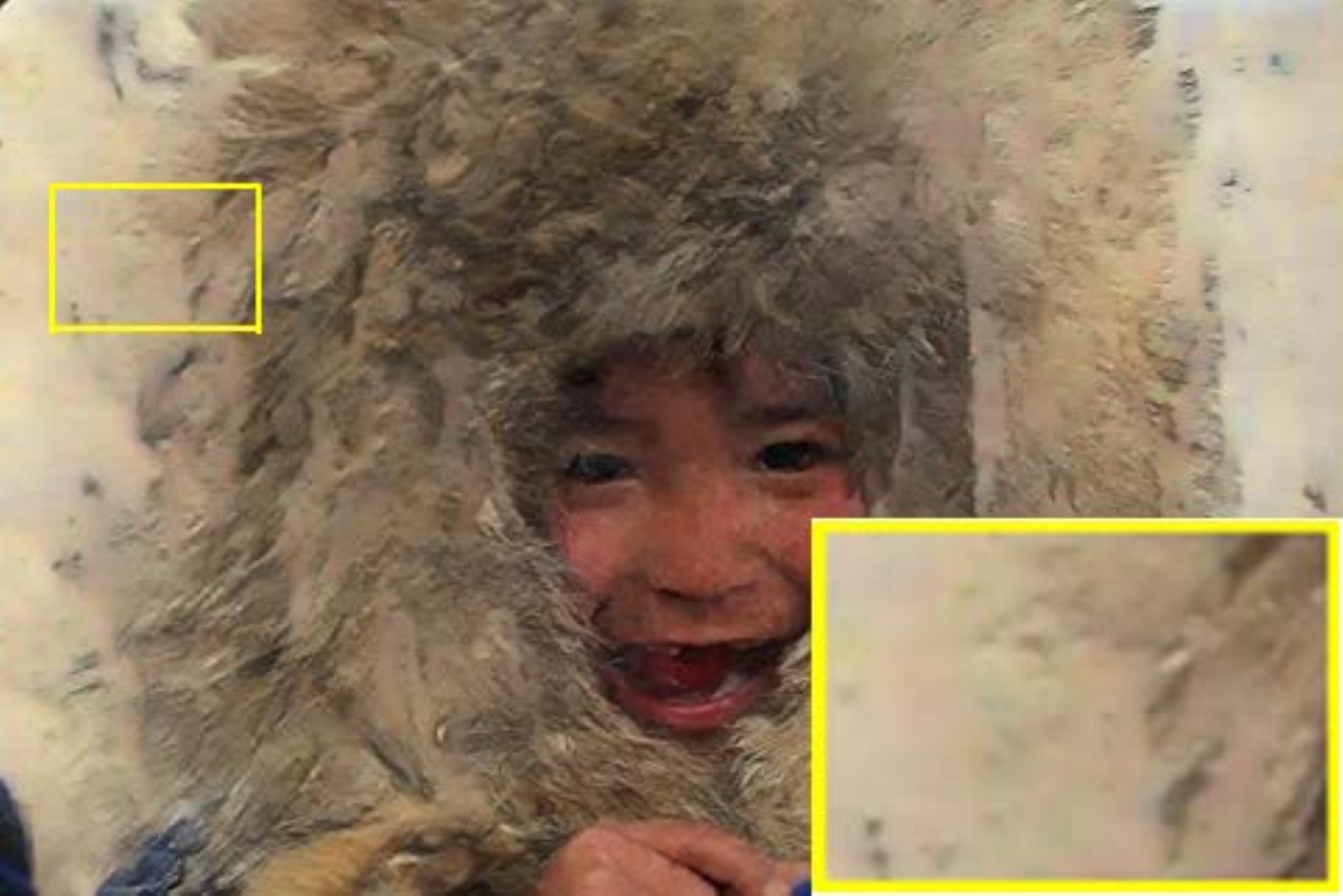}&\hspace{-4mm}
			\includegraphics[width = 0.106\linewidth]{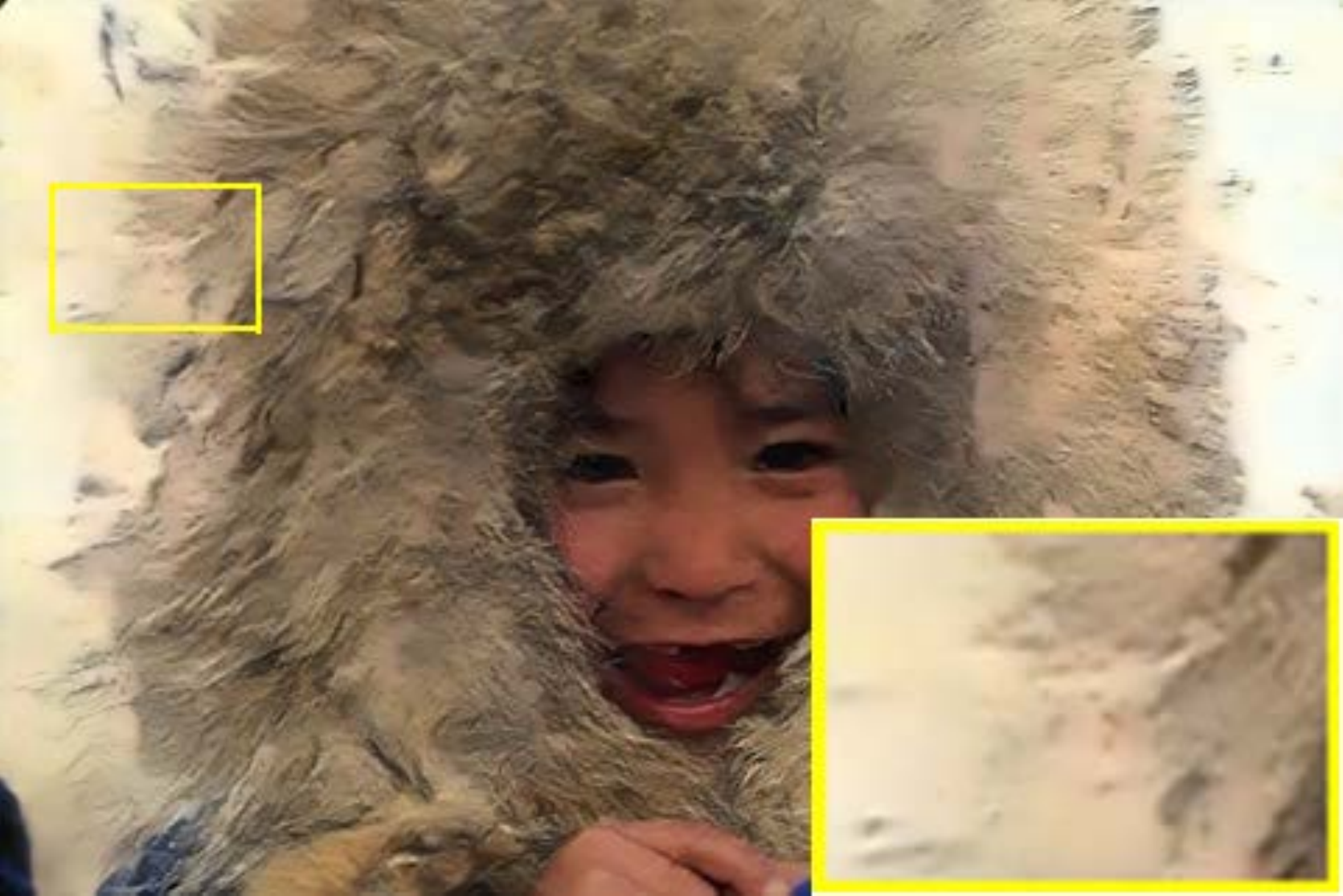}&\hspace{-4mm}
			\includegraphics[width = 0.106\linewidth]{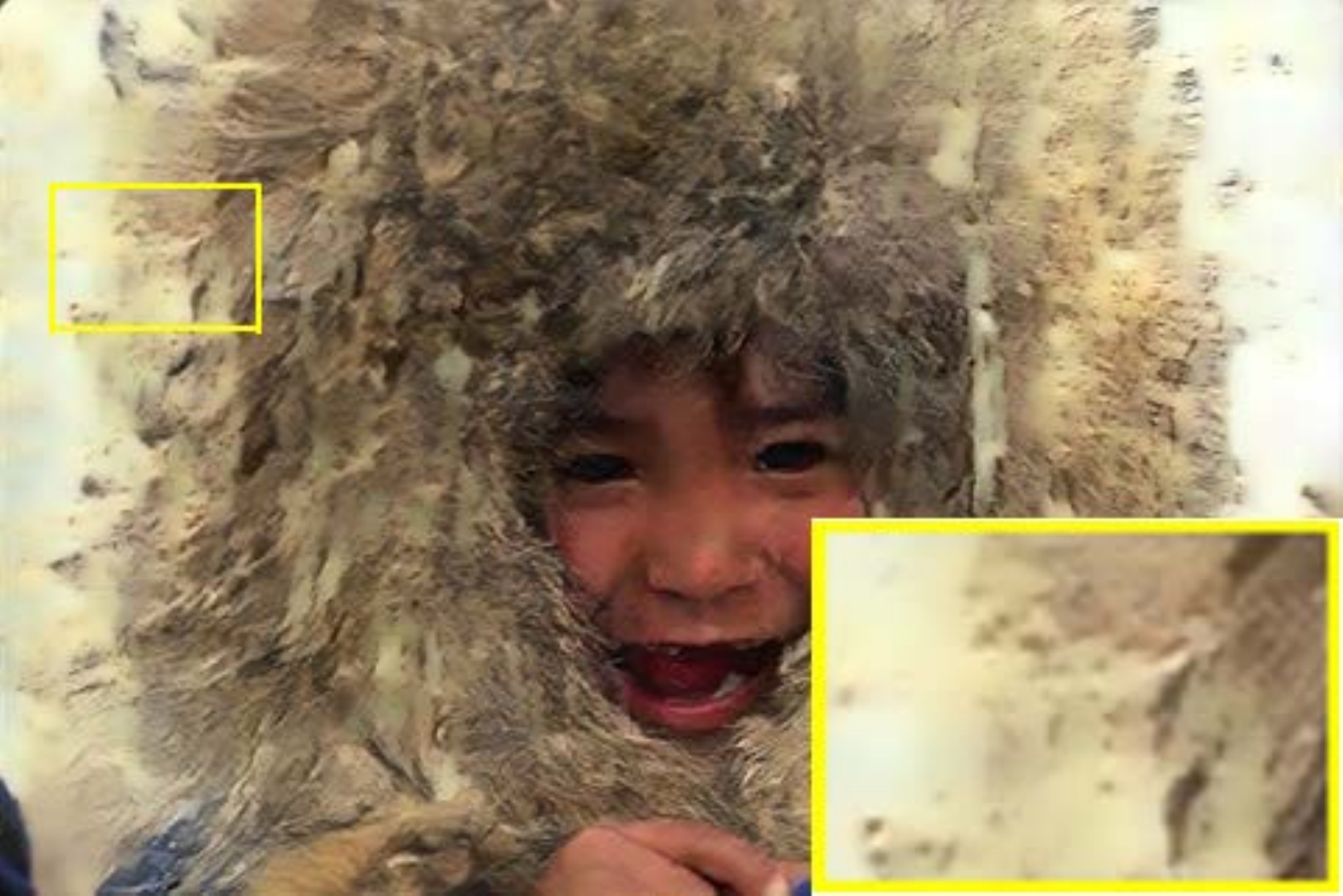}&\hspace{-4mm}
			\includegraphics[width = 0.106\linewidth]{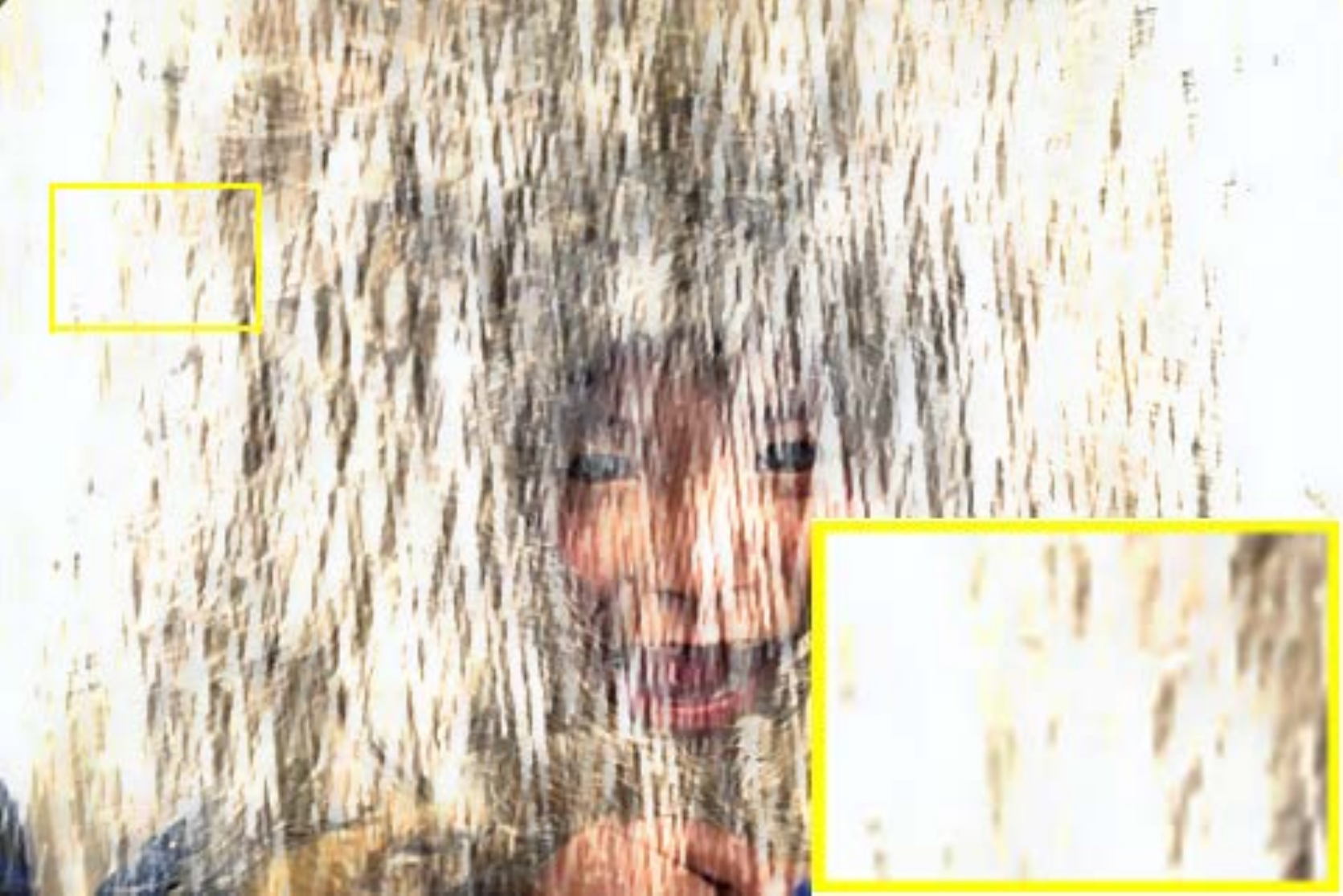}&\hspace{-4mm}
			\includegraphics[width = 0.106\linewidth]{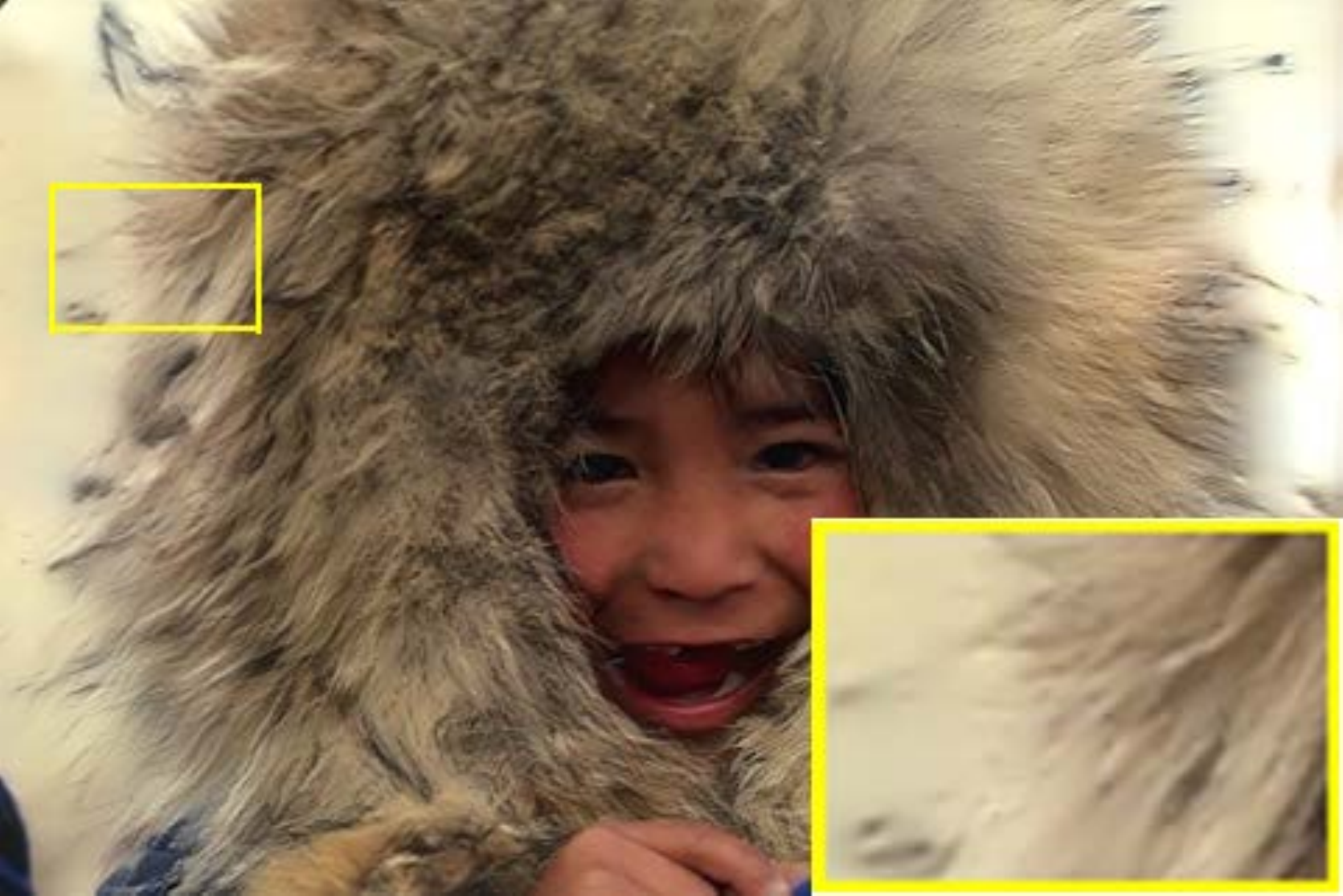}&\hspace{-4mm}
			\includegraphics[width = 0.106\linewidth]{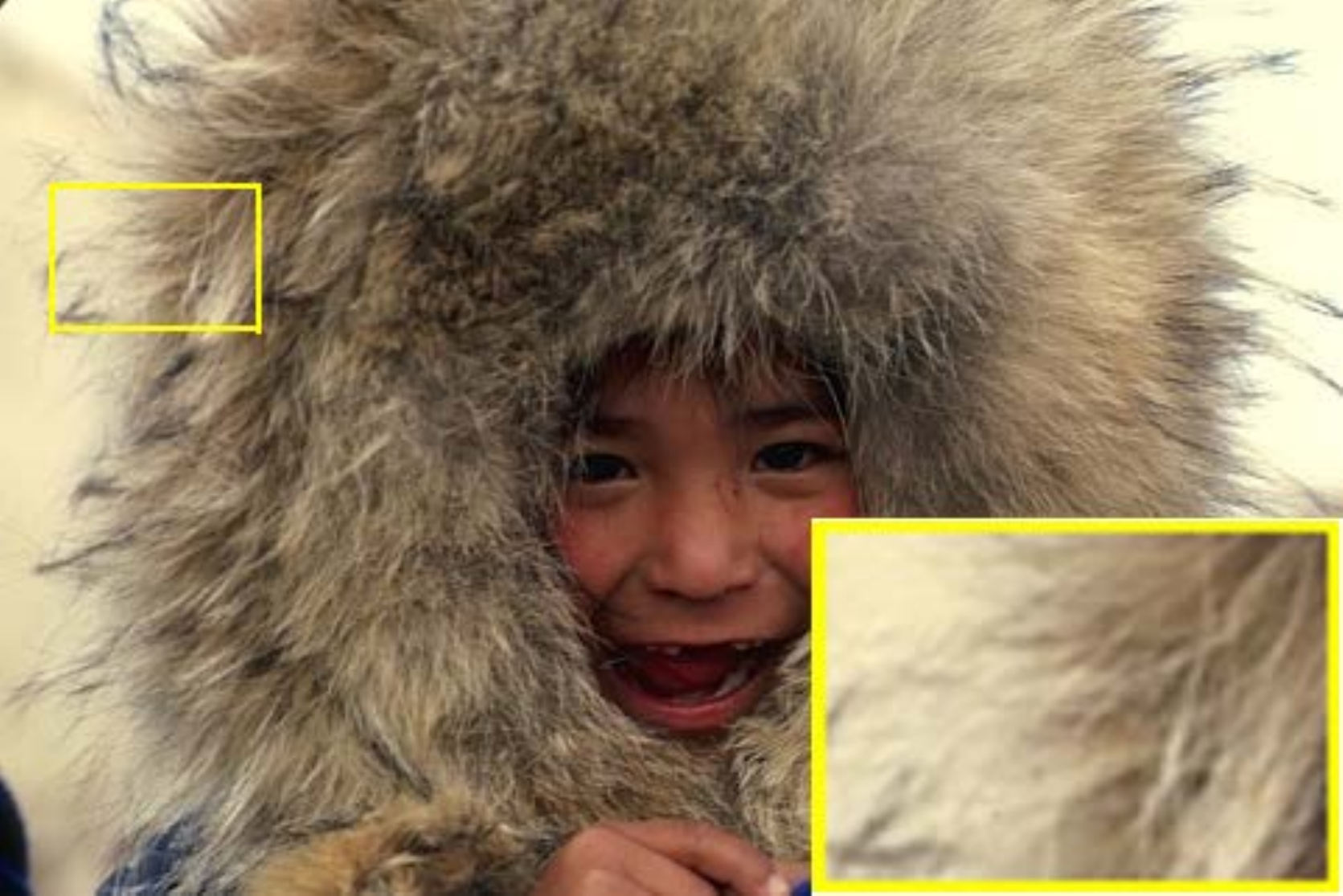}
			\\
			\includegraphics[width = 0.106\linewidth]{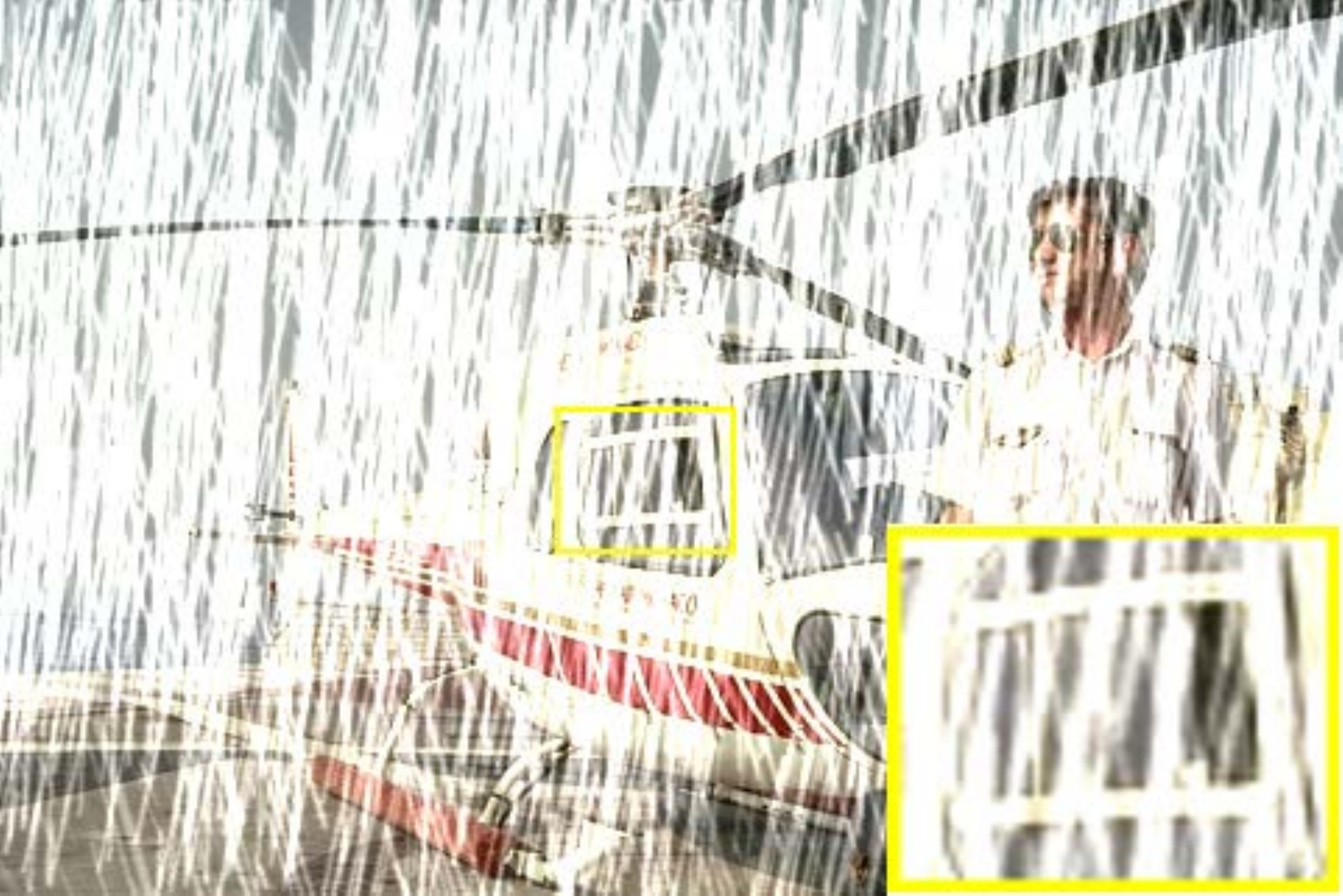}&\hspace{-4mm}
			\includegraphics[width = 0.106\linewidth]{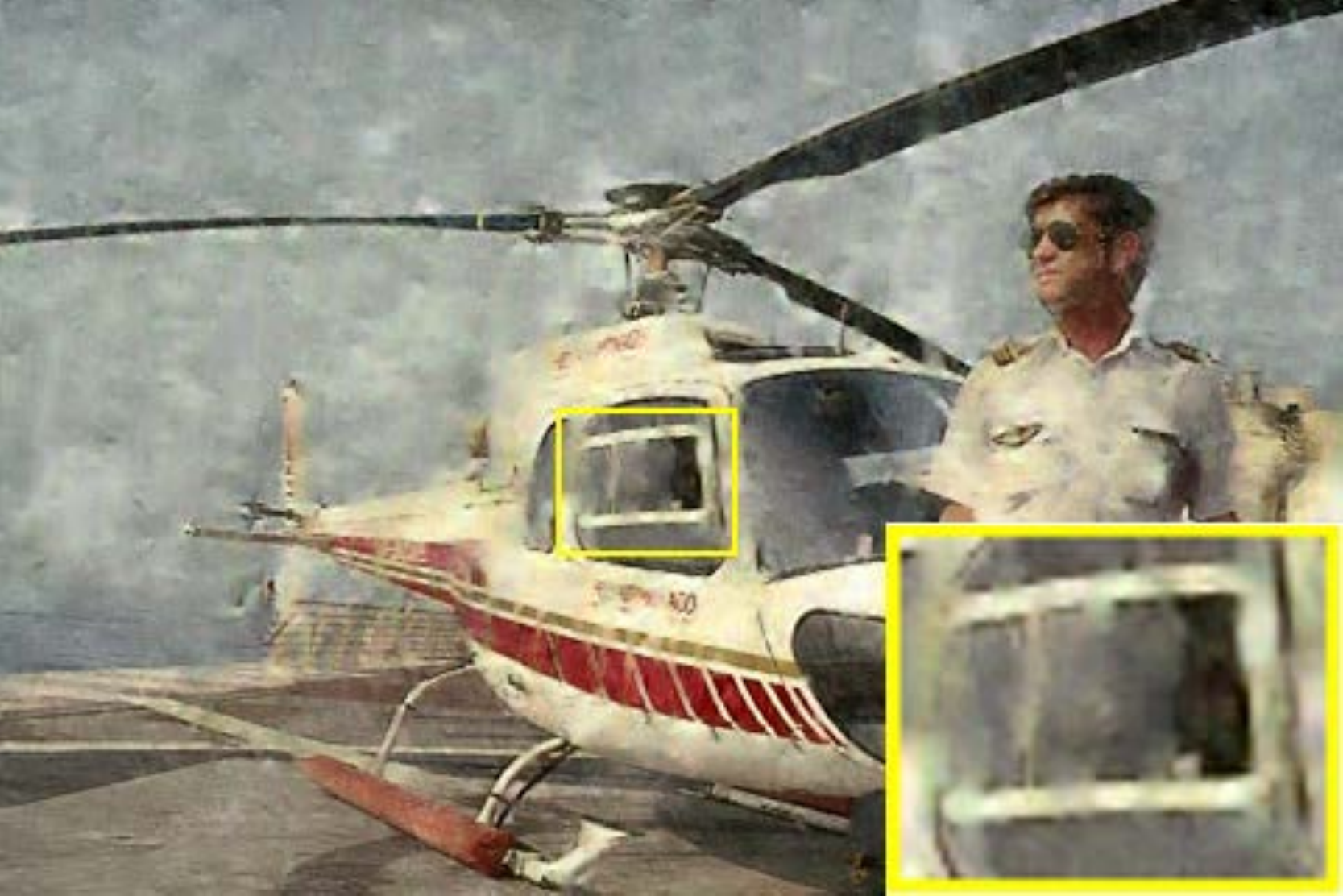}&\hspace{-4mm}
			\includegraphics[width = 0.106\linewidth]{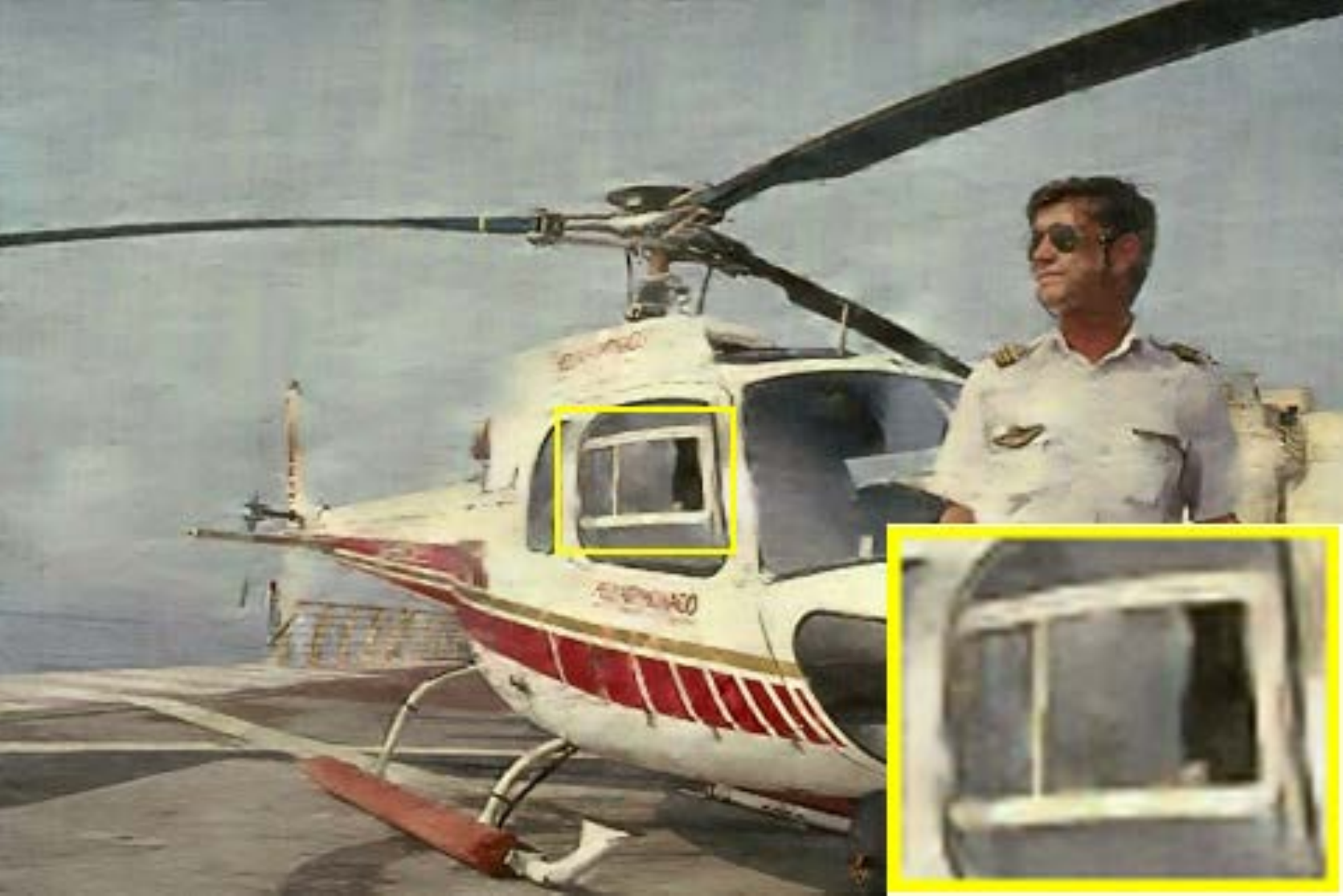}&\hspace{-4mm}
			\includegraphics[width = 0.106\linewidth]{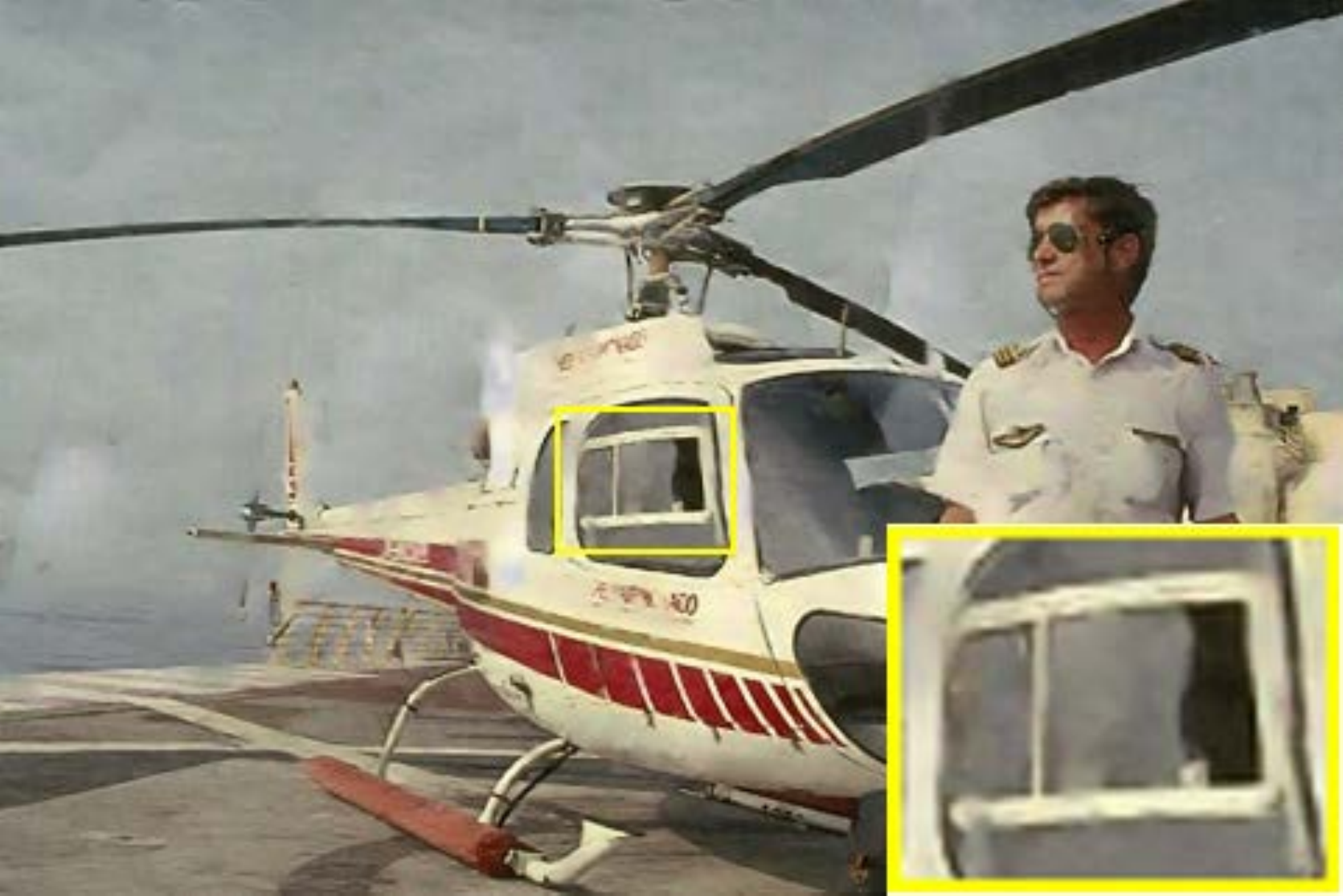}&\hspace{-4mm}
			\includegraphics[width = 0.106\linewidth]{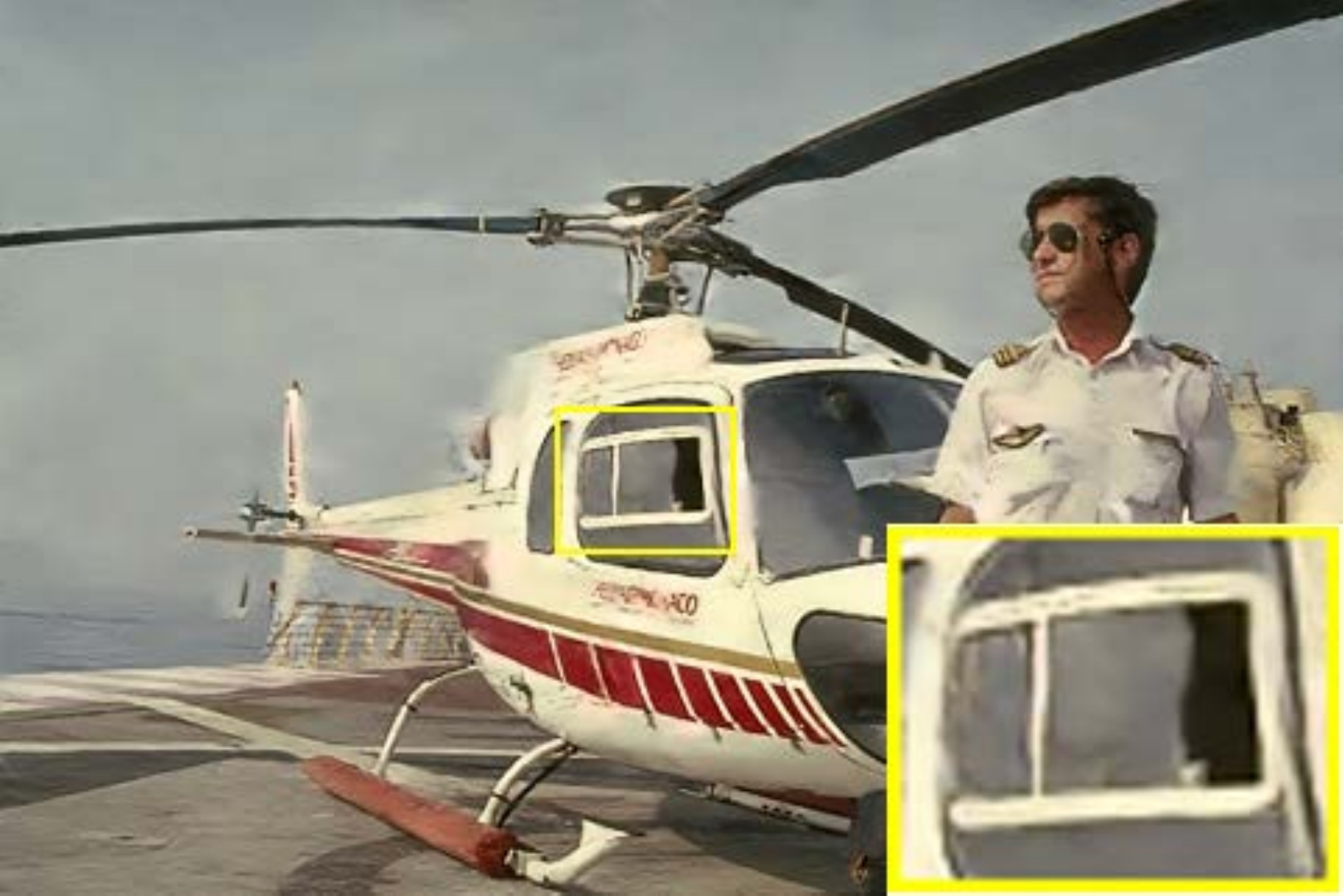}&\hspace{-4mm}
			\includegraphics[width = 0.106\linewidth]{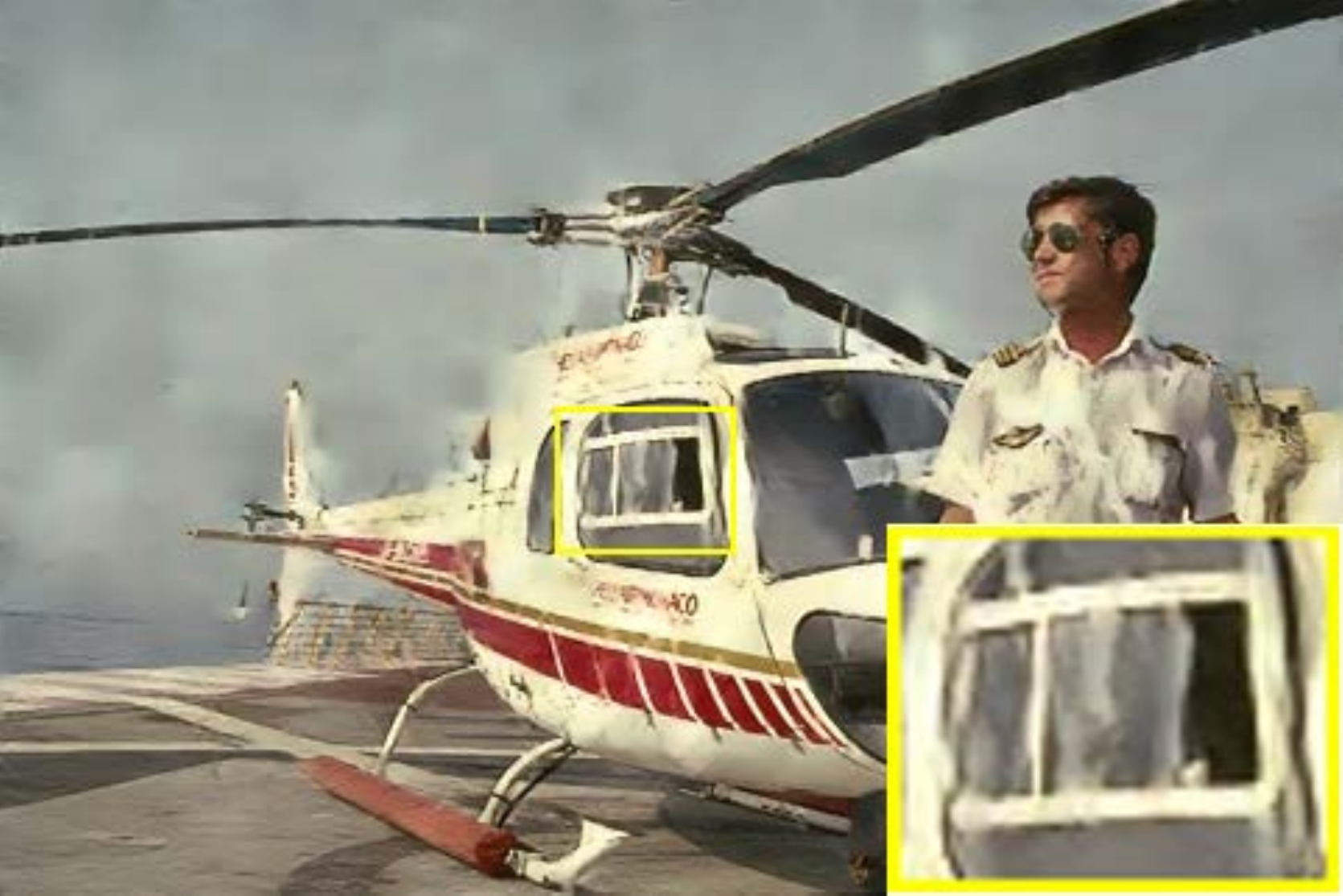}&\hspace{-4mm}
			\includegraphics[width = 0.106\linewidth]{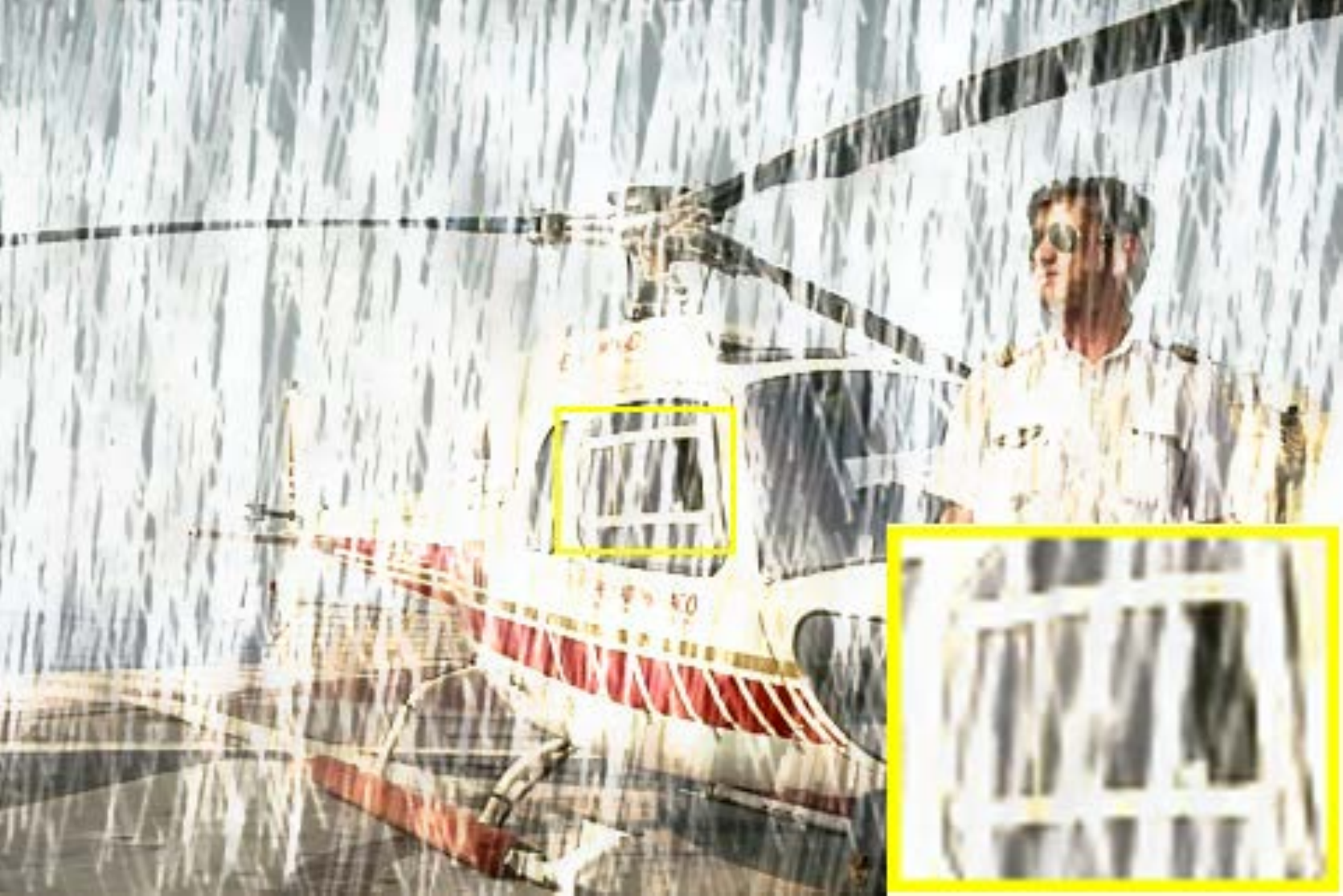}&\hspace{-4mm}
			\includegraphics[width = 0.106\linewidth]{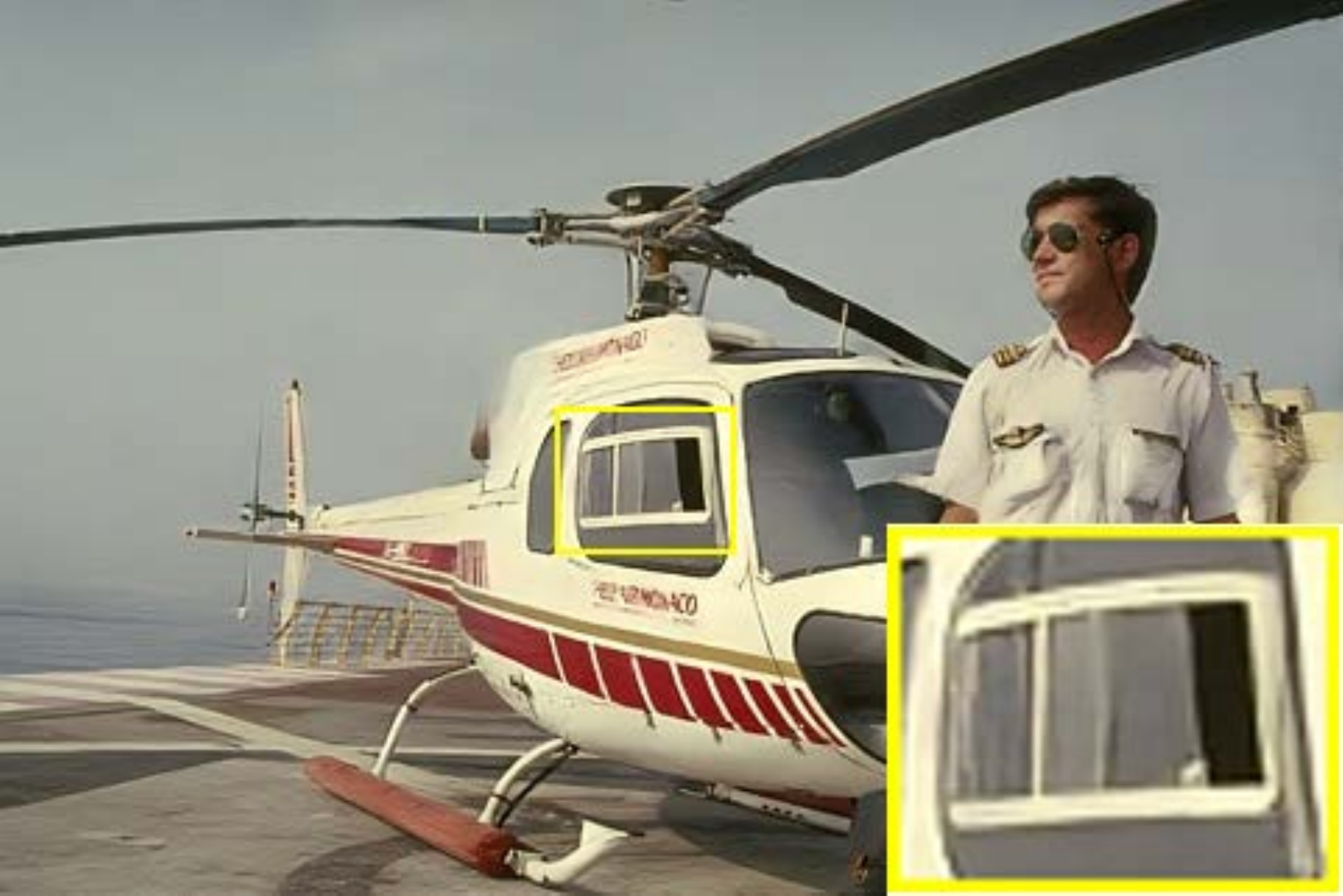}&\hspace{-4mm}
			\includegraphics[width = 0.106\linewidth]{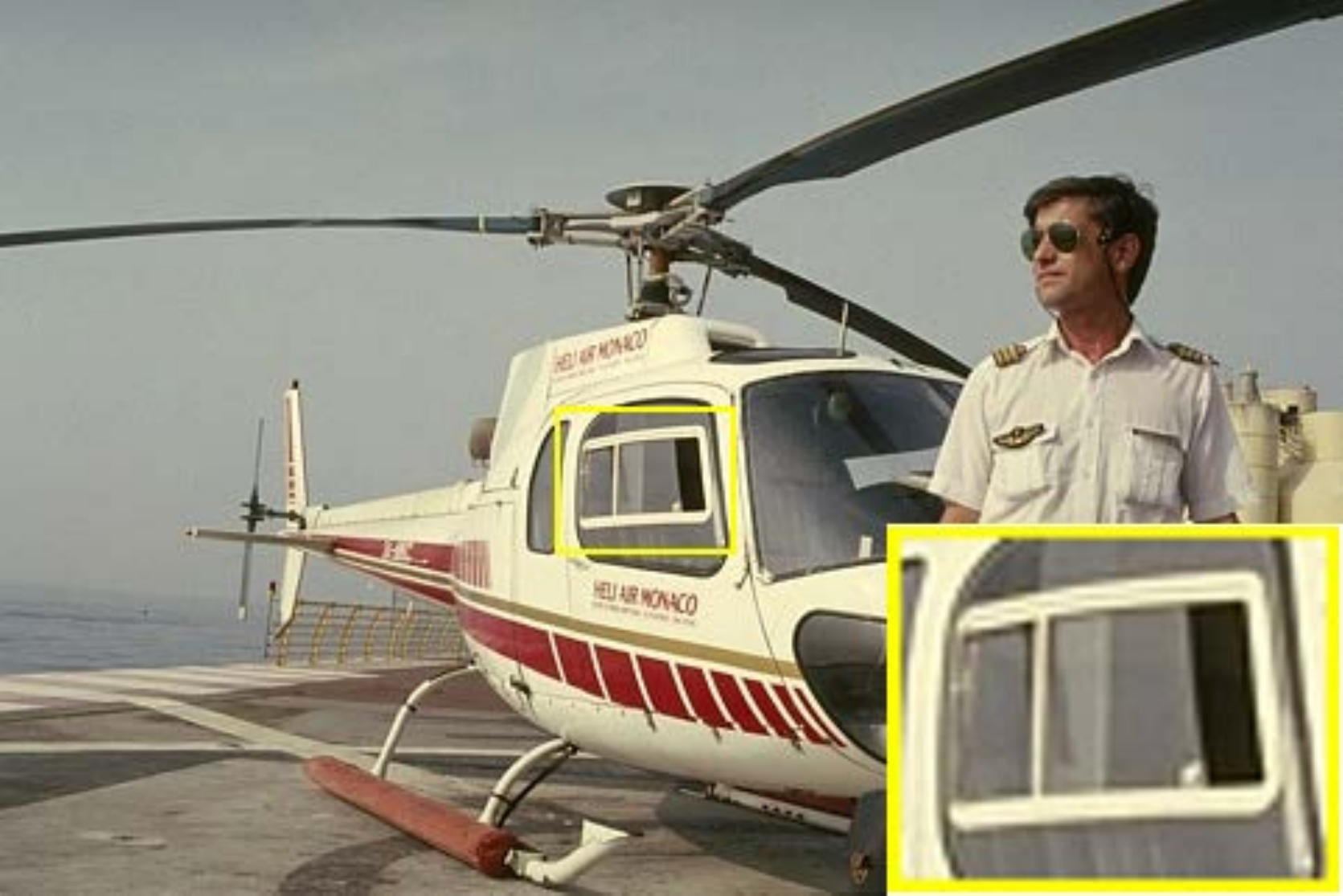}
			\\
			\includegraphics[width = 0.106\linewidth]{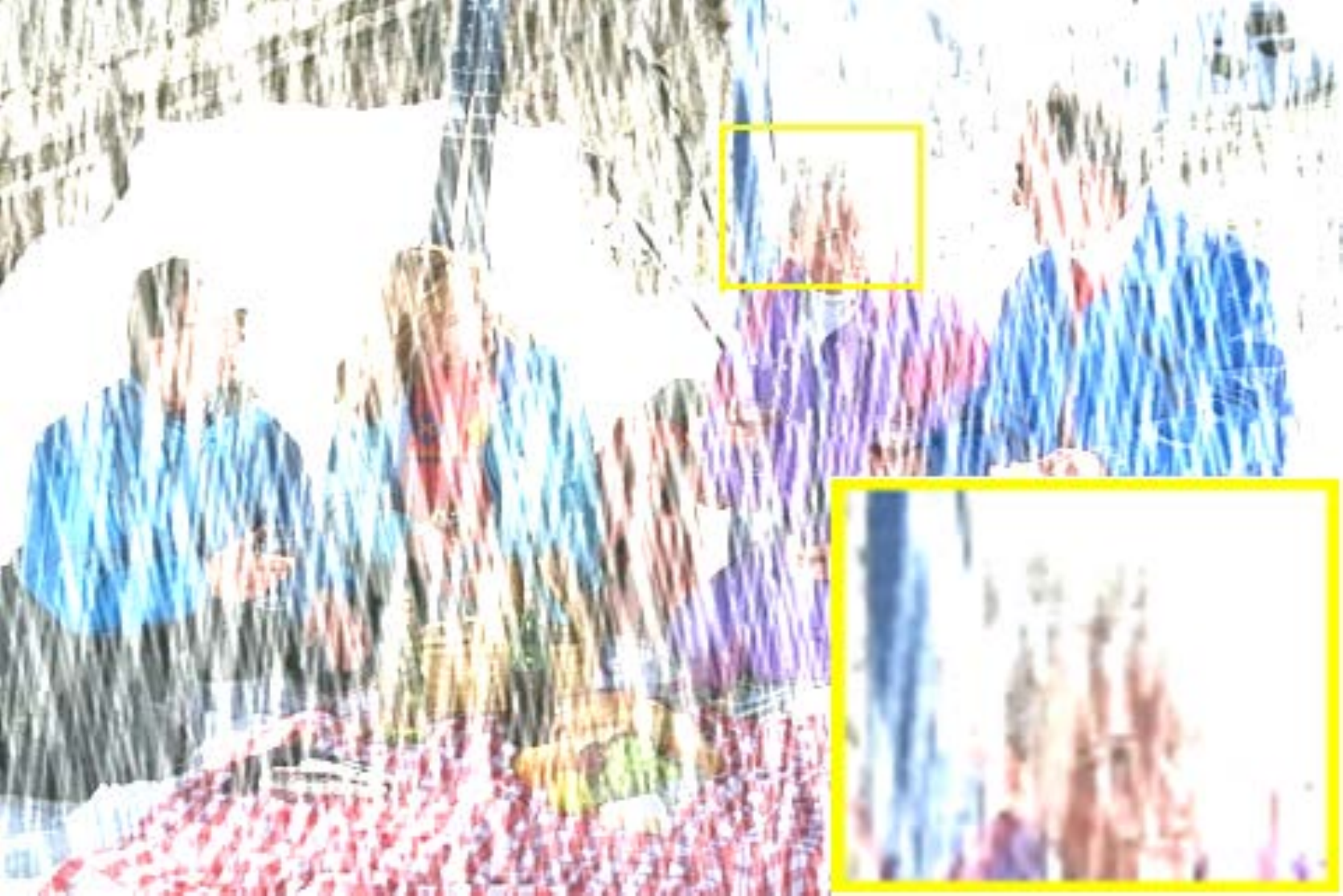}&\hspace{-4mm}
			\includegraphics[width = 0.106\linewidth]{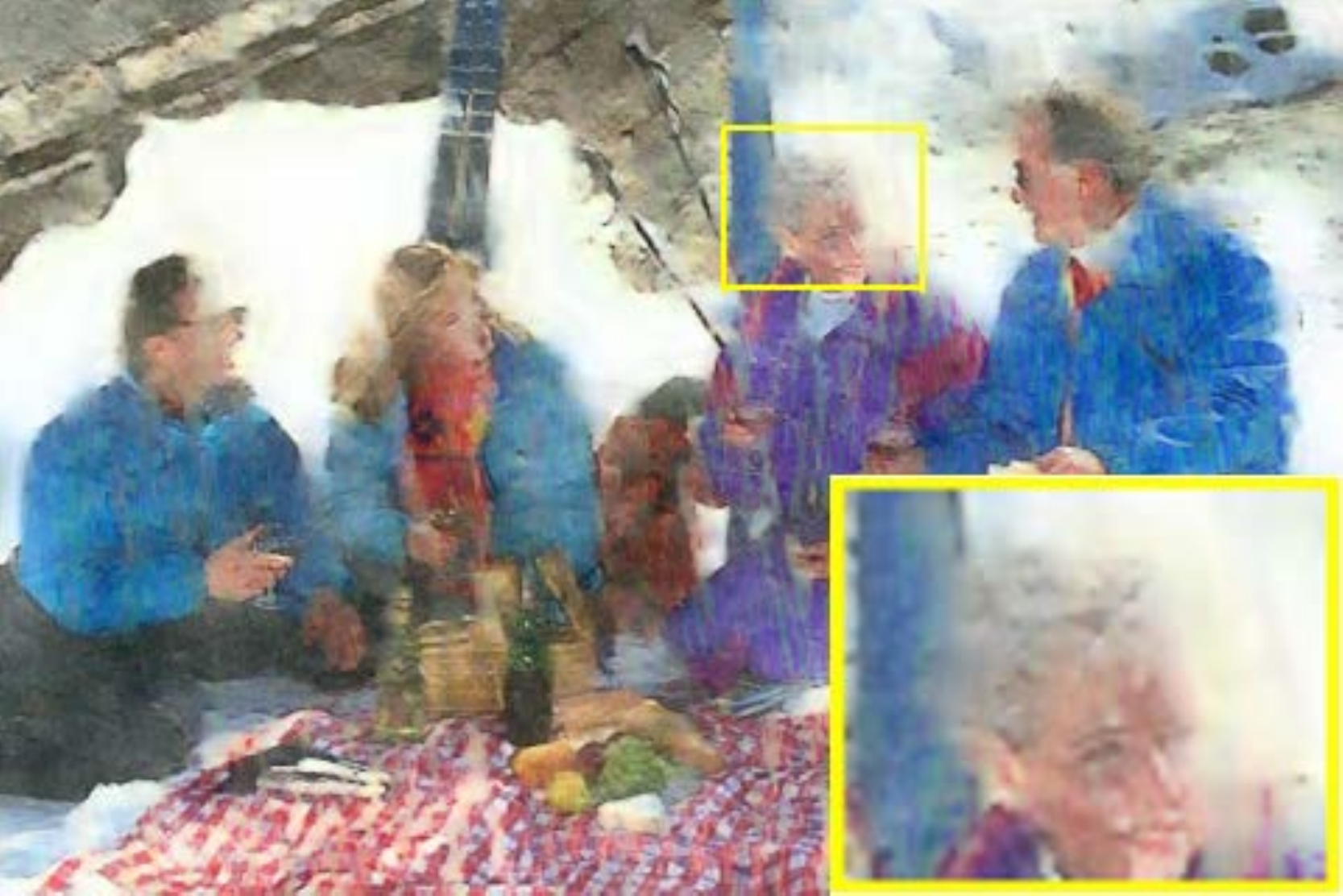}&\hspace{-4mm}
			\includegraphics[width = 0.106\linewidth]{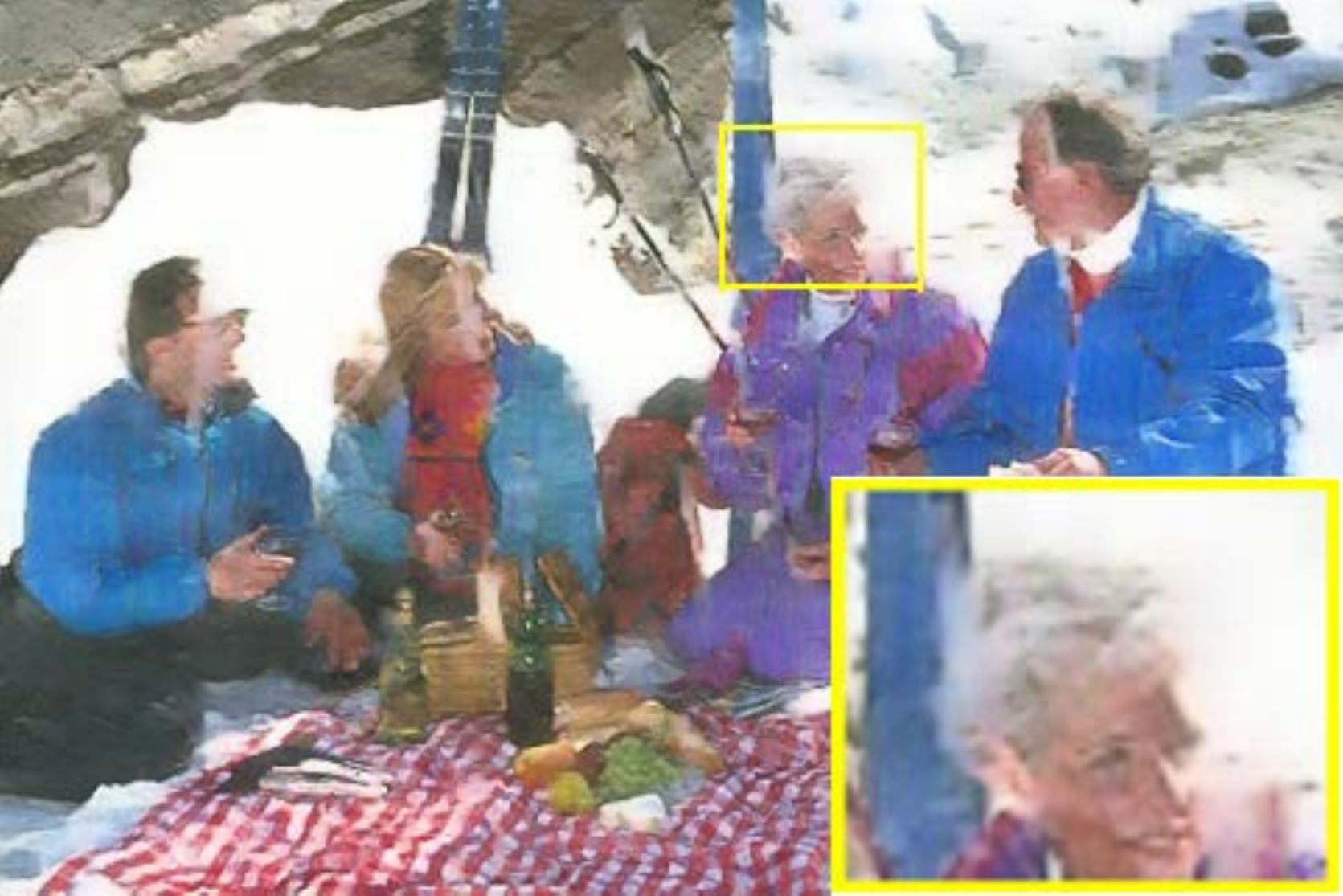}&\hspace{-4mm}
			\includegraphics[width = 0.106\linewidth]{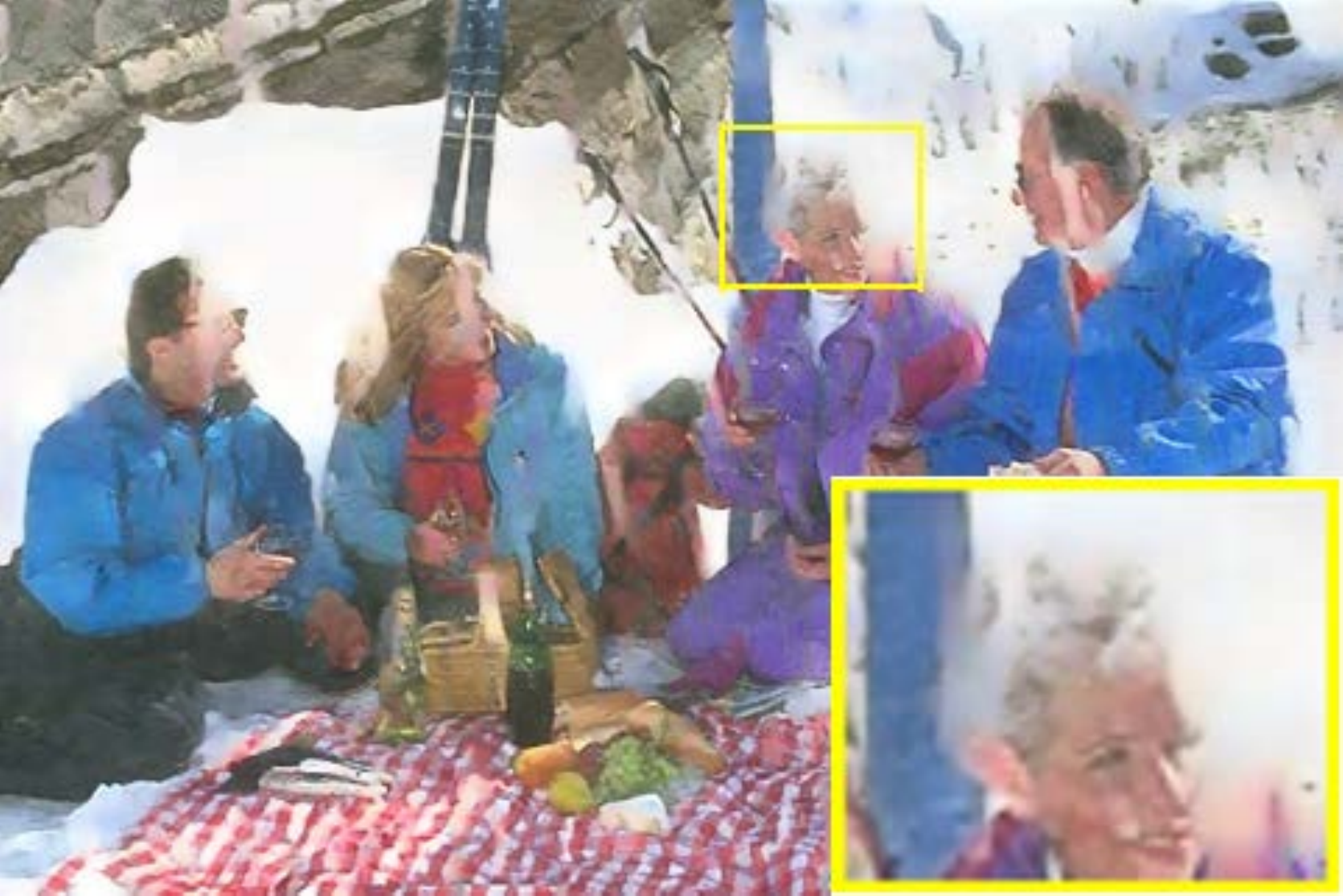}&\hspace{-4mm}
			\includegraphics[width = 0.106\linewidth]{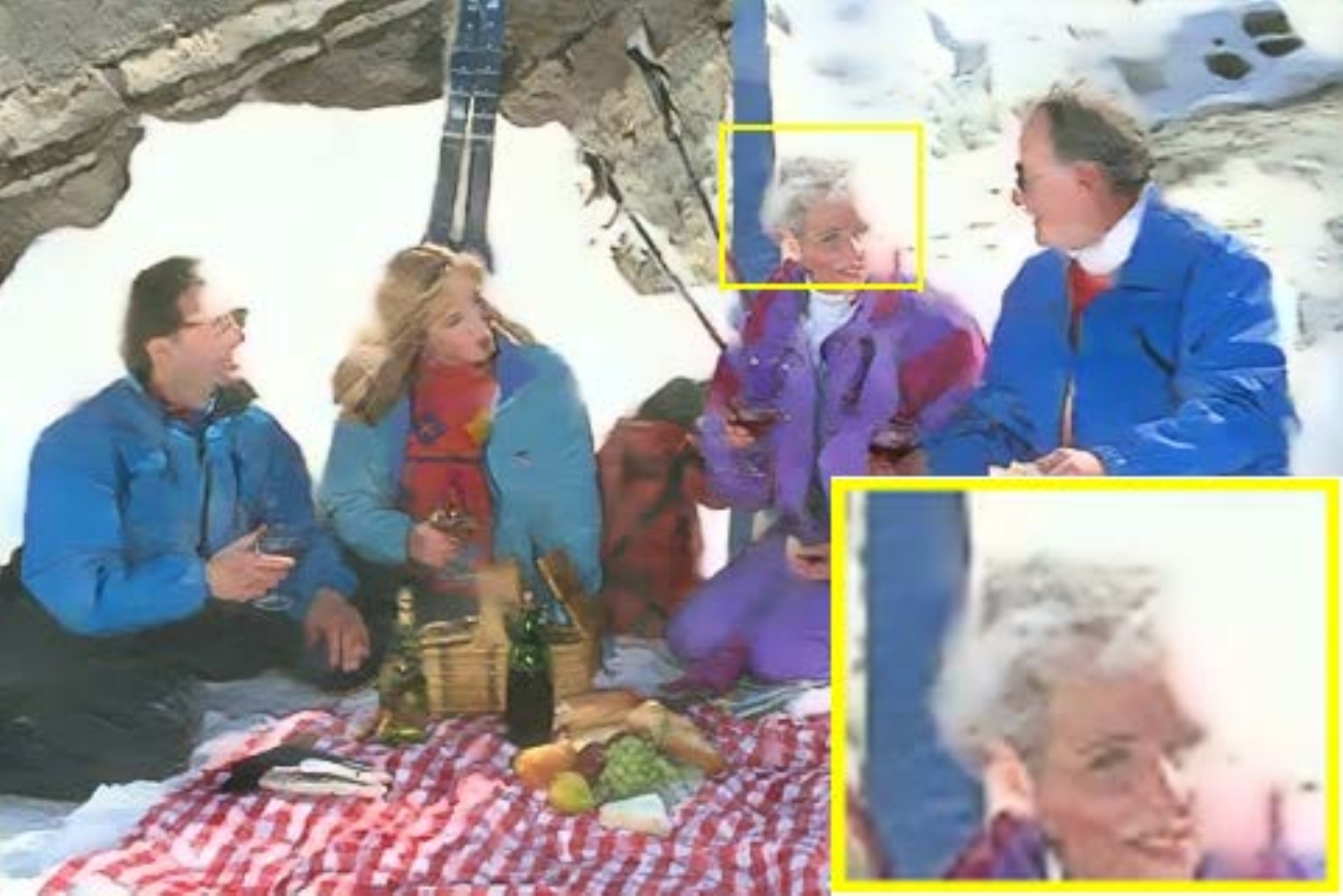}&\hspace{-4mm}
			\includegraphics[width = 0.106\linewidth]{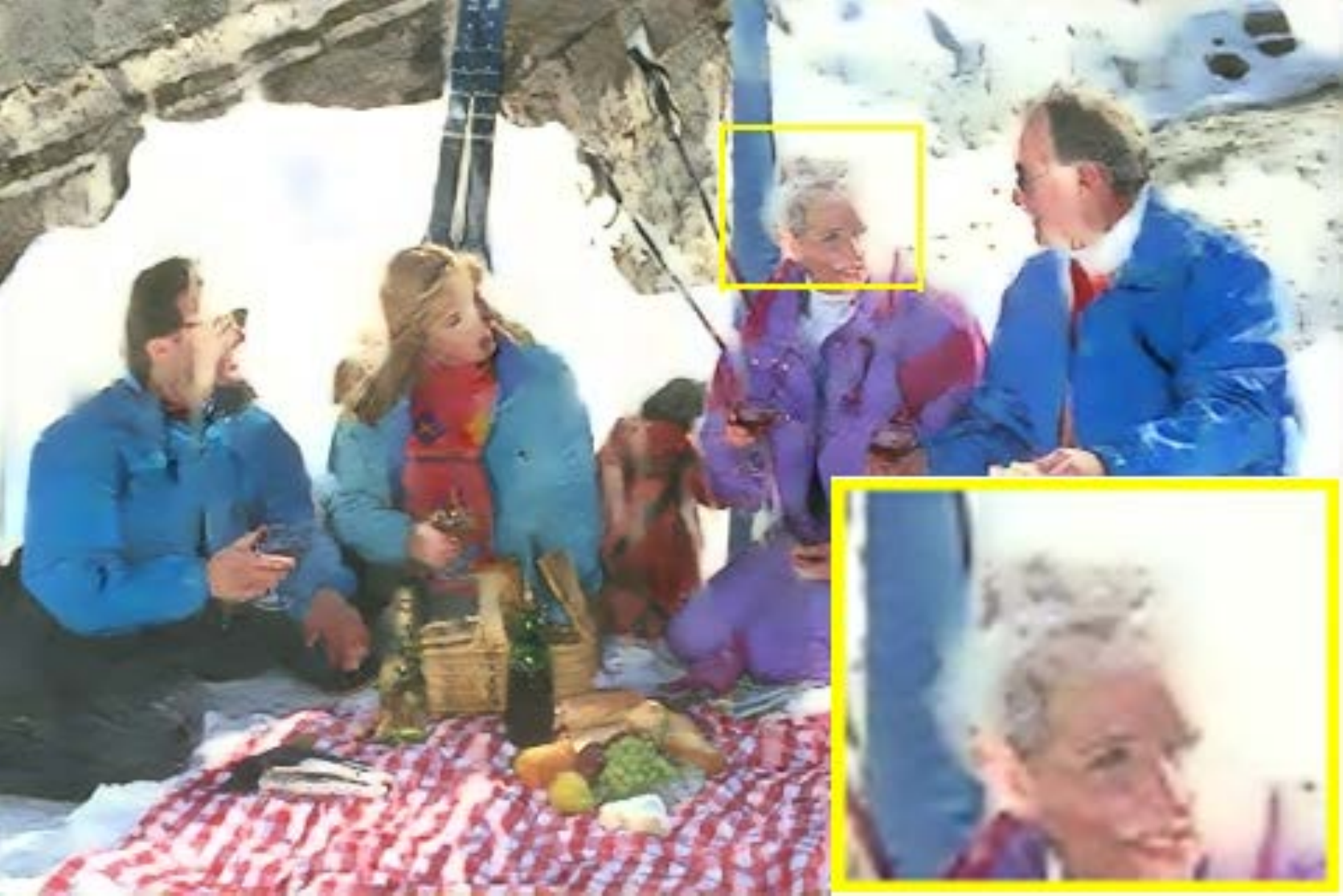}&\hspace{-4mm}
			\includegraphics[width = 0.106\linewidth]{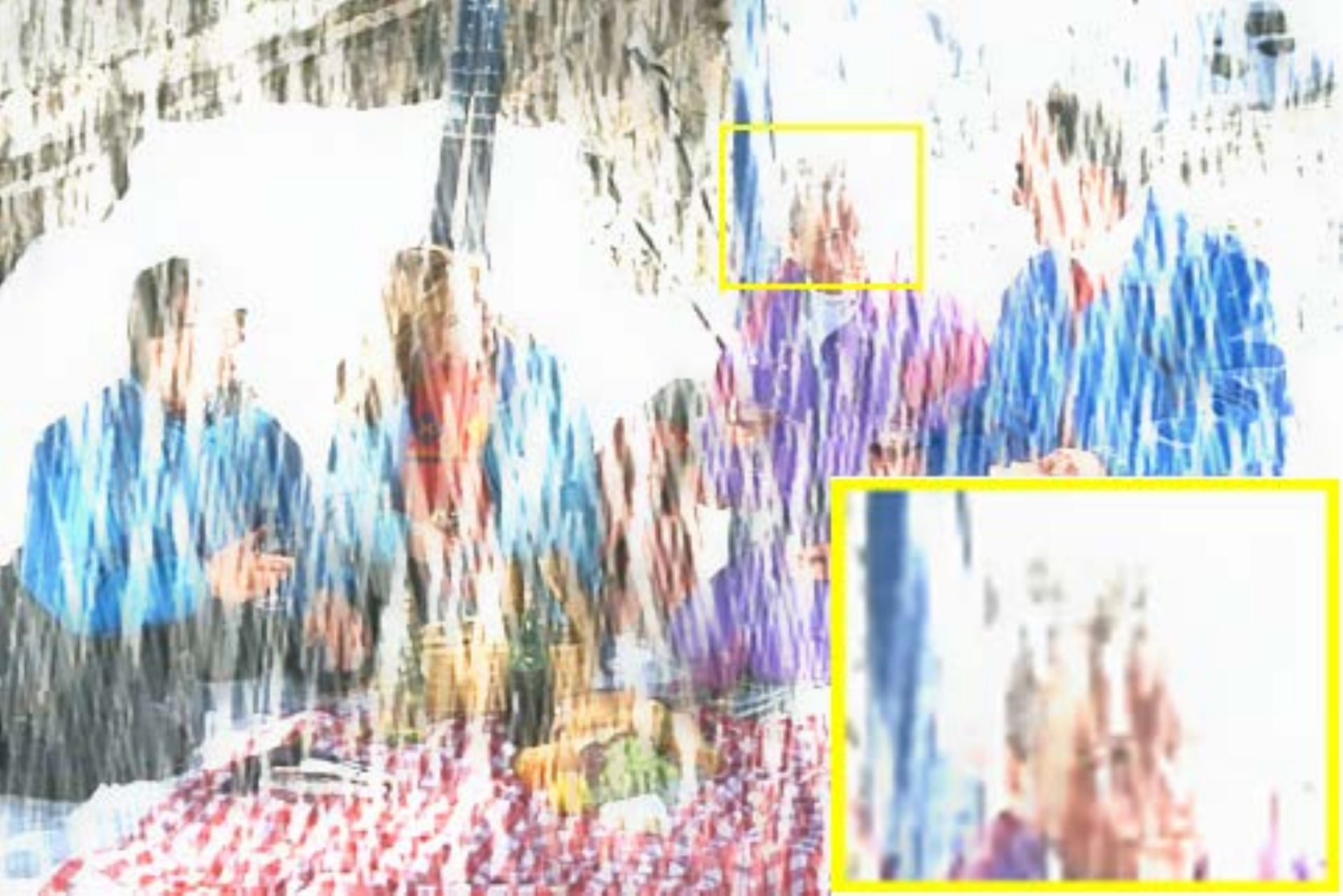}&\hspace{-4mm}
			\includegraphics[width = 0.106\linewidth]{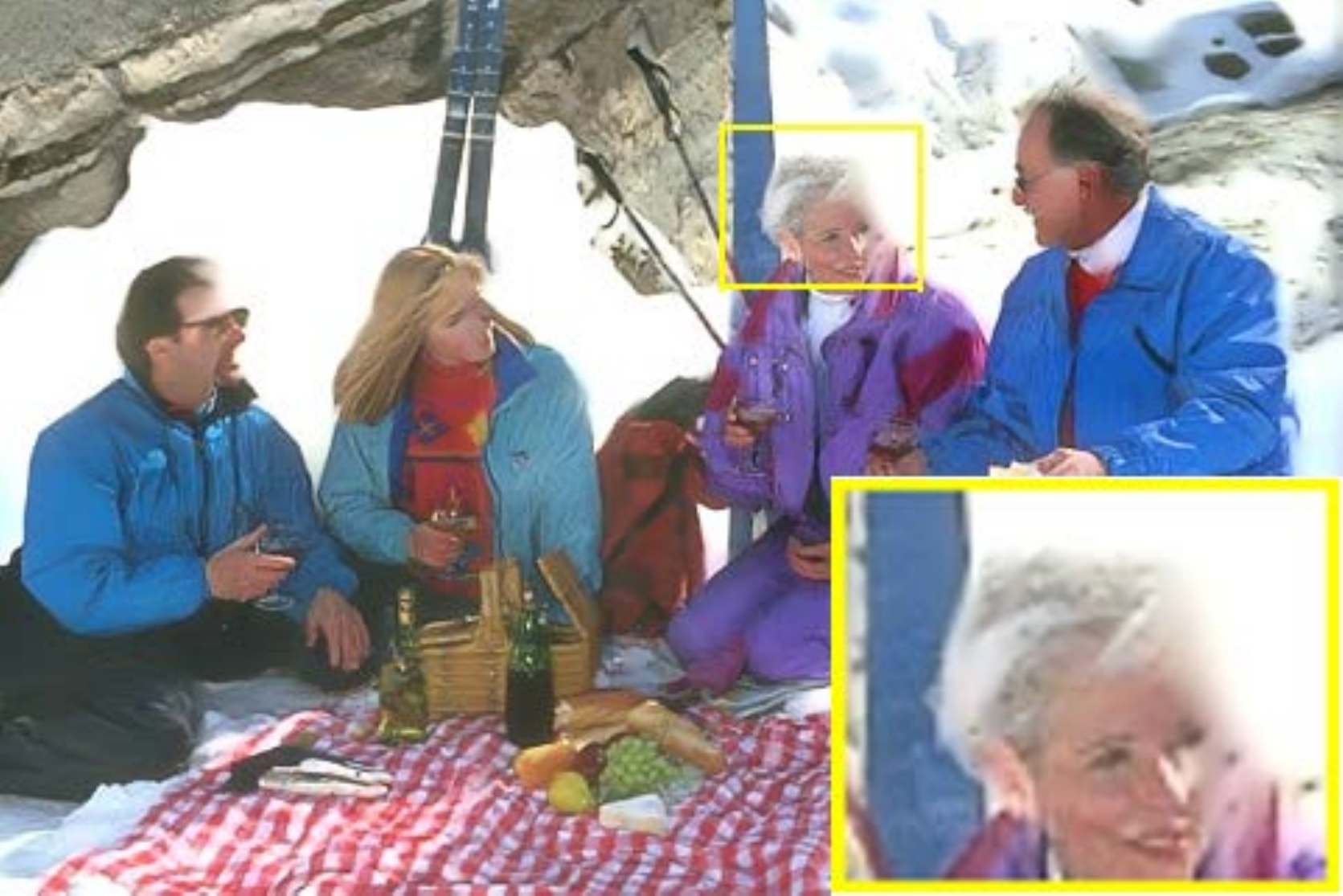}&\hspace{-4mm}
			\includegraphics[width = 0.106\linewidth]{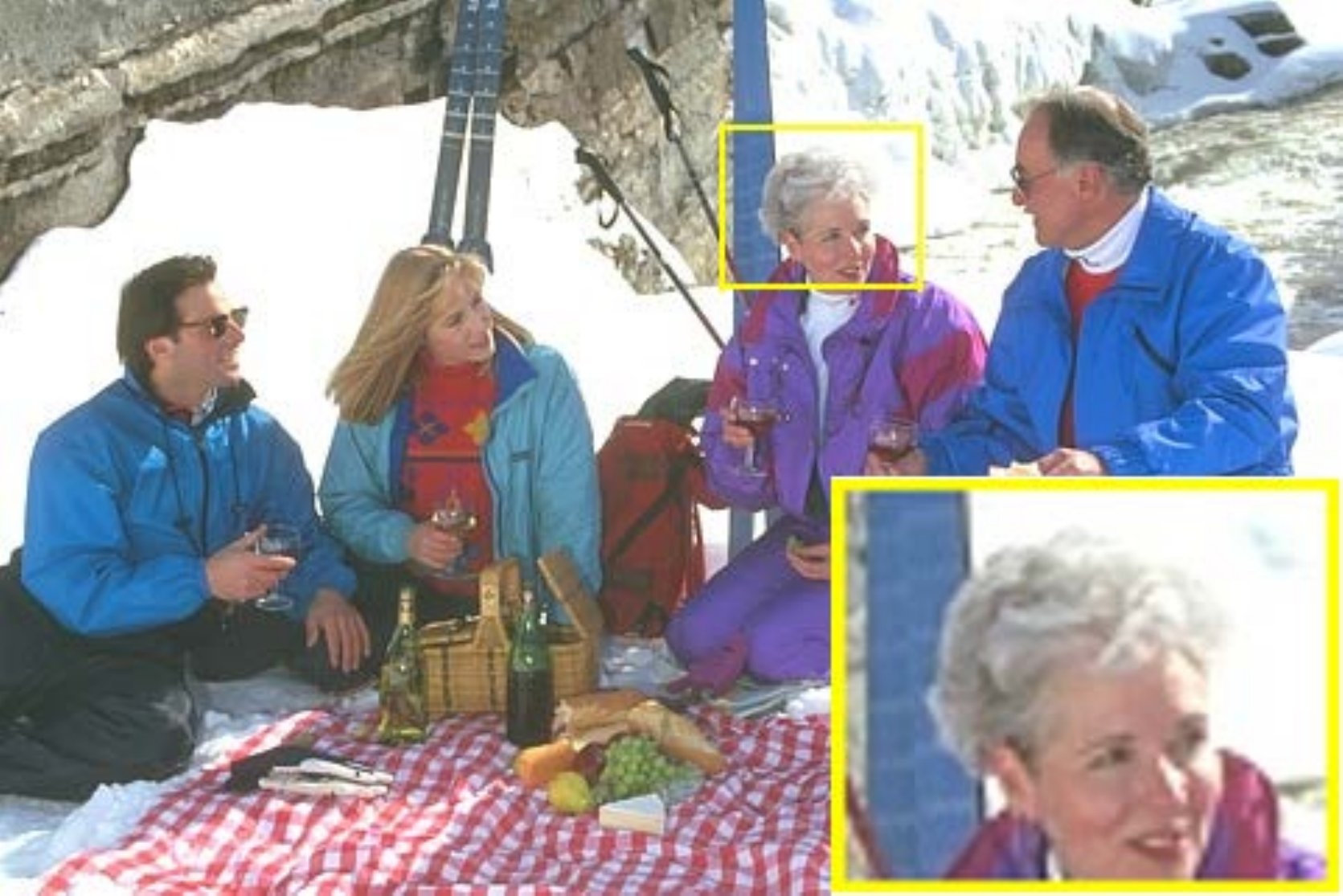}
			\\
			\includegraphics[width = 0.106\linewidth]{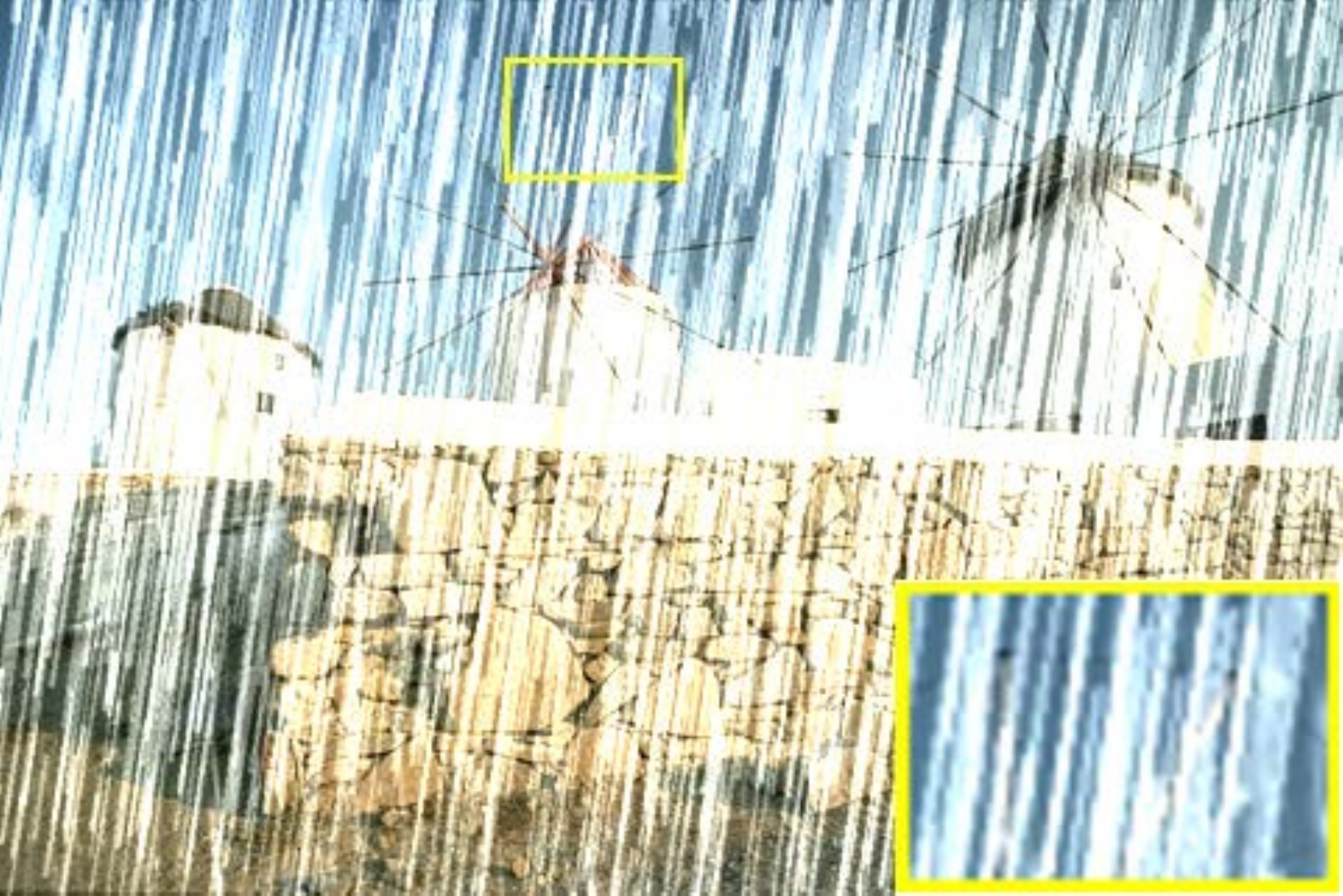}&\hspace{-4mm}
			\includegraphics[width = 0.106\linewidth]{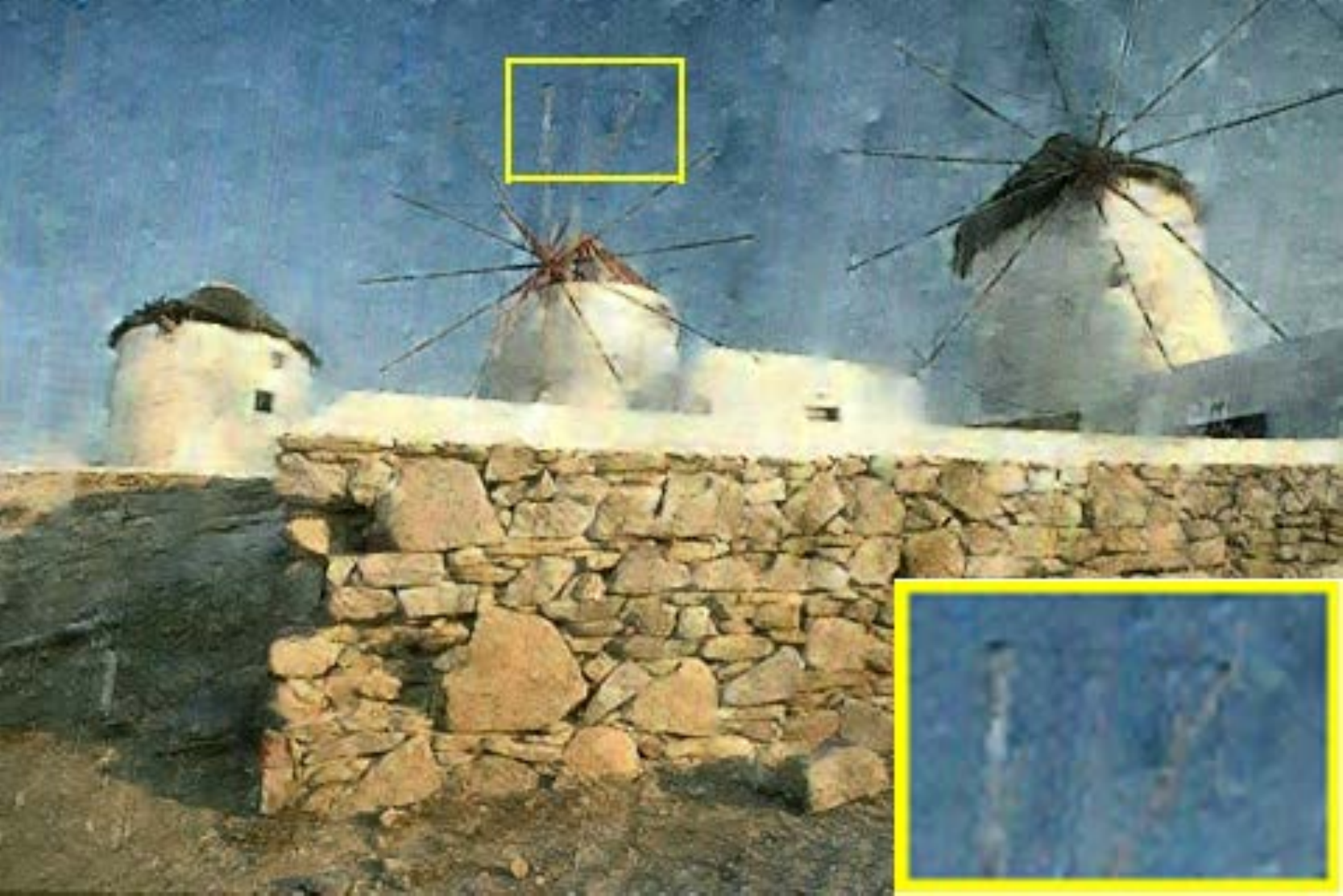}&\hspace{-4mm}
			\includegraphics[width = 0.106\linewidth]{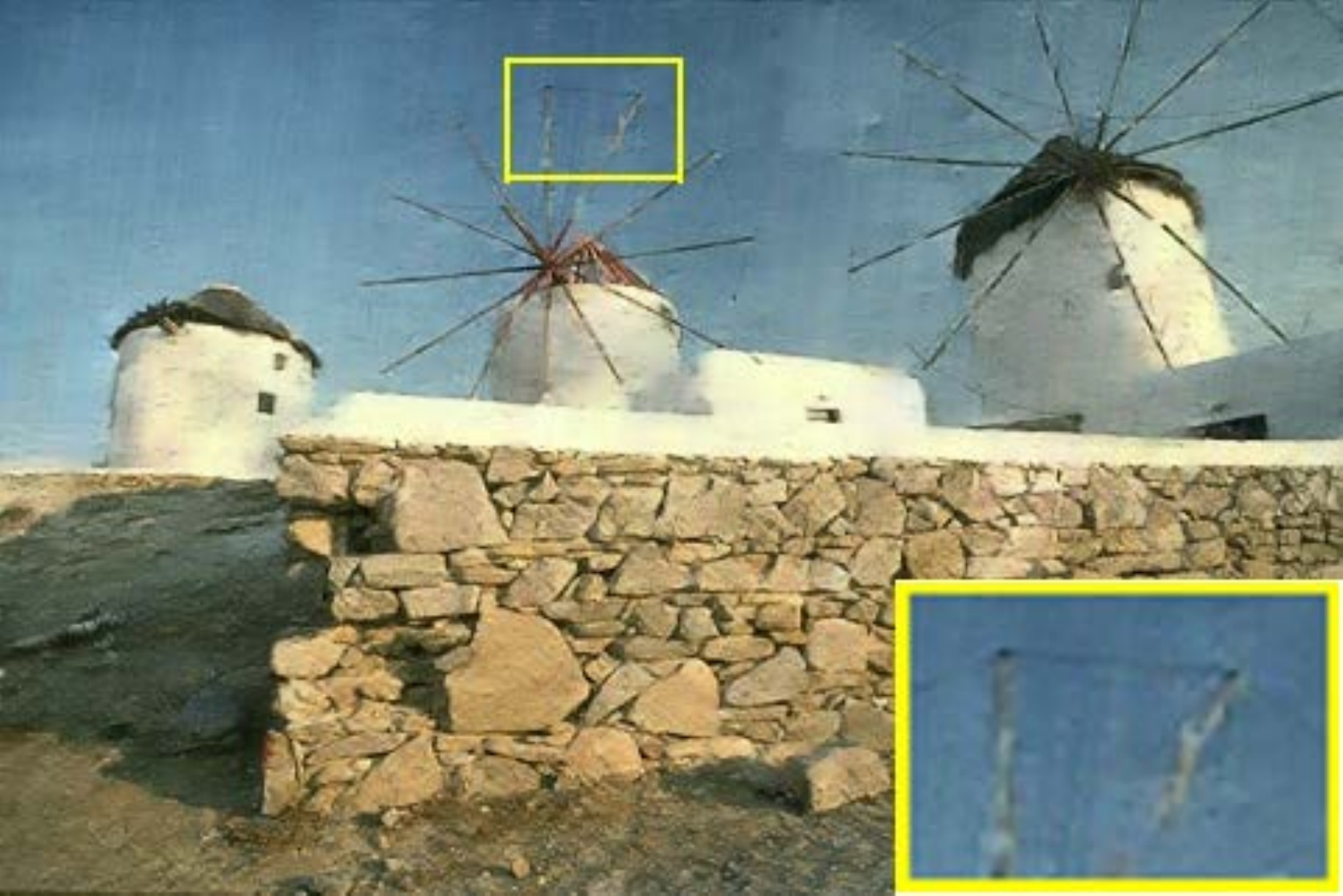}&\hspace{-4mm}
			\includegraphics[width = 0.106\linewidth]{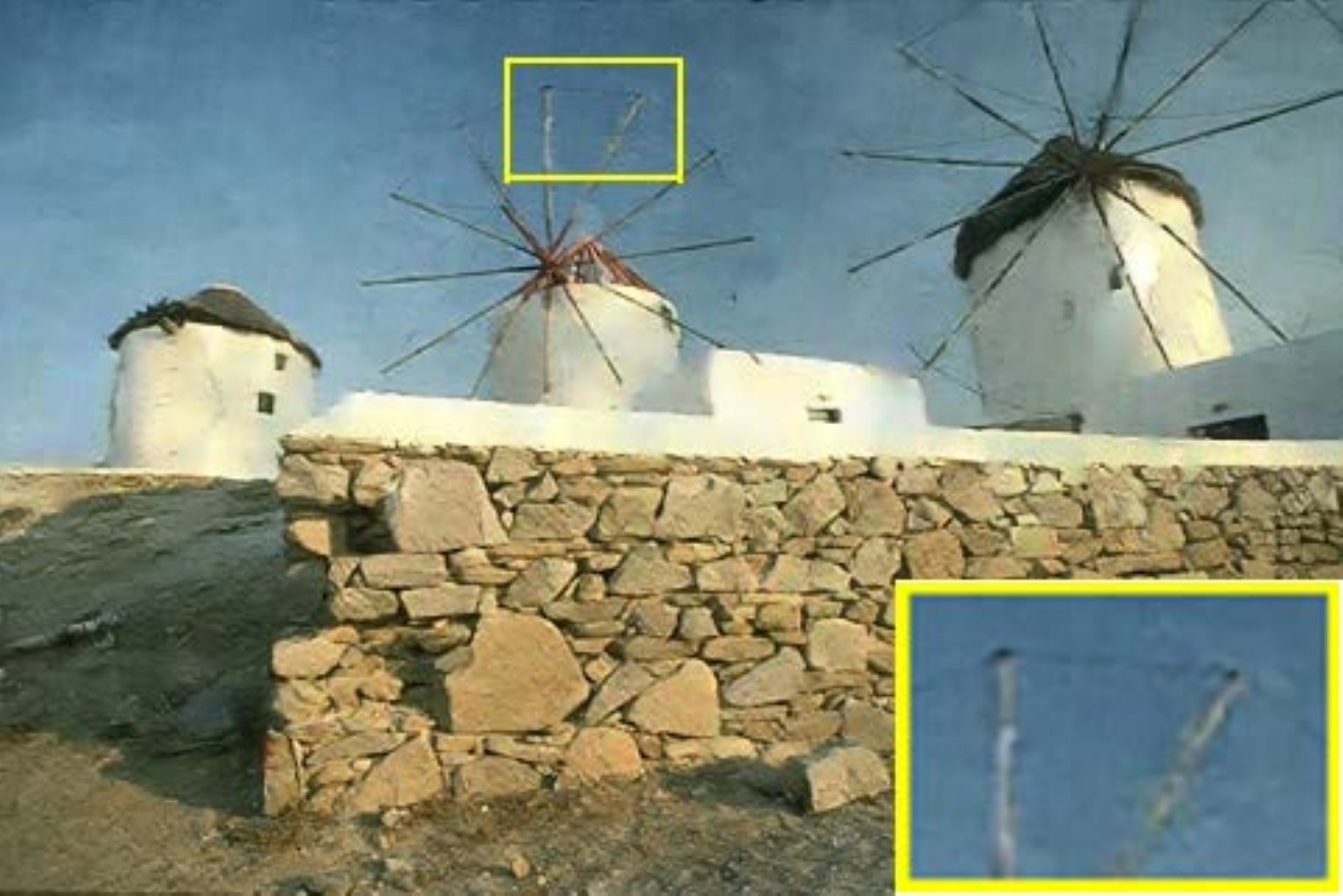}&\hspace{-4mm}
			\includegraphics[width = 0.106\linewidth]{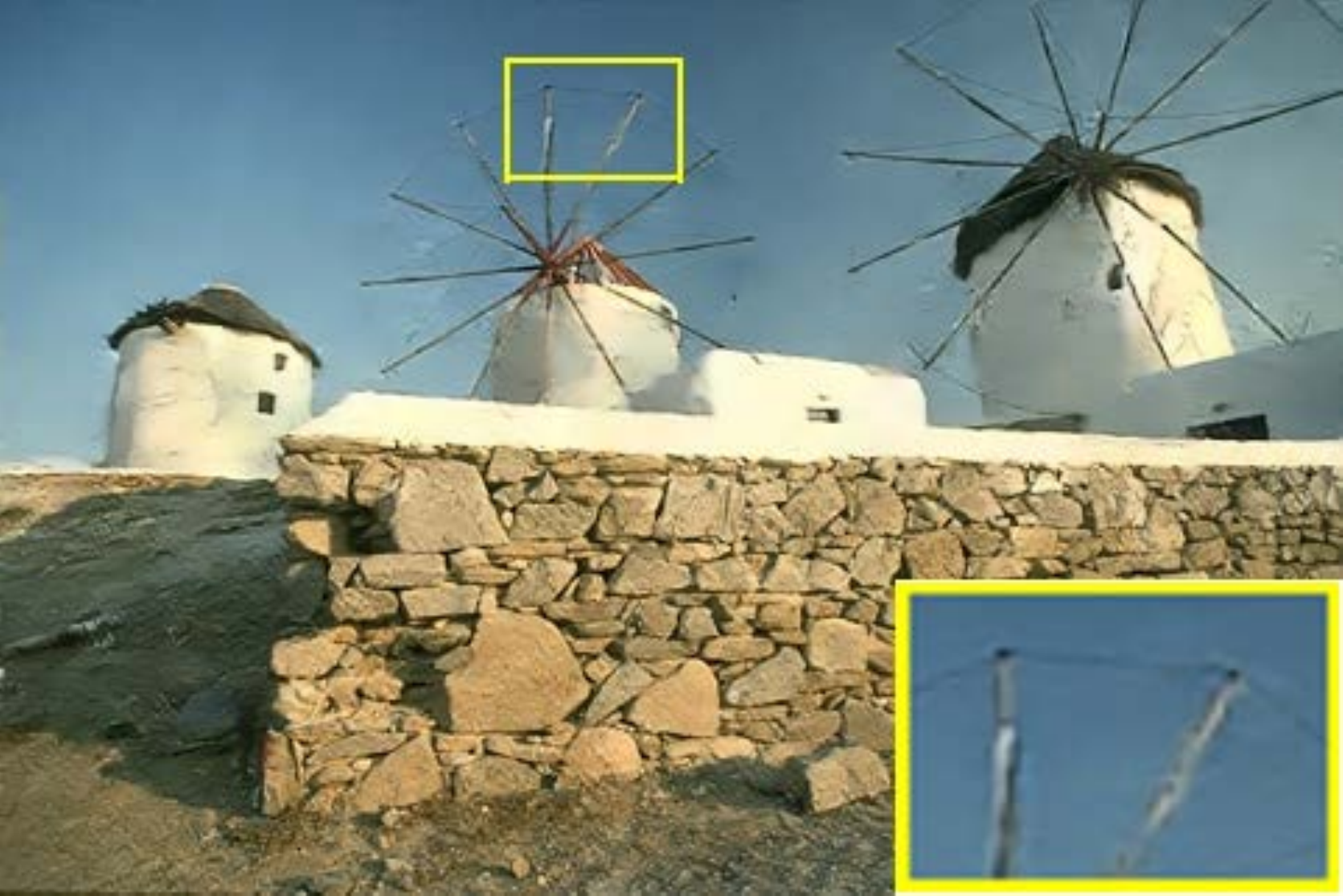}&\hspace{-4mm}
			\includegraphics[width = 0.106\linewidth]{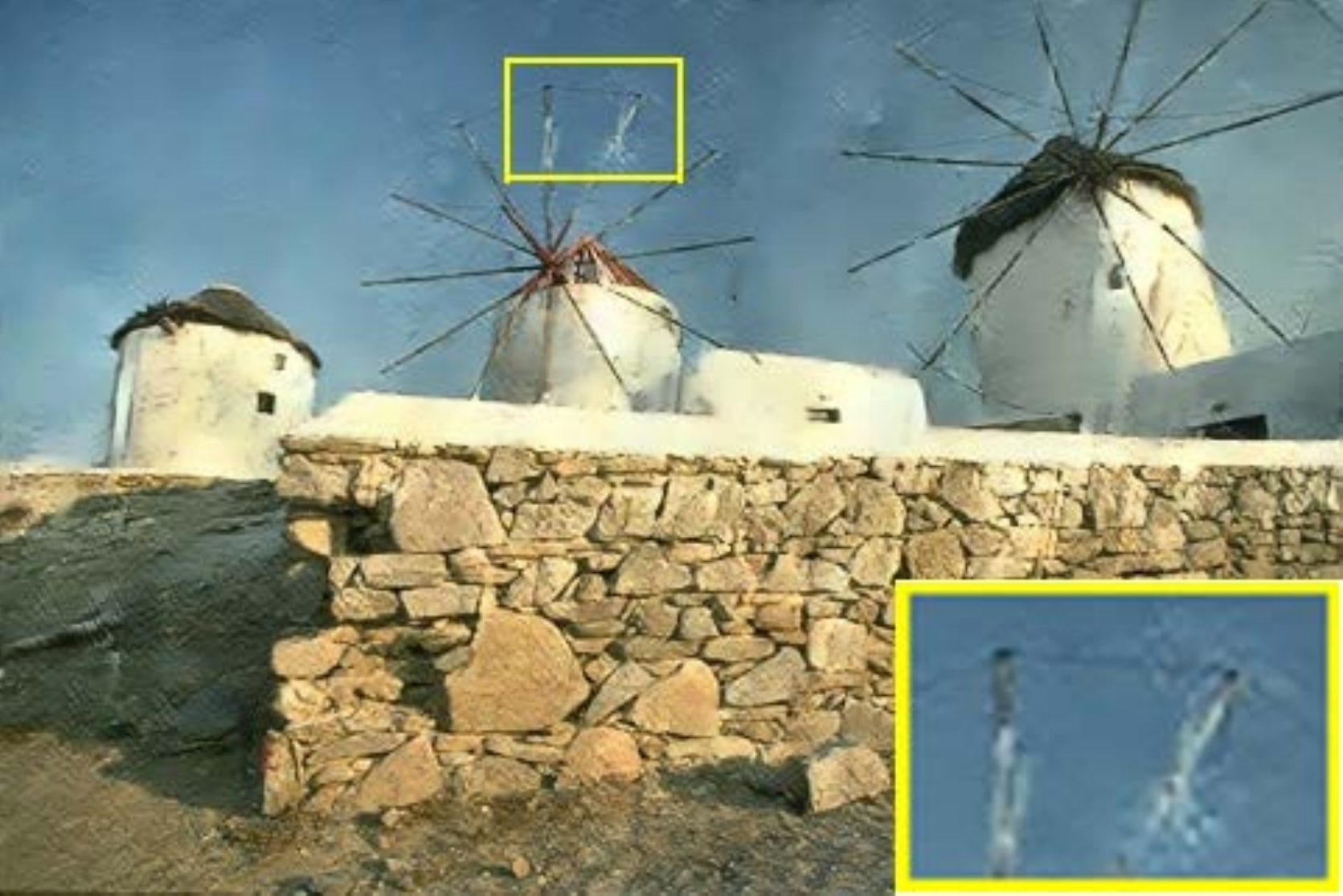}&\hspace{-4mm}
			\includegraphics[width = 0.106\linewidth]{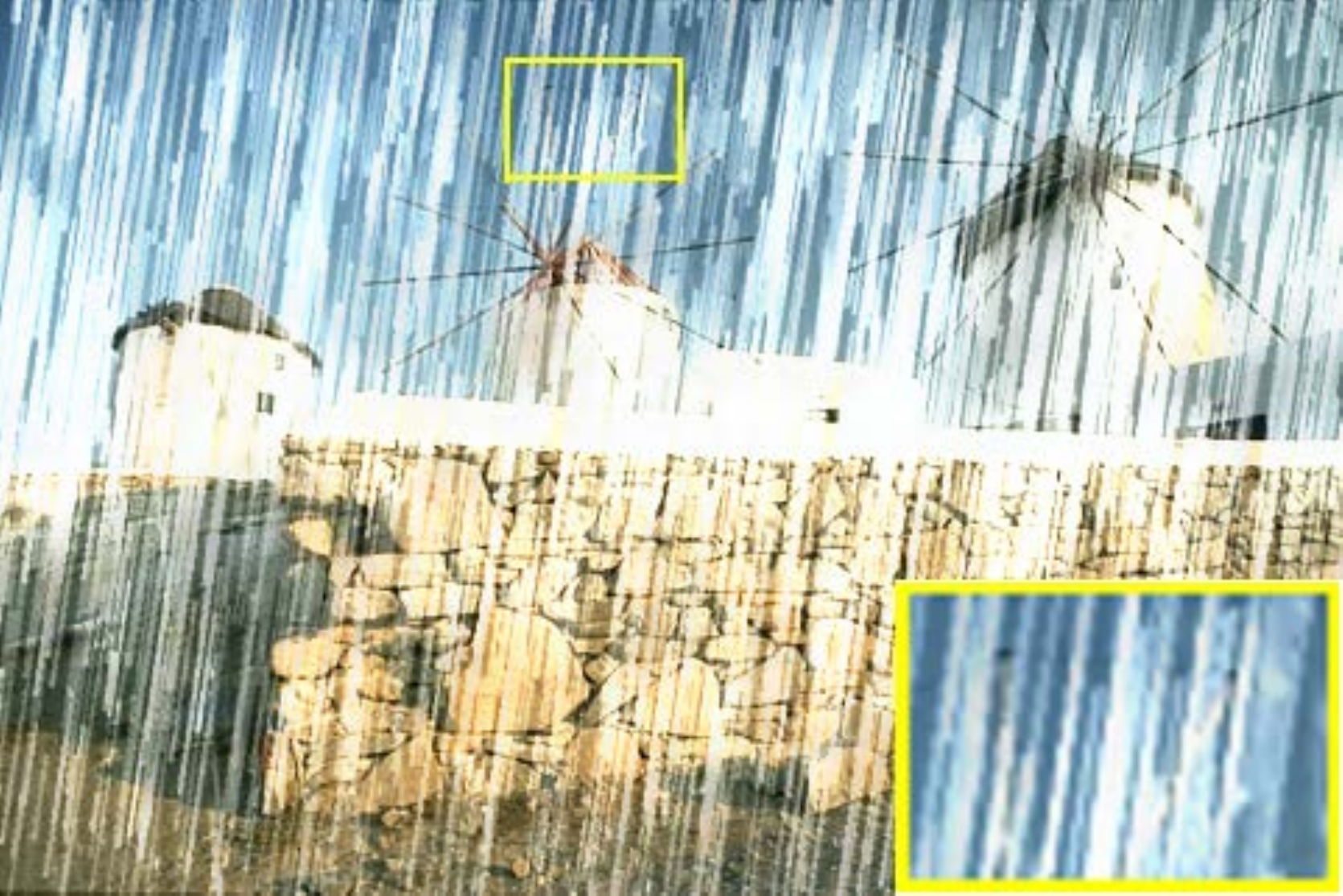}&\hspace{-4mm}
			\includegraphics[width = 0.106\linewidth]{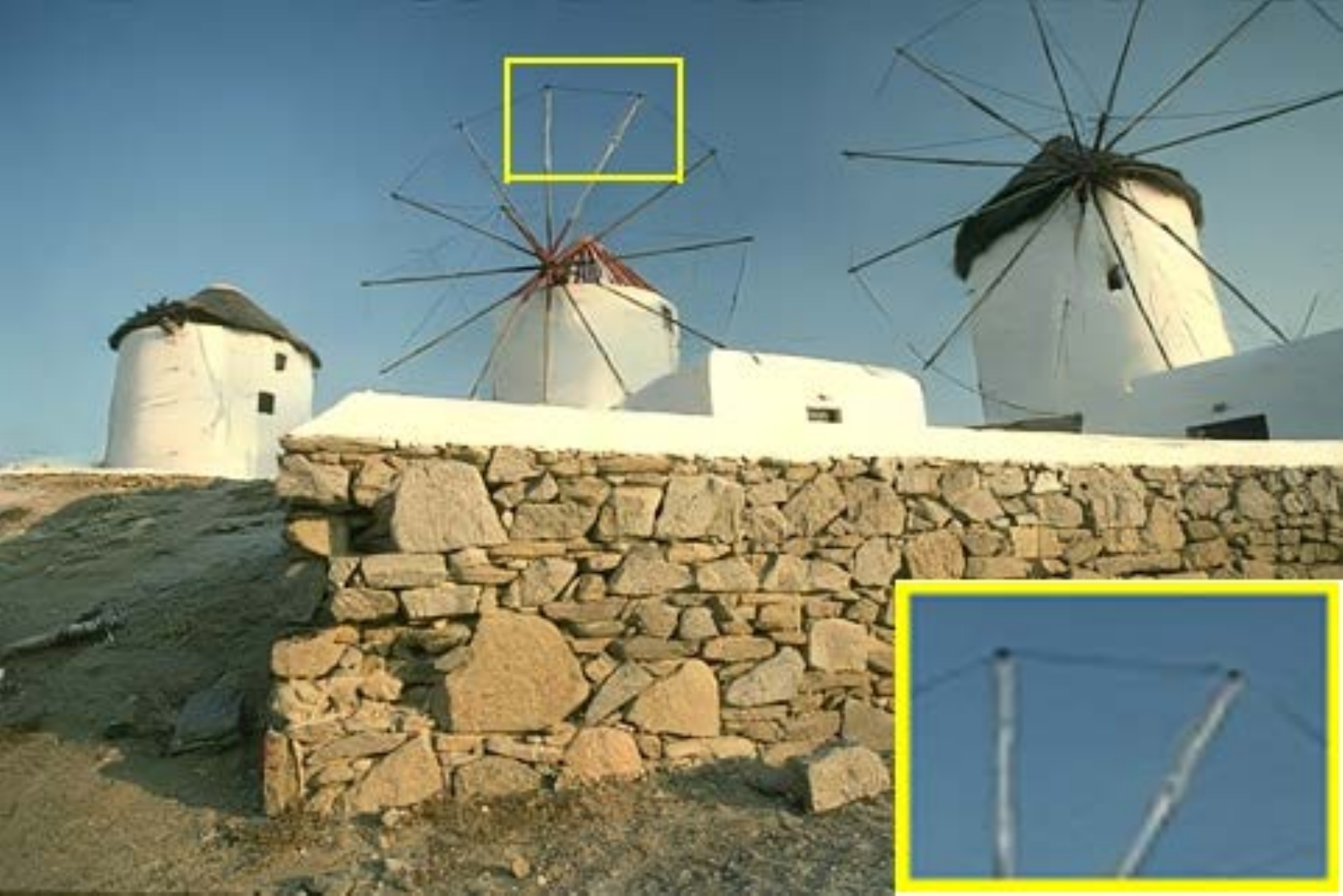}&\hspace{-4mm}
			\includegraphics[width = 0.106\linewidth]{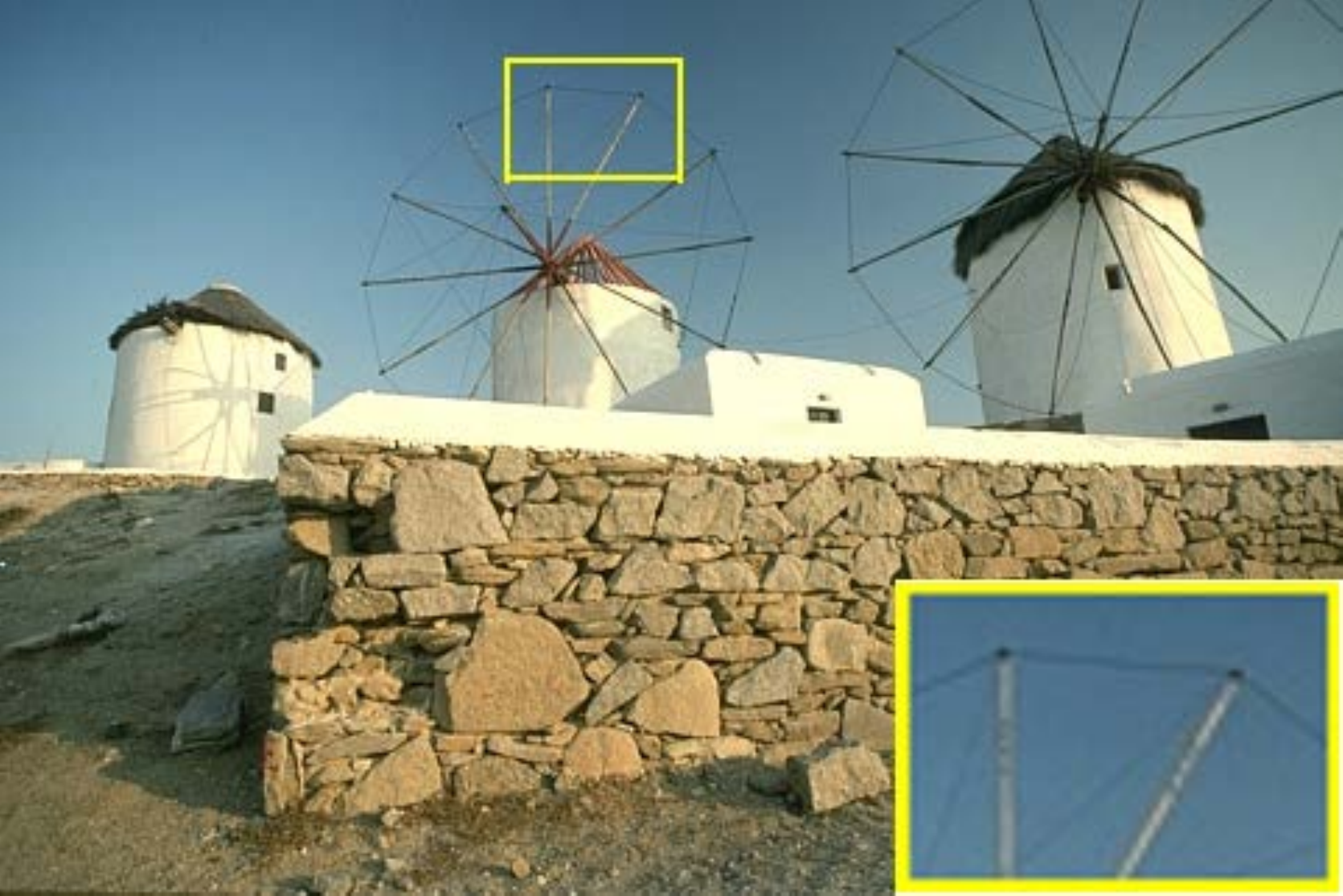}
			\\
			\includegraphics[width = 0.106\linewidth]{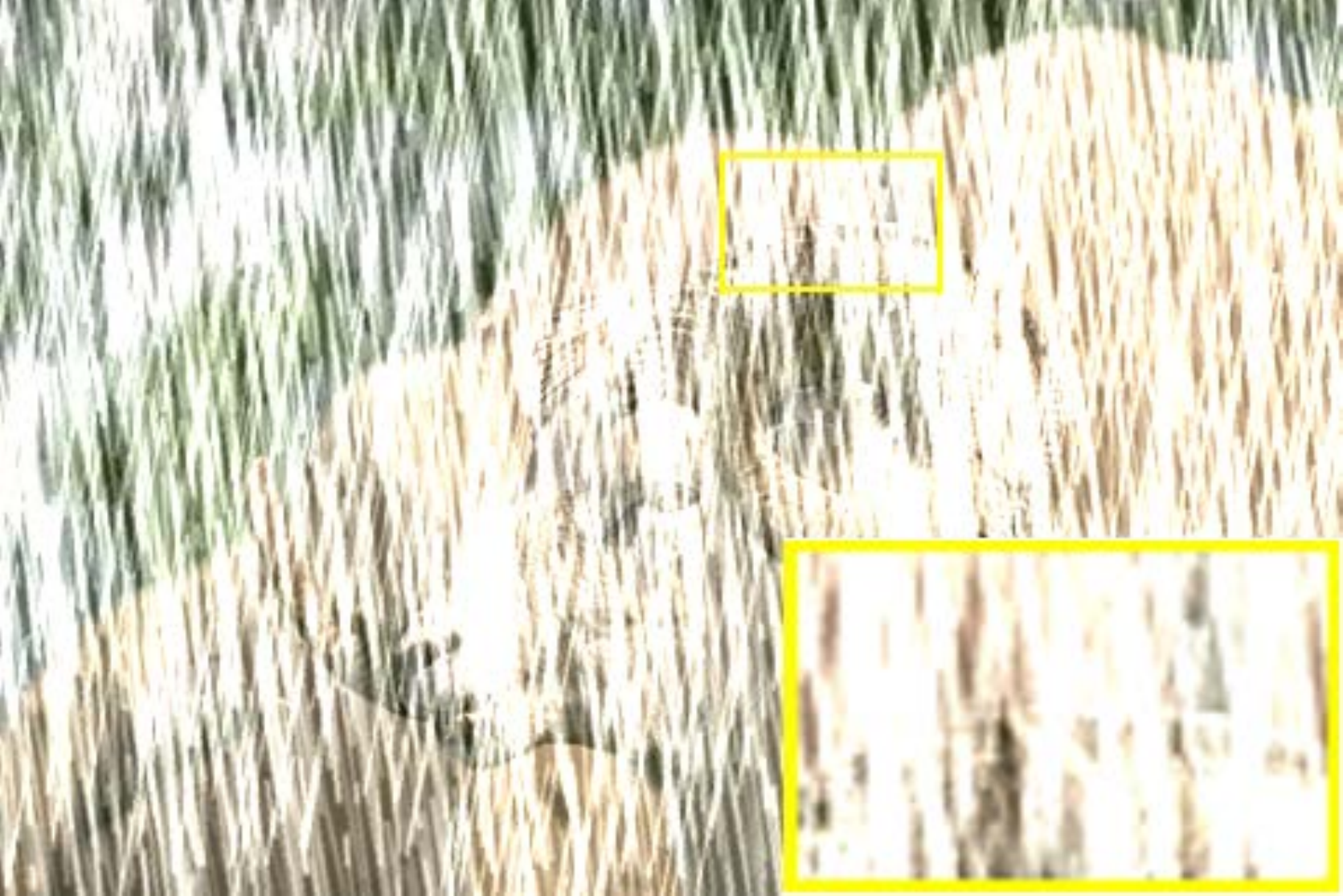}&\hspace{-4mm}
			\includegraphics[width = 0.106\linewidth]{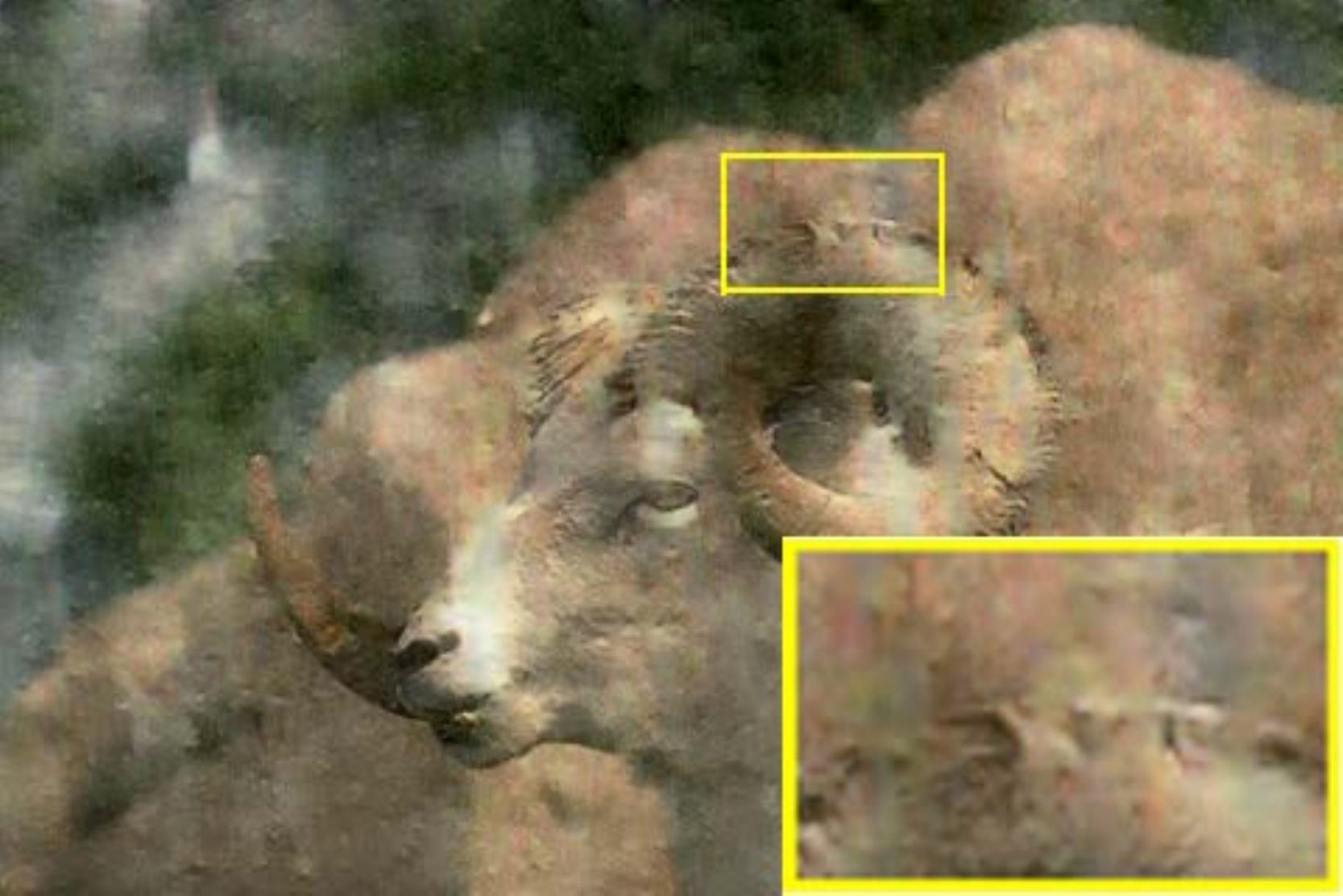}&\hspace{-4mm}
			\includegraphics[width = 0.106\linewidth]{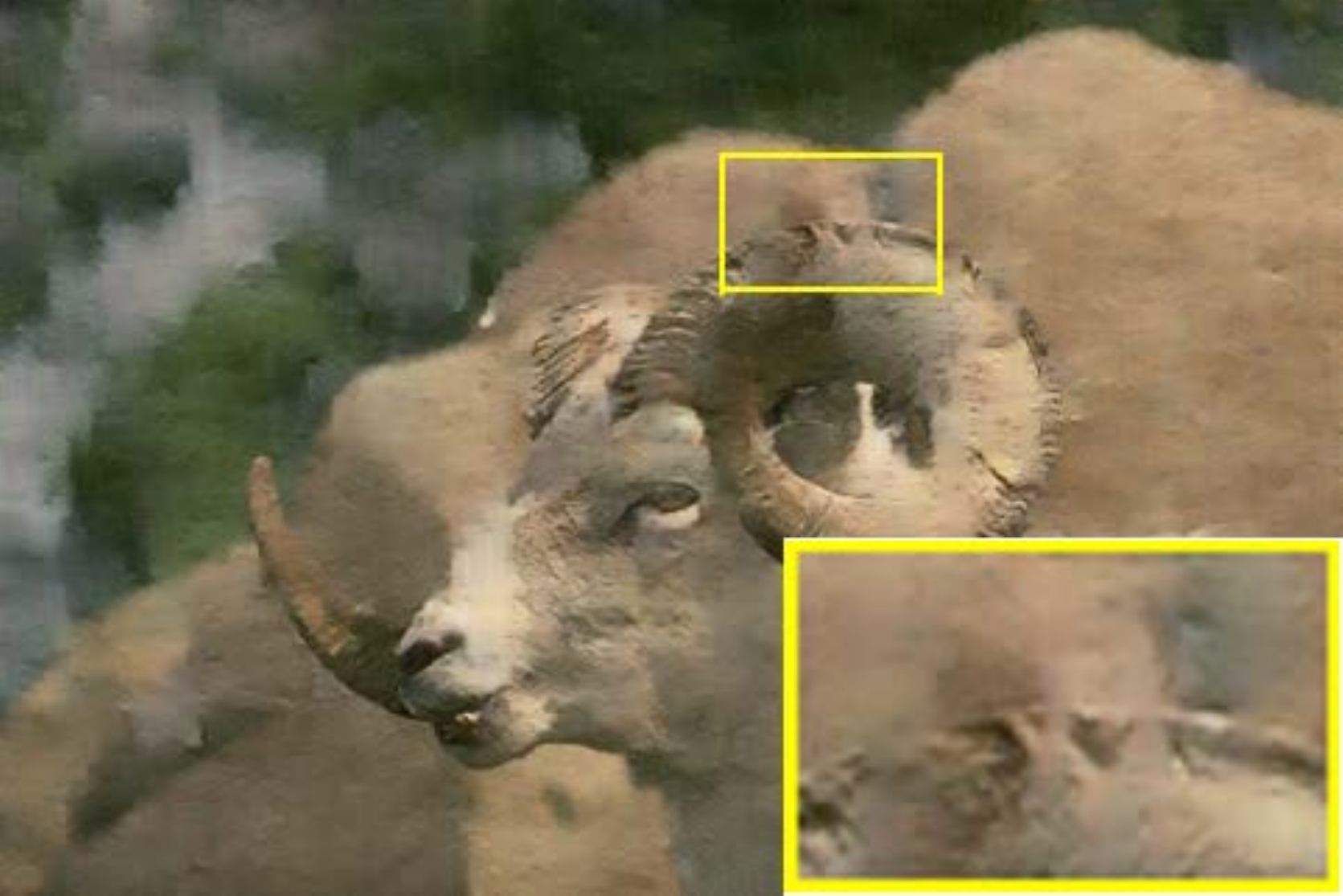}&\hspace{-4mm}
			\includegraphics[width = 0.106\linewidth]{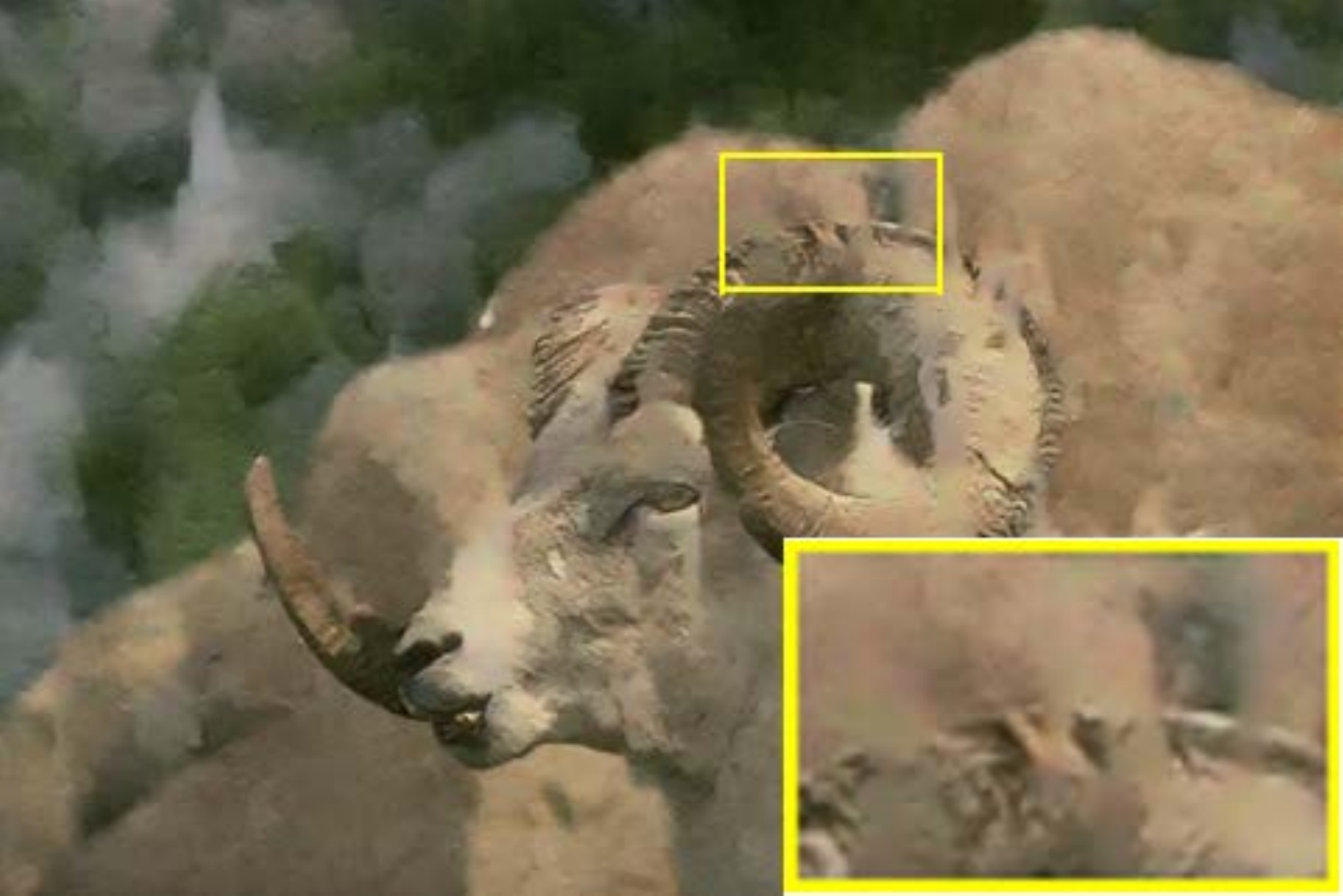}&\hspace{-4mm}
			\includegraphics[width = 0.106\linewidth]{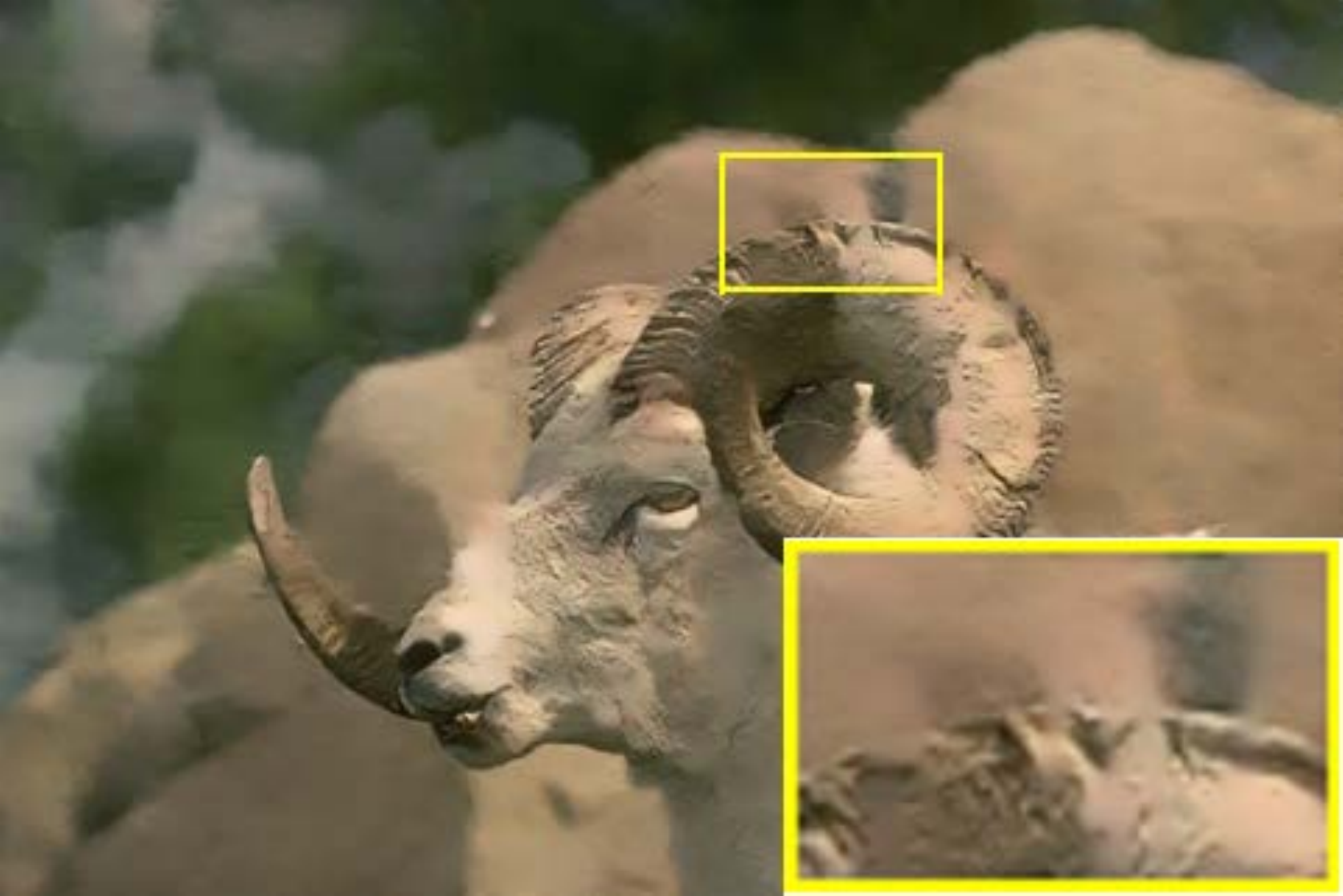}&\hspace{-4mm}
			\includegraphics[width = 0.106\linewidth]{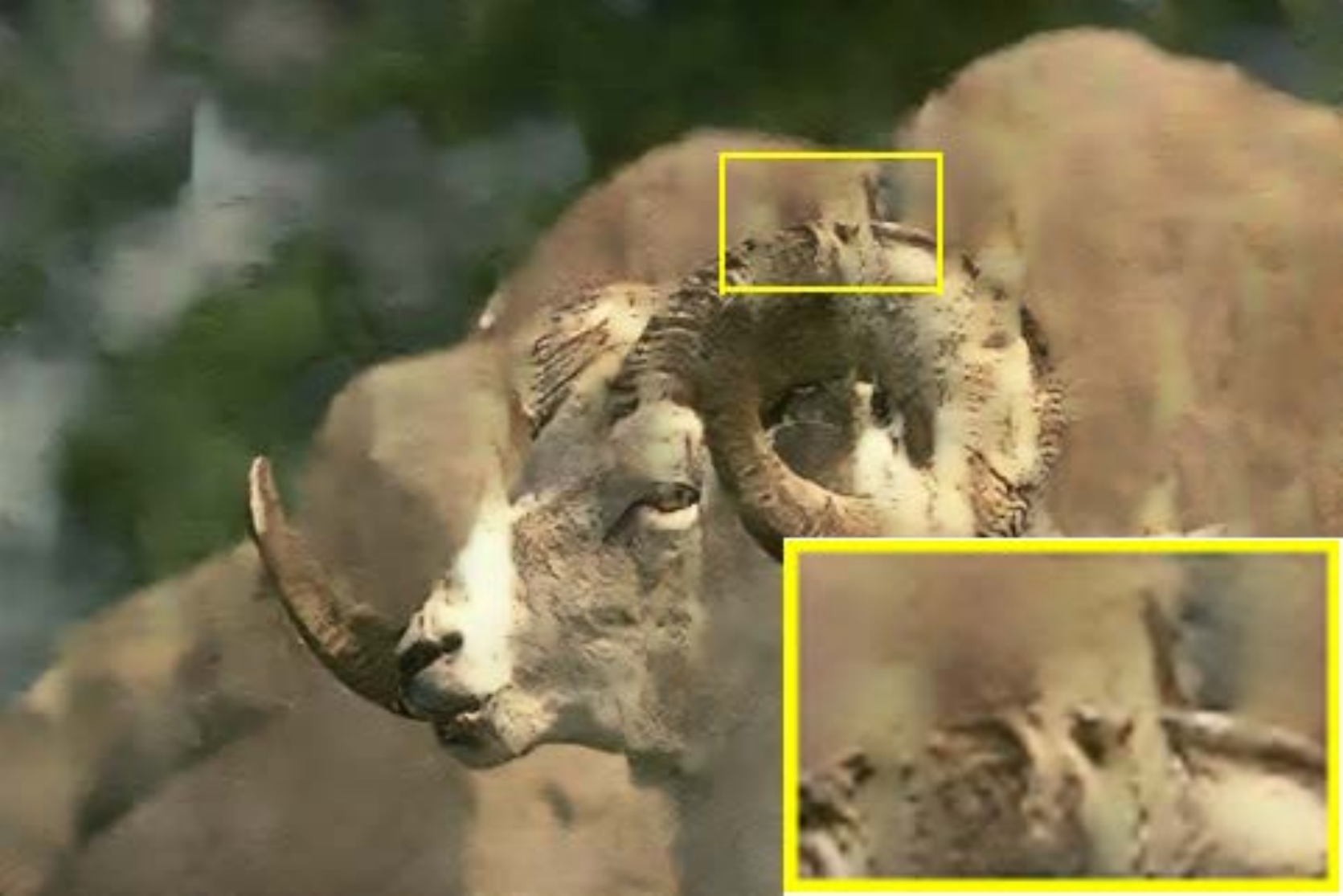}&\hspace{-4mm}
			\includegraphics[width = 0.106\linewidth]{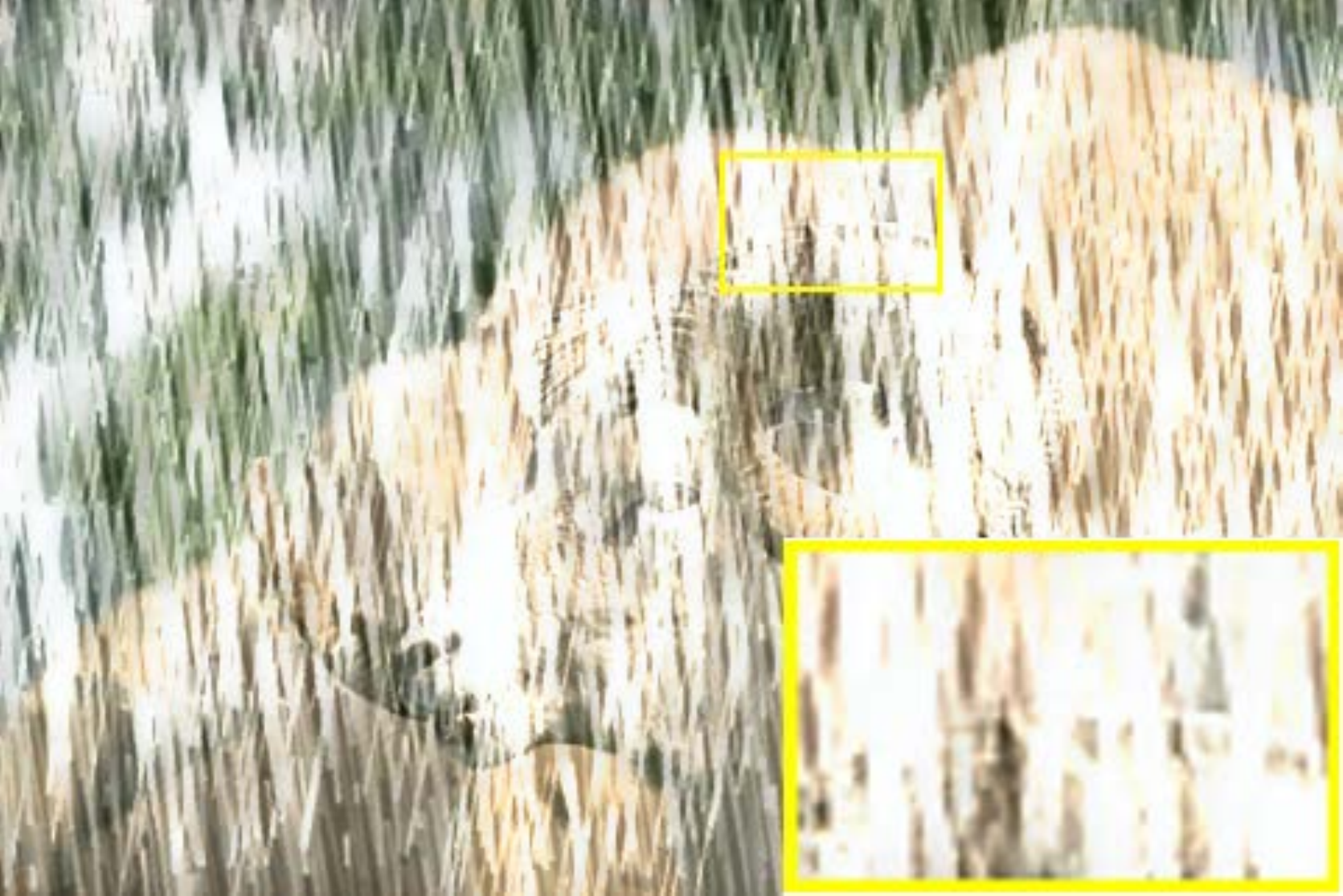}&\hspace{-4mm}
			\includegraphics[width = 0.106\linewidth]{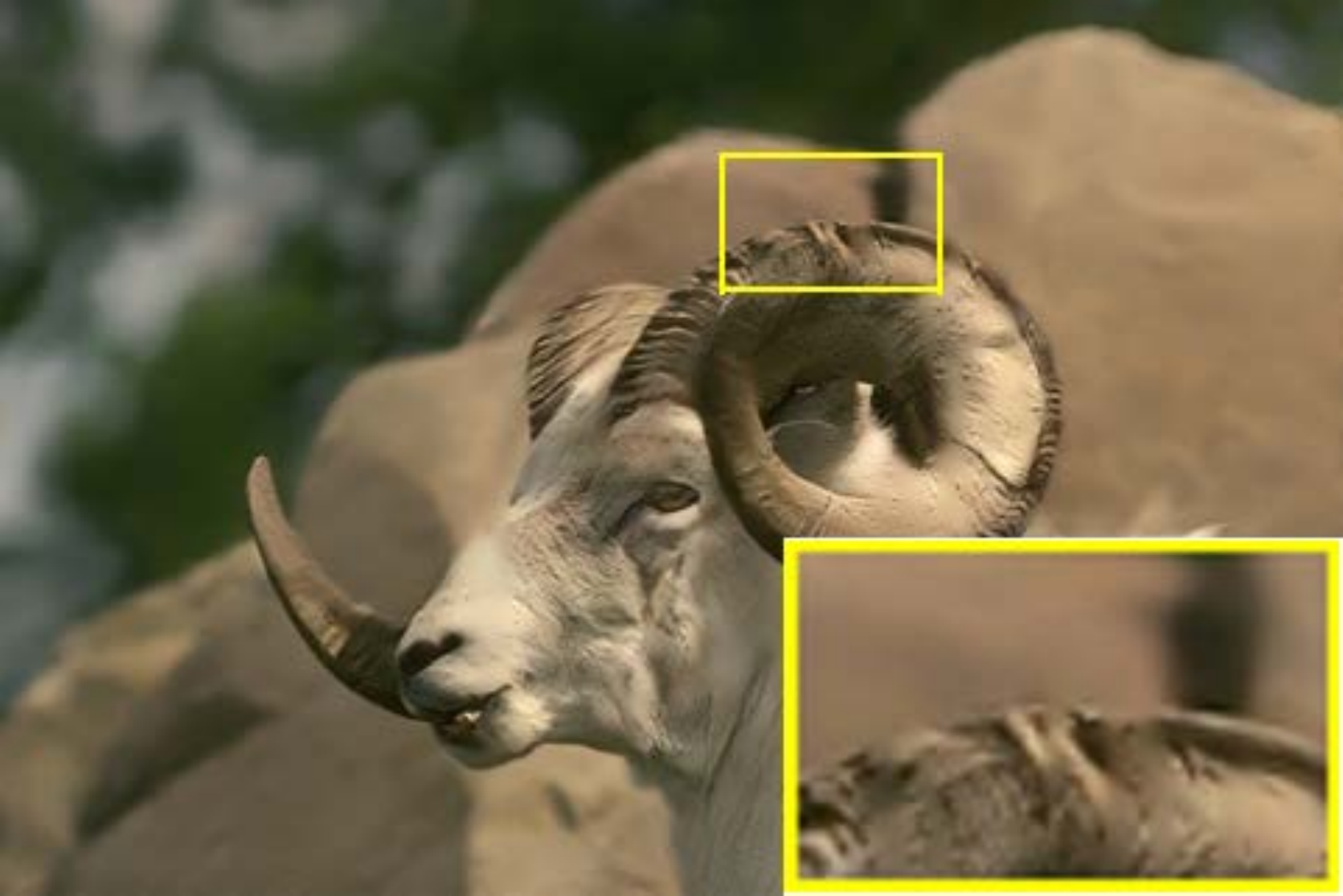}&\hspace{-4mm}
			\includegraphics[width = 0.106\linewidth]{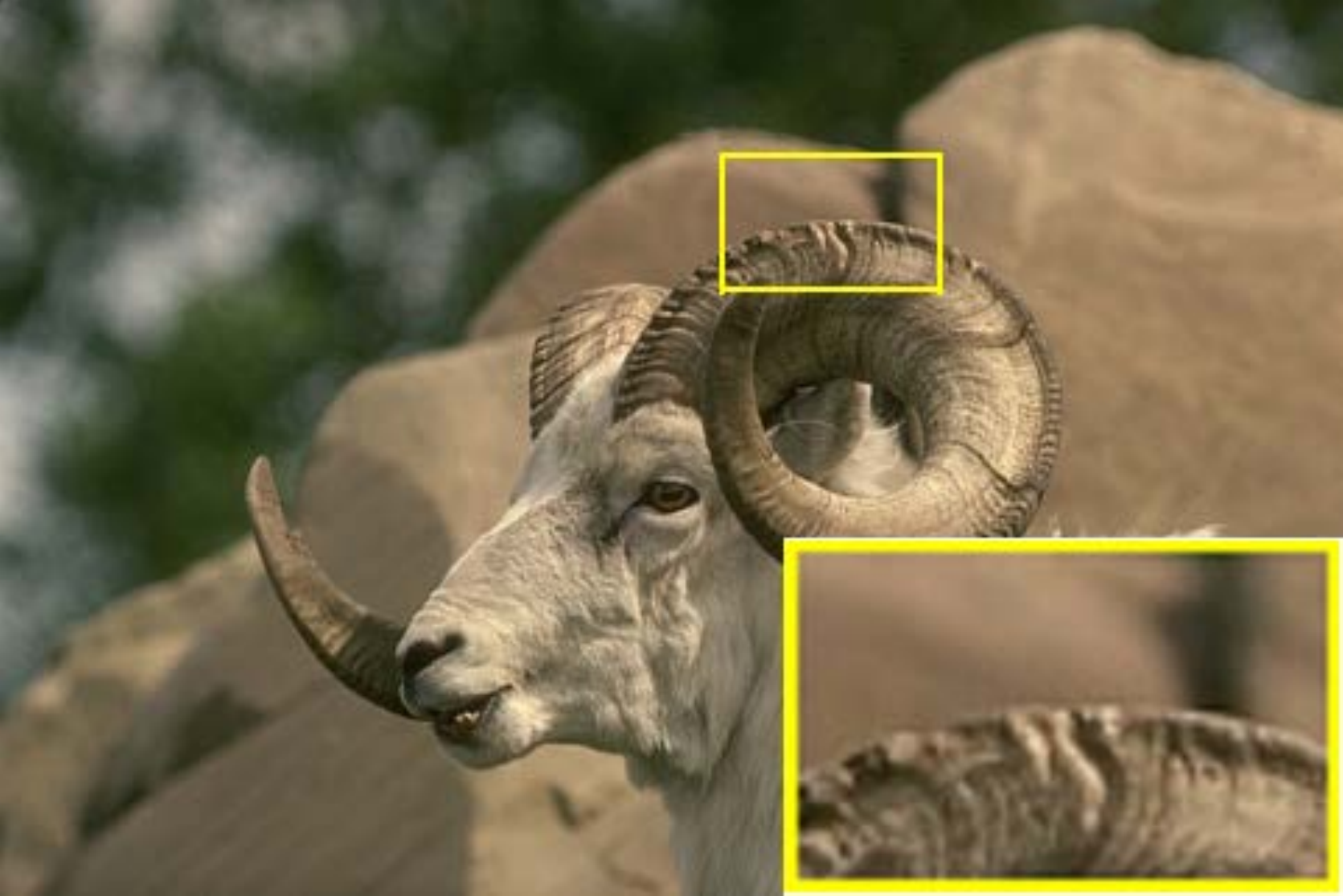}\\
			(a) Input&\hspace{-4mm} (b) DDN&\hspace{-4mm} (c) RESCAN&\hspace{-4mm} (d) REHEN&\hspace{-4mm} (e) PreNet&\hspace{-4mm} (f) SpaNet&\hspace{-4mm} (g) SSIR&\hspace{-4mm} (h) JDNet&\hspace{-4mm} (i) GT\\
		\end{tabular}
	\end{center}
	\caption{Examples about the comparison of our method with other methods on Rain100H dataset.}
	\label{fig:syn-compared}
\end{figure*}

\begin{figure*}[t]
	\begin{center}
		\begin{tabular}{ccccccccc}
			\includegraphics[width = 0.12\linewidth]{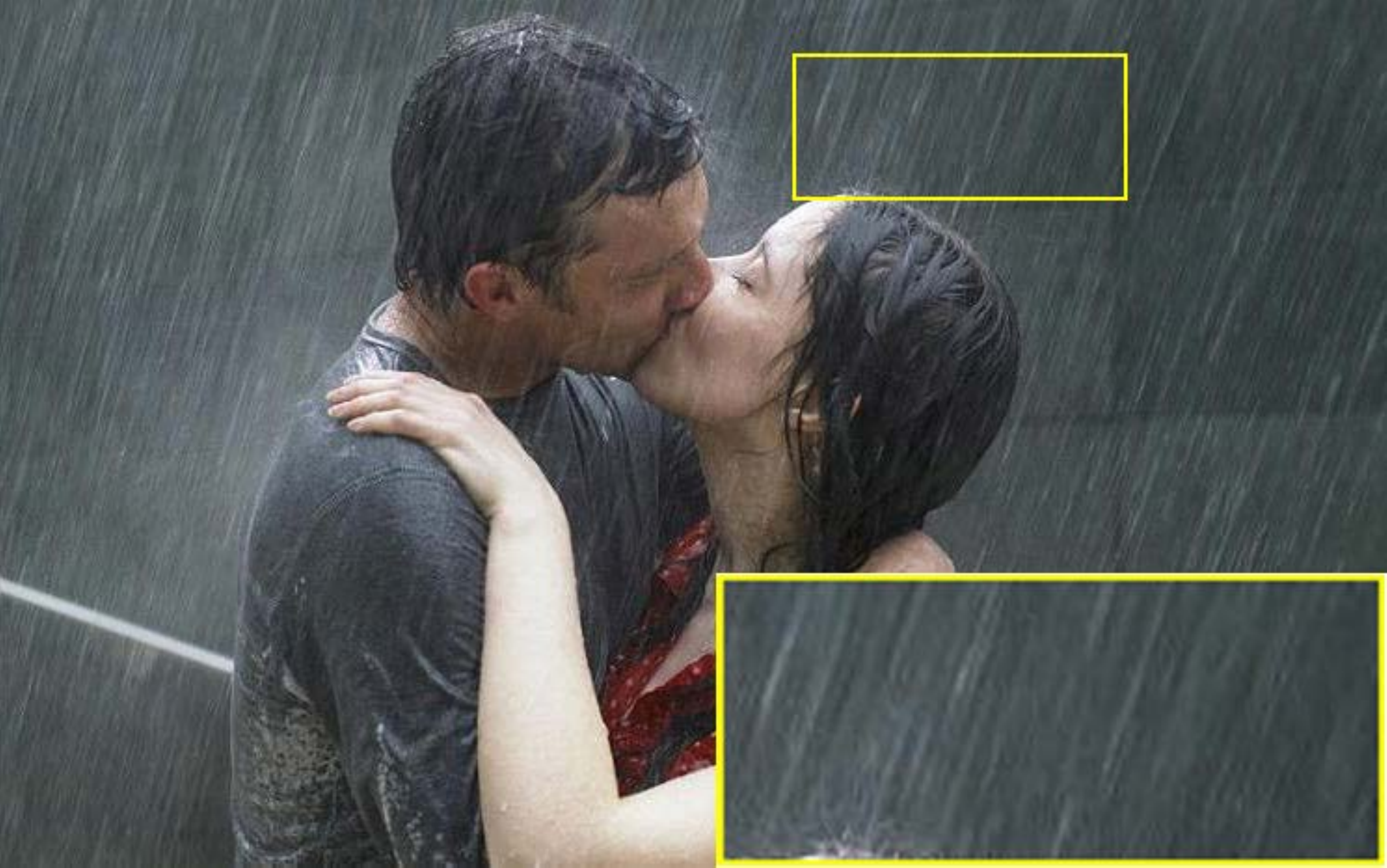}&\hspace{-4mm}
			\includegraphics[width = 0.12\linewidth]{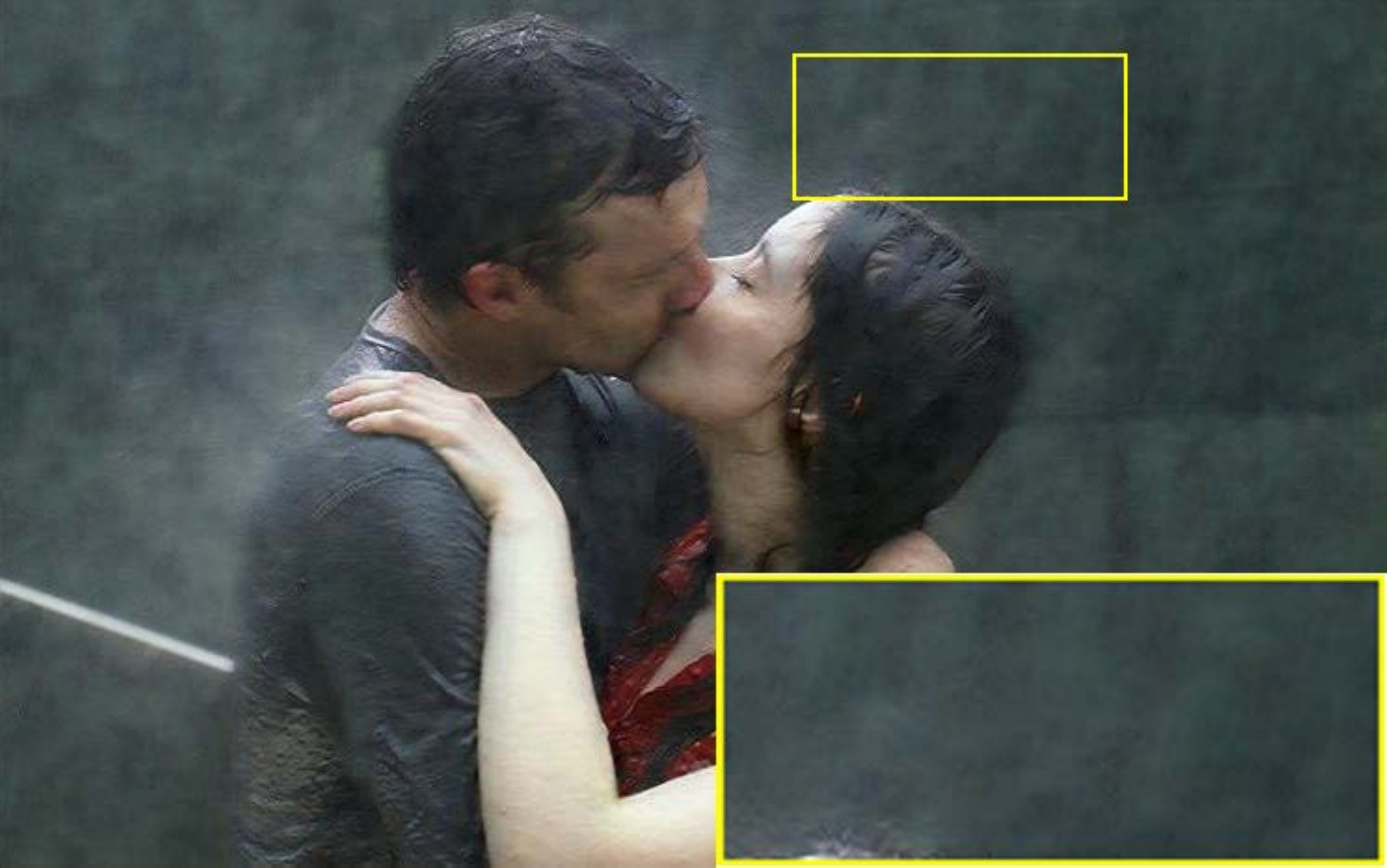}&\hspace{-4mm}
			\includegraphics[width = 0.12\linewidth]{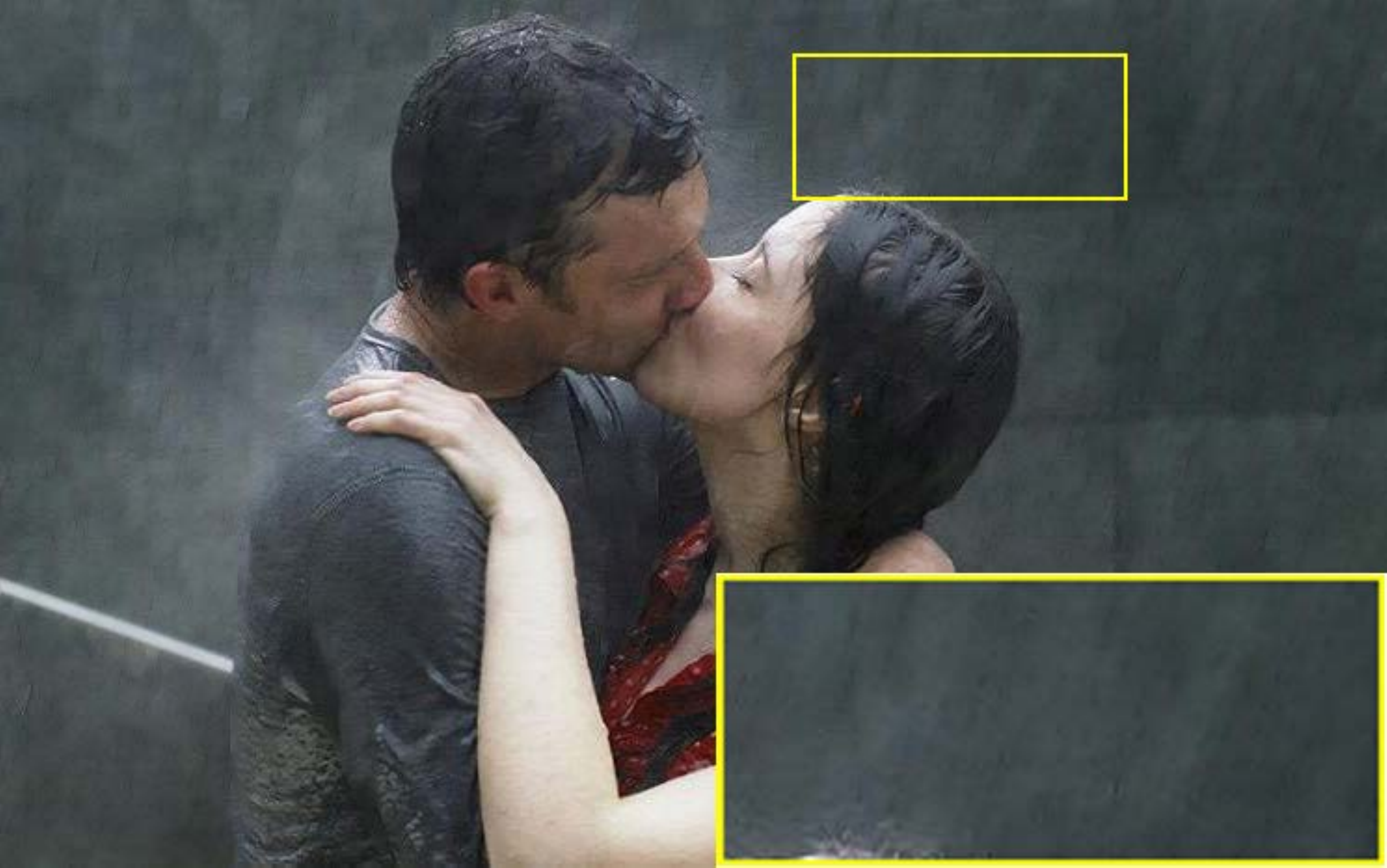}&\hspace{-4mm}
			\includegraphics[width = 0.12\linewidth]{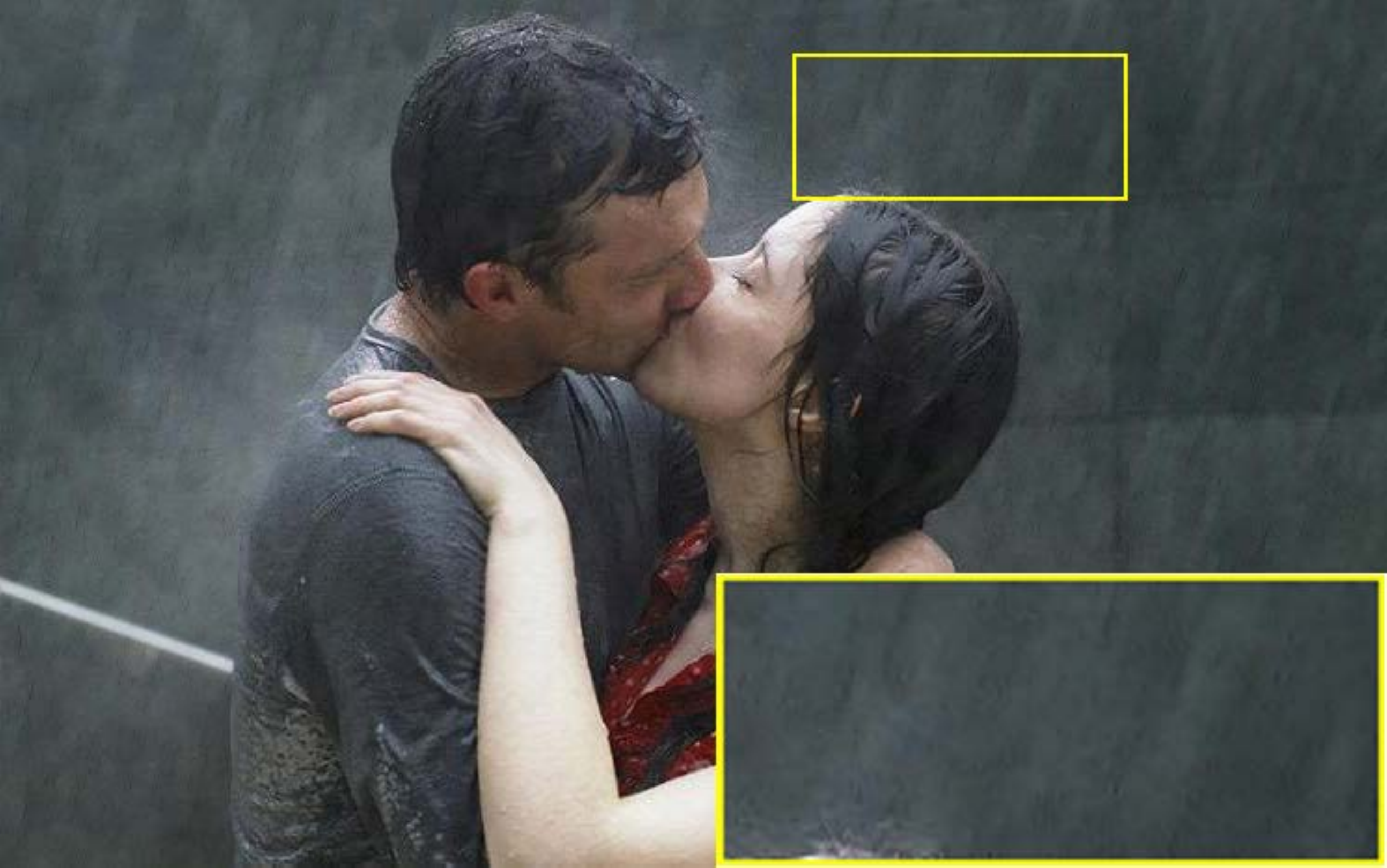}&\hspace{-4mm}
			\includegraphics[width = 0.12\linewidth]{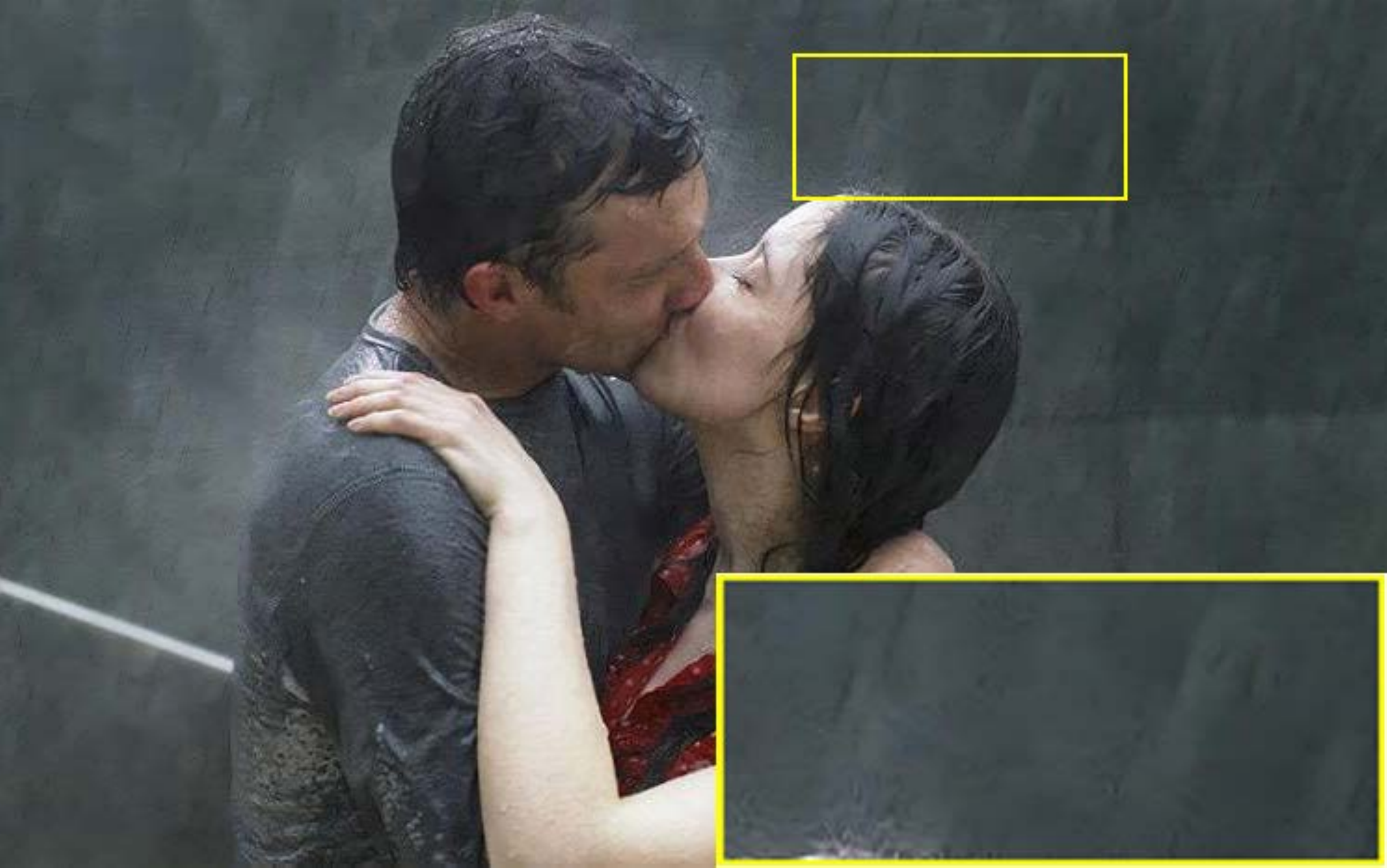}&\hspace{-4mm}
			\includegraphics[width = 0.12\linewidth]{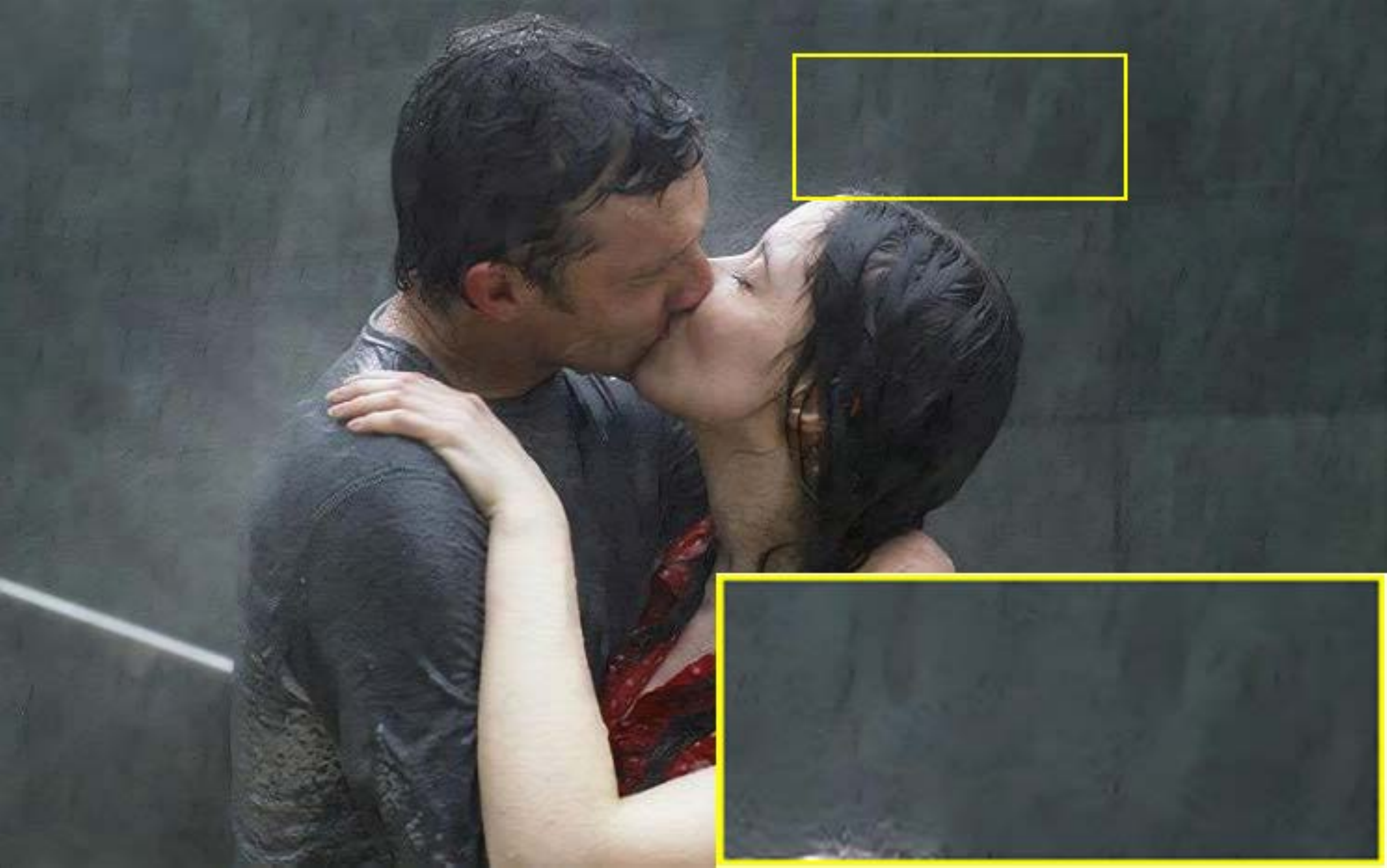}&\hspace{-4mm}
			\includegraphics[width = 0.12\linewidth]{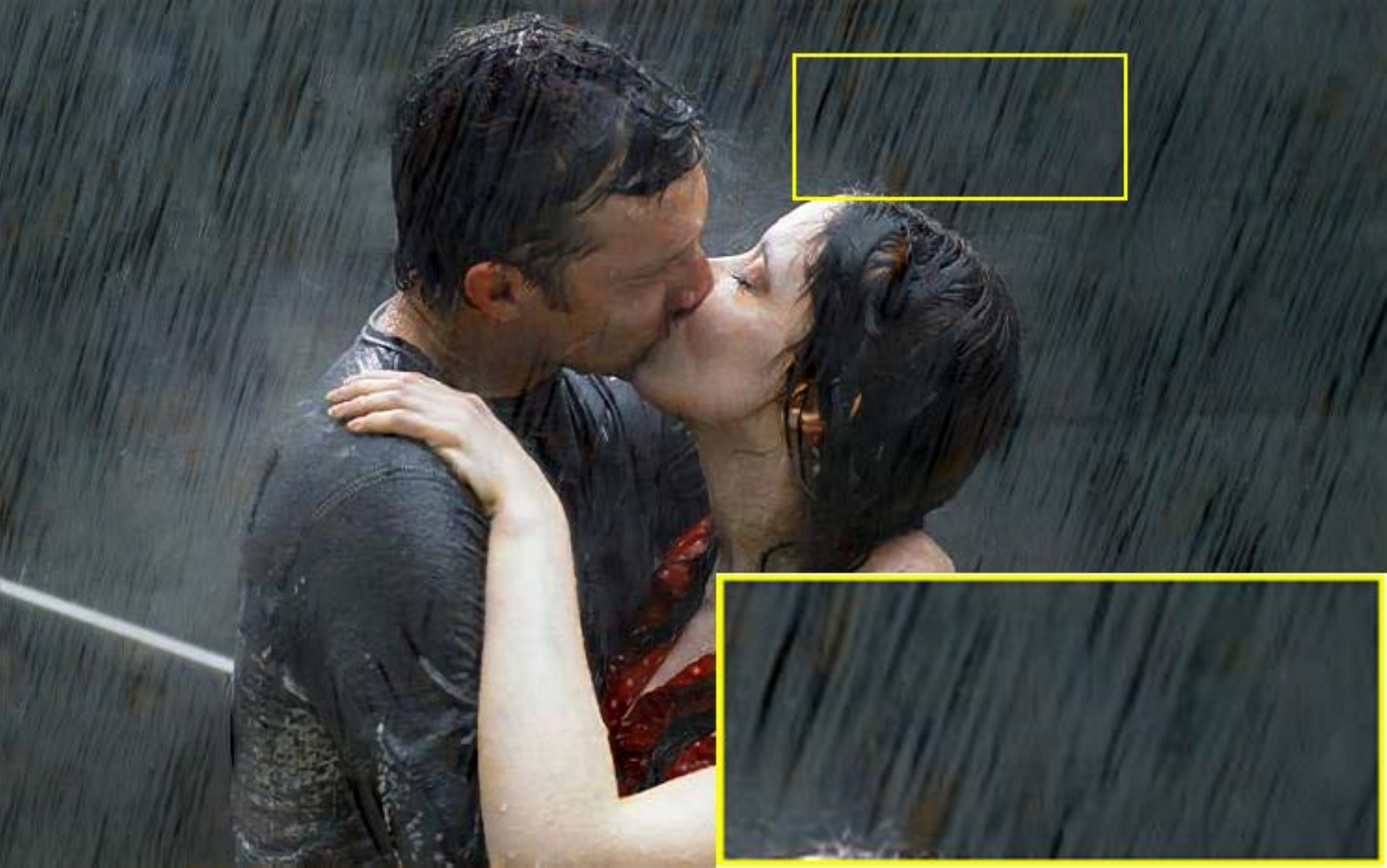}&\hspace{-4mm}
			\includegraphics[width = 0.12\linewidth]{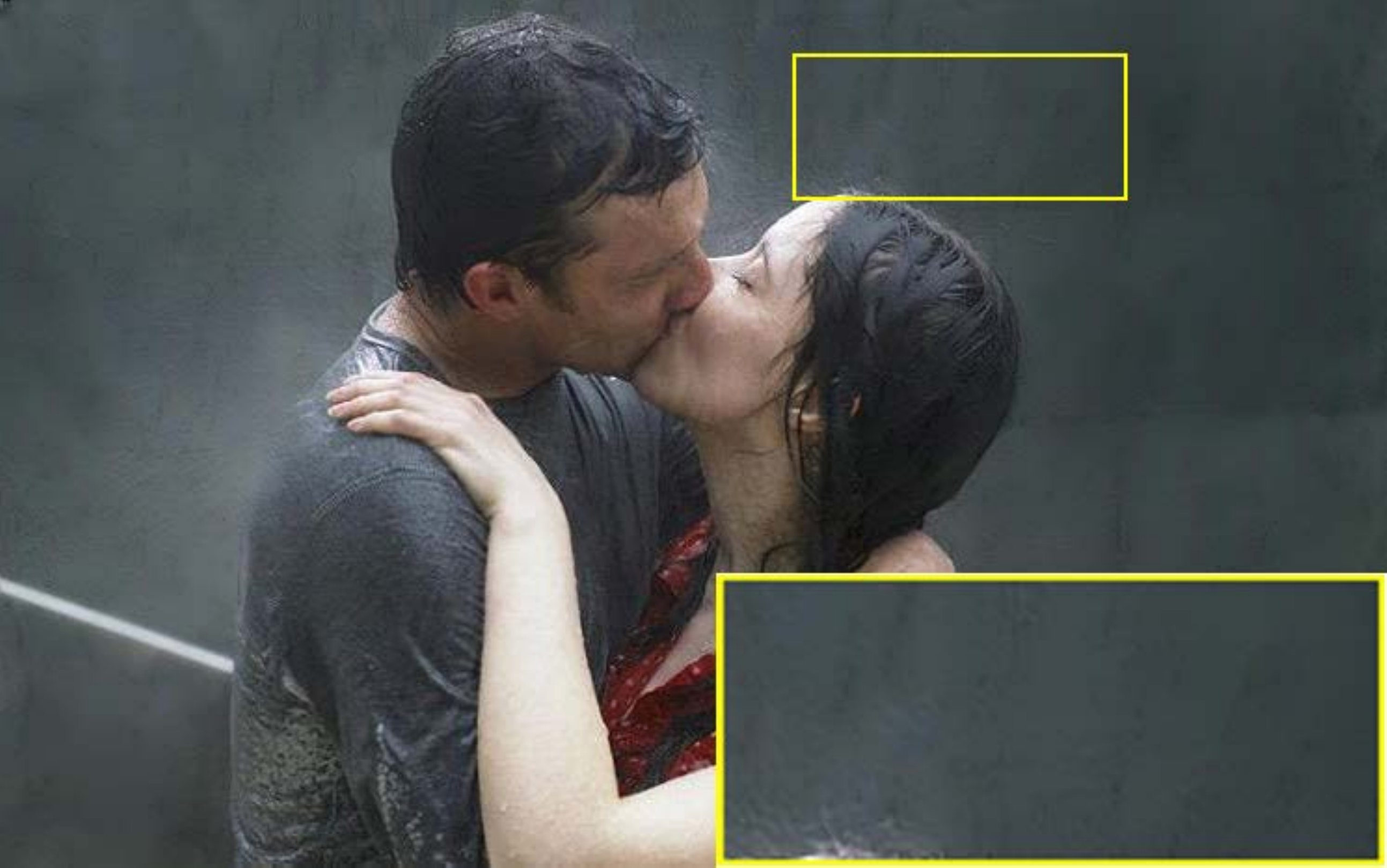}
			\\
			\includegraphics[width = 0.12\linewidth]{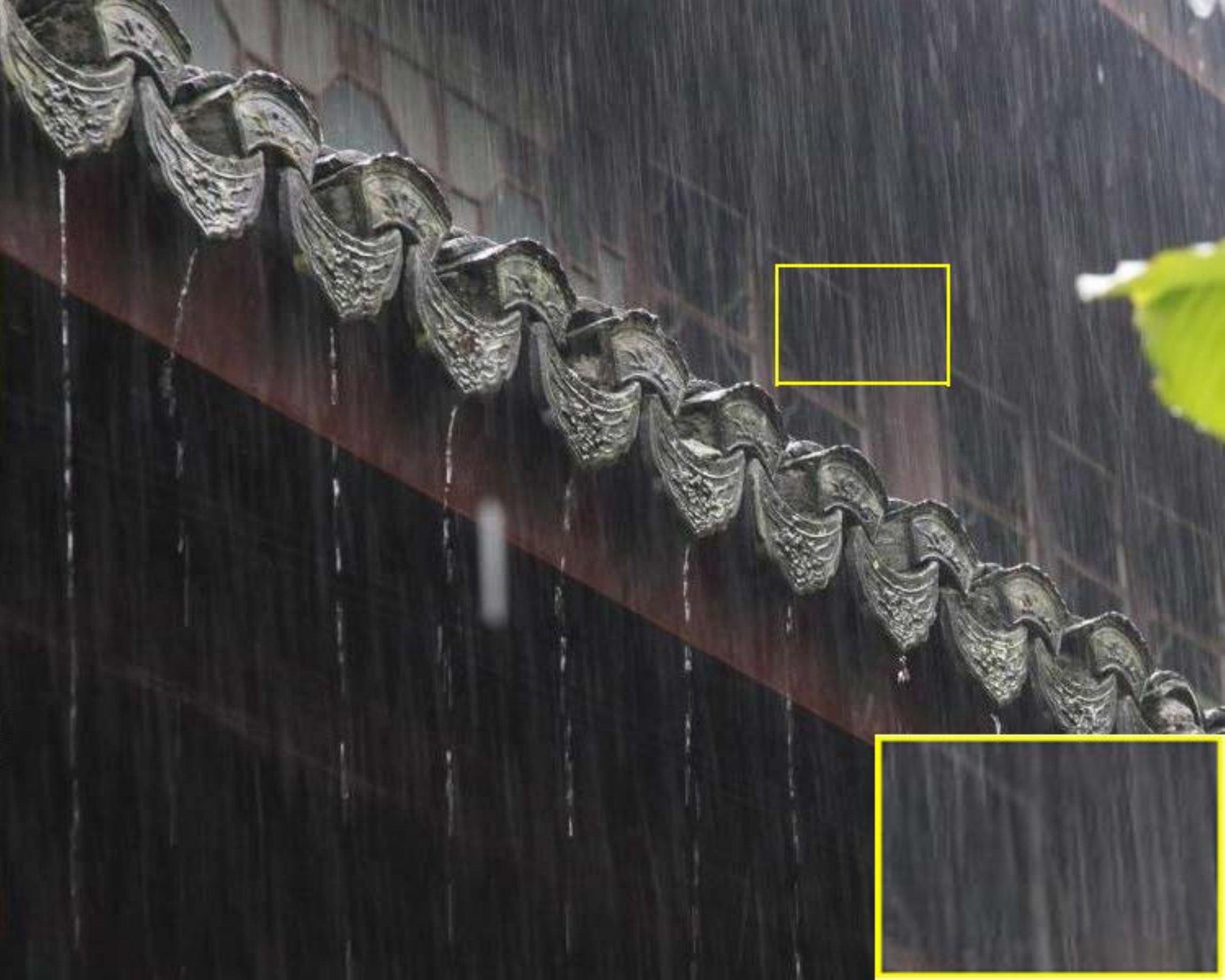}&\hspace{-4mm}
			\includegraphics[width = 0.12\linewidth]{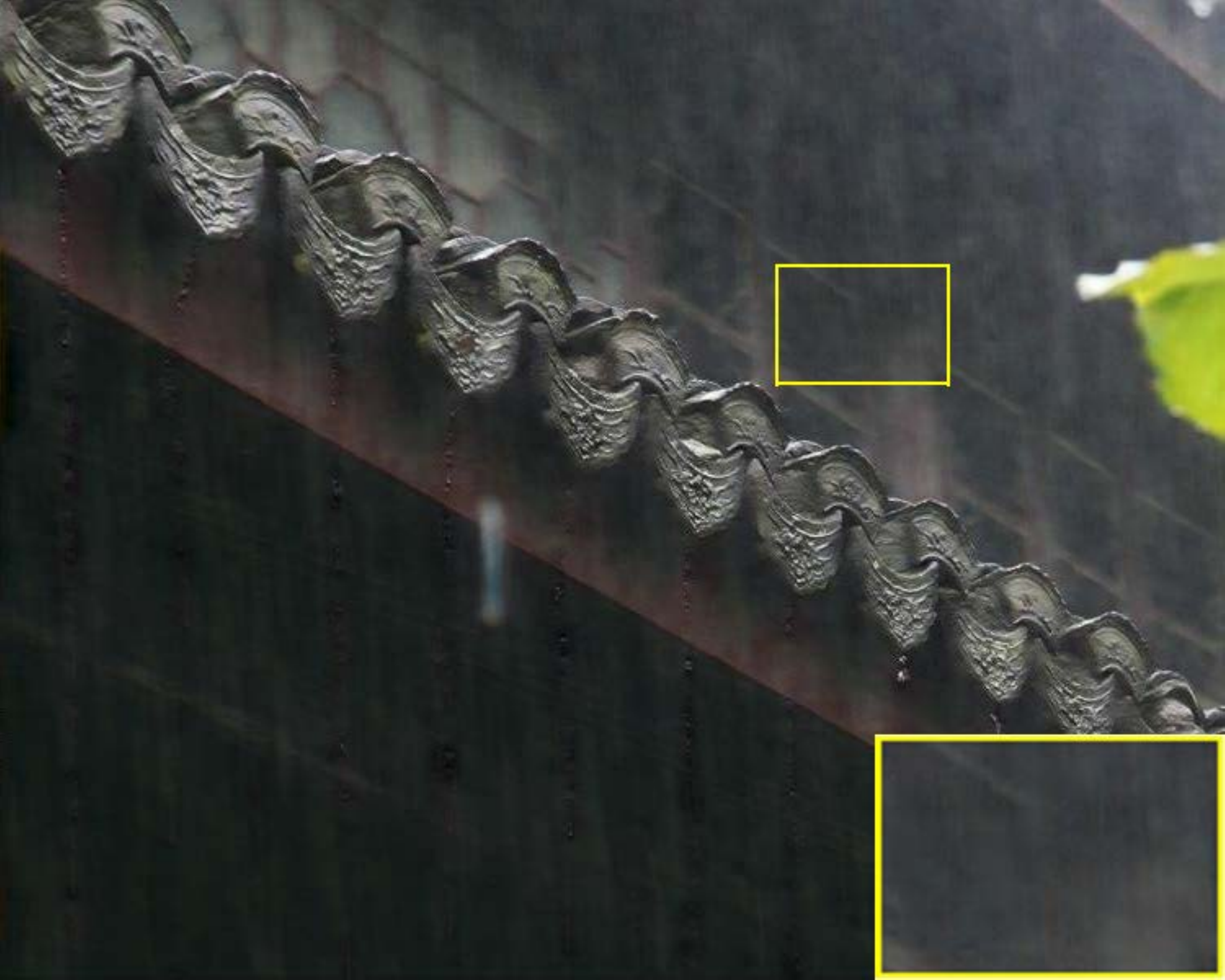}&\hspace{-4mm}
			\includegraphics[width = 0.12\linewidth]{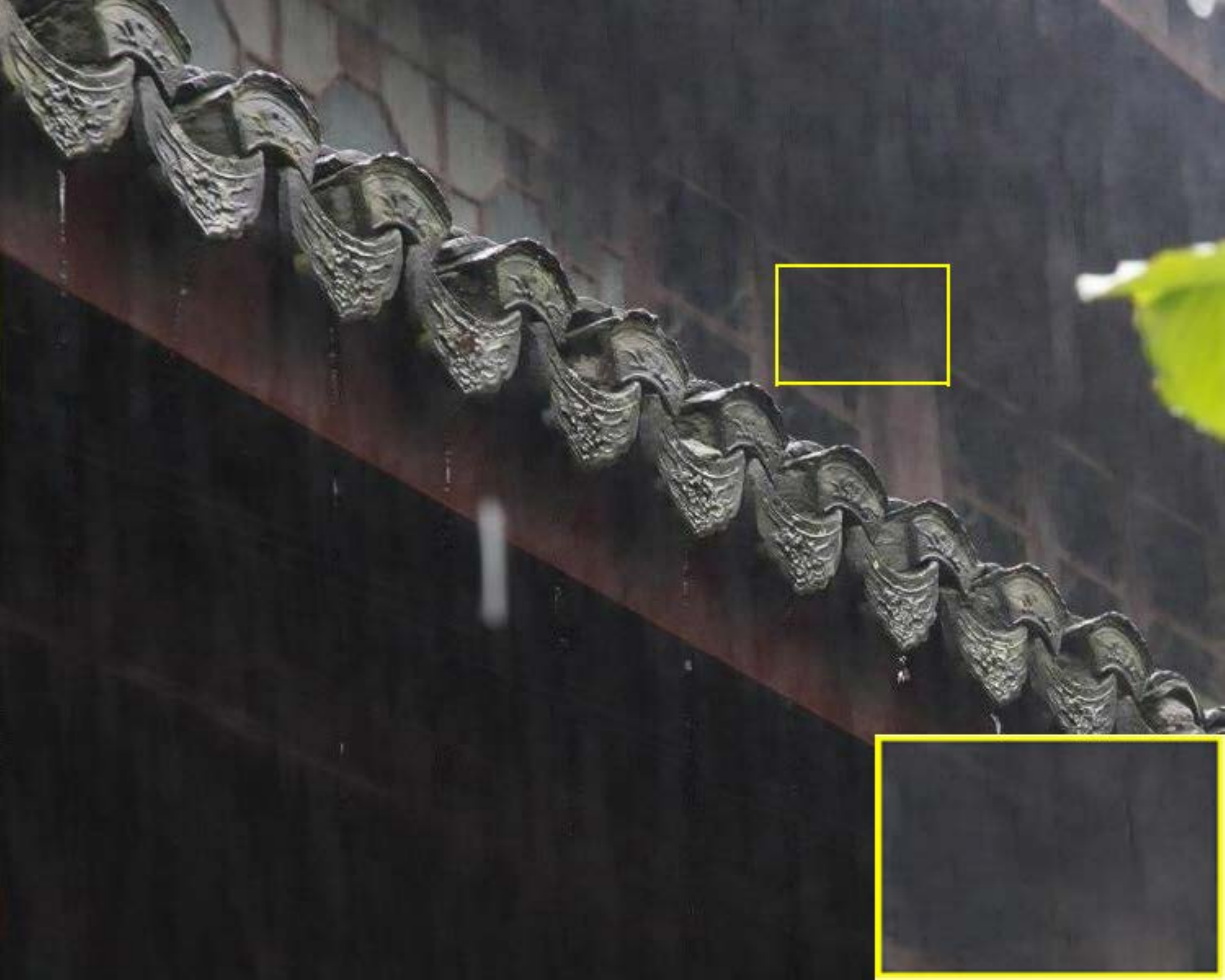}&\hspace{-4mm}
			\includegraphics[width = 0.12\linewidth]{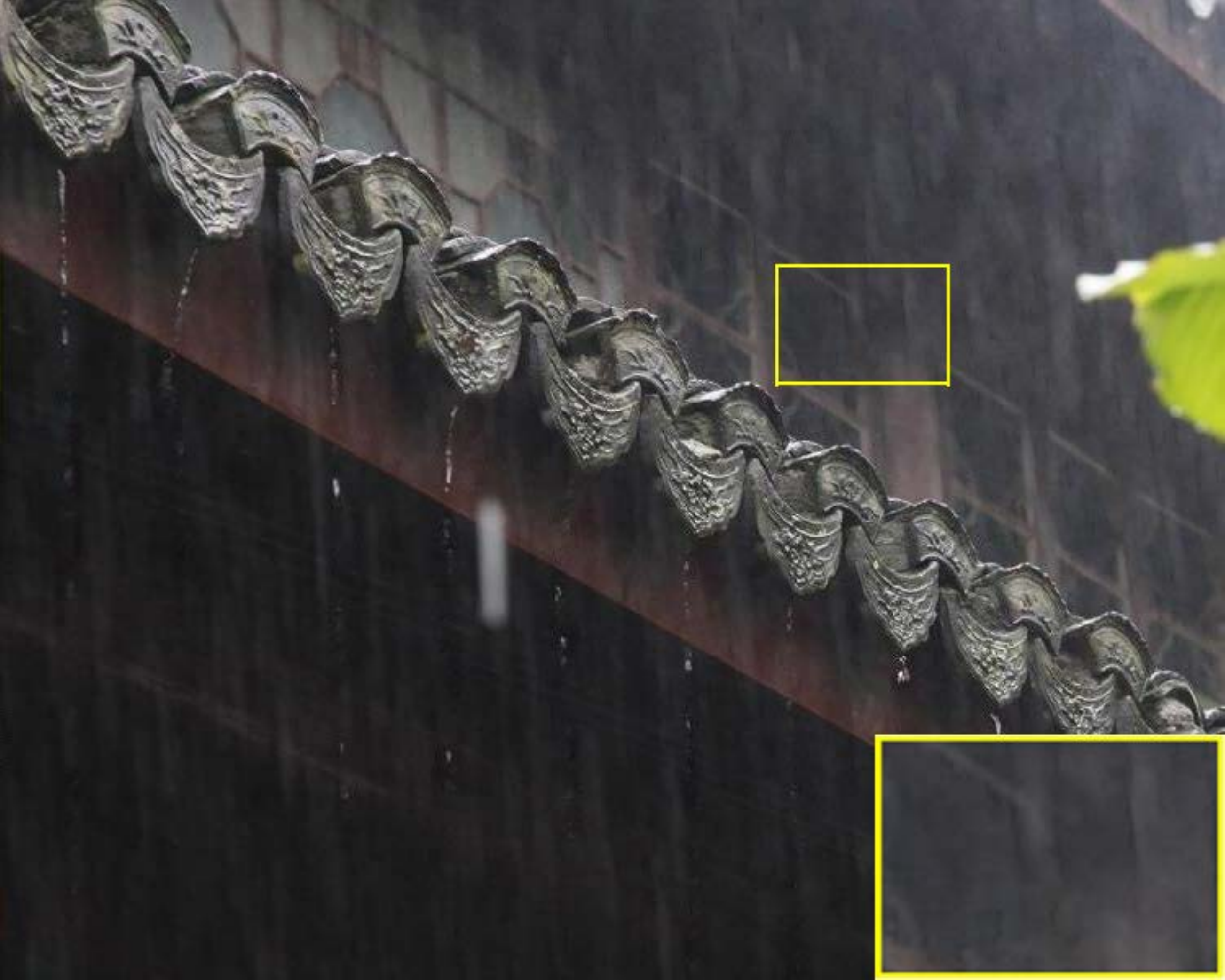}&\hspace{-4mm}
			\includegraphics[width = 0.12\linewidth]{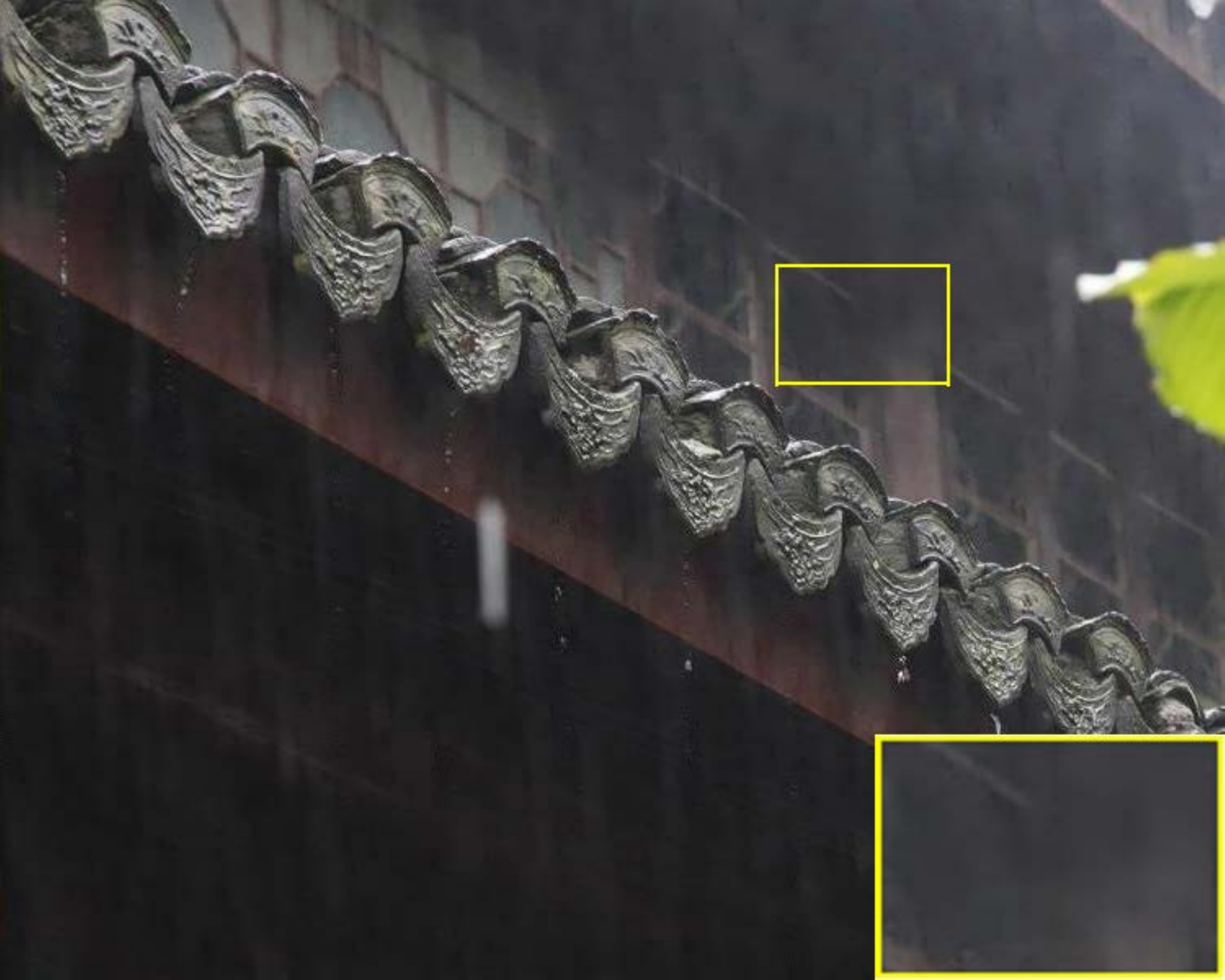}&\hspace{-4mm}
			\includegraphics[width = 0.12\linewidth]{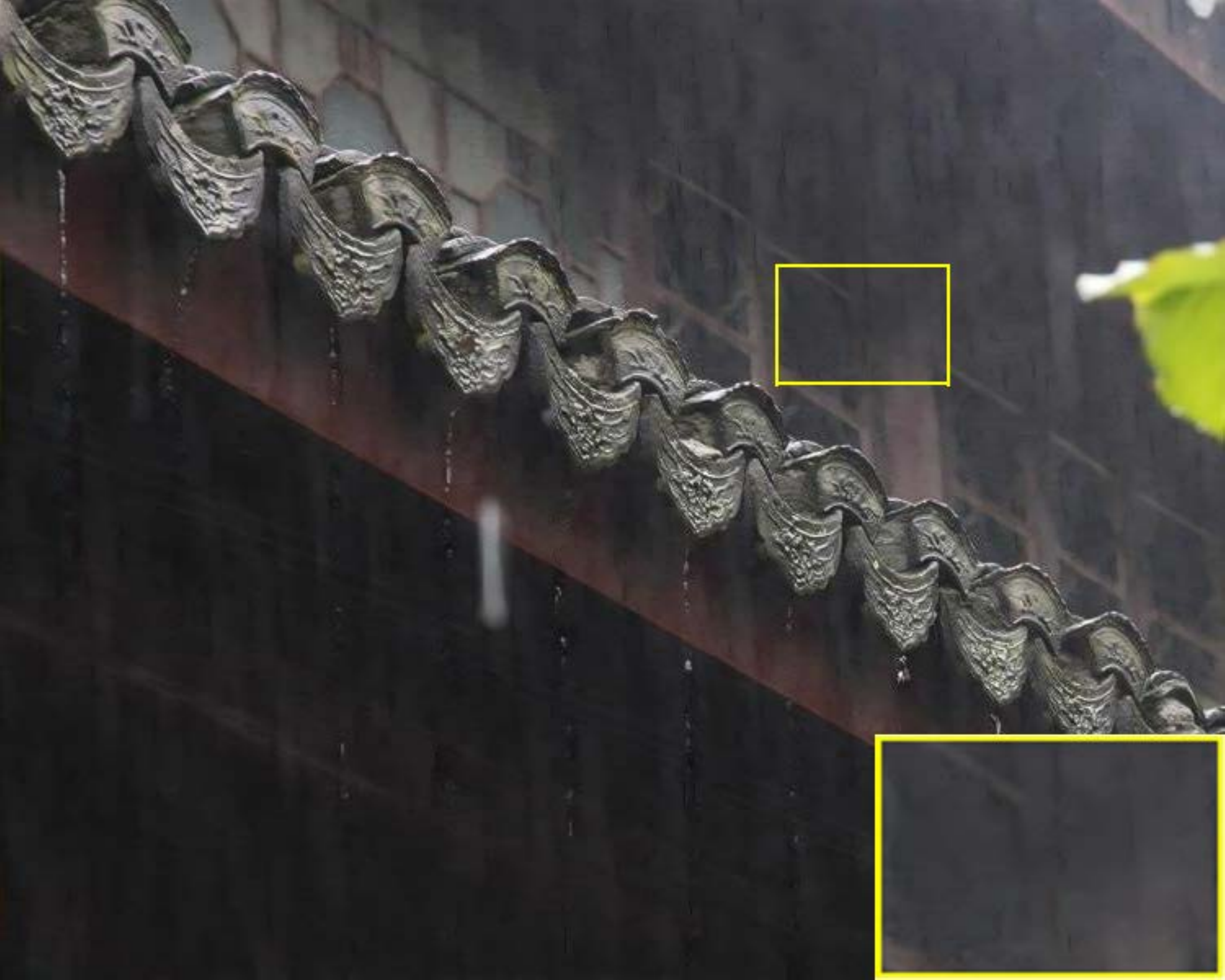}&\hspace{-4mm}
			\includegraphics[width = 0.12\linewidth]{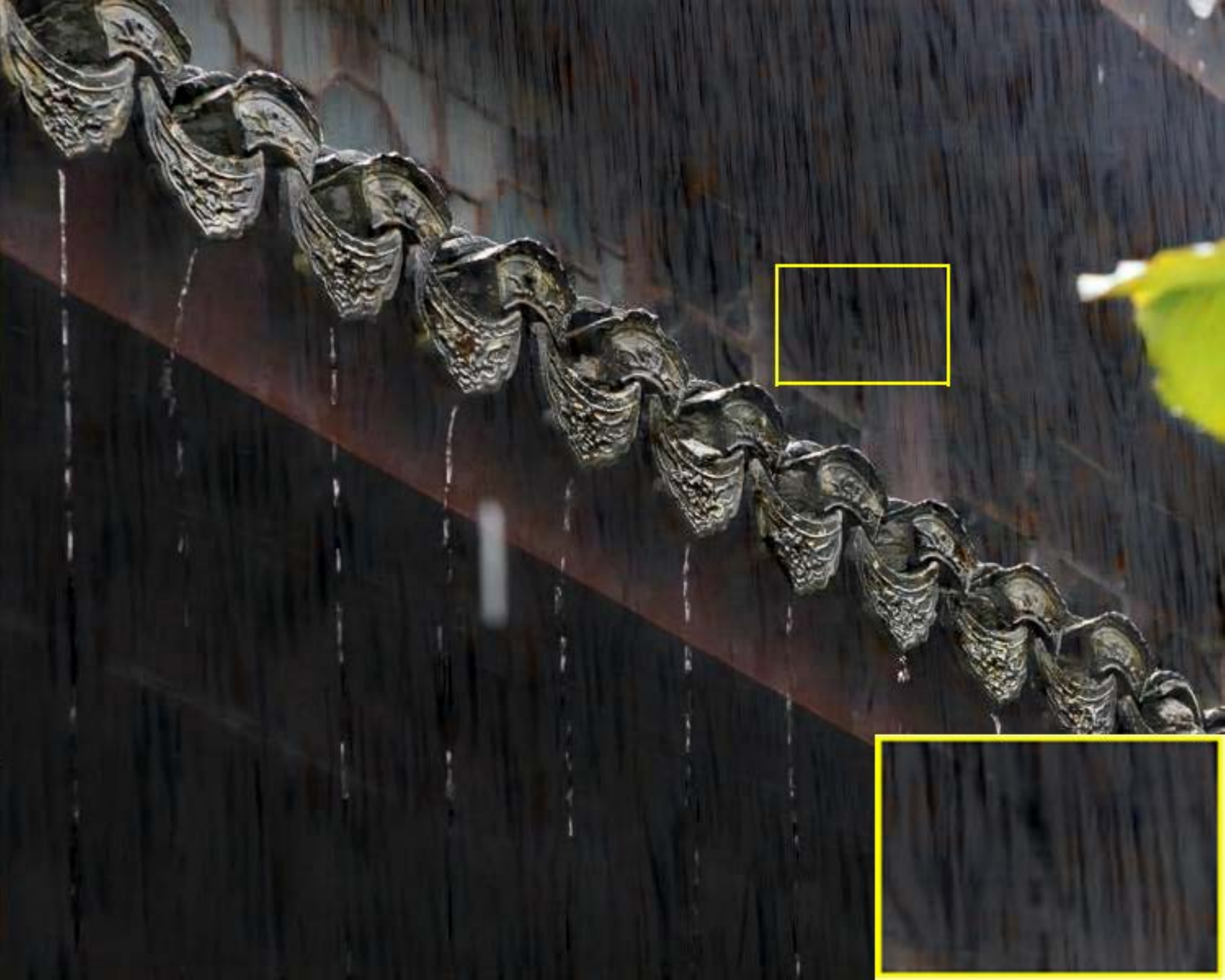}&\hspace{-4mm}
			\includegraphics[width = 0.12\linewidth]{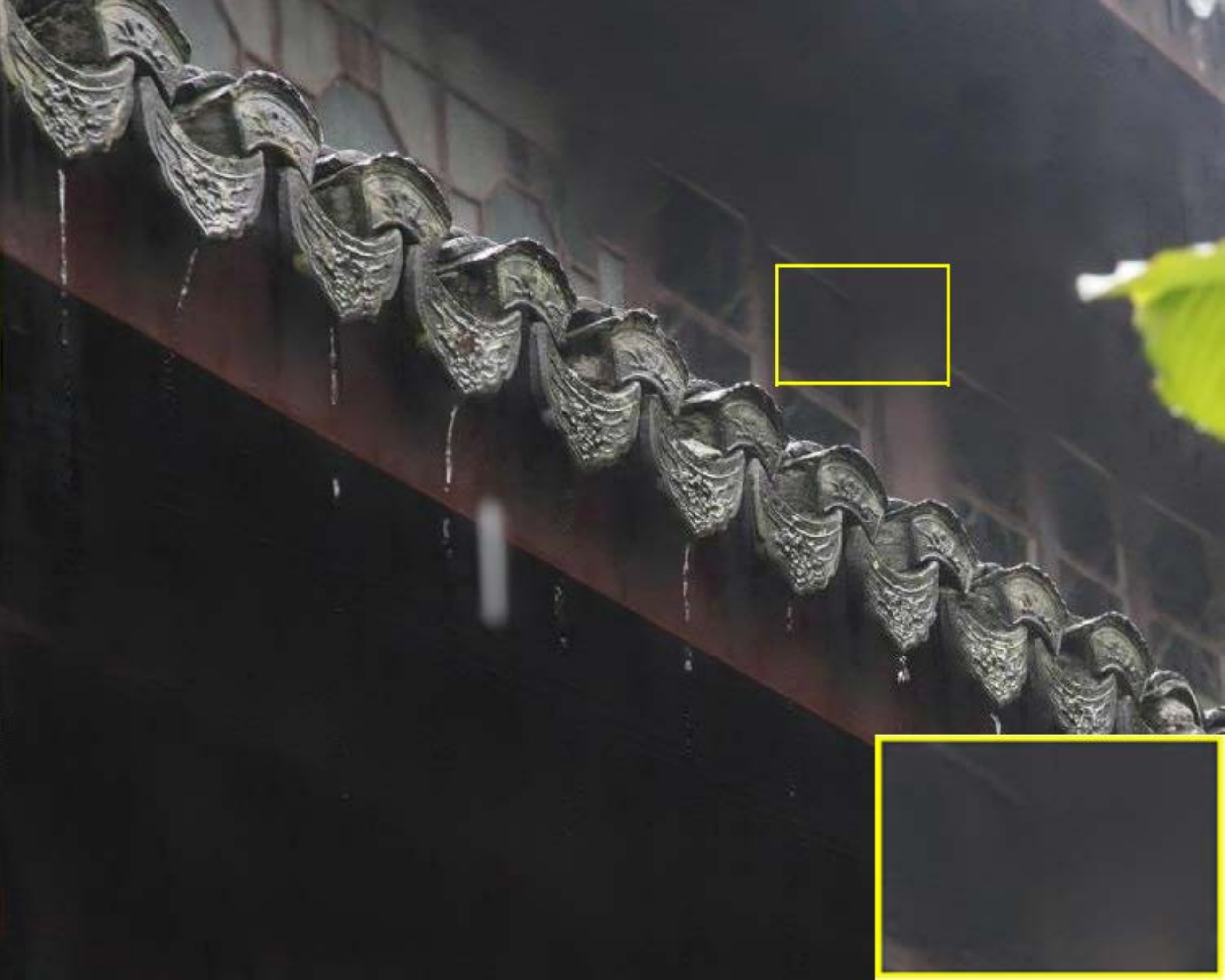}
			\\
			\includegraphics[width = 0.12\linewidth]{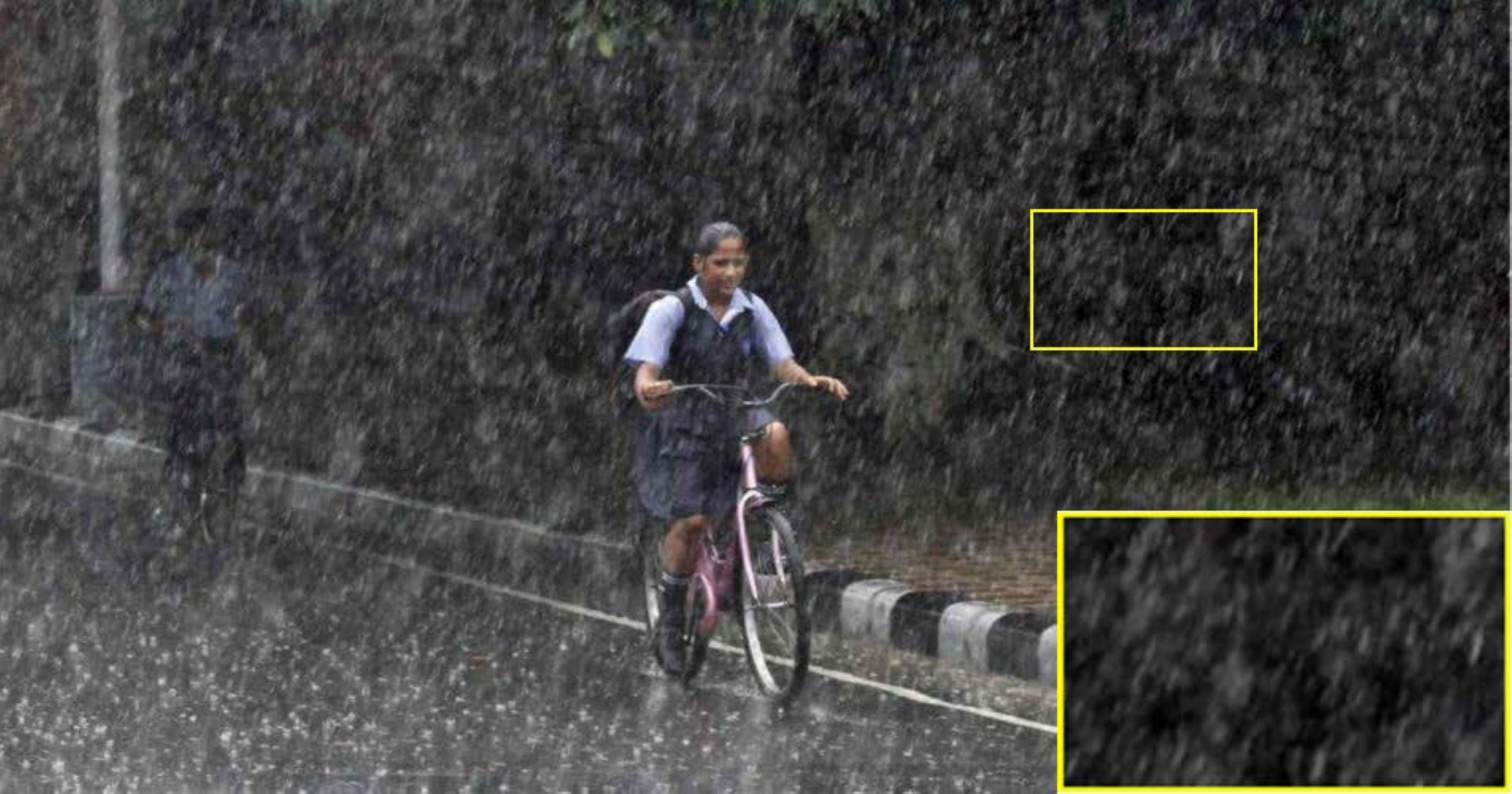}&\hspace{-4mm}
			\includegraphics[width = 0.12\linewidth]{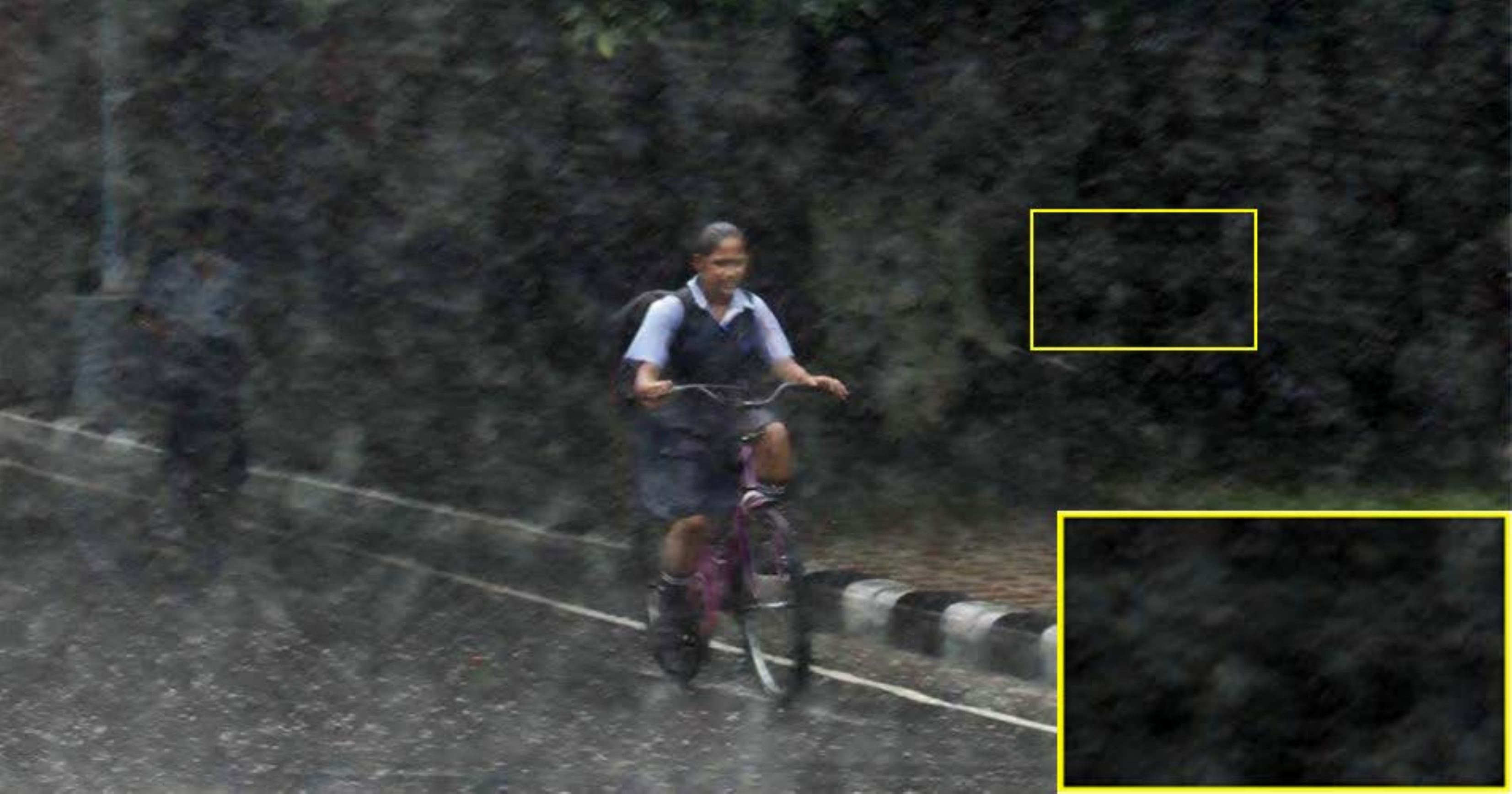}&\hspace{-4mm}
			\includegraphics[width = 0.12\linewidth]{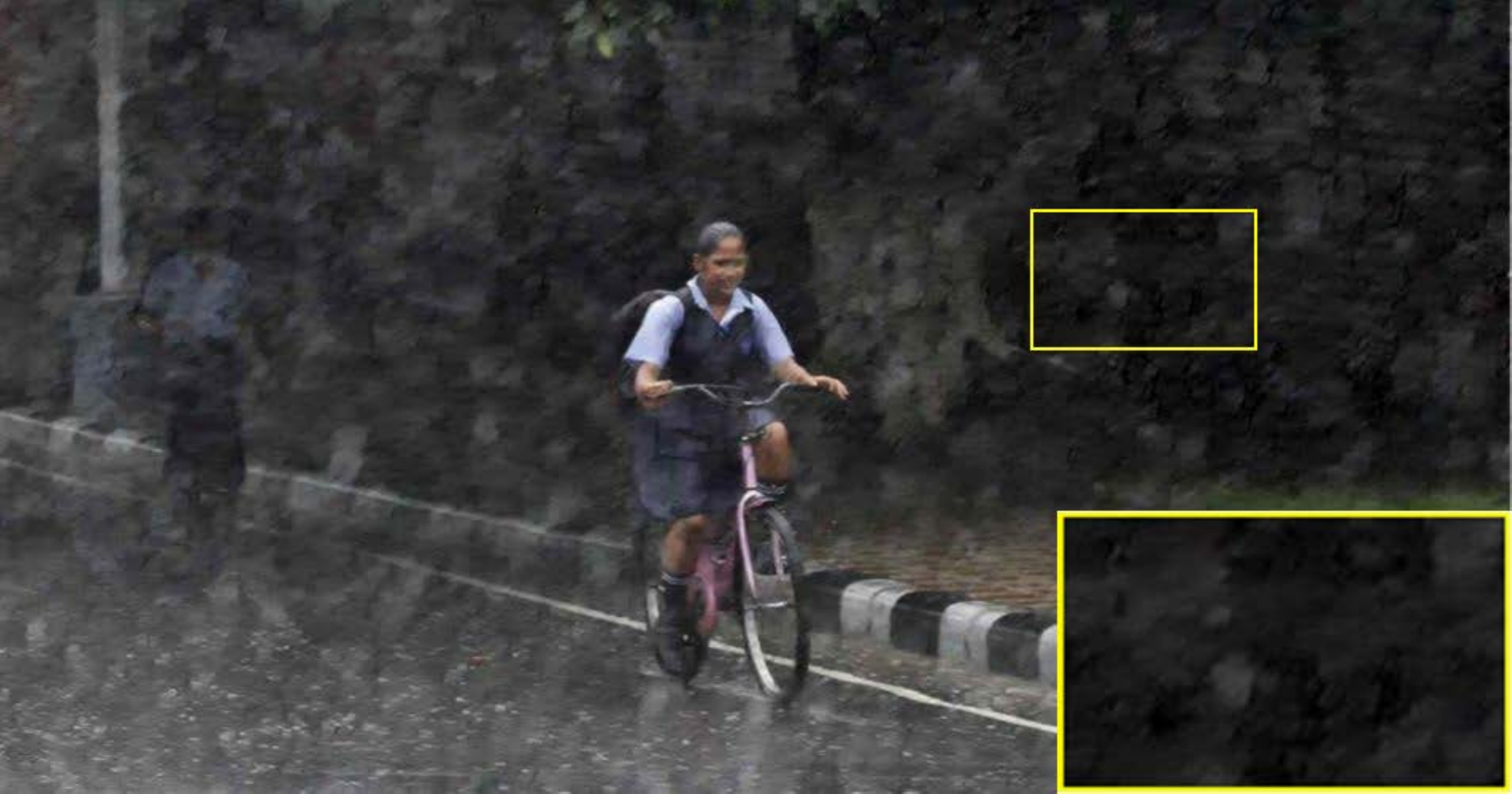}&\hspace{-4mm}
			\includegraphics[width = 0.12\linewidth]{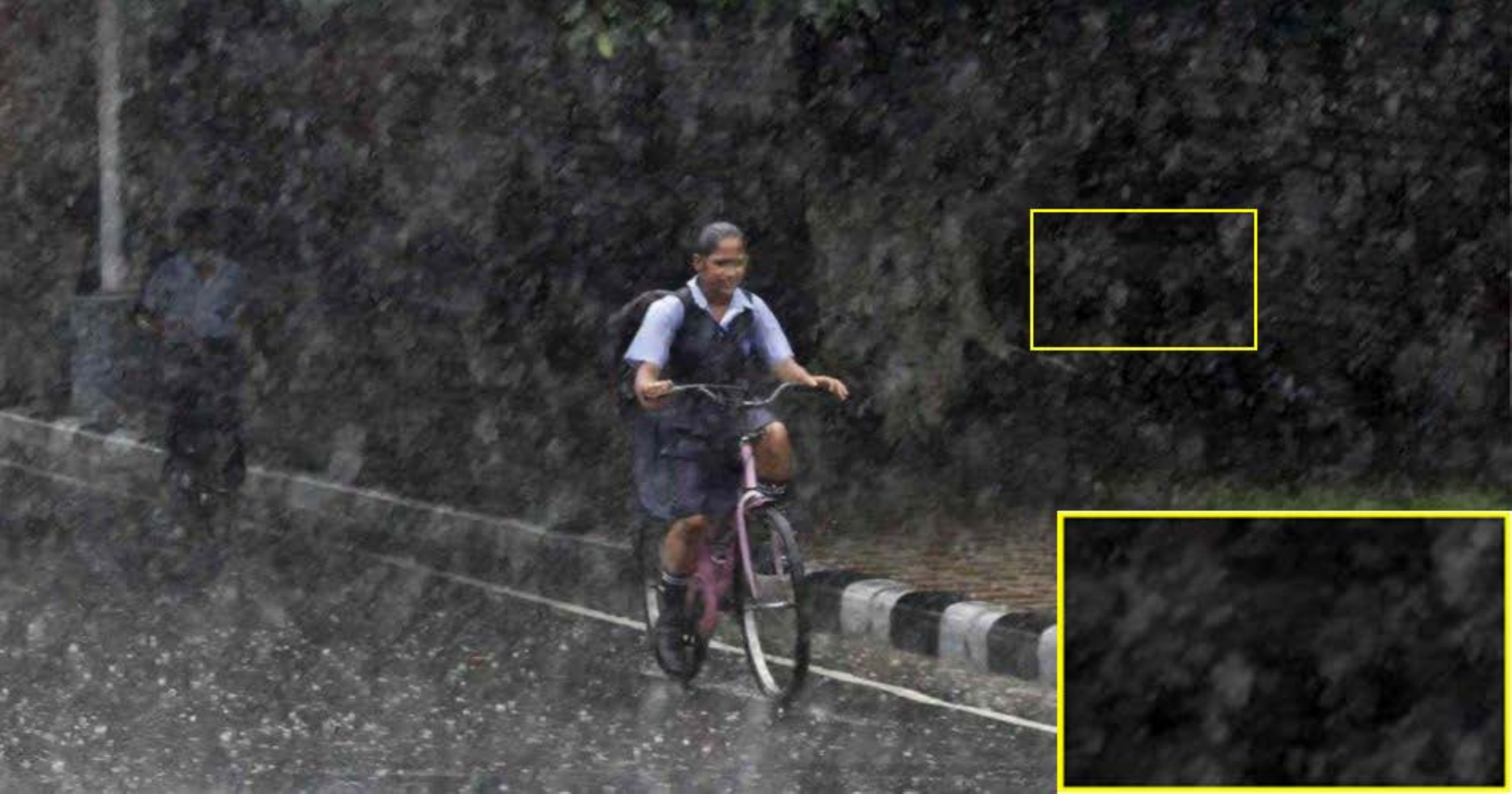}&\hspace{-4mm}
			\includegraphics[width = 0.12\linewidth]{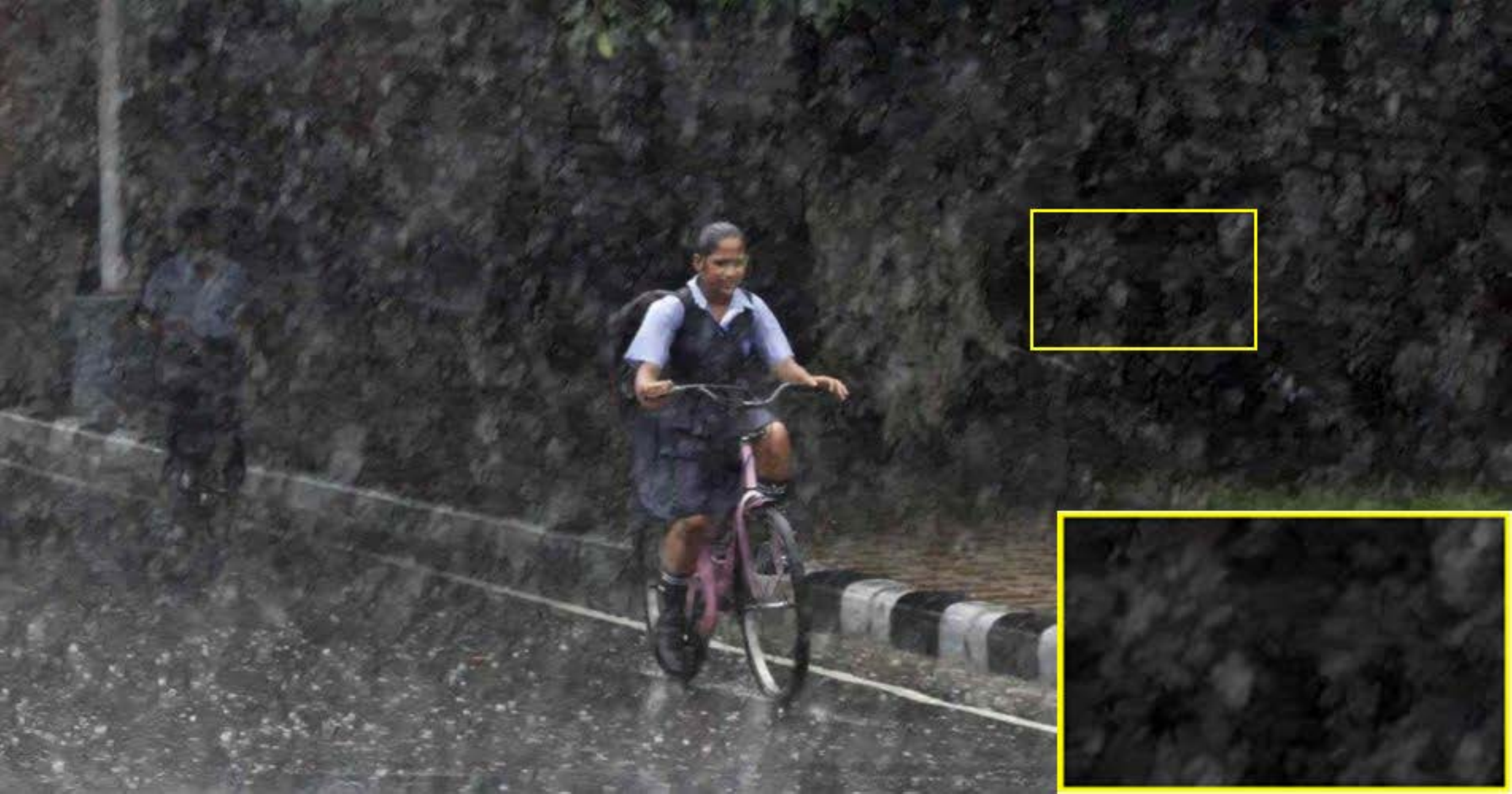}&\hspace{-4mm}
			\includegraphics[width = 0.12\linewidth]{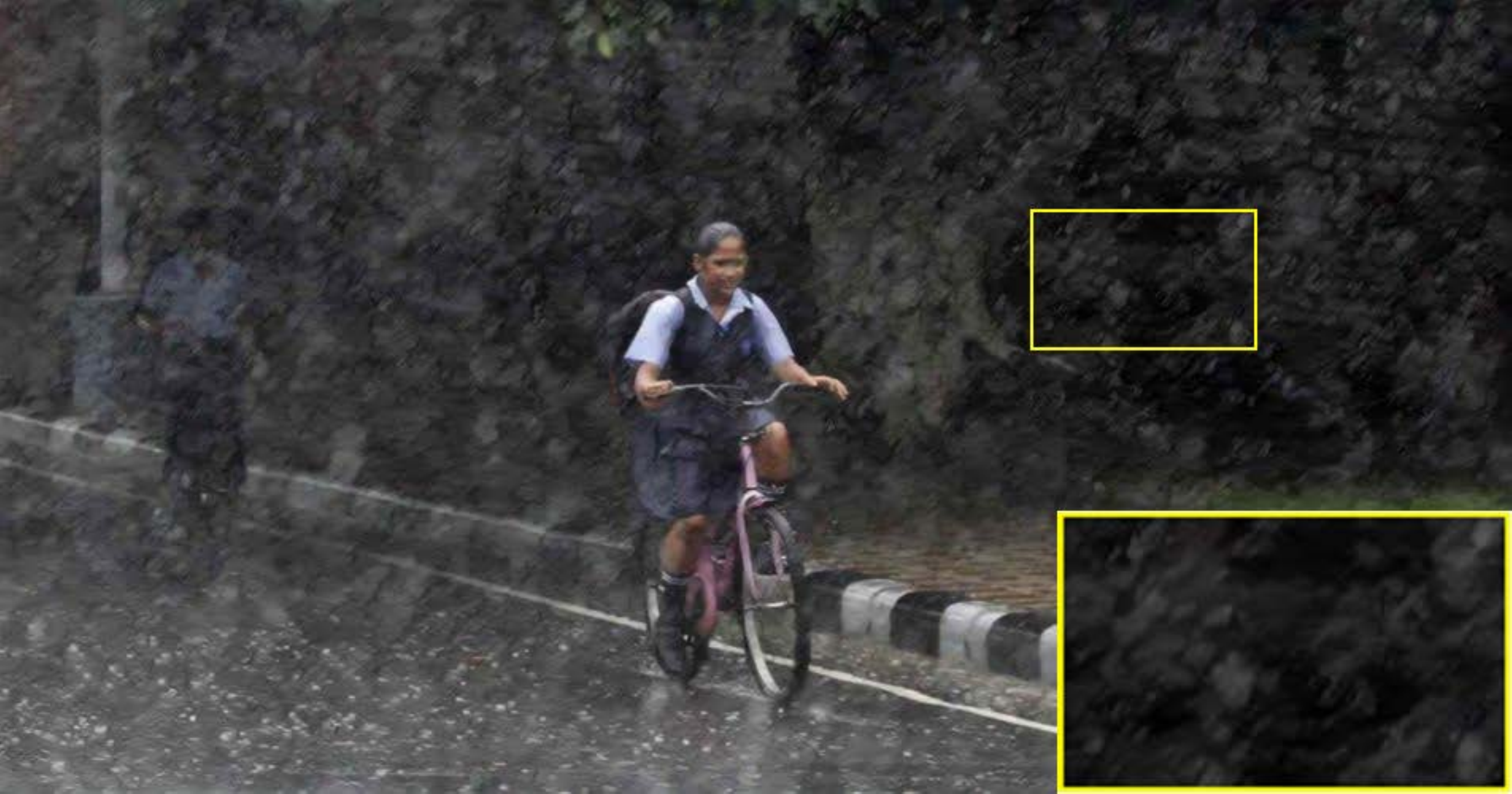}&\hspace{-4mm}
			\includegraphics[width = 0.12\linewidth]{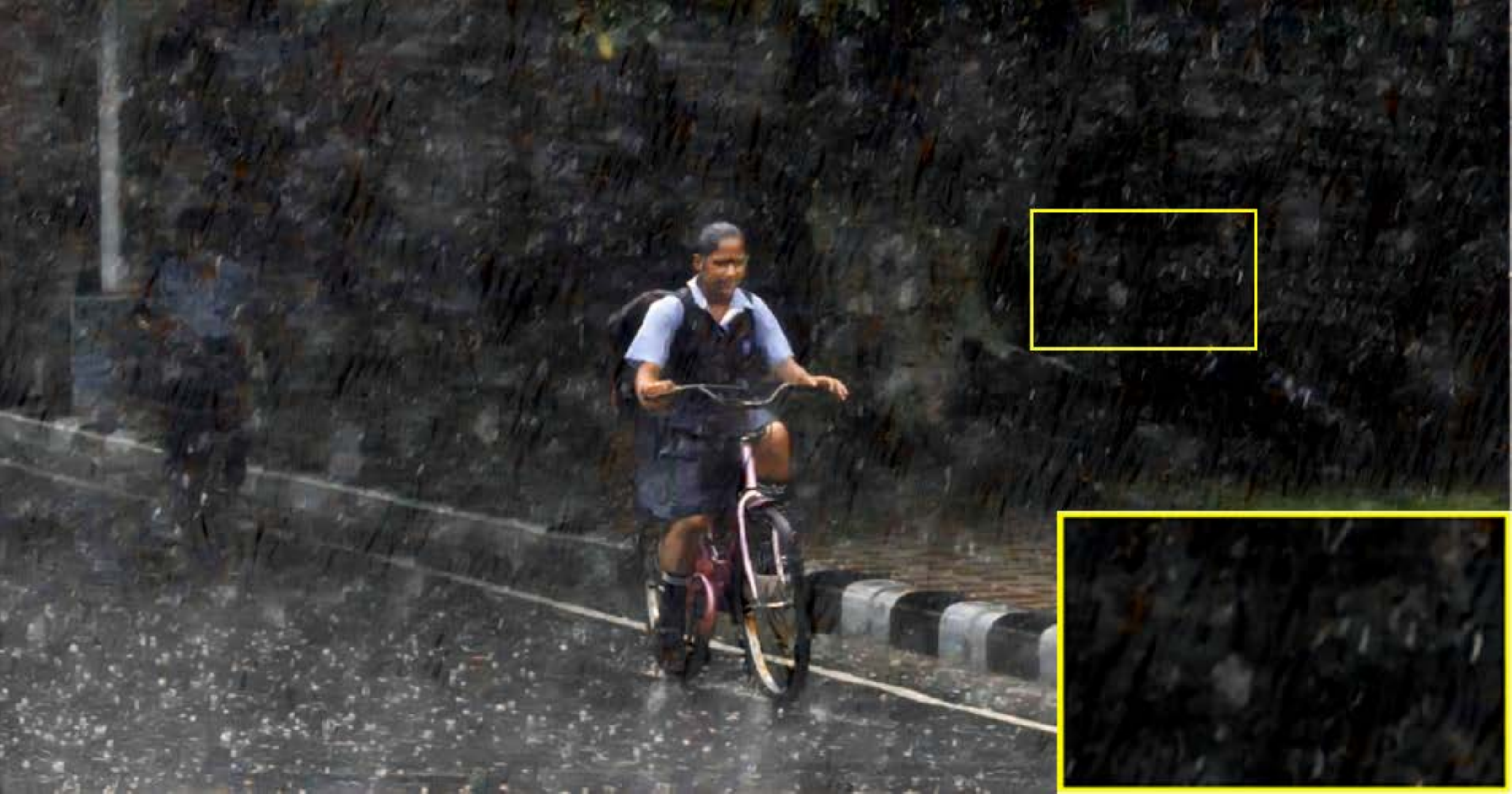}&\hspace{-4mm}
			\includegraphics[width = 0.12\linewidth]{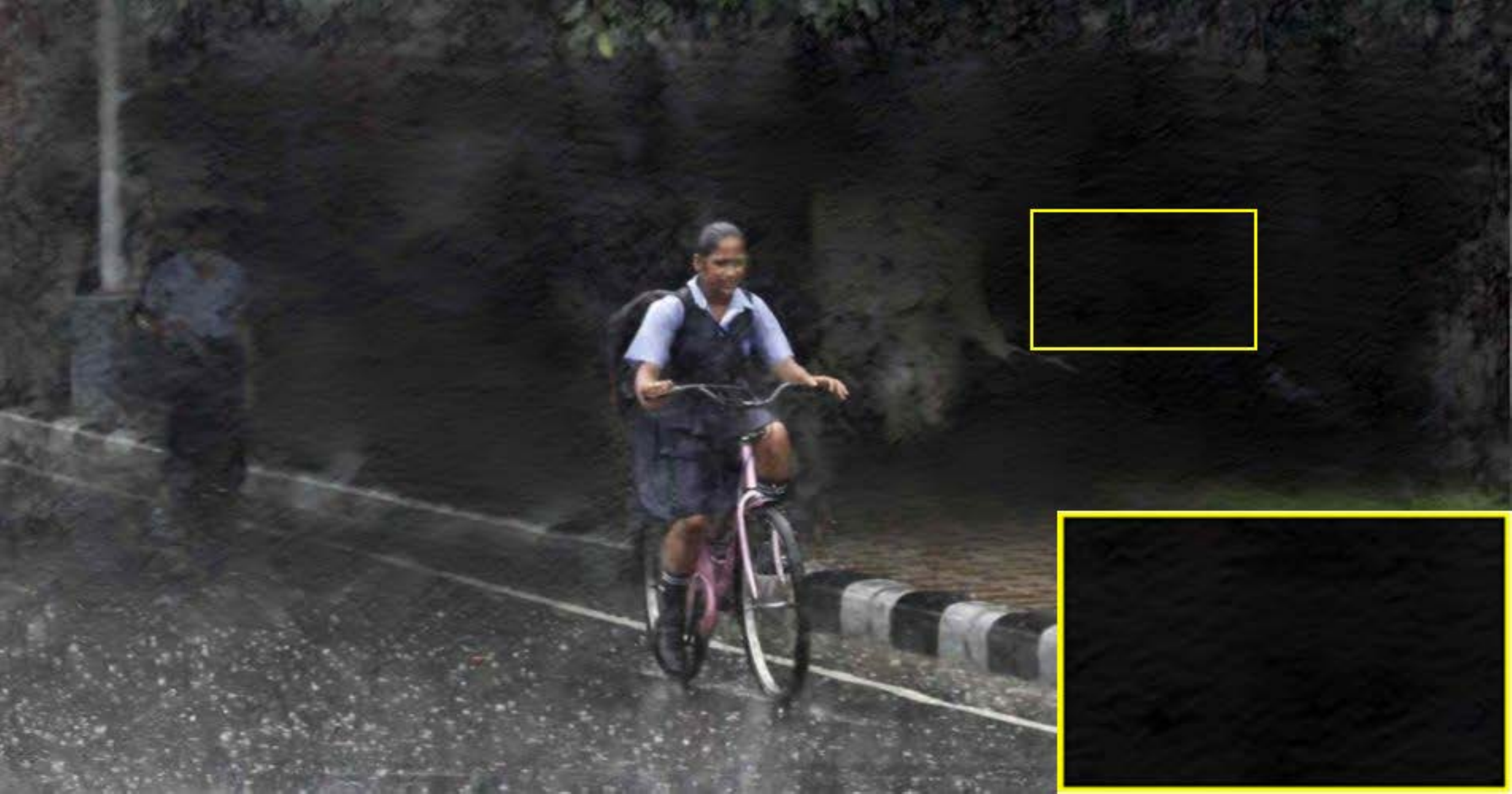}
			\\
			\includegraphics[width = 0.12\linewidth]{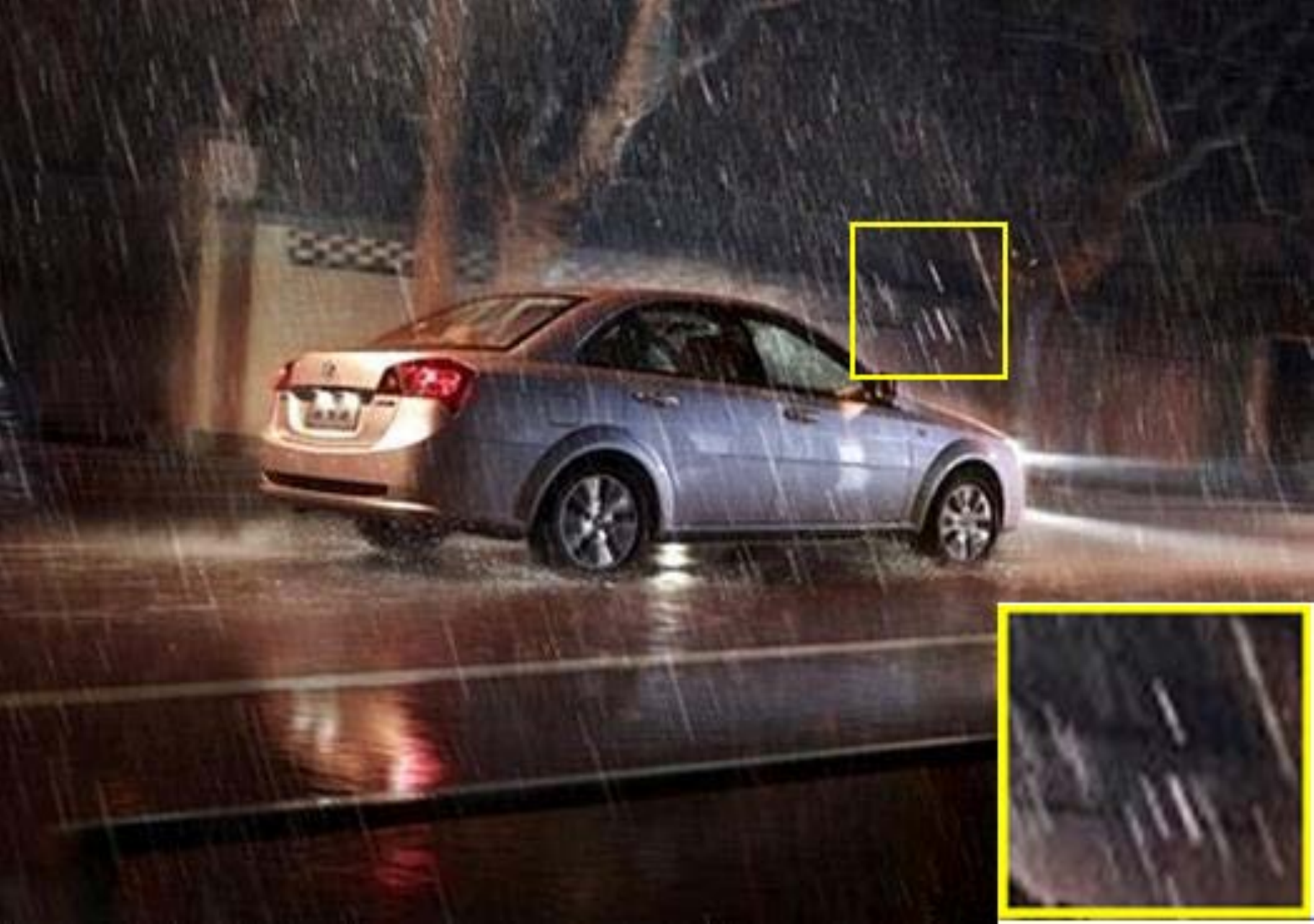}&\hspace{-4mm}
			\includegraphics[width = 0.12\linewidth]{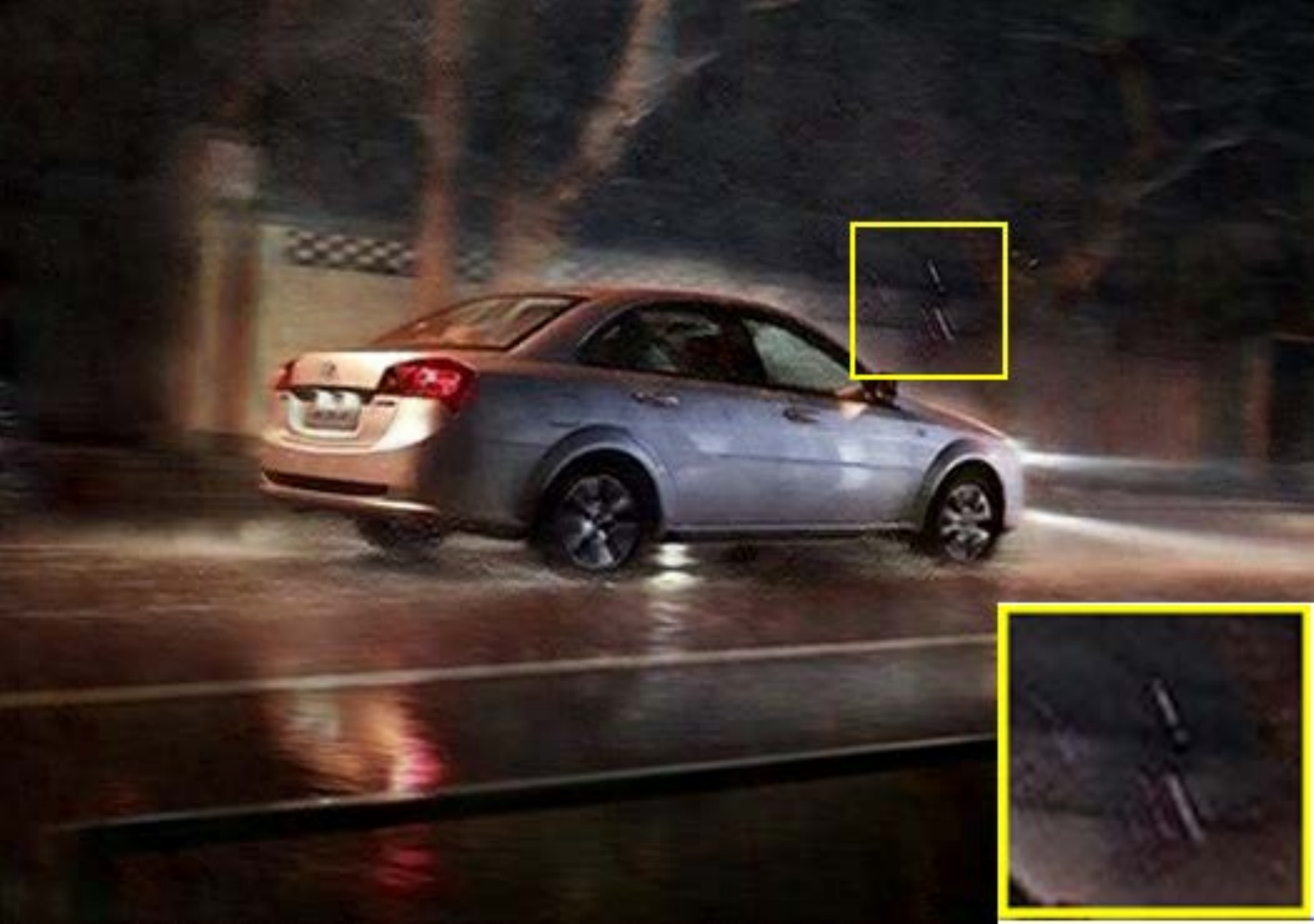}&\hspace{-4mm}
			\includegraphics[width = 0.12\linewidth]{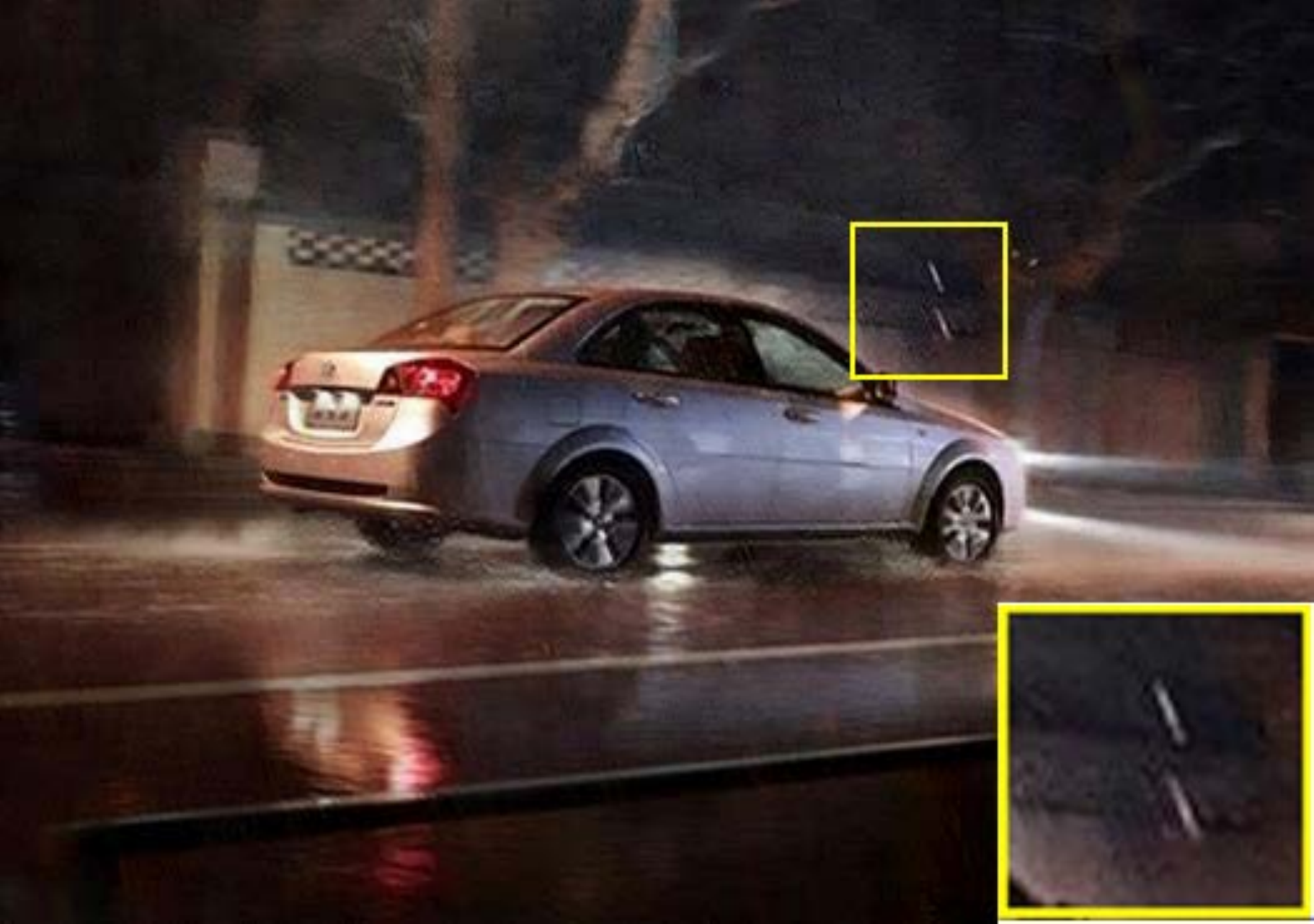}&\hspace{-4mm}
			\includegraphics[width = 0.12\linewidth]{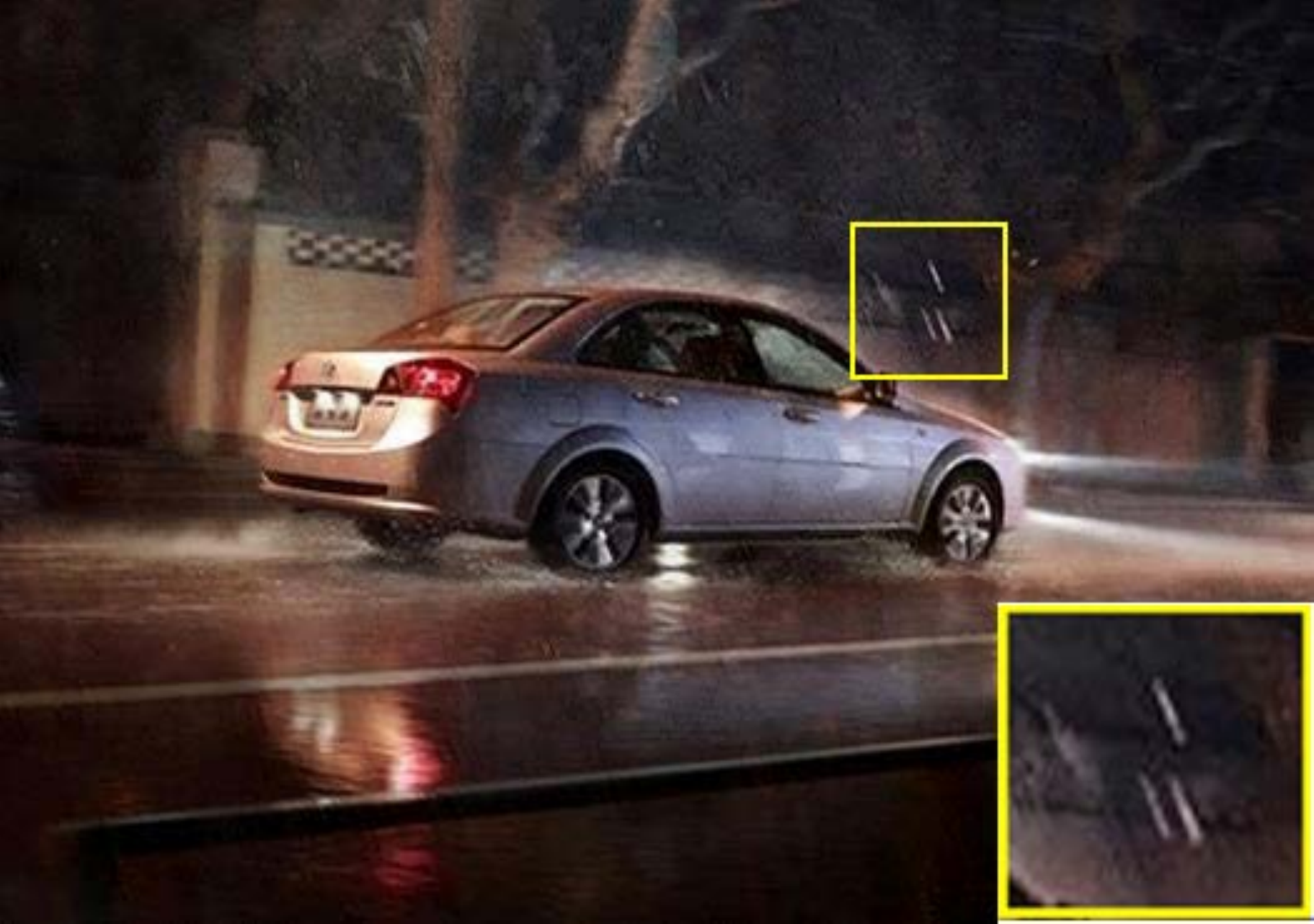}&\hspace{-4mm}
			\includegraphics[width = 0.12\linewidth]{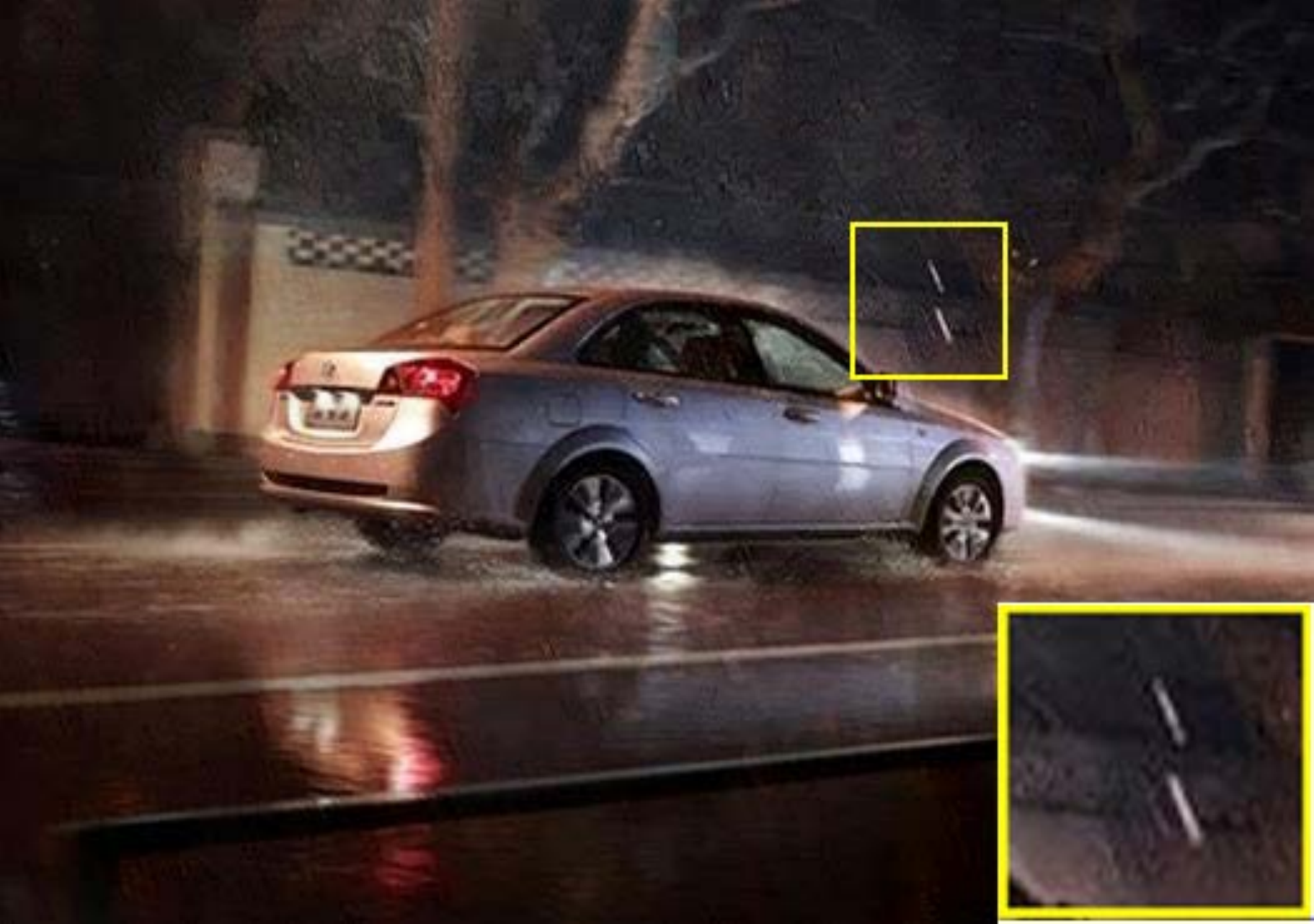}&\hspace{-4mm}
			\includegraphics[width = 0.12\linewidth]{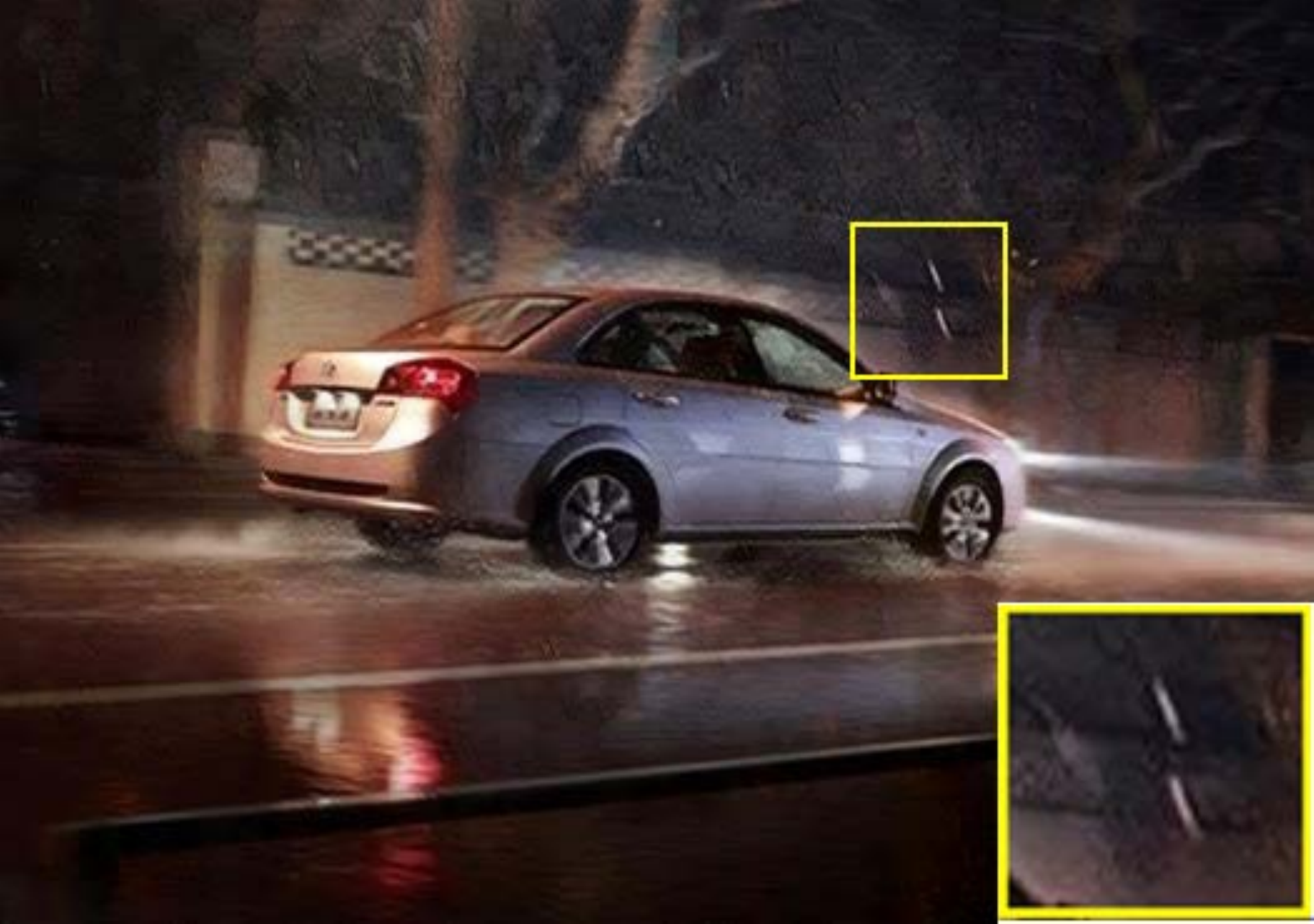}&\hspace{-4mm}
			\includegraphics[width = 0.12\linewidth]{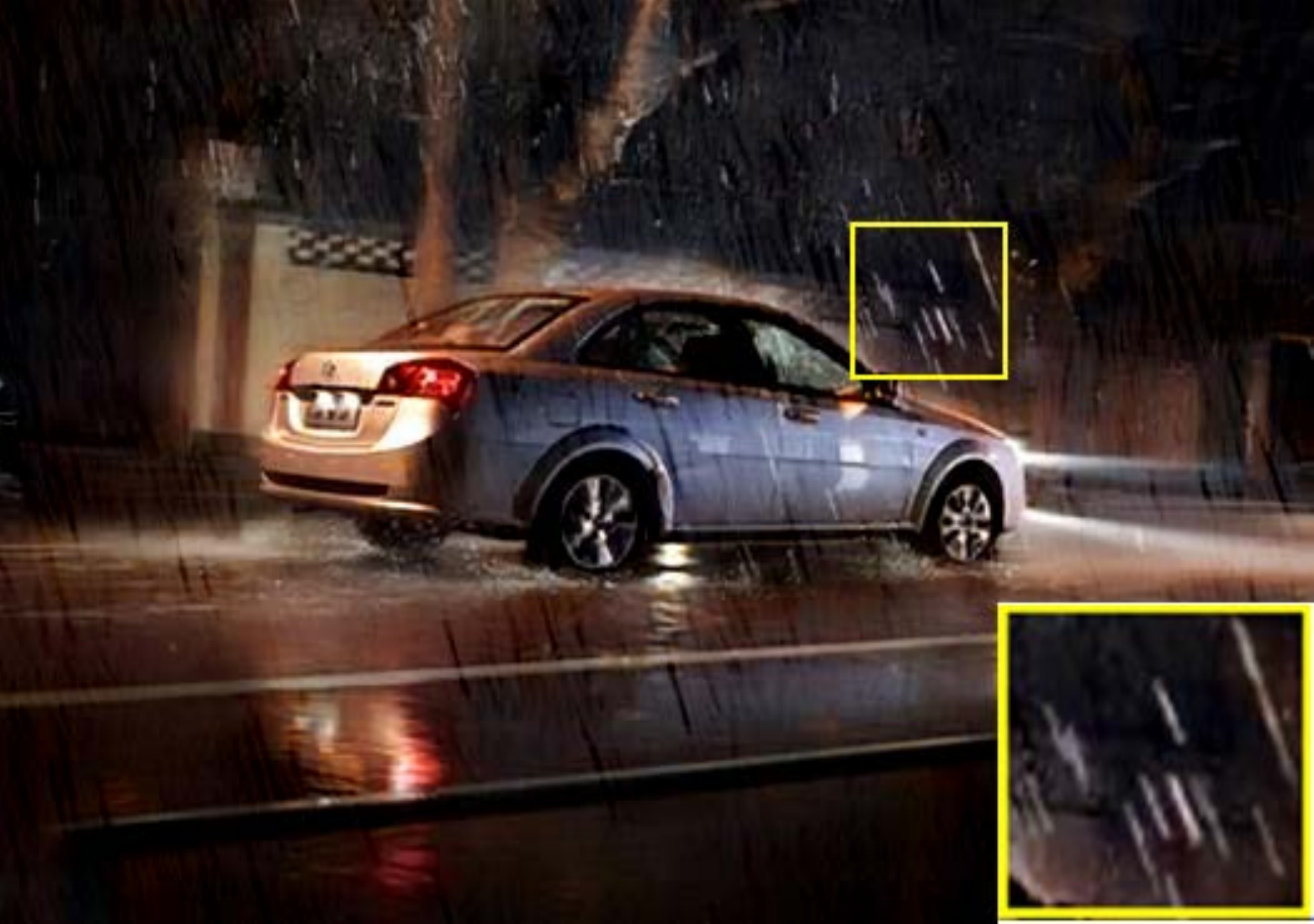}&\hspace{-4mm}
			\includegraphics[width = 0.12\linewidth]{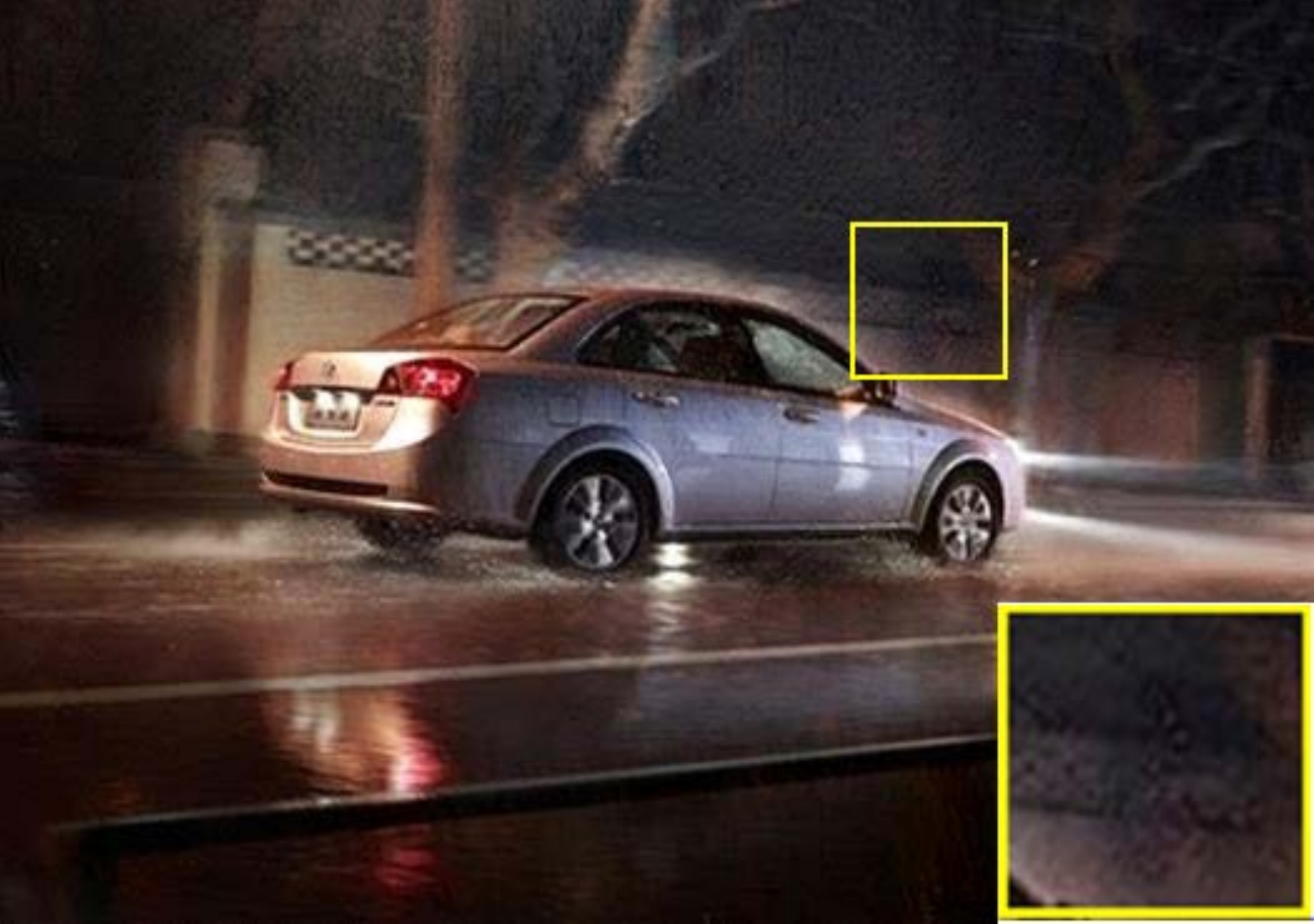}
			\\
			\includegraphics[width = 0.12\linewidth]{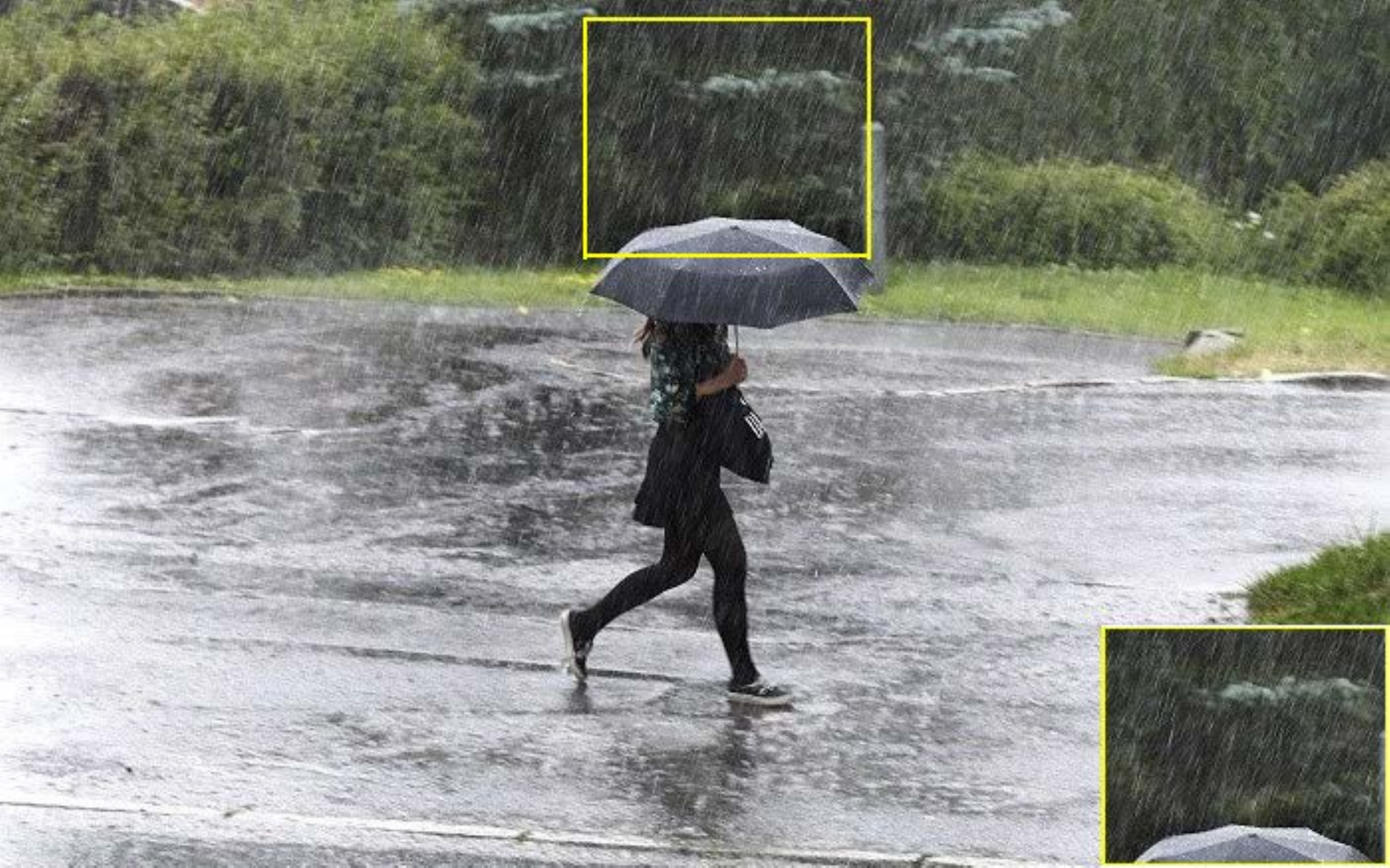}&\hspace{-4mm}
			\includegraphics[width = 0.12\linewidth]{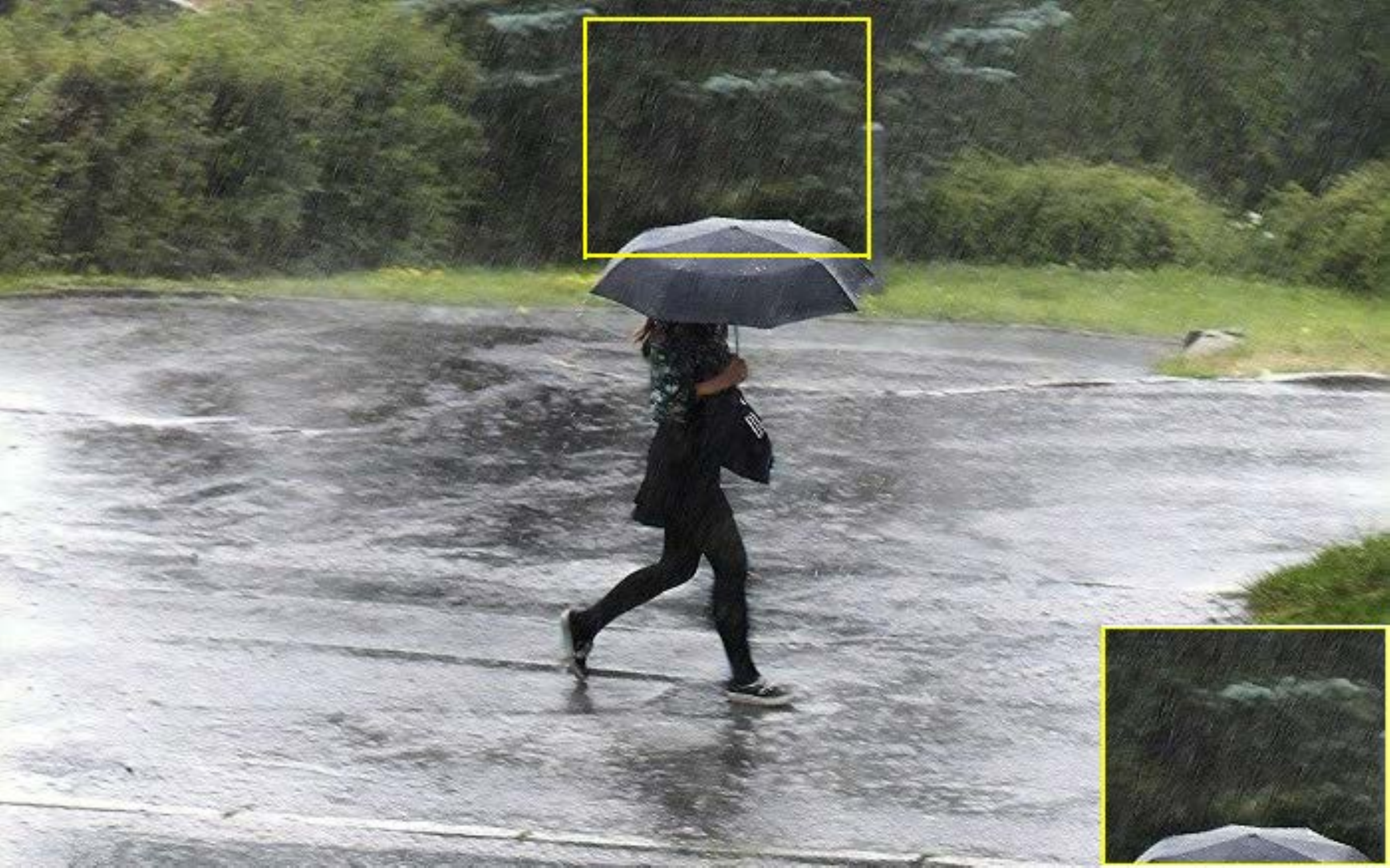}&\hspace{-4mm}
			\includegraphics[width = 0.12\linewidth]{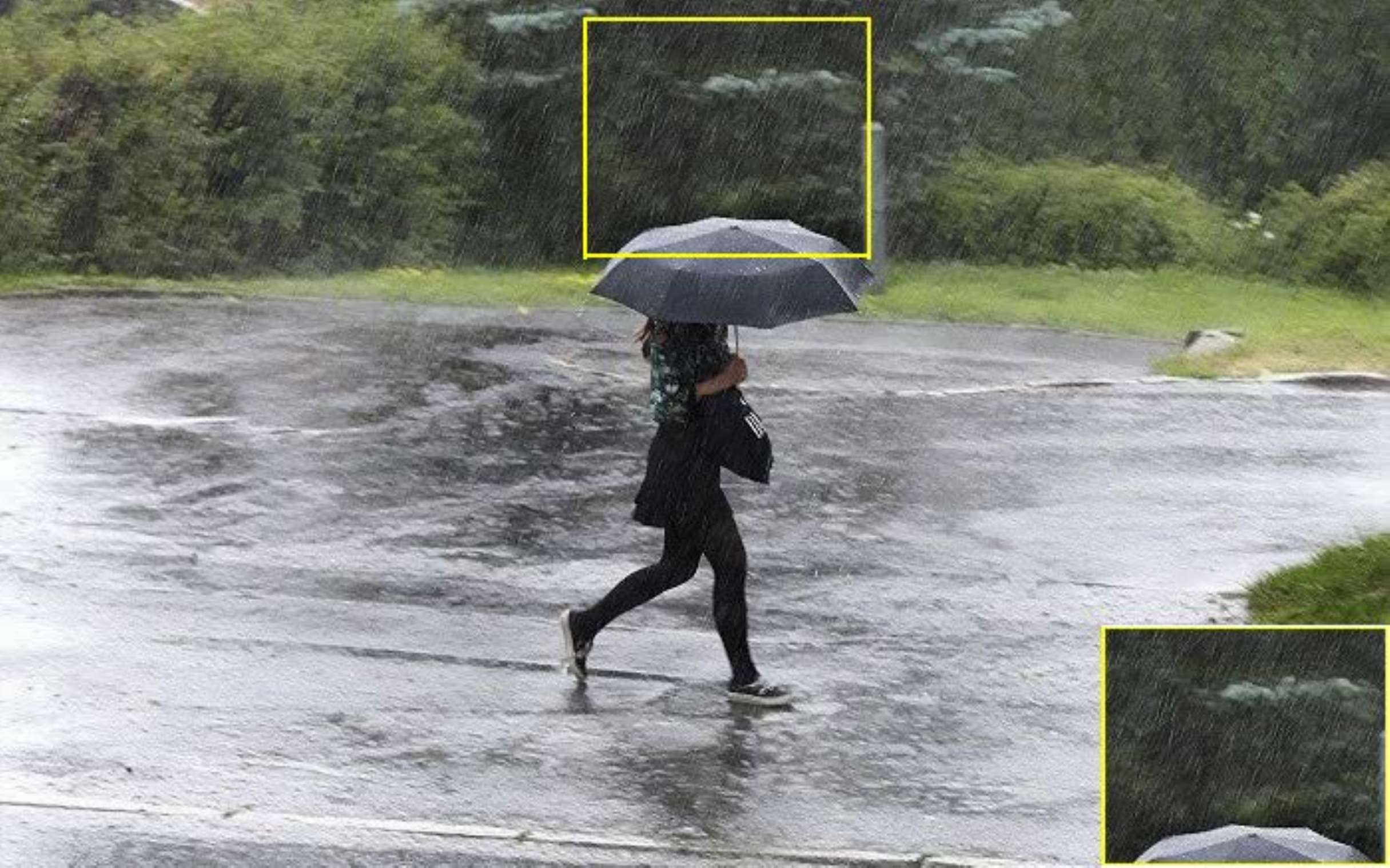}&\hspace{-4mm}
			\includegraphics[width = 0.12\linewidth]{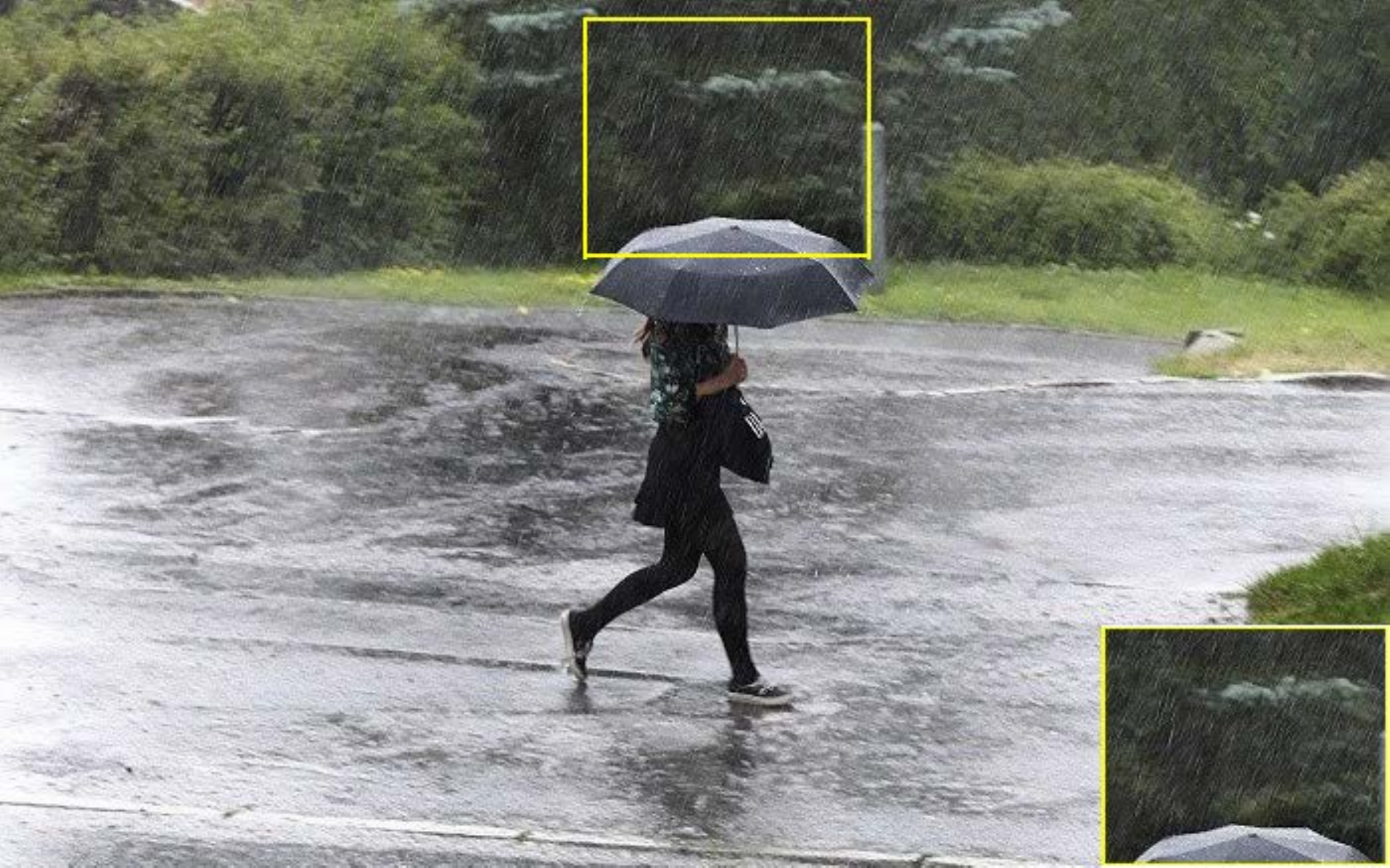}&\hspace{-4mm}
			\includegraphics[width = 0.12\linewidth]{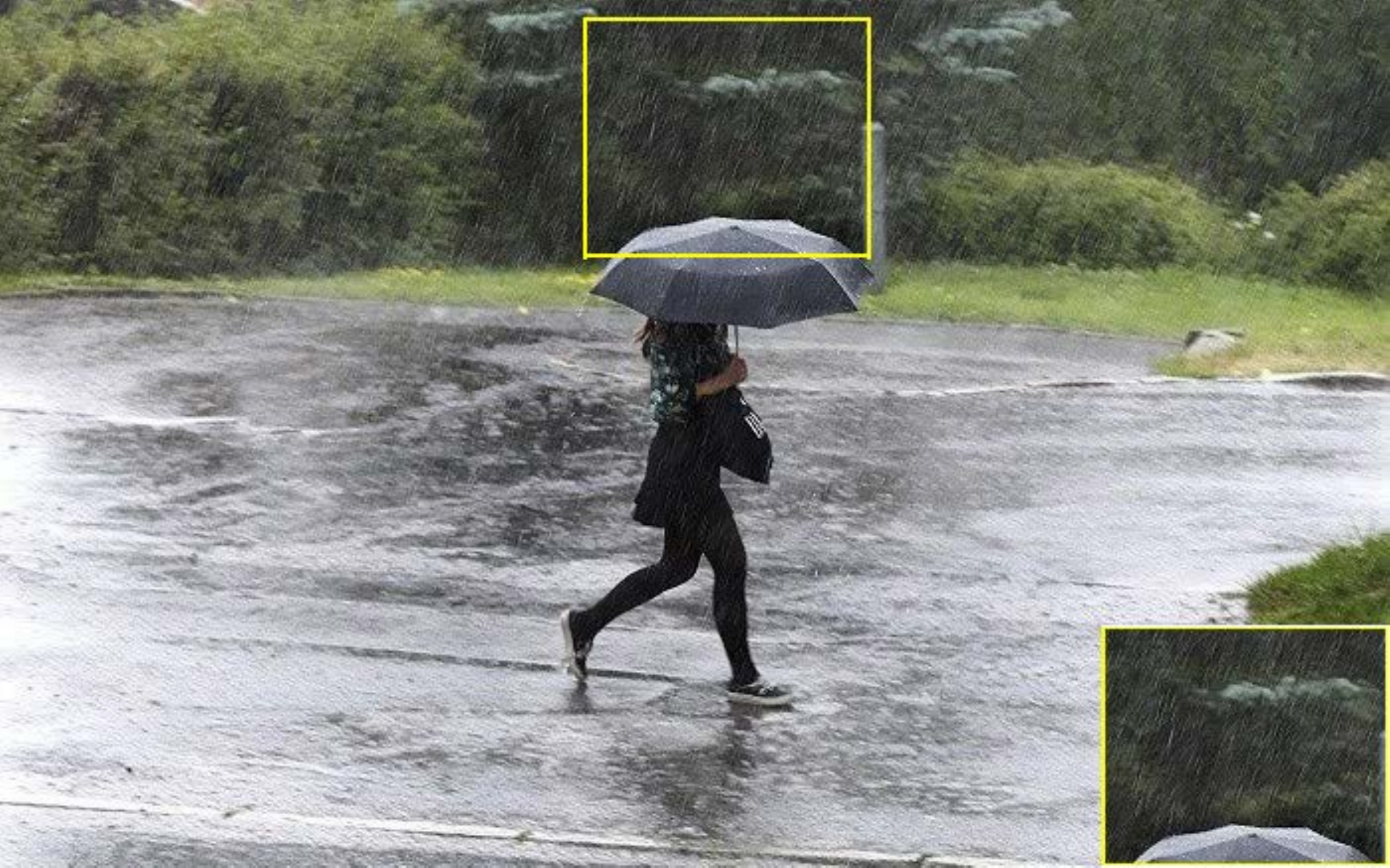}&\hspace{-4mm}
			\includegraphics[width = 0.12\linewidth]{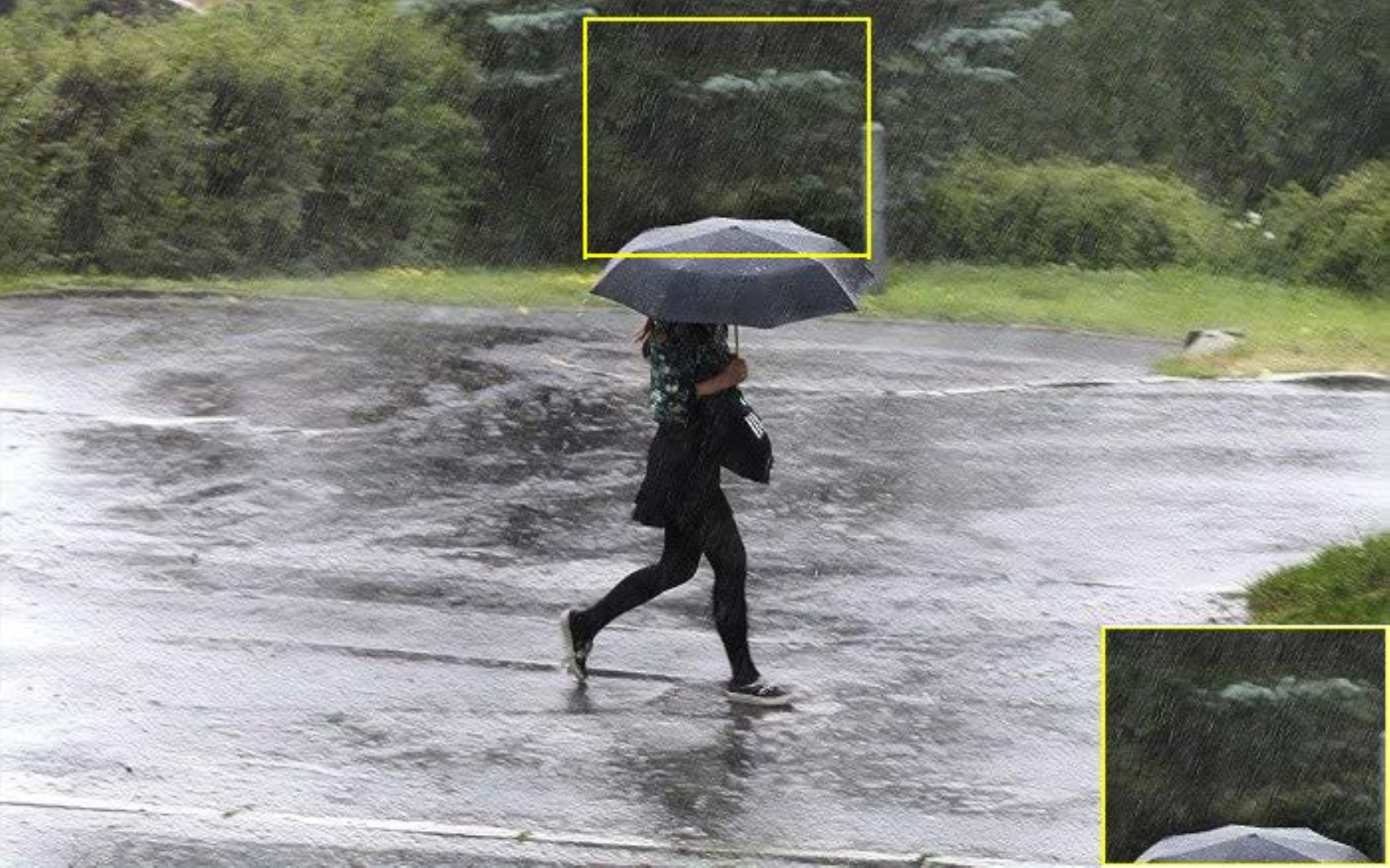}&\hspace{-4mm}
			\includegraphics[width = 0.12\linewidth]{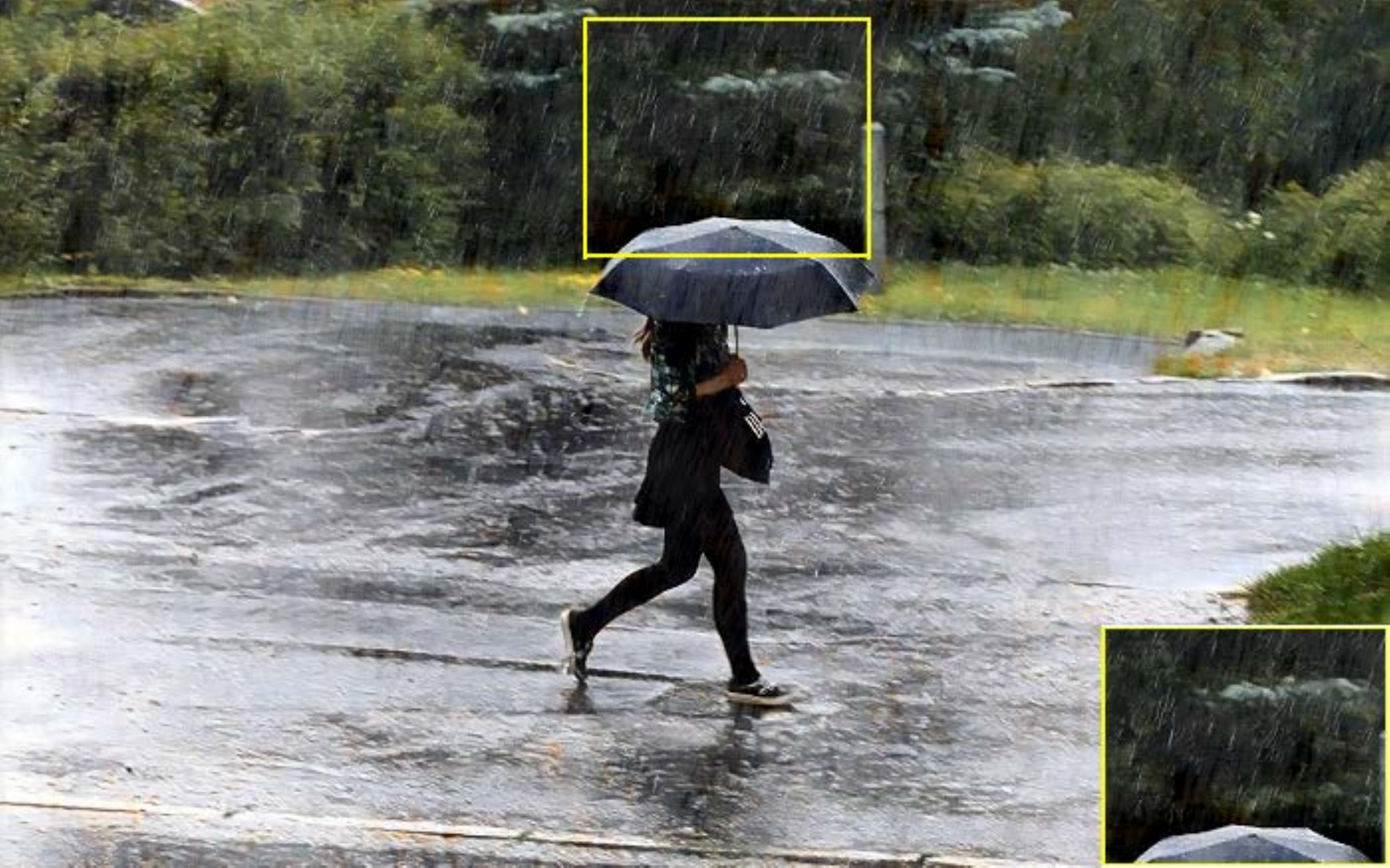}&\hspace{-4mm}
			\includegraphics[width = 0.12\linewidth]{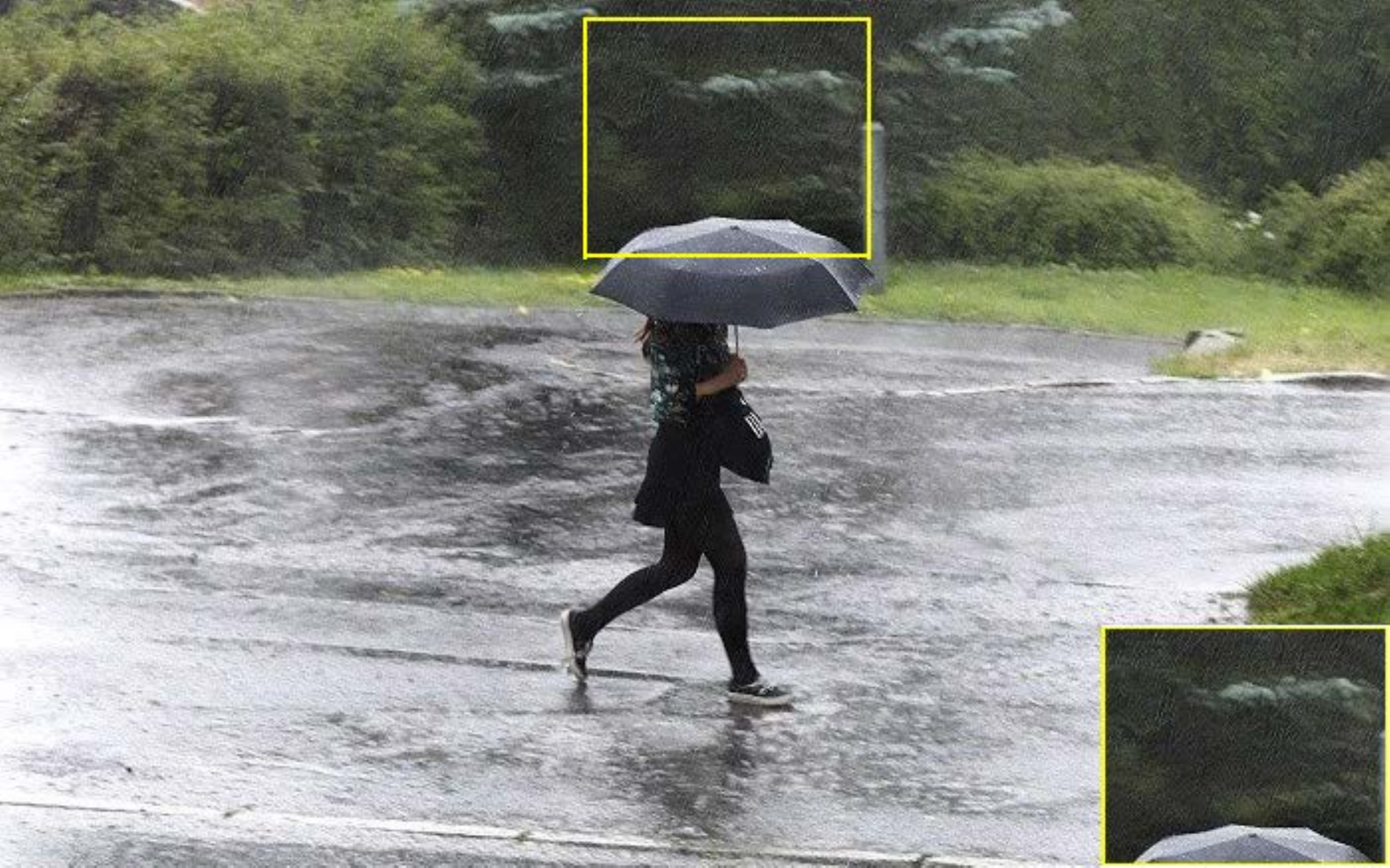}
			\\
			\includegraphics[width = 0.12\linewidth]{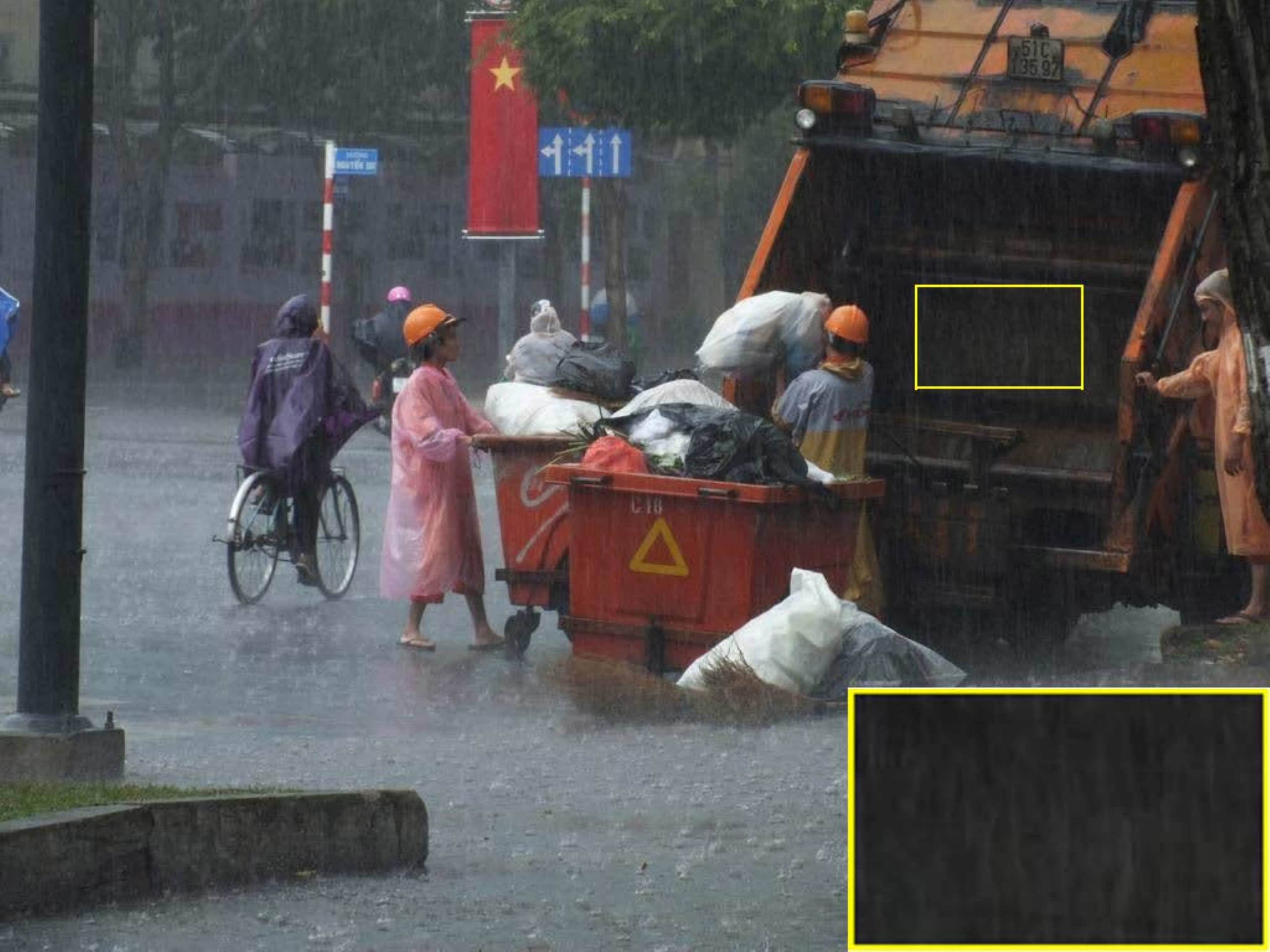}&\hspace{-4mm}
			\includegraphics[width = 0.12\linewidth]{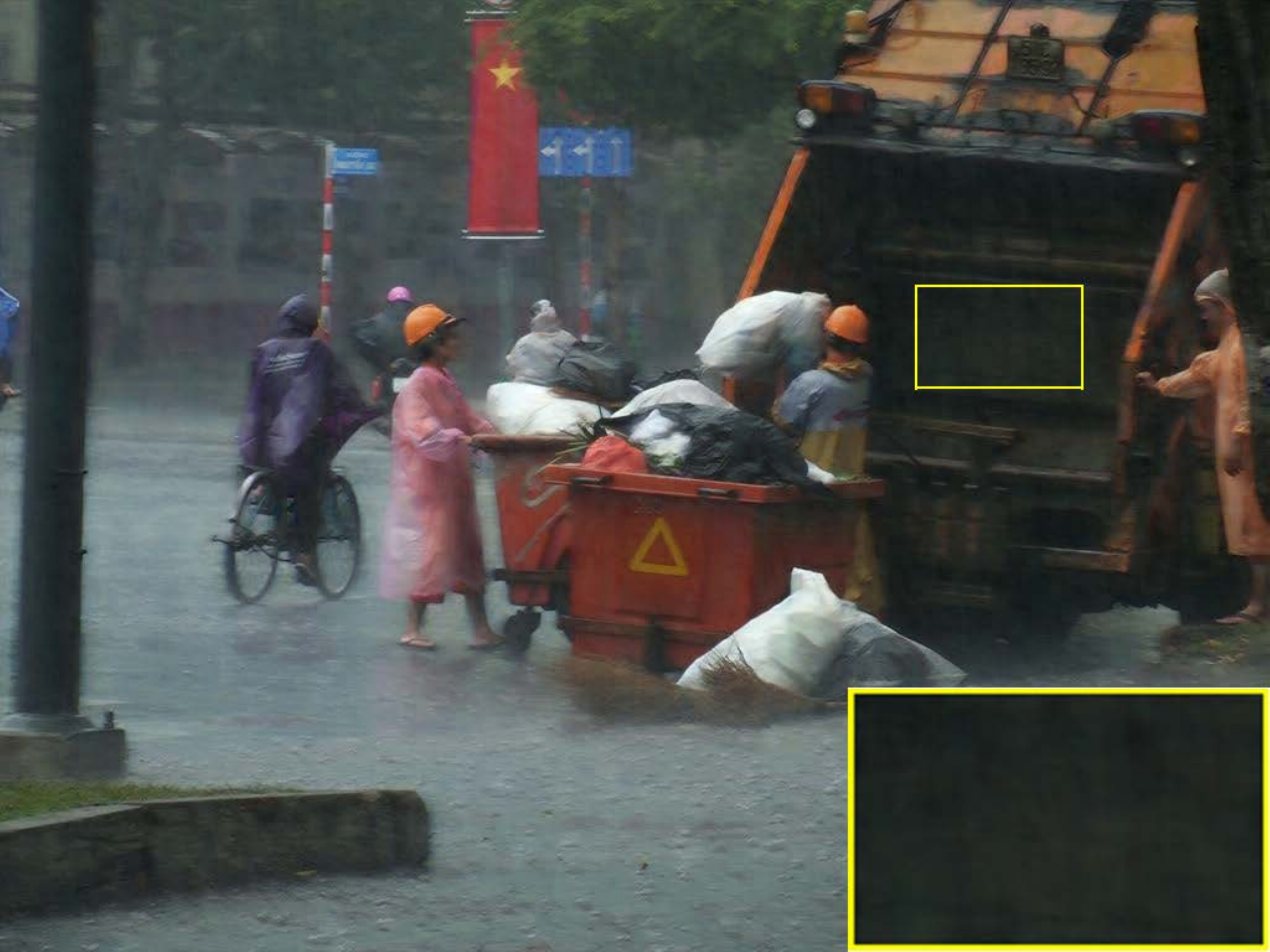}&\hspace{-4mm}
			\includegraphics[width = 0.12\linewidth]{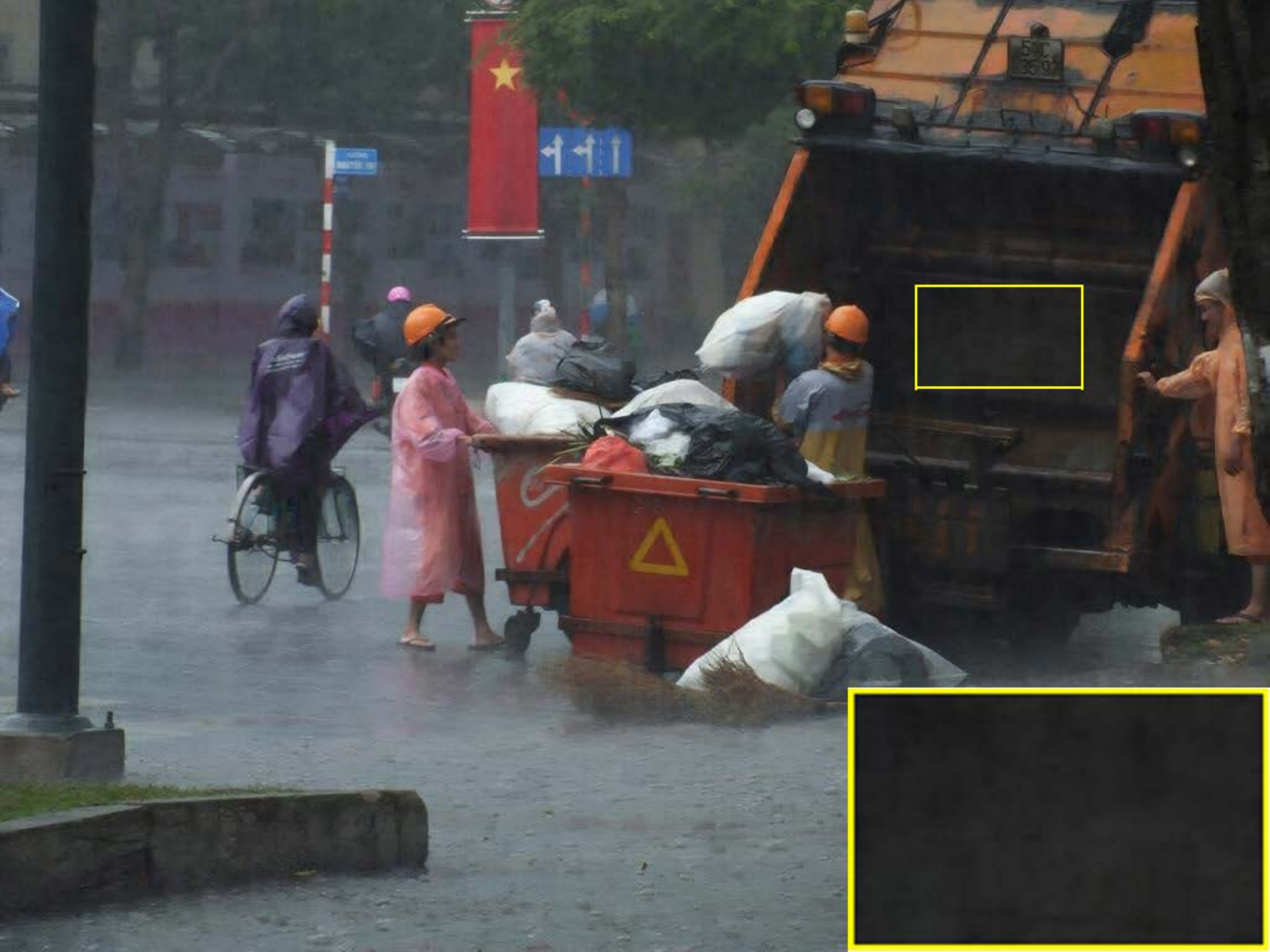}&\hspace{-4mm}
			\includegraphics[width = 0.12\linewidth]{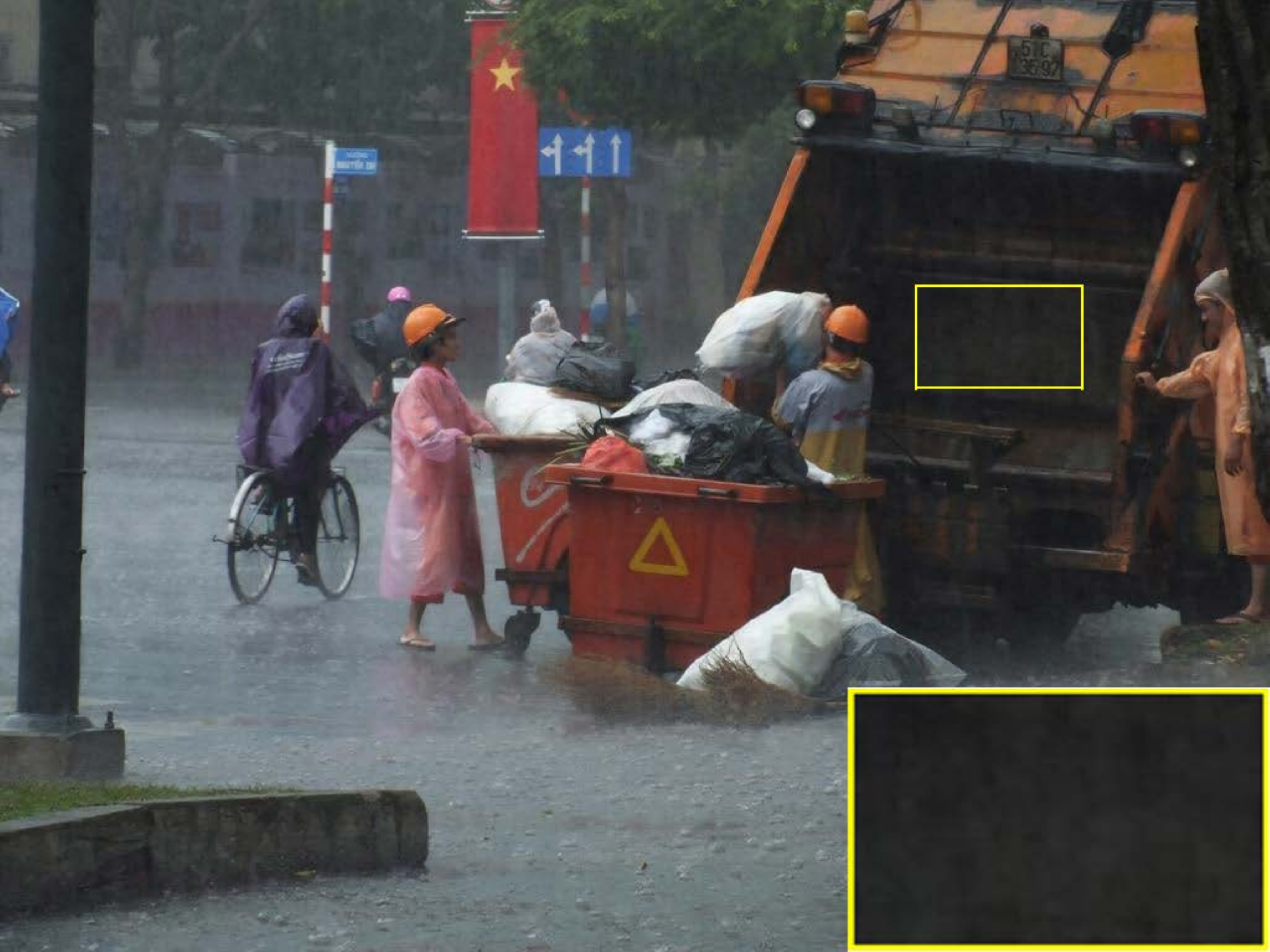}&\hspace{-4mm}
			\includegraphics[width = 0.12\linewidth]{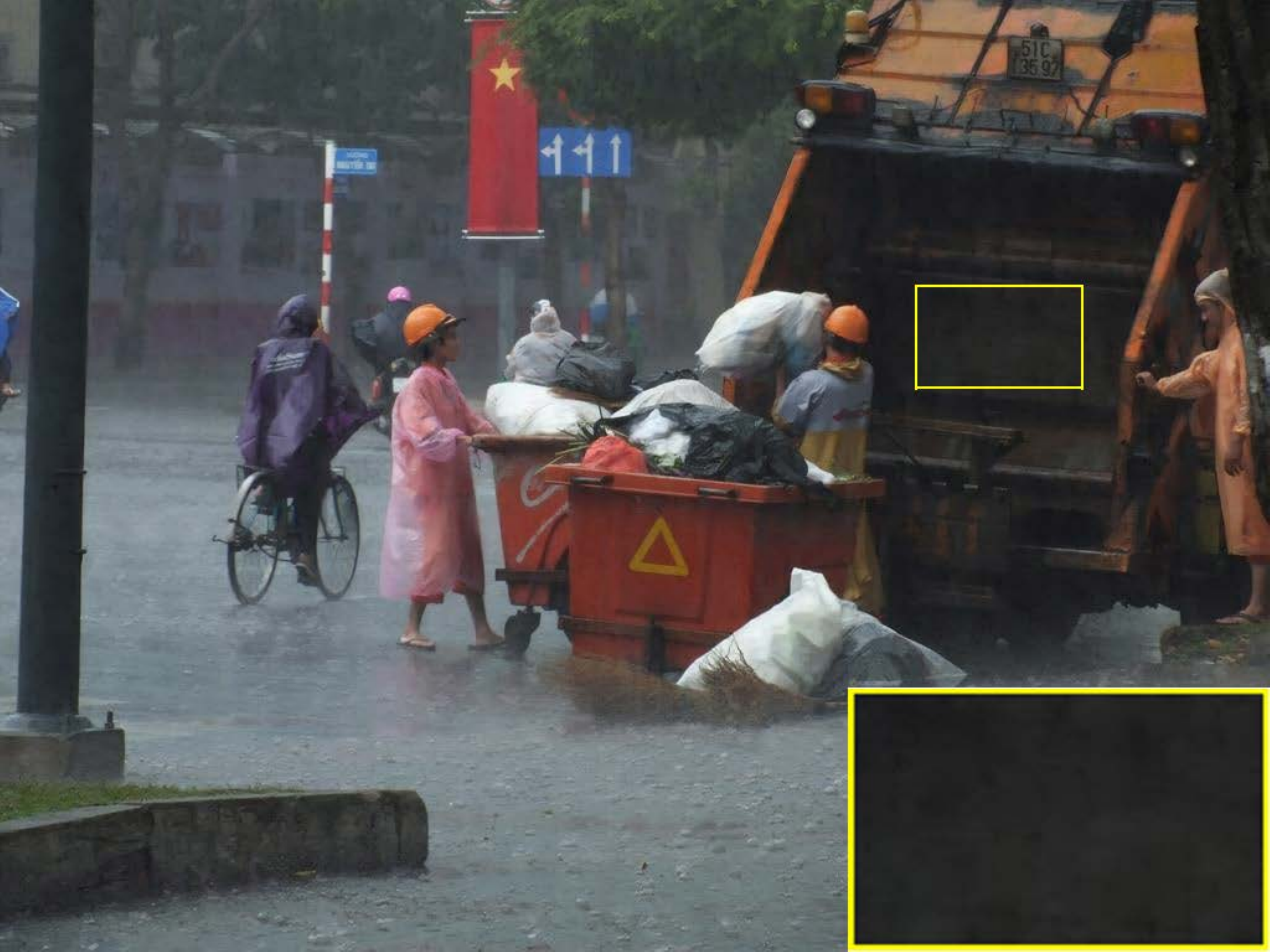}&\hspace{-4mm}
			\includegraphics[width = 0.12\linewidth]{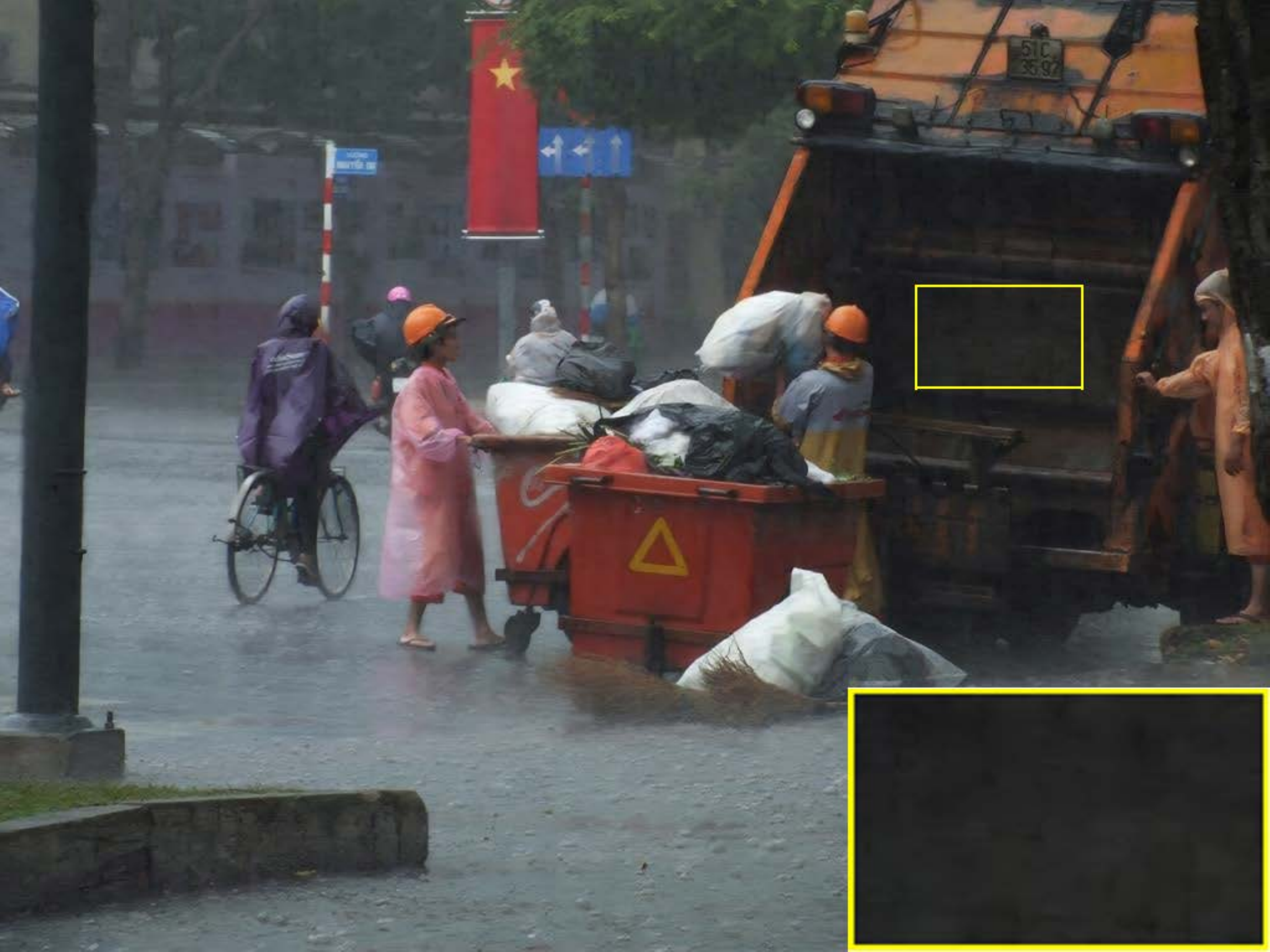}&\hspace{-4mm}
			\includegraphics[width = 0.12\linewidth]{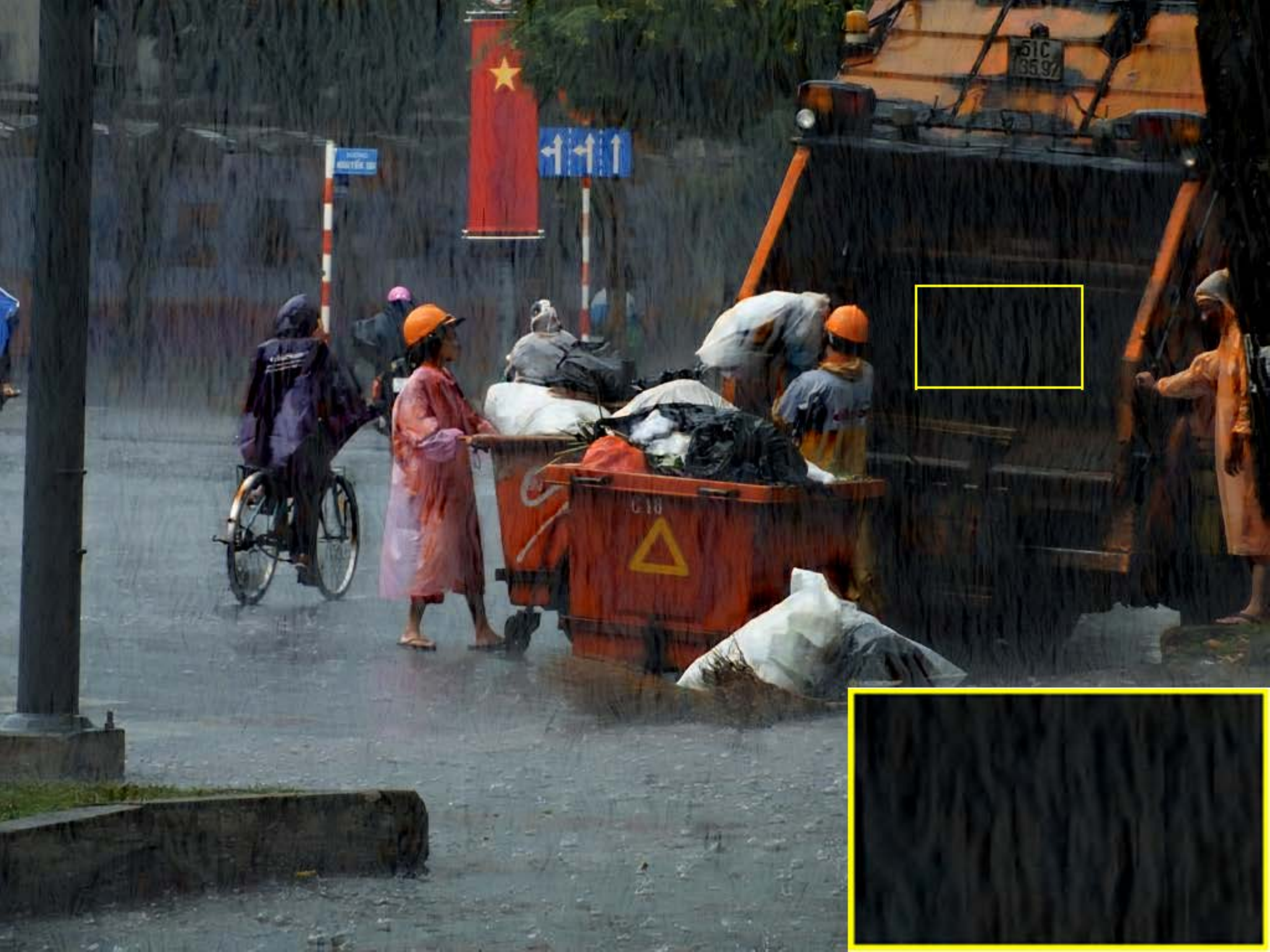}&\hspace{-4mm}
			\includegraphics[width = 0.12\linewidth]{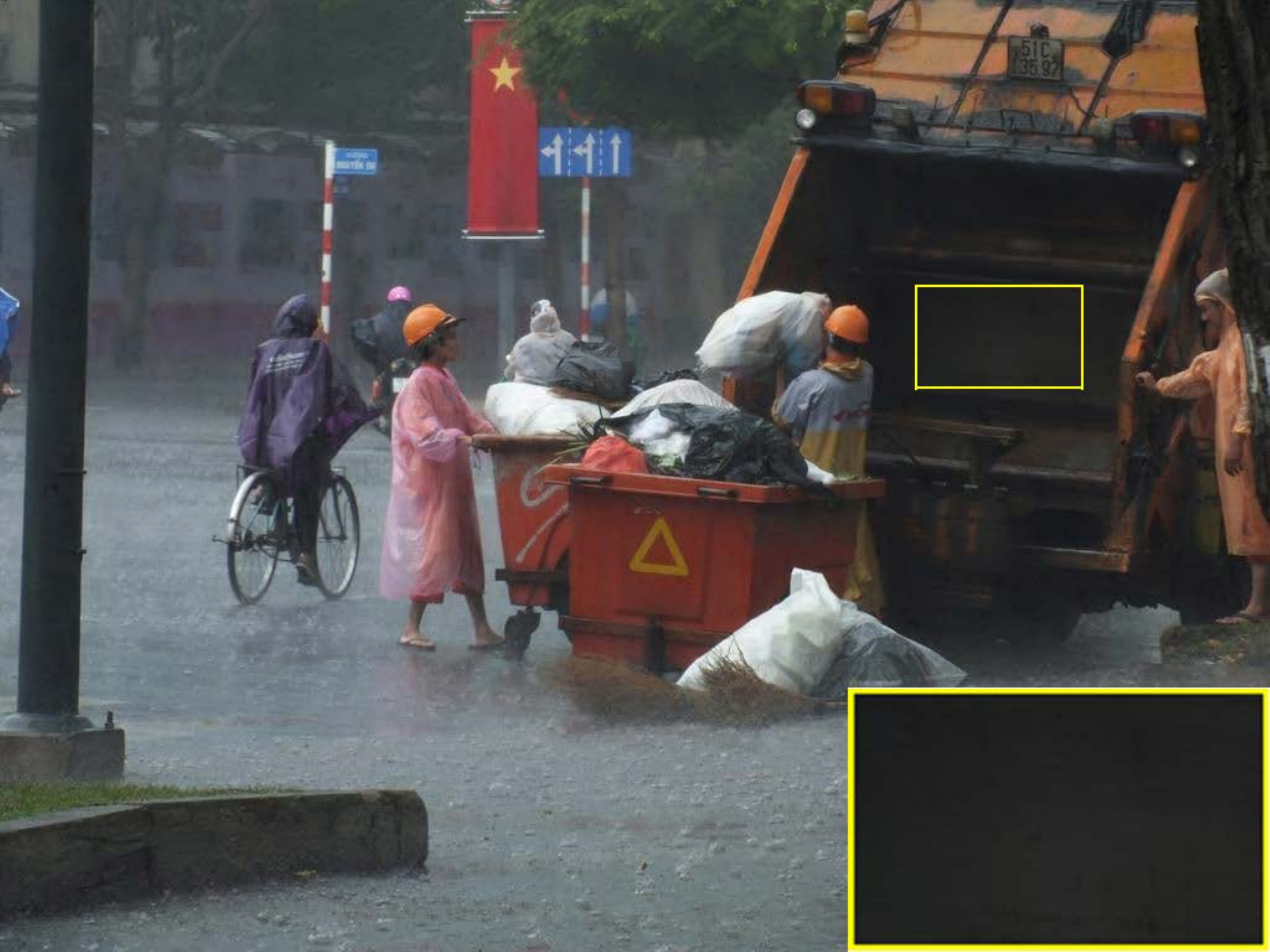}
			\\
			\includegraphics[width = 0.12\linewidth]{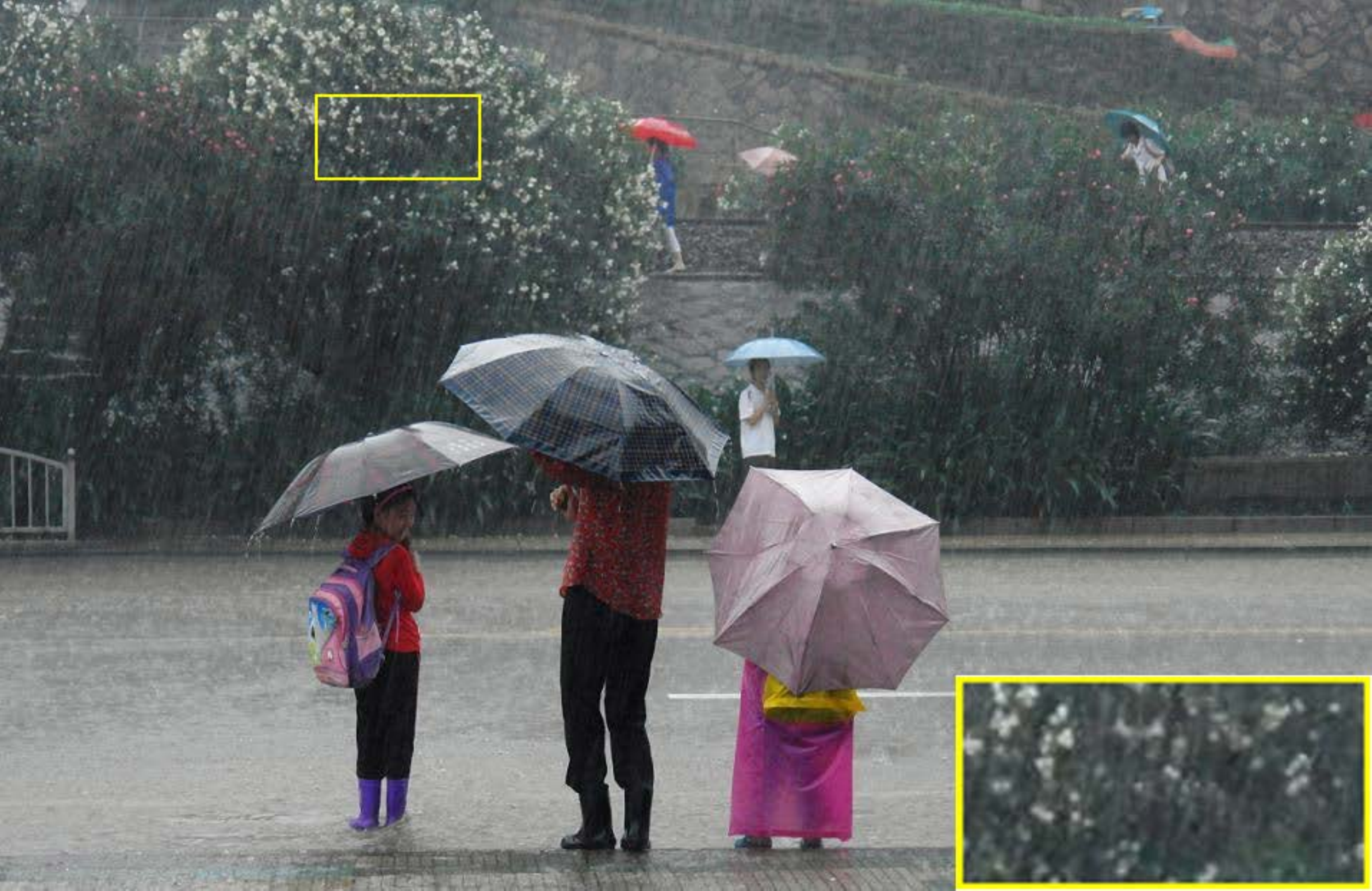}&\hspace{-4mm}
			\includegraphics[width = 0.12\linewidth]{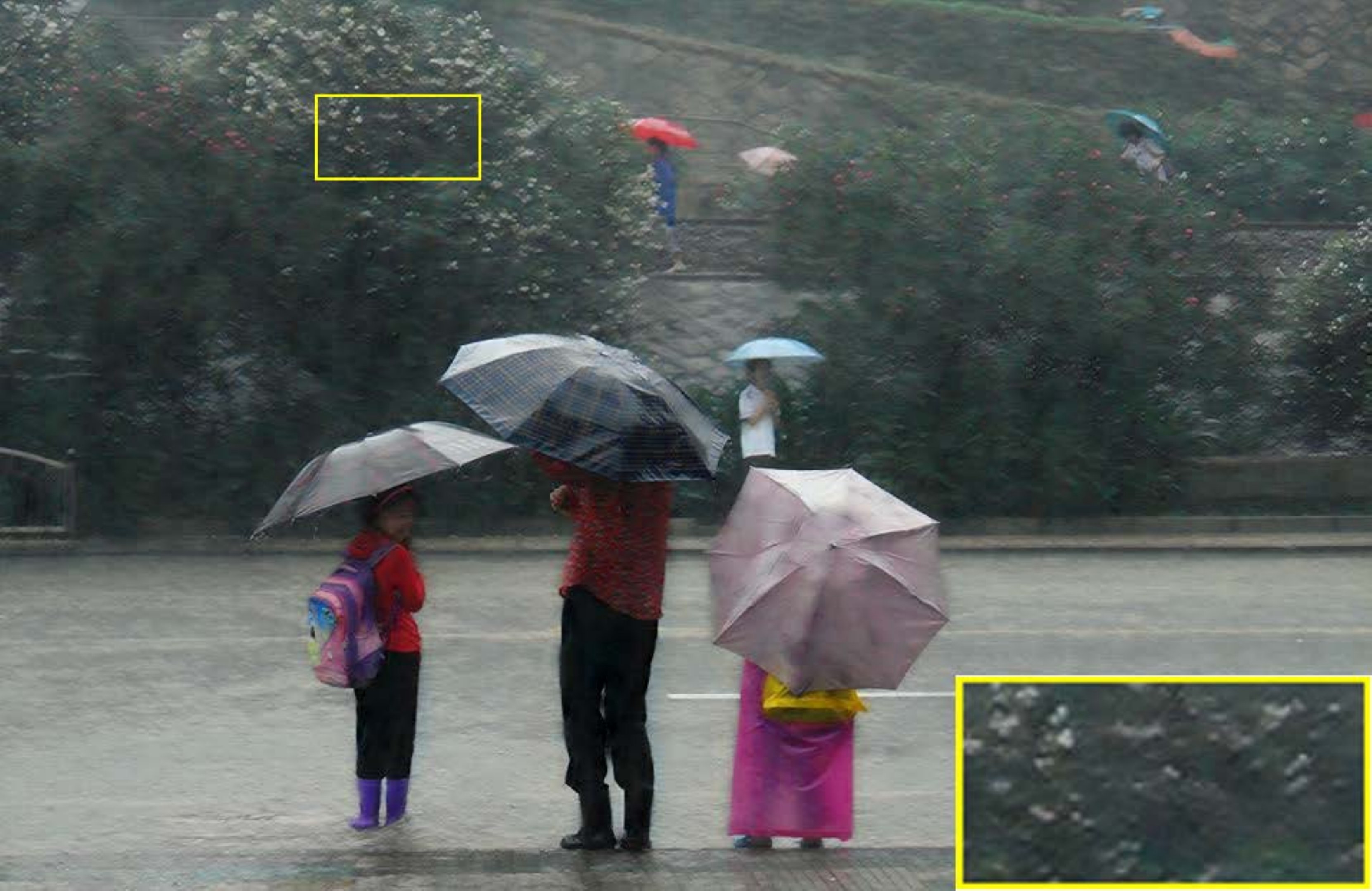}&\hspace{-4mm}
			\includegraphics[width = 0.12\linewidth]{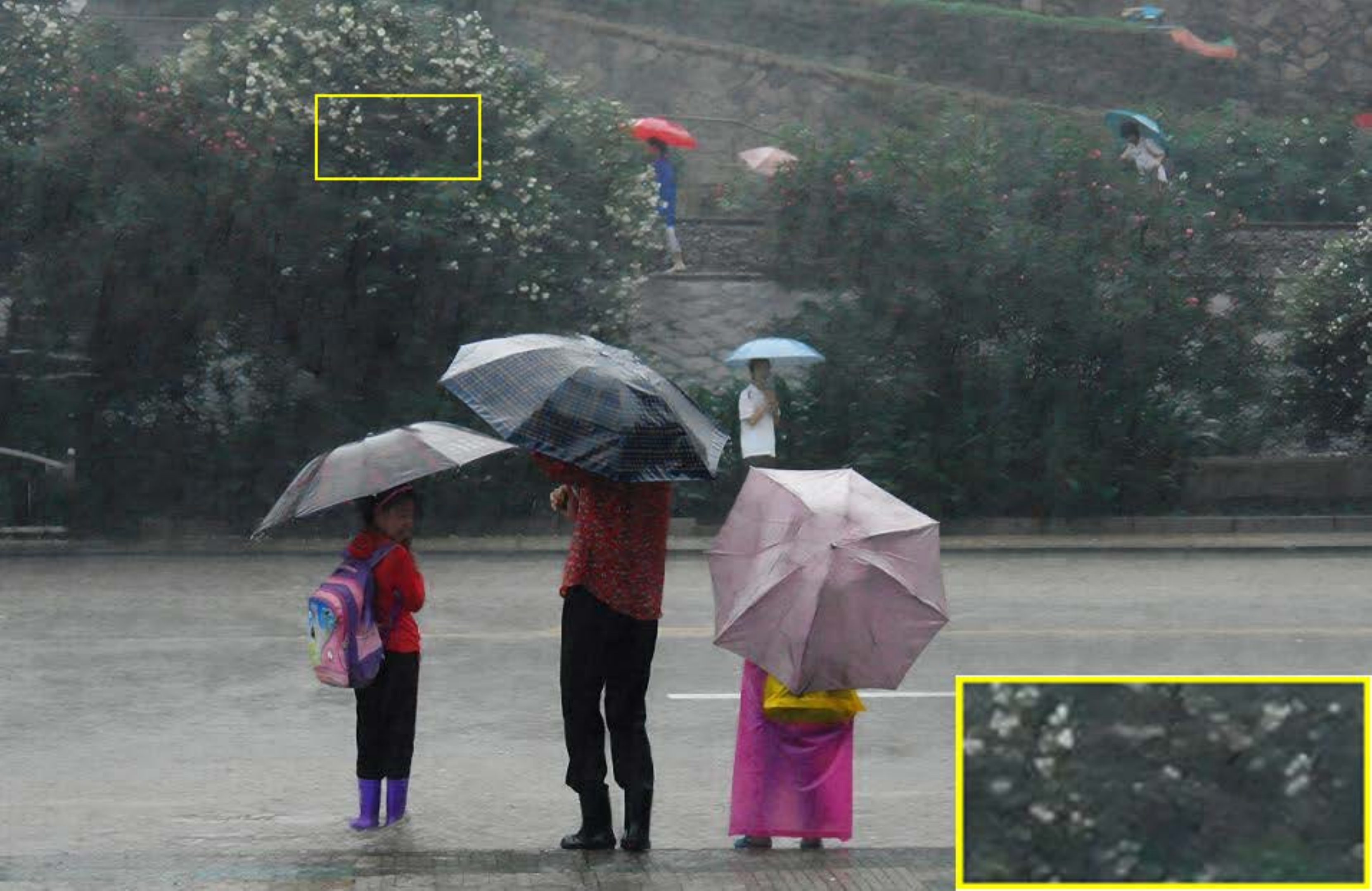}&\hspace{-4mm}
			\includegraphics[width = 0.12\linewidth]{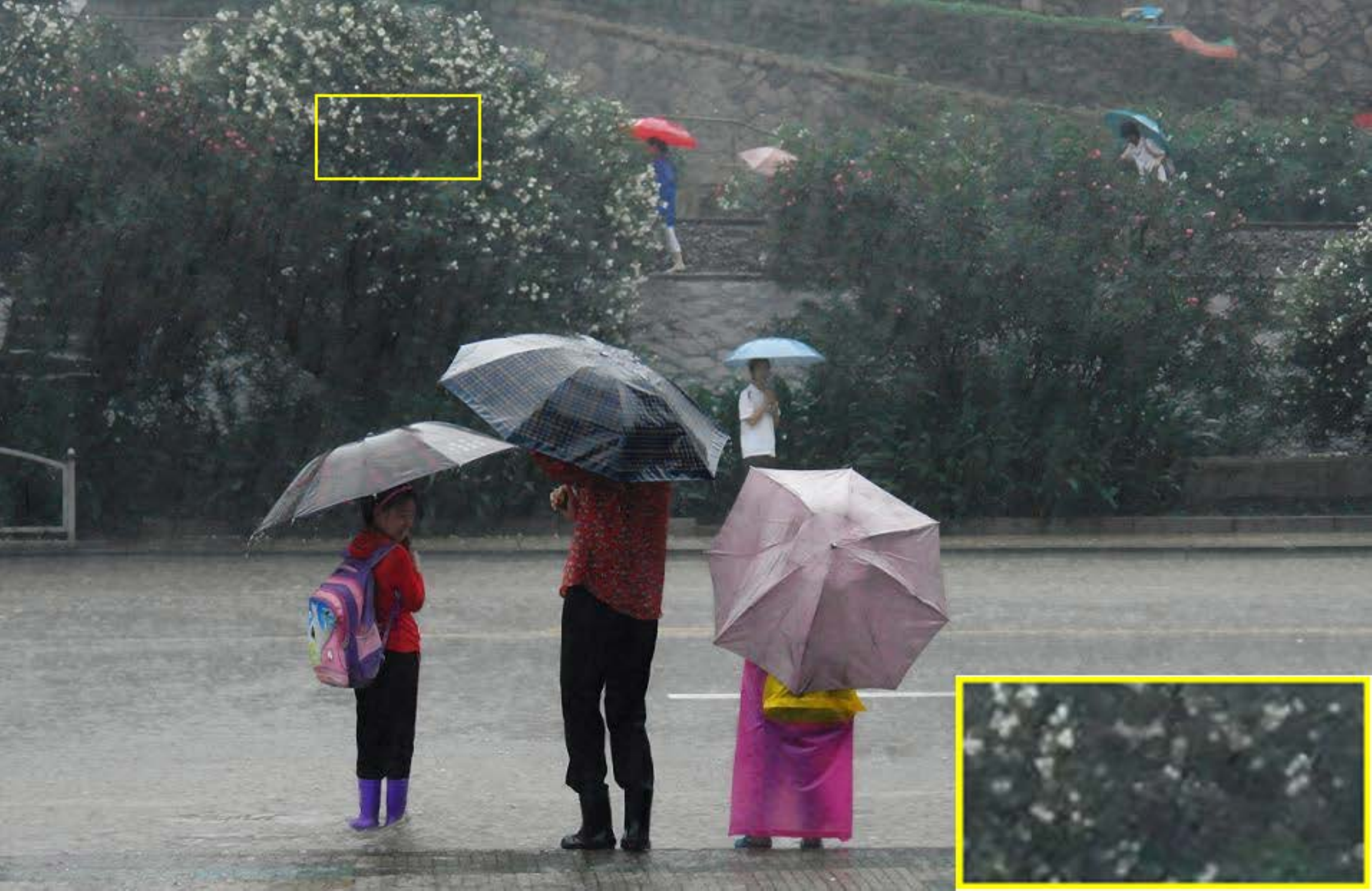}&\hspace{-4mm}
			\includegraphics[width = 0.12\linewidth]{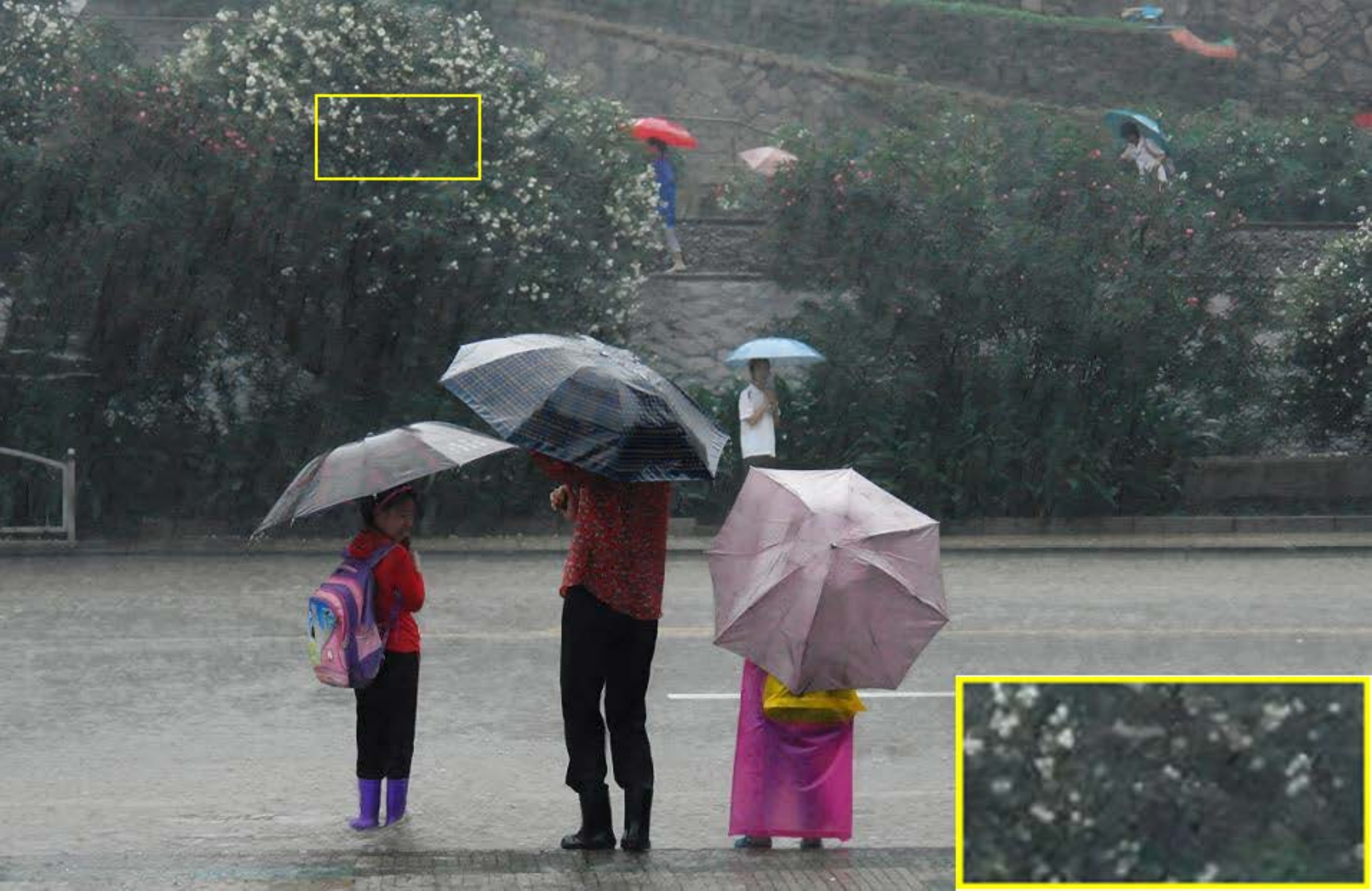}&\hspace{-4mm}
			\includegraphics[width = 0.12\linewidth]{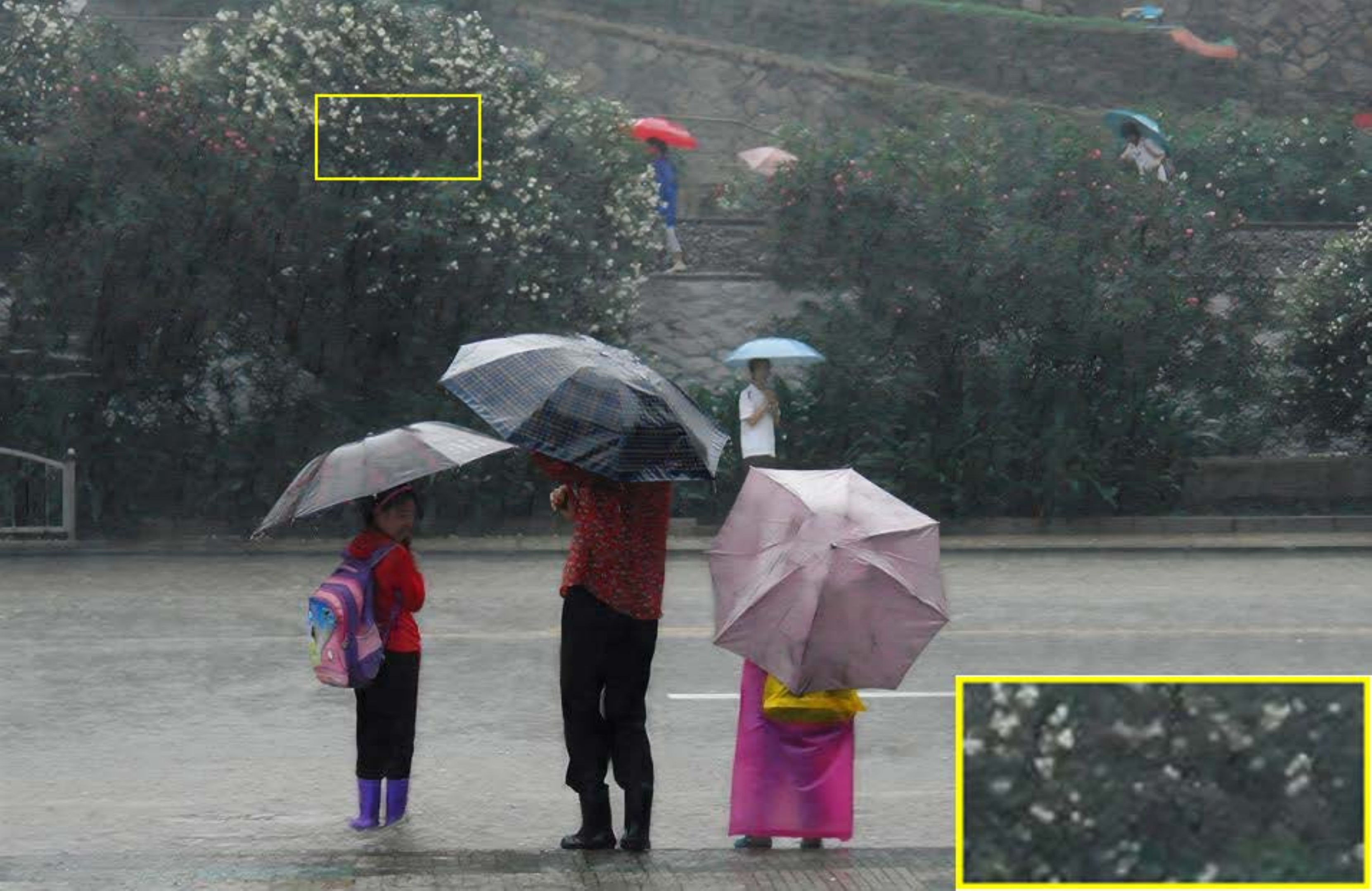}&\hspace{-4mm}
			\includegraphics[width = 0.12\linewidth]{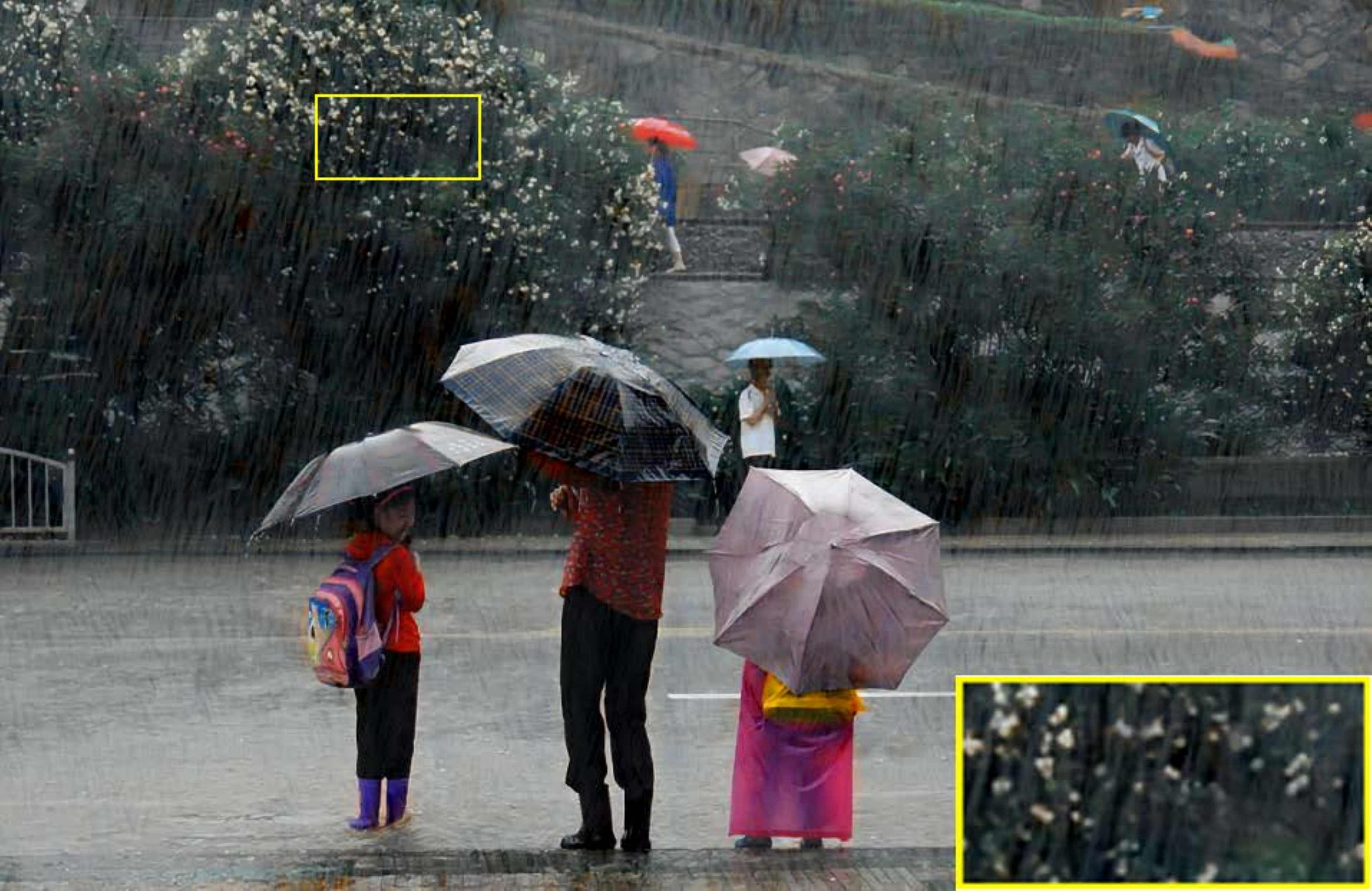}&\hspace{-4mm}
			\includegraphics[width = 0.12\linewidth]{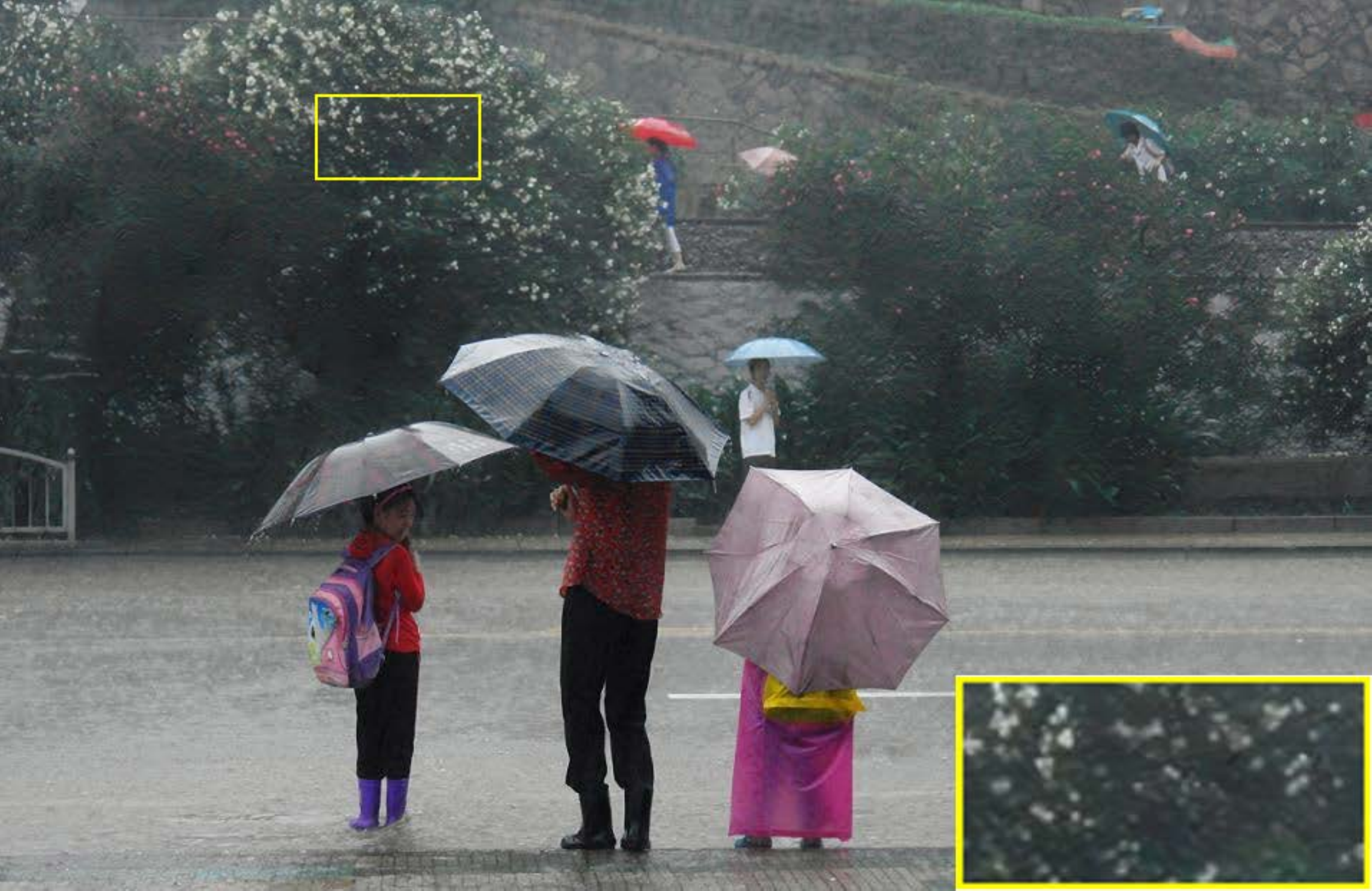}
			\\
			\includegraphics[width = 0.12\linewidth]{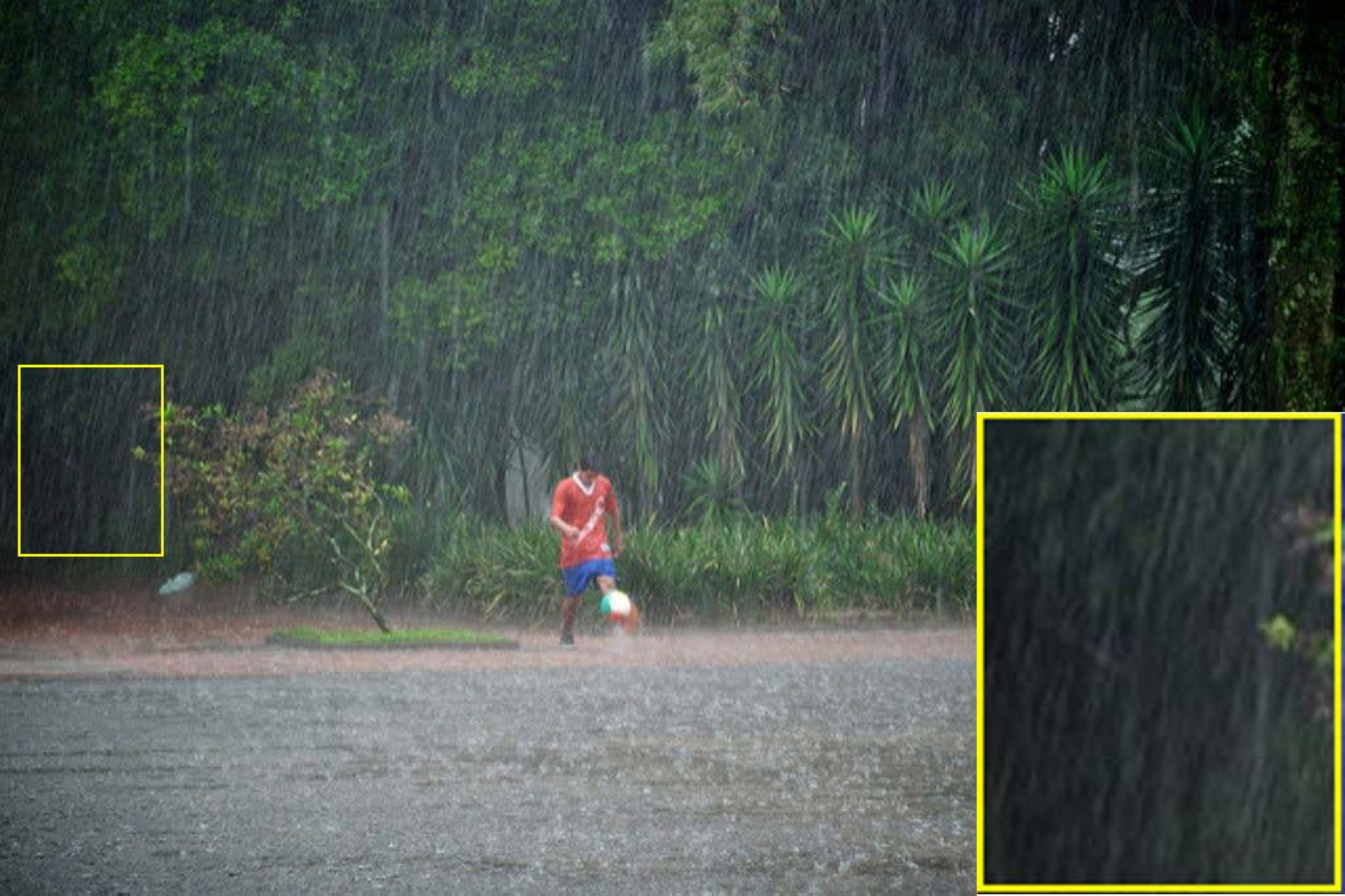}&\hspace{-4mm}
			\includegraphics[width = 0.12\linewidth]{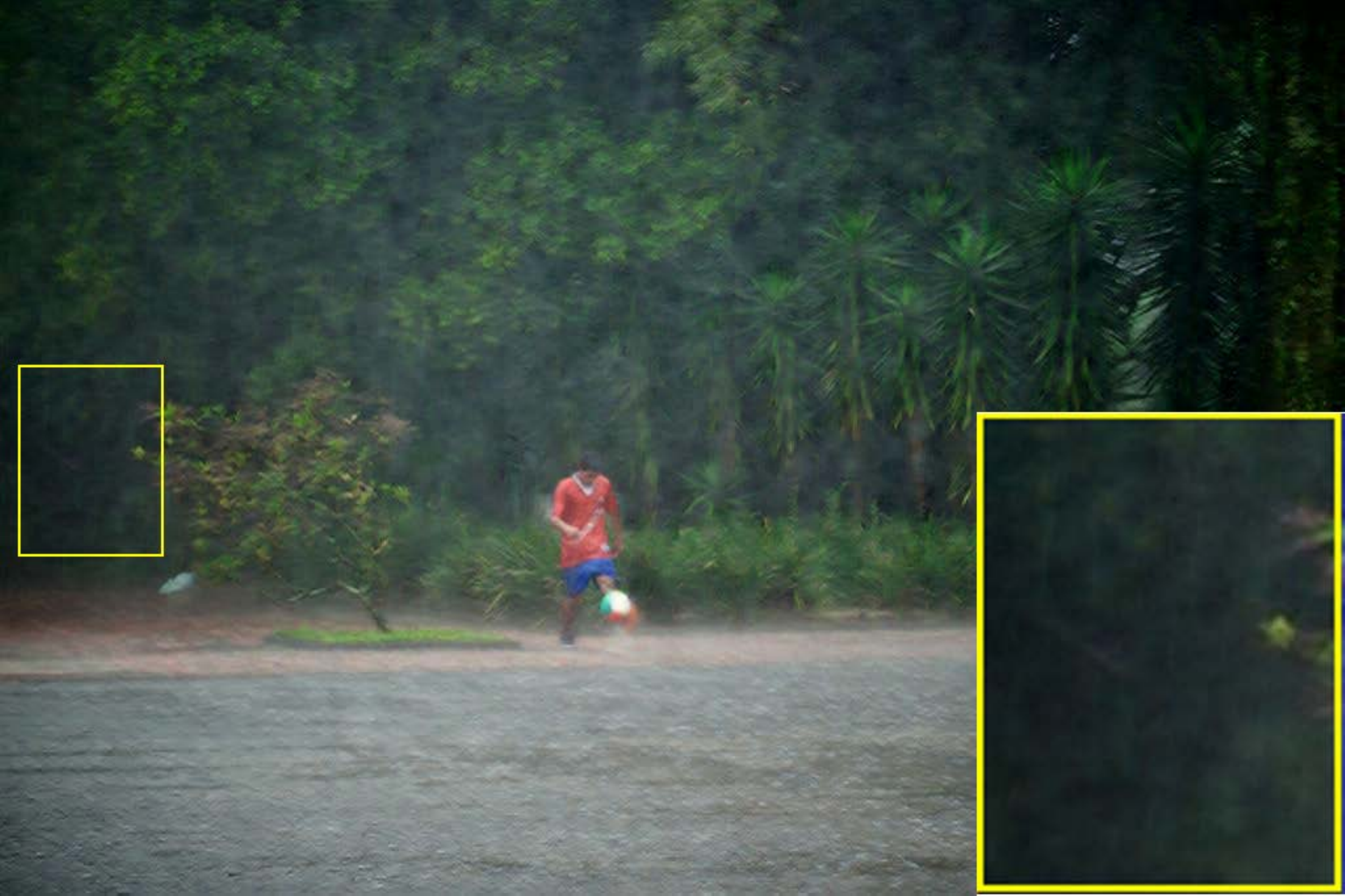}&\hspace{-4mm}
			\includegraphics[width = 0.12\linewidth]{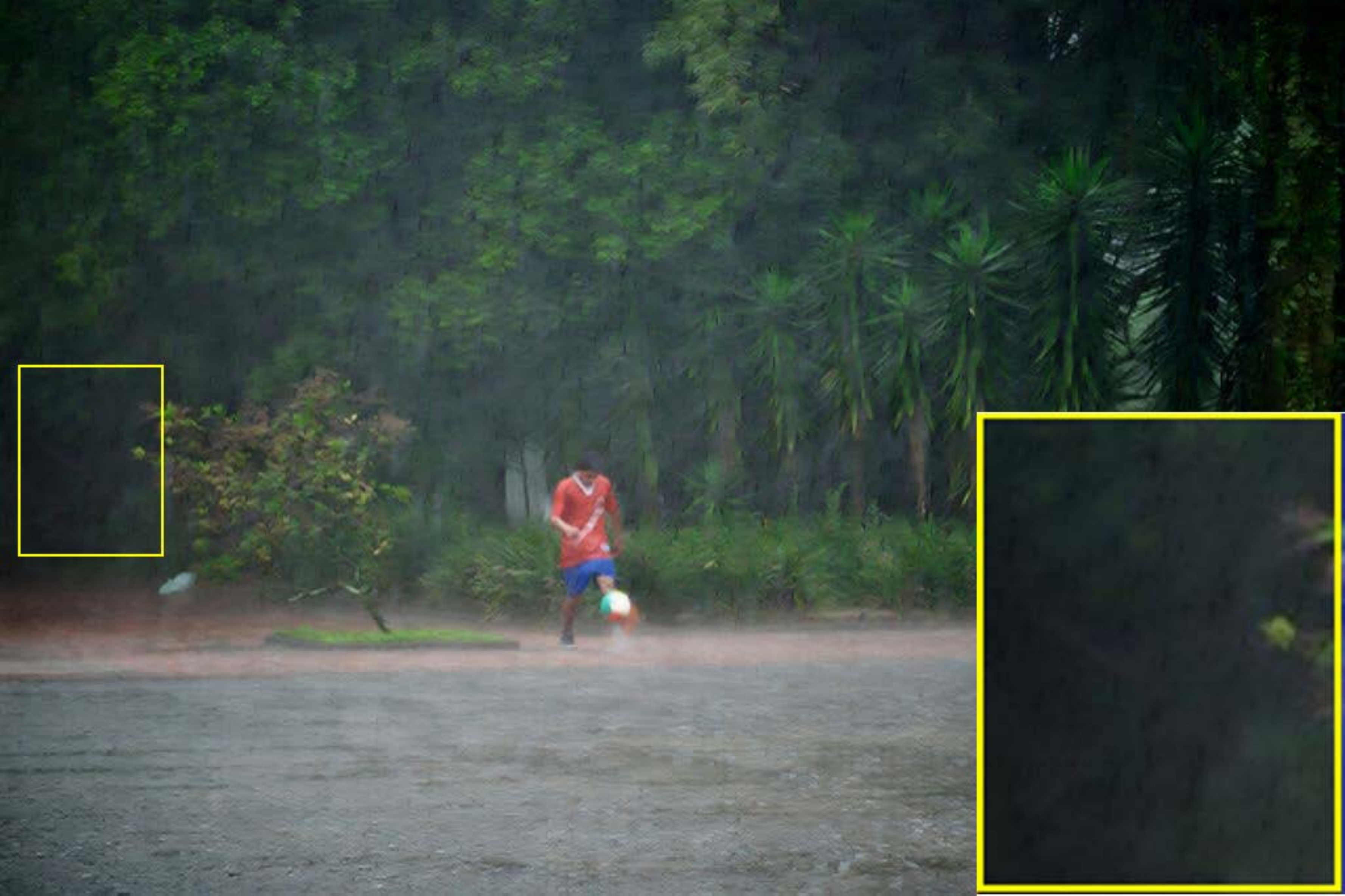}&\hspace{-4mm}
			\includegraphics[width = 0.12\linewidth]{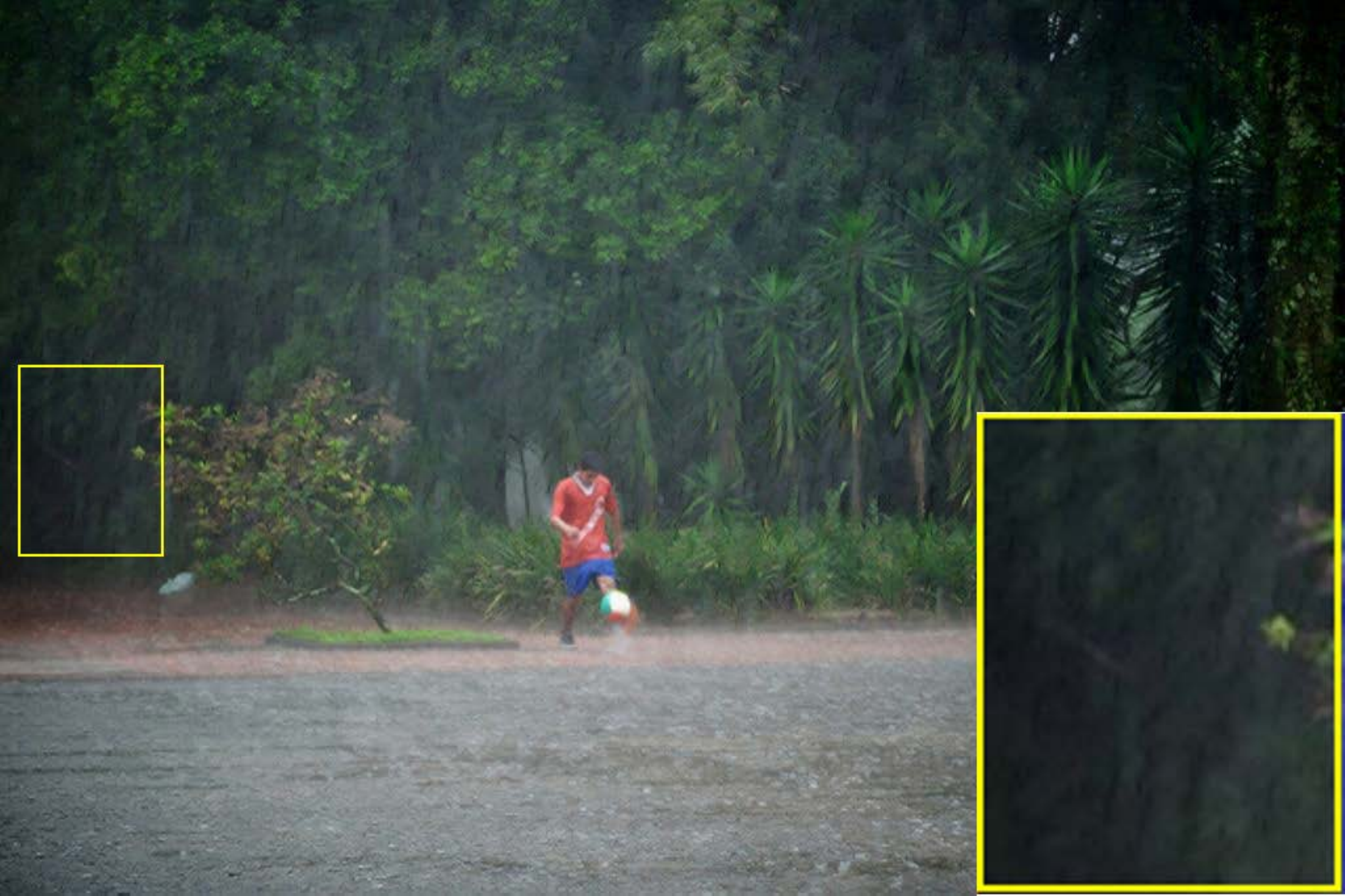}&\hspace{-4mm}
			\includegraphics[width = 0.12\linewidth]{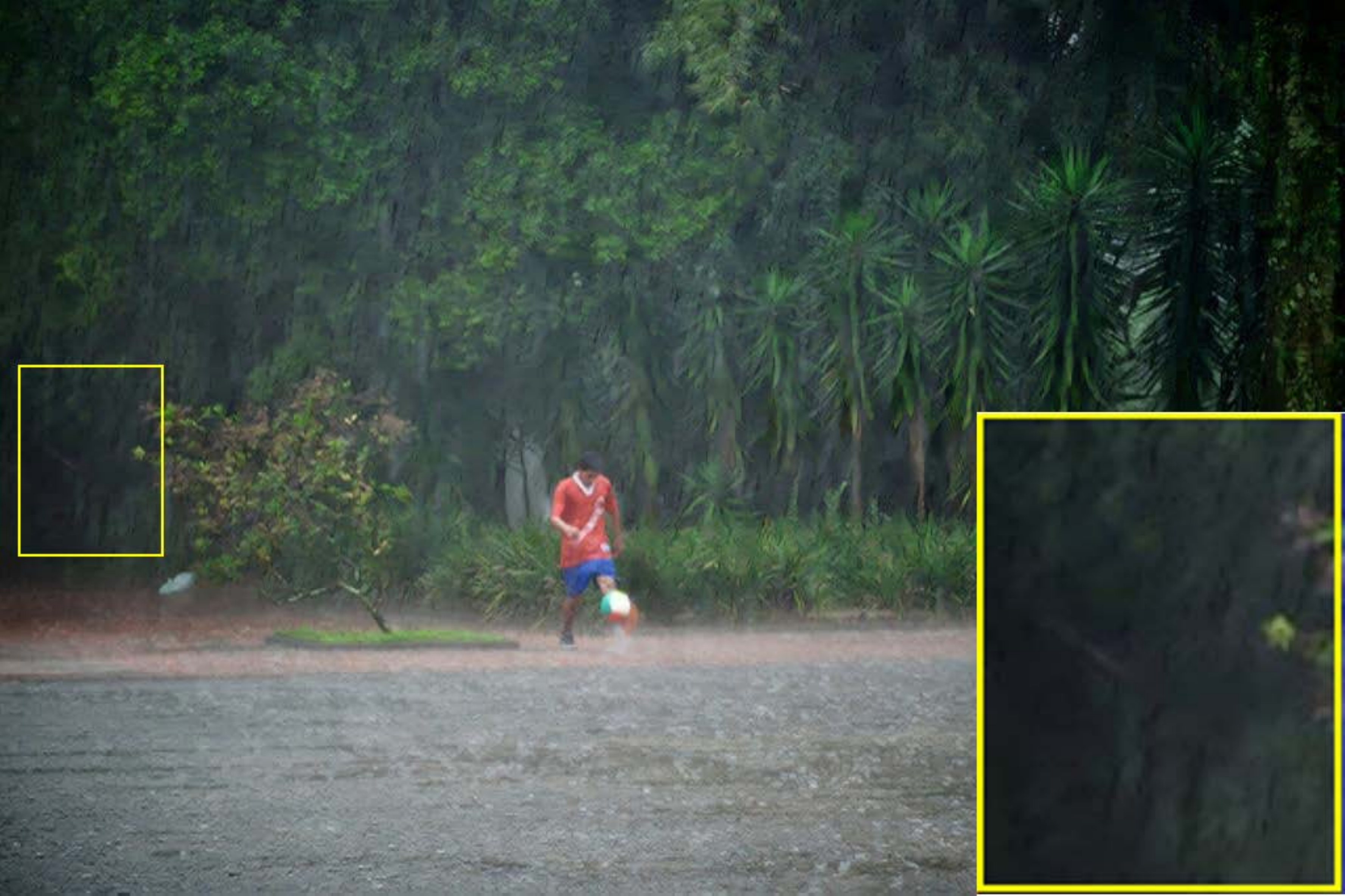}&\hspace{-4mm}
			\includegraphics[width = 0.12\linewidth]{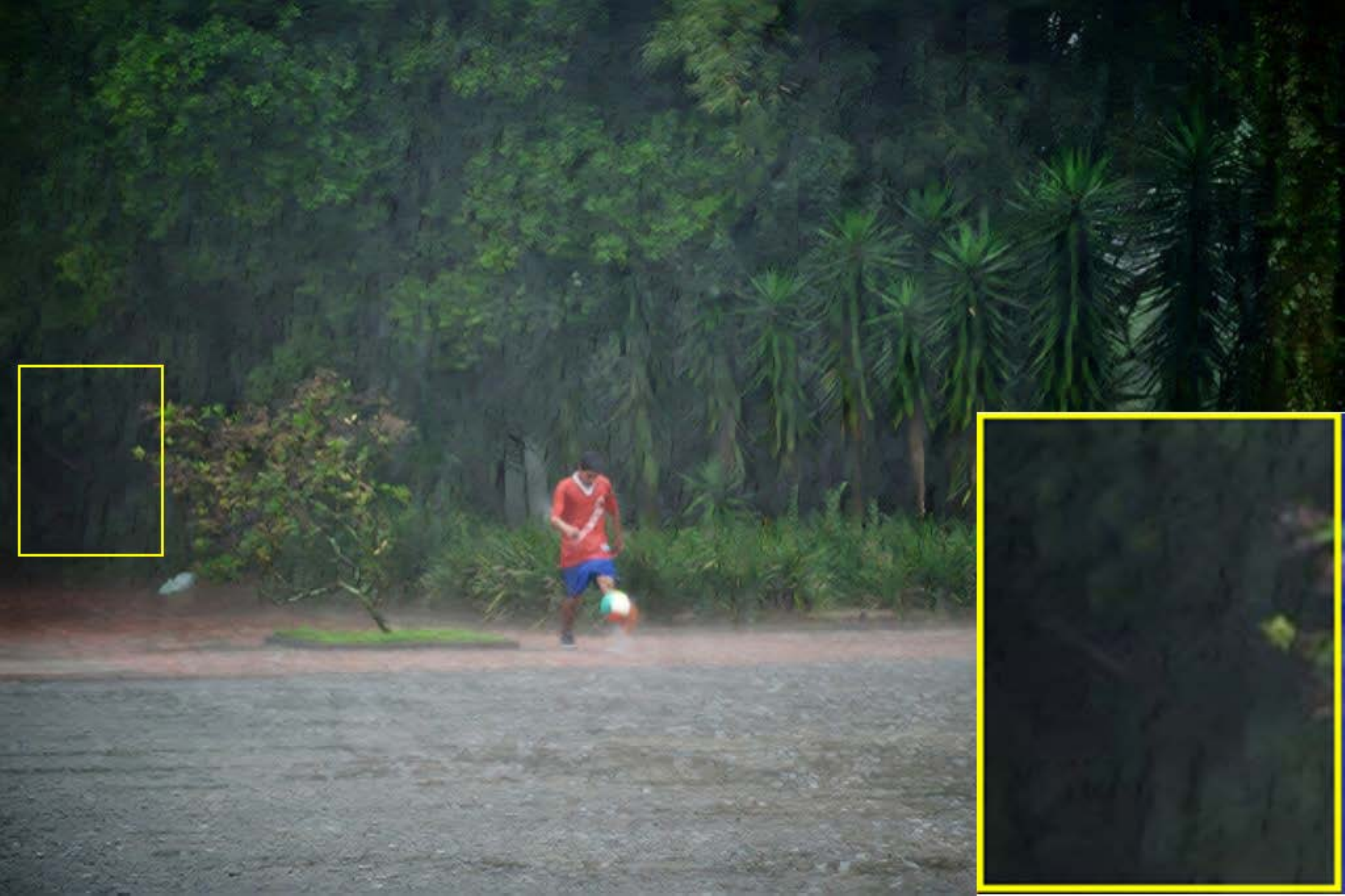}&\hspace{-4mm}
			\includegraphics[width = 0.12\linewidth]{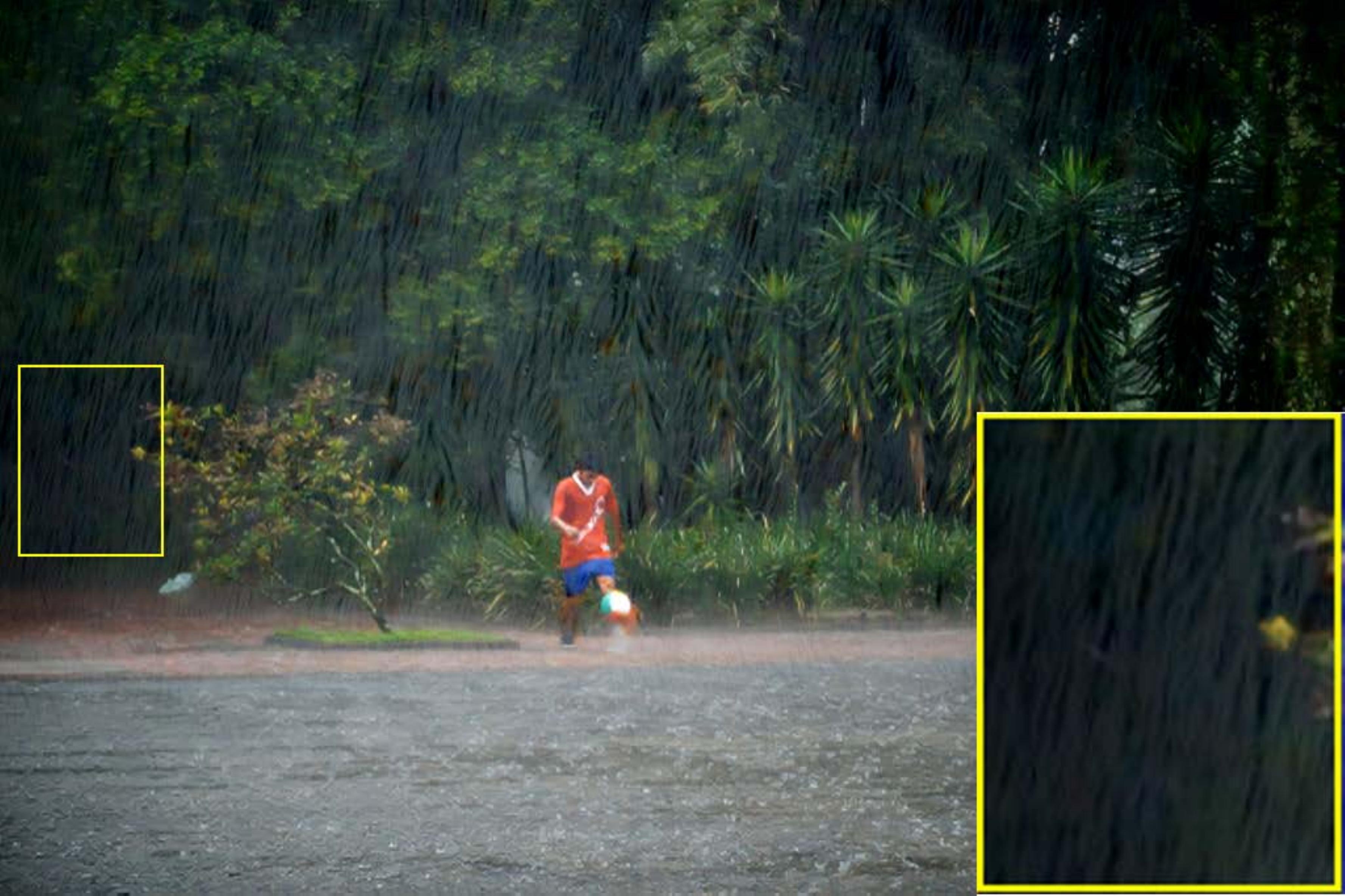}&\hspace{-4mm}
			\includegraphics[width = 0.12\linewidth]{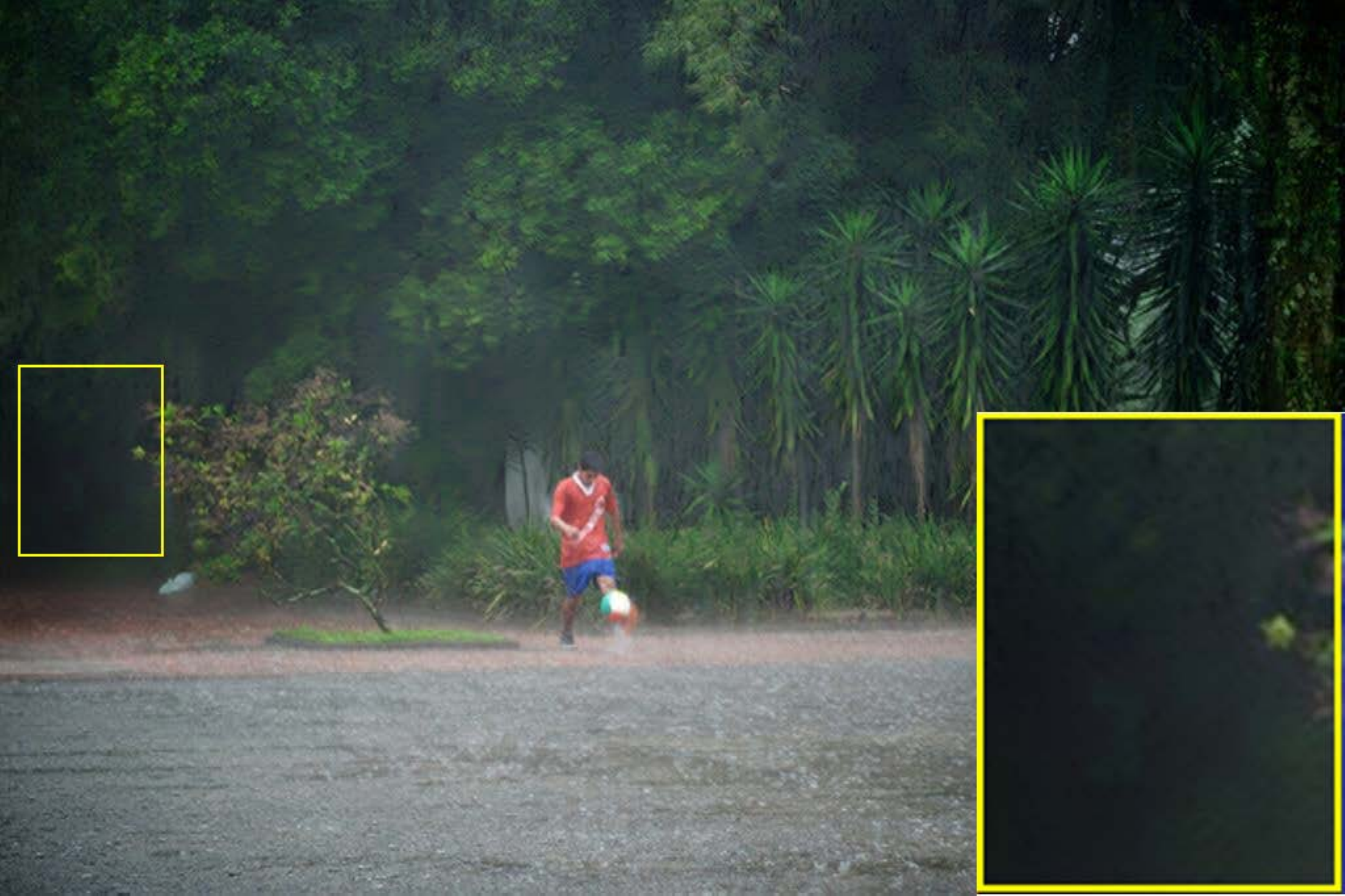}\\
			(a) Input&\hspace{-4mm} (b) DDN&\hspace{-4mm} (c) RESCAN&\hspace{-4mm} (d) REHEN&\hspace{-4mm} (e) PreNet&\hspace{-4mm} (f) SpaNet&\hspace{-4mm} (g) SSIR&\hspace{-4mm} (h) JDNet\\
		\end{tabular}
	\end{center}
	\caption{Examples about the comparison of our method with other methods on real-world datasets.}
	\label{fig:nature-compared}
\end{figure*}

\subsection{Loss Fuction}

SSIM(Structural Similarity)~\cite{ssim} is an image quality evaluation metric in the range of $[0,1]$.
It measures the similarity of two images from brightness, contrast, and structure. When the two images are identical, SSIM is 1.
The effectiveness of negative SSIM loss for image deraining has been confirmed in \cite{derain_prenet_Ren_2019_CVPR}.

Hence, we use negative $SSIM$ loss as the loss function:
\begin{equation}
\mathcal{L}=-SSIM(\tilde{B},B),
\end{equation}
where $\tilde{B}$ and $B$ are the deraining result and corresponding ground-truth, respectively.

\subsection{Training Details}

We use the PyTorch framework to train and test the proposed method.
We trained the network for 1000 epochs with 32 joint units.
Each pair of training samples will be randomly cropped to $64\times64$ pixels.
Adam optimizer~\cite{adam} is used with a learning rate of $5\times10^{-4}$ which is divided by 10 after the $600th$ epoch and the $800th$ epoch.
Both $n$ in the Scale-Aggregation module and $r$ in the Self-Calibrated convolution are set to 4, while the channel dimensionality of the entire network is 32.
We train these networks on a PC with two NVIDIA GTX 1080Ti GPUs.

\section{Experimental Results}

In this section, we will conduct training and corresponding tests on the synthetic datasets Rain100H\cite{derain_jorder_yang}, Rain100L\cite{derain_jorder_yang}, respectively.
Rain100H and Rain100L contain 1800 pairs of training images and 200 pairs of testing images.
Rain100H represents heavy rain and Rain100L represents light rain.
The model trained under the Rain100H dataset is used to test the real-world datasets\cite{derain_analysis,derain_jorder_yang,derain_cgan_zhang} and Rain12~\cite{derain_lp_li}.

In order to clearly show the superiority of our proposed method in terms of quantity and quality, we compare it with the state-of-the-art single image deraining methods published in ACM'MM\cite{derain_acmmm19_rehen}, CVPR\cite{derain_ddn_fu,derain_prenet_Ren_2019_CVPR,derain_2019_CVPR_spa,derain-cvpr19-semi}, and ECCV\cite{derain_rescan_li} in the last three years.

\subsection{Results and Analysis}
\textbf{Quantitative Comparison.}
We compare our proposed method with DDN~\cite{derain_ddn_fu}, RESCAN~\cite{derain_rescan_li}, REHEN~\cite{derain_acmmm19_rehen}, PreNet~\cite{derain_prenet_Ren_2019_CVPR}, SpaNet~\cite{derain_2019_CVPR_spa}, and SSIR~\cite{derain-cvpr19-semi} under the two metrics of SSIM~\cite{ssim} and PSNR~\cite{psnr}.

We train our models on the synthetic datasets Rain100H and Rain100L, and compare the quantitative results obtained with the training methods under the corresponding dataset, respectively.
The metric results of our and compared deraining methods are shown in Tab.~\ref{tab: the results in synthetic datasets}.
It can be clearly seen that the proposed method has achieved the highest SSIM and PSNR in all datasets.
Compared with other state-of-the-art methods, our approach has a big improvement.

\textbf{Qualitative Comparison.}
In Fig.~\ref{fig:syn-compared}, we show some synthetic examples on the Rain100H dataset.
We can see that the proposed model can restore cleaner and clearer results, while other approaches also hand down some artifacts or remaining rain streaks.
Especially, SSIR~\cite{derain-cvpr19-semi} fails to work on synthetic datasets.

Furthermore, we provide more examples of real-world datasets to illustrate the superiority of the proposed method.
Fig.~\ref{fig:nature-compared} shows the results and we can observe that the proposed method is able to remove all of rain streaks and can preserve better details.
However, other state-of-the-art methods hand down lots of rain streaks so that they fail to restore clear and clean rain-free images.
Especially, SSIR~\cite{derain-cvpr19-semi} always fails to work and even leaves behind traces of rain streaks.

To sum up, our proposed is a more robust deraining method that not only can adapt various rainy conditions that can remove most of rain streaks, but also can better preserve image details and texture information, which benefits from our designed three learned parts: Self-Attention module, Scale-Aggregation module and Self-Calibrated convolution.
This also illustrates our method has better deraining power.

\subsection{Ablation Study}
In this section, we analyze the importance of different modules.
Especially, we utilize the Scale-Aggregation as the baseline module.
And for different modules, their abbreviations are as follows:
\begin{itemize}
	\label{itemize}
	\item $R_{1}:$ The Scale-Aggregation module as the baseline without Self-Attention module and Self-Calibrated convolution.
	\item $R_{2}:$ Baseline with Self-Calibrated convolution.
	\item $R_{3}:$ Baseline with Self-Attention module and Self-Calibrated convolution, i.e., our proposed final joint unit.
\end{itemize}
The results of different modules are shown in Tab.~\ref{tab: Ablation of the results in syn datasets}.
We can see that the Self-Calibrated convolution can improve the deraining results compared with the baseline module, and the Self-Attention module further boosts the deraining performance compared with the baseline module with Self-Calibrated convolution.
Compared with the baseline module, the introduced Self-Calibrated convolution and Self-Attention module improve the SSIM about 0.01.

\begin{table}[!h]
	\caption{The results of different modules on Rain100H. The best results are highlighted in boldface.}
	\centering
	\scalebox{0.96}{
		\begin{tabular}{ccccccccccc}
			\toprule
			Modules      & SC conv & Self-Attention & SSIM  &PSNR \\
			\midrule
			$R_1$              &                      &                 &0.9130  &29.3357       \\
			$R_2$              &$\surd$         &                 &0.9219   &\textbf{30.0307}  \\
			$R_3$               &$\surd$         &$\surd$  &\textbf{0.9221}  &30.0160  \\
			\bottomrule
	\end{tabular}}
	\label{tab: Ablation of the results in syn datasets}
\end{table}

\begin{figure*}[!t]
	\begin{center}
		\begin{tabular}{ccccccccc}
			\includegraphics[width = 0.19\linewidth]{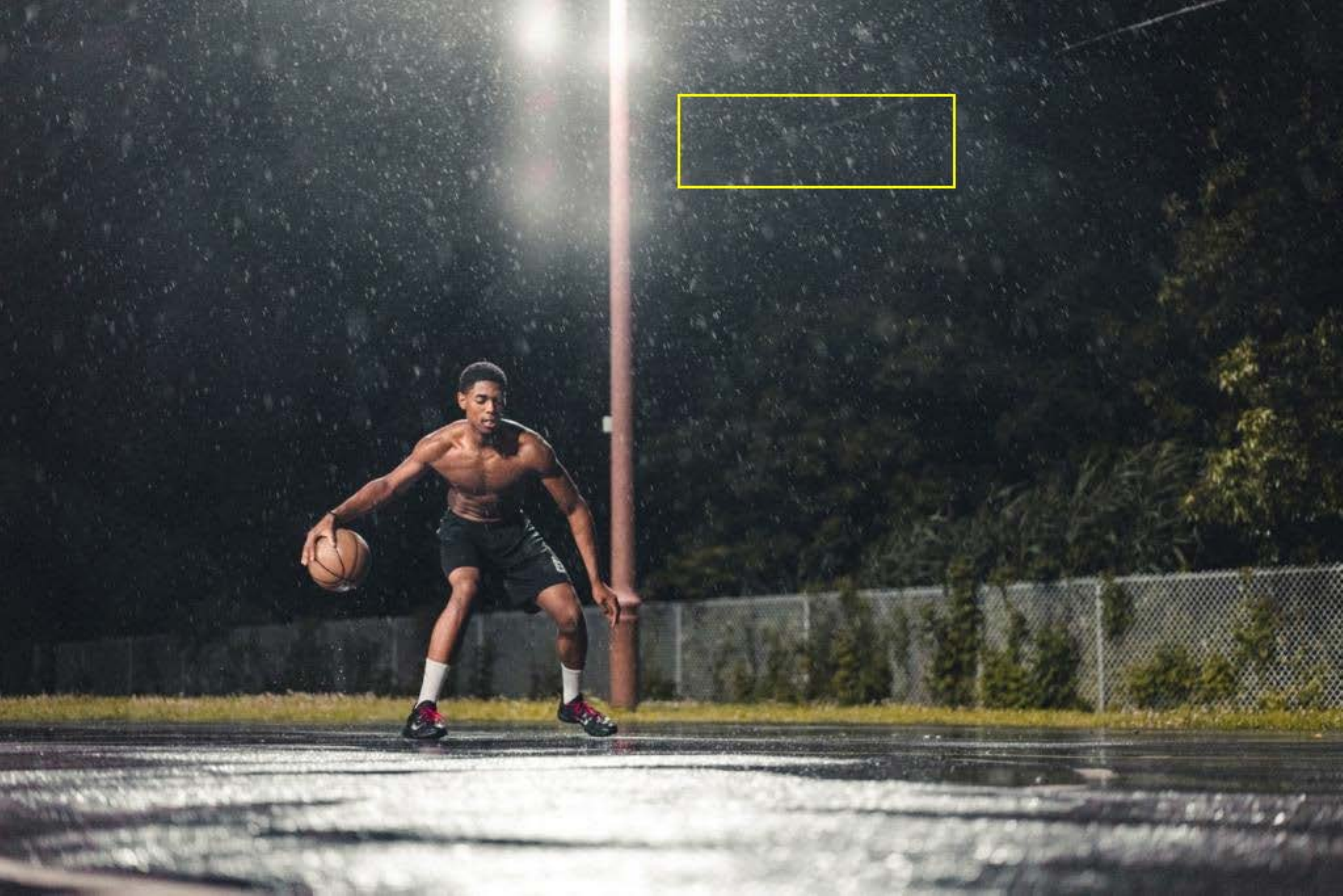}&\hspace{-4mm}
			\includegraphics[width = 0.19\linewidth]{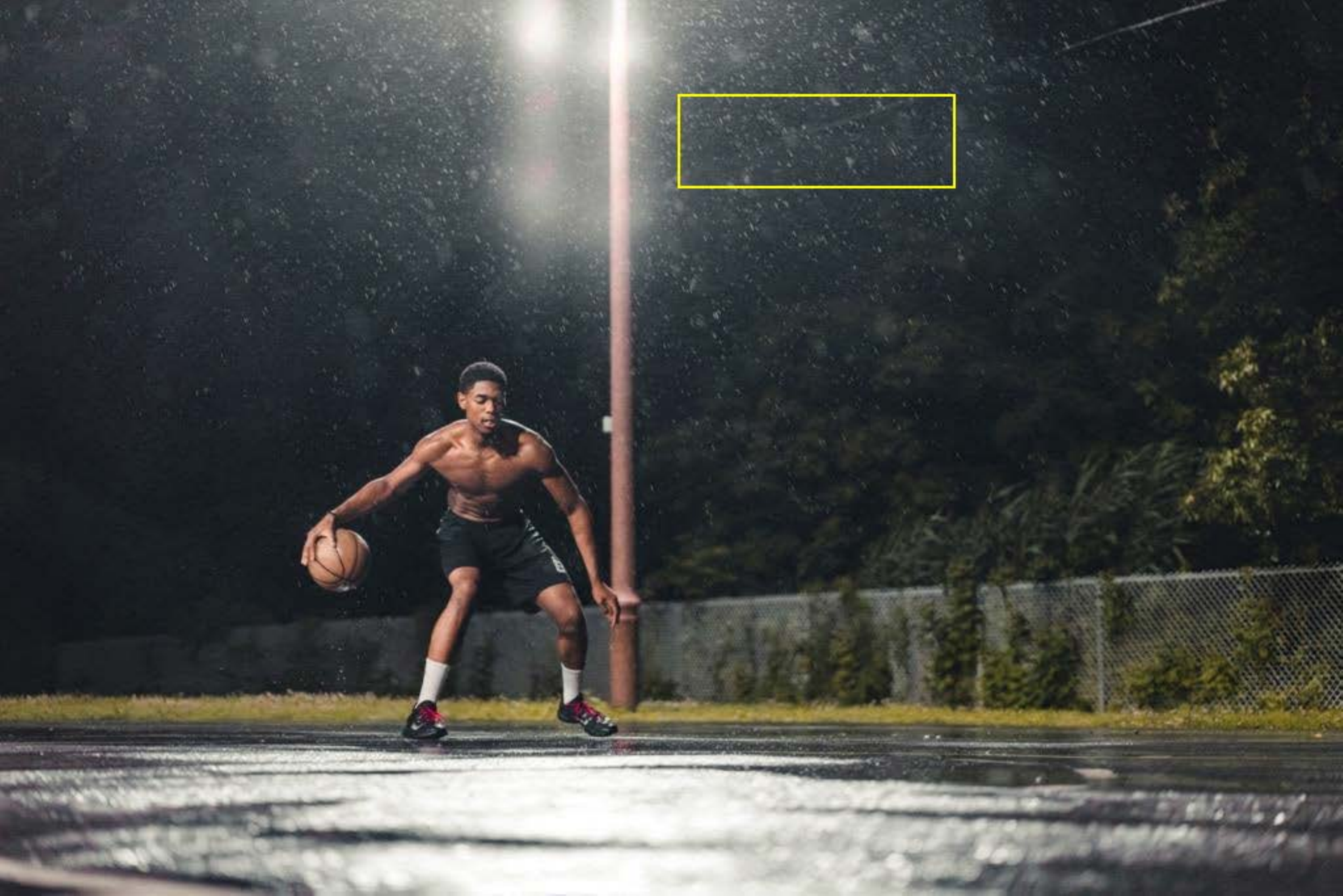}&\hspace{-4mm}
			\includegraphics[width = 0.19\linewidth]{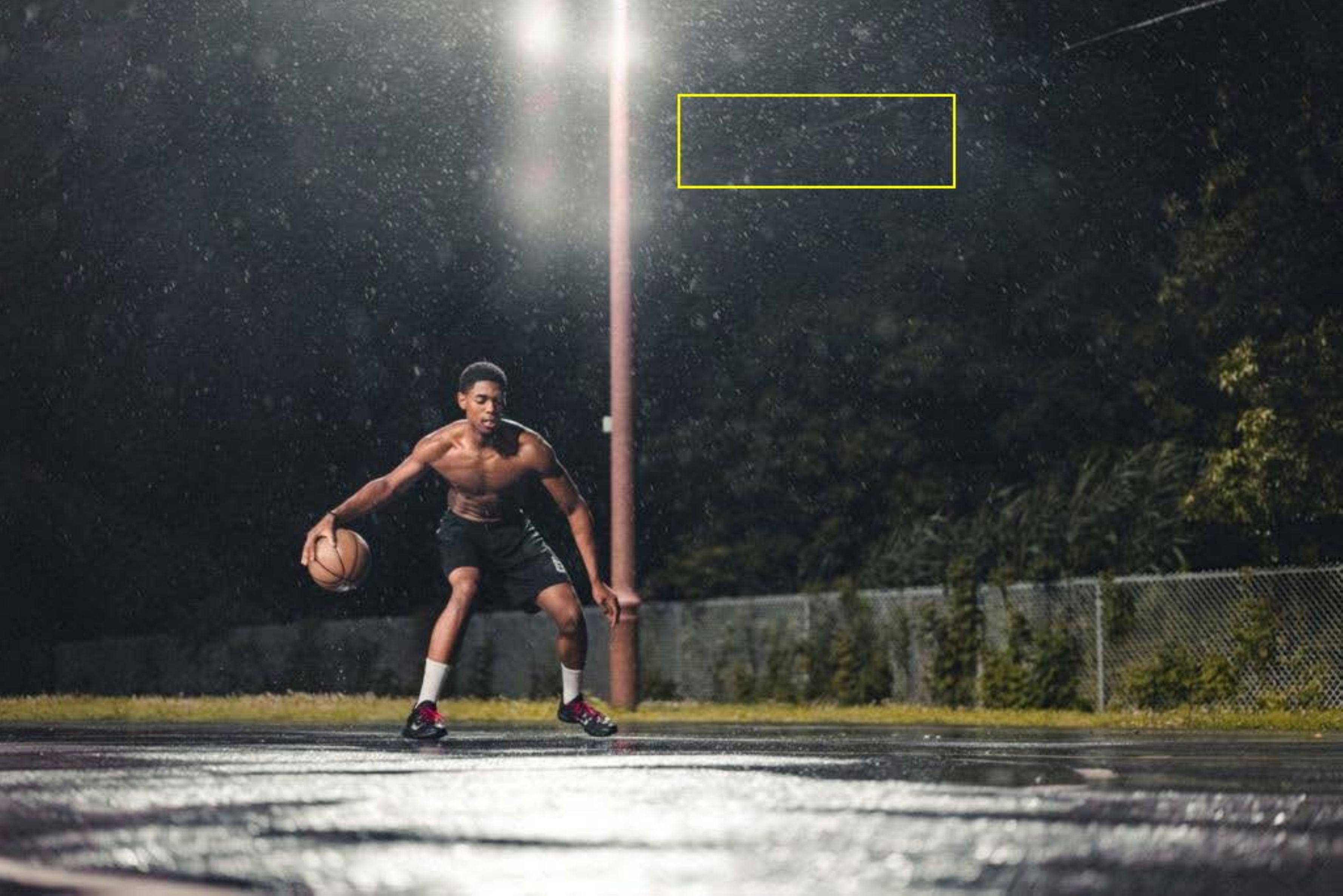}&\hspace{-4mm}
			\includegraphics[width = 0.19\linewidth]{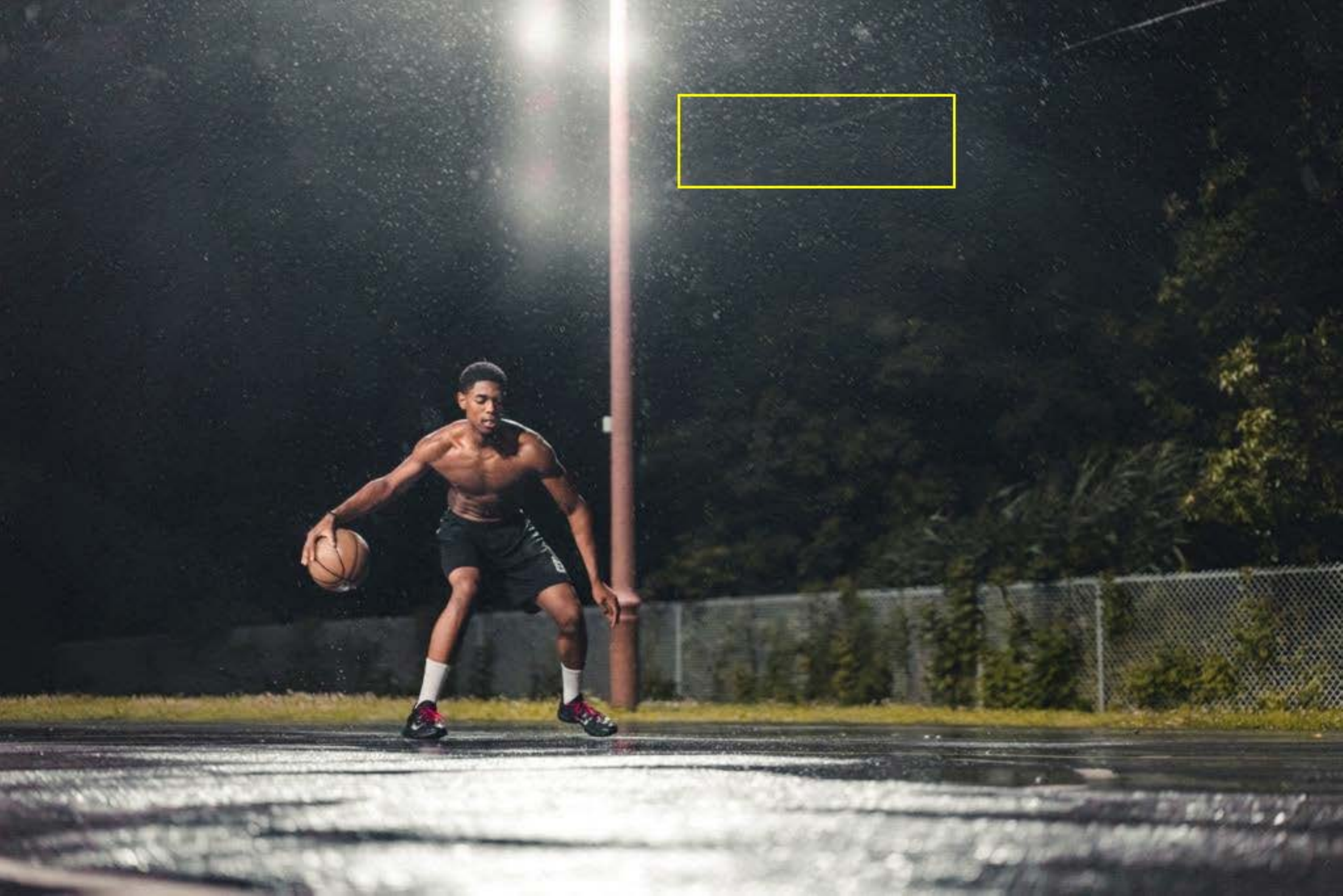}
			\\
			\includegraphics[width = 0.19\linewidth]{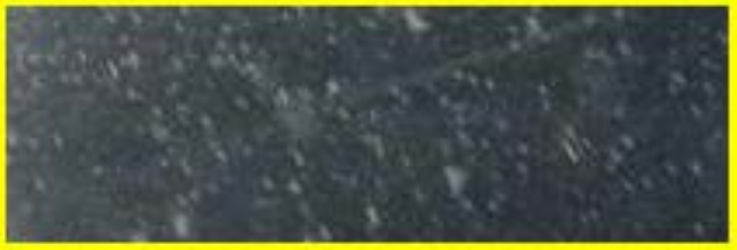}&\hspace{-4mm}
			\includegraphics[width = 0.19\linewidth]{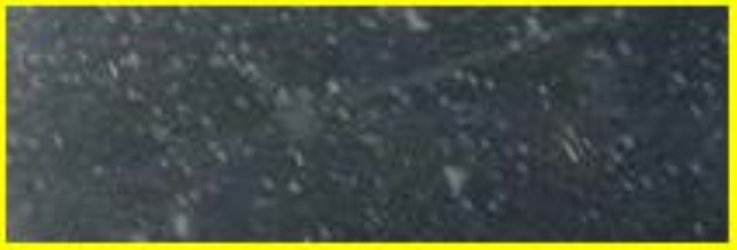}&\hspace{-4mm}
			\includegraphics[width = 0.19\linewidth]{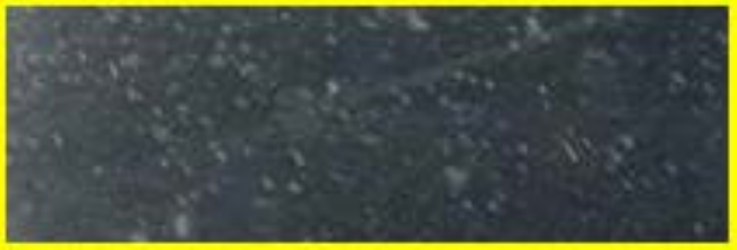}&\hspace{-4mm}
			\includegraphics[width = 0.19\linewidth]{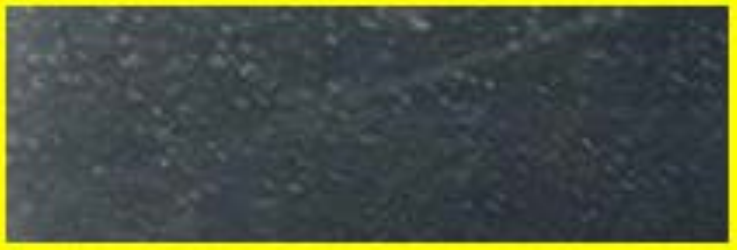}
			\\
			\includegraphics[width = 0.19\linewidth]{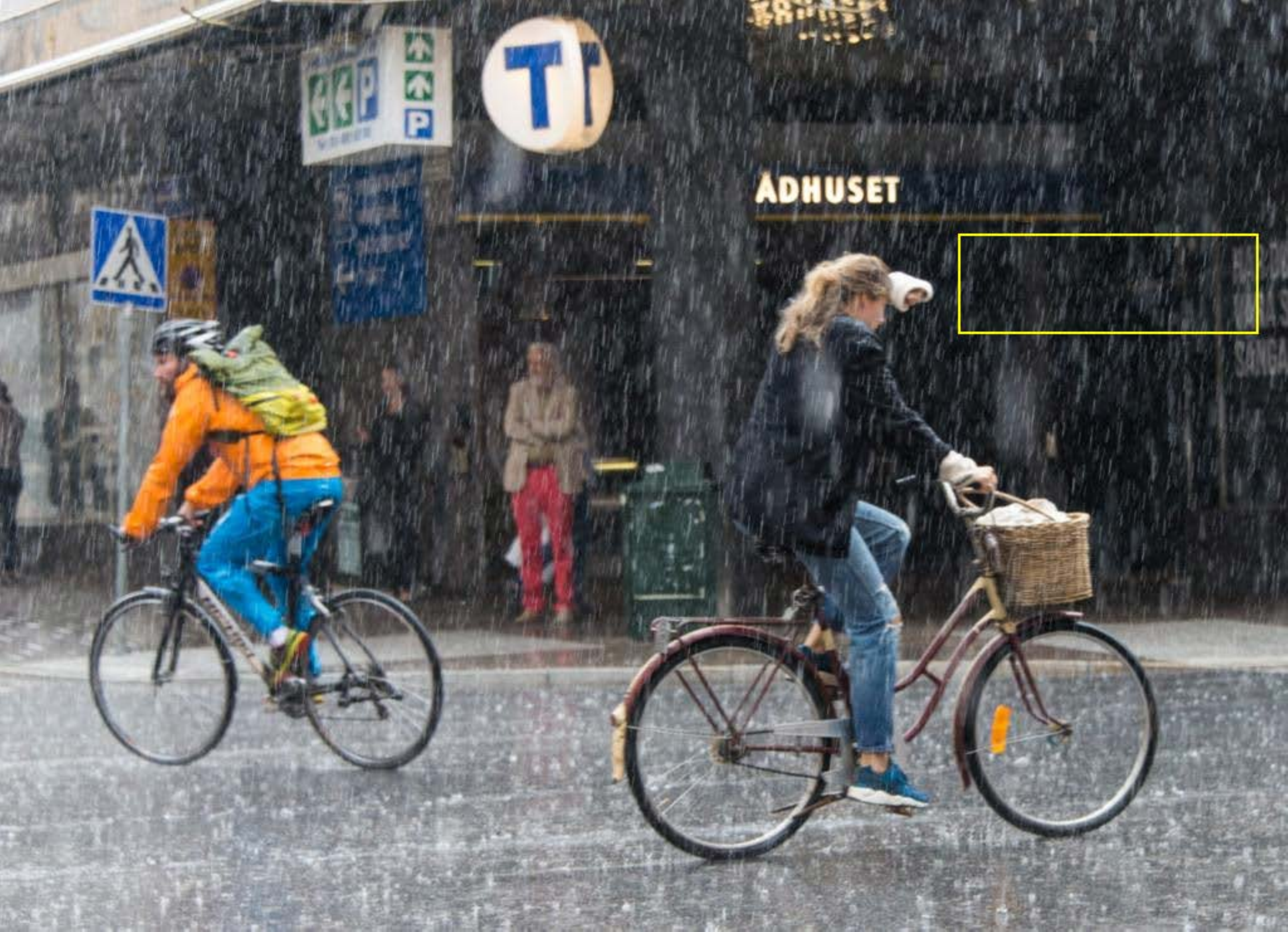}&\hspace{-4mm}
			\includegraphics[width = 0.19\linewidth]{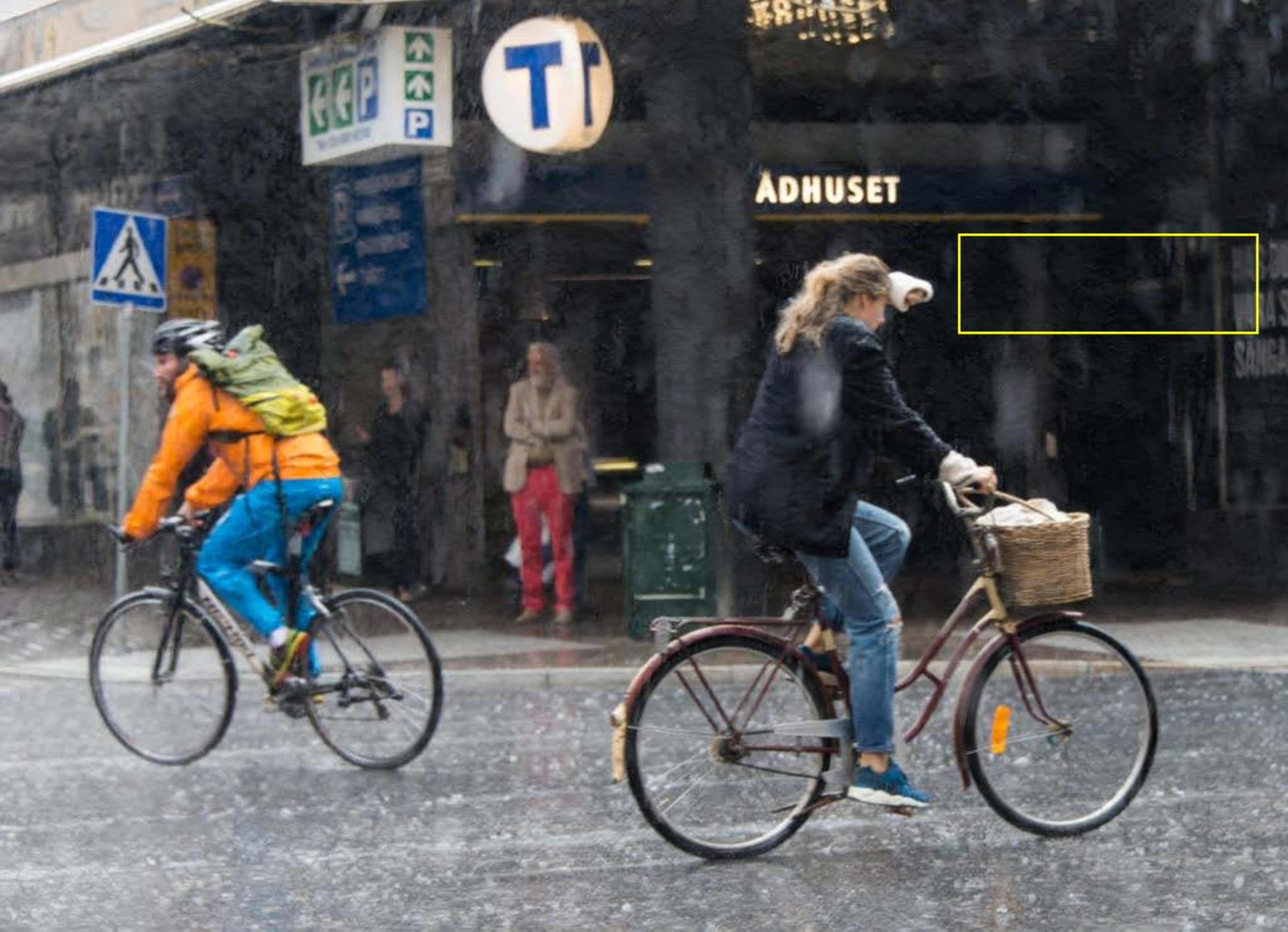}&\hspace{-4mm}
			\includegraphics[width = 0.19\linewidth]{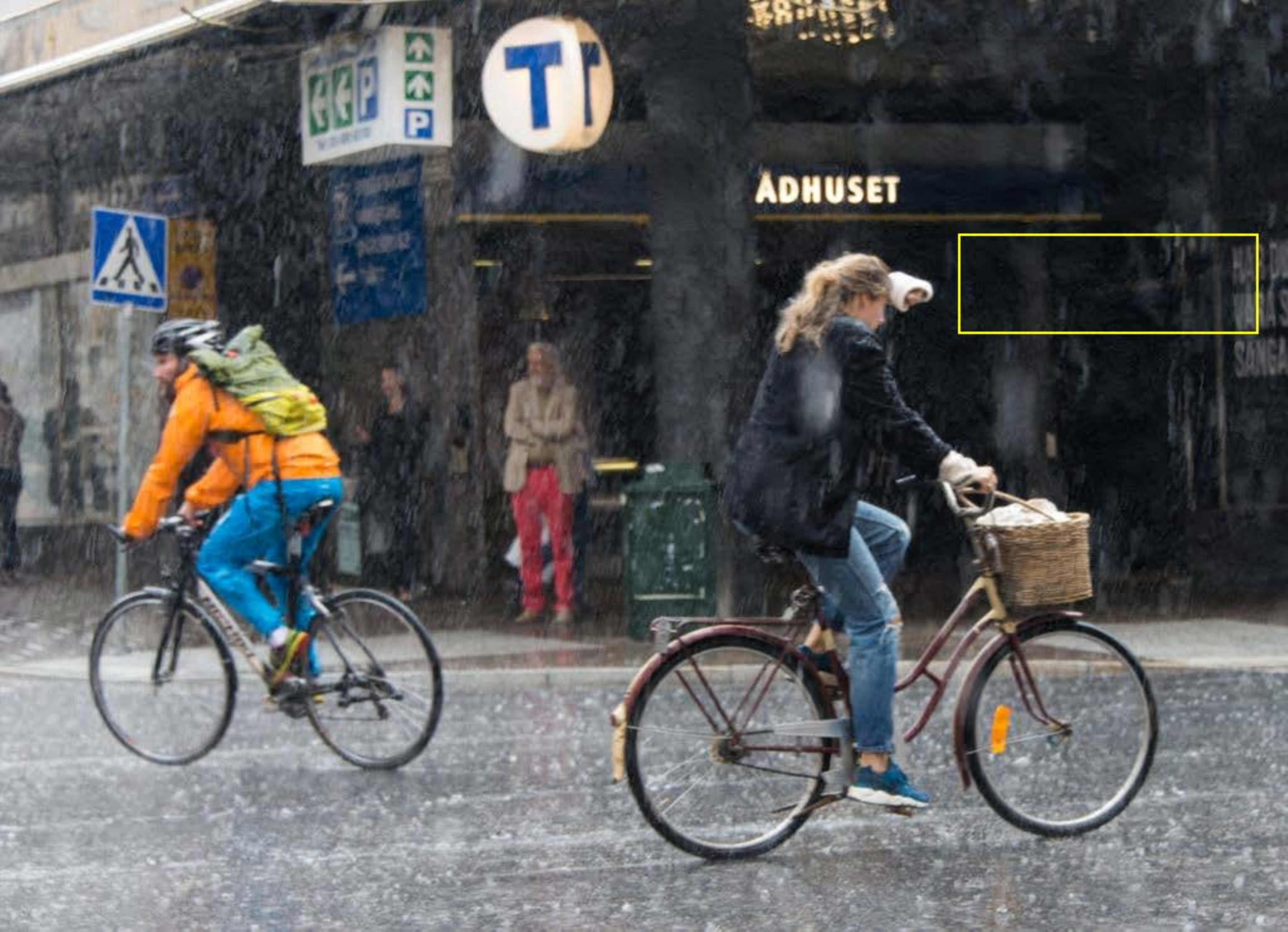}&\hspace{-4mm}
			\includegraphics[width = 0.19\linewidth]{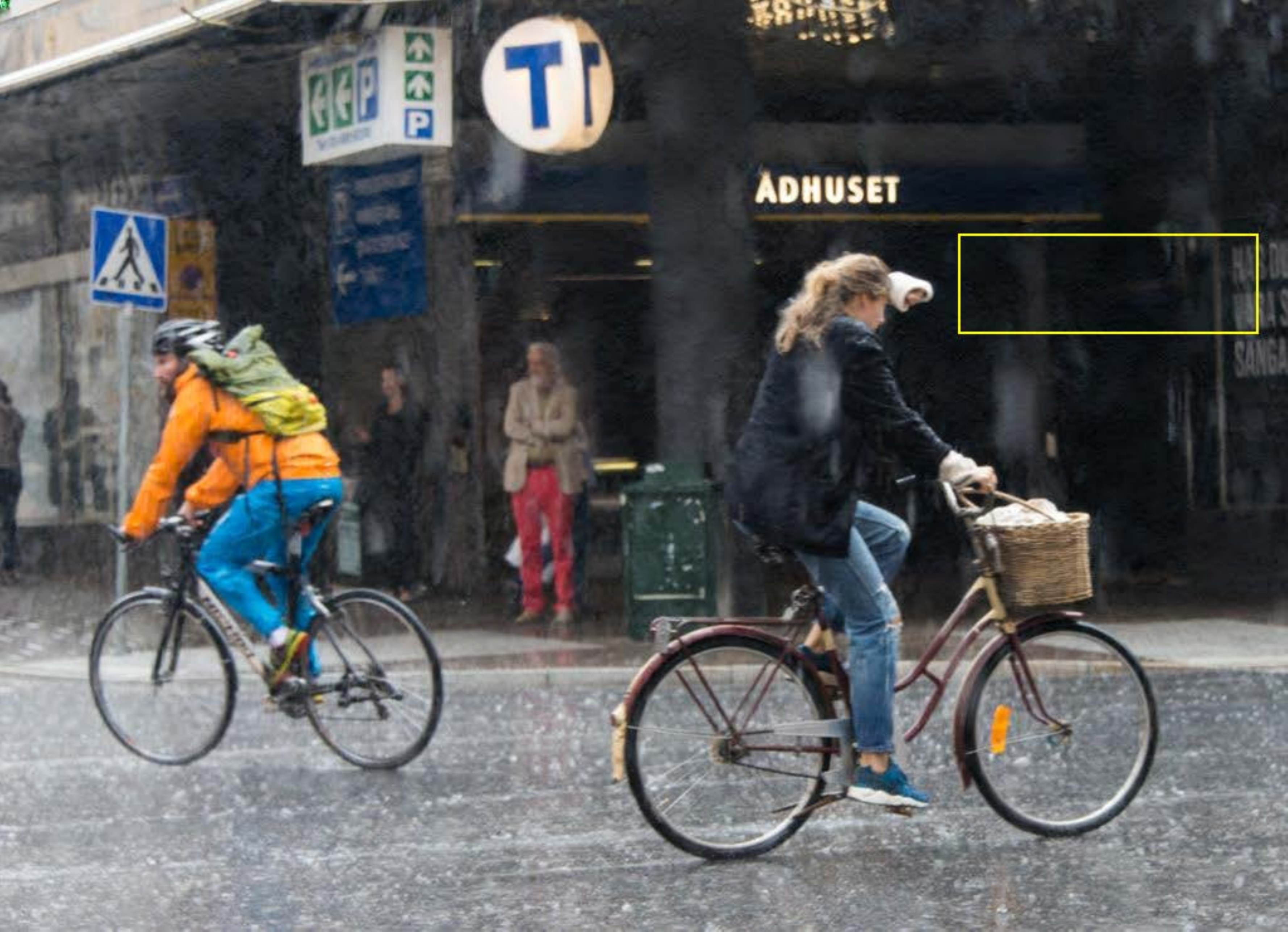}
			\\
			\includegraphics[width = 0.19\linewidth]{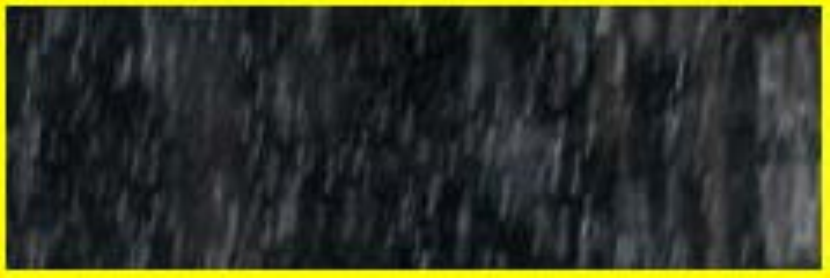}&\hspace{-4mm}
			\includegraphics[width = 0.19\linewidth]{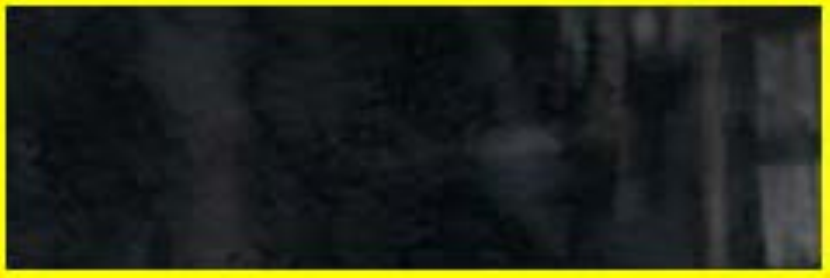}&\hspace{-4mm}
			\includegraphics[width = 0.19\linewidth]{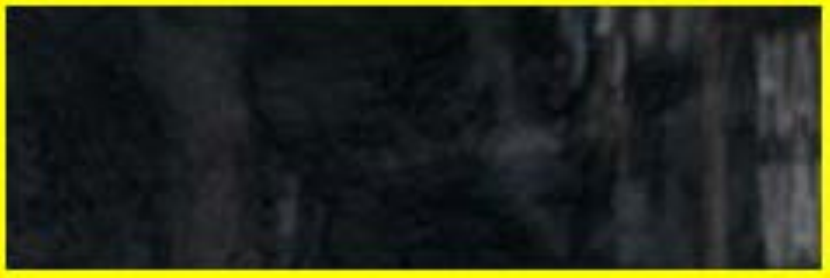}&\hspace{-4mm}
			\includegraphics[width = 0.19\linewidth]{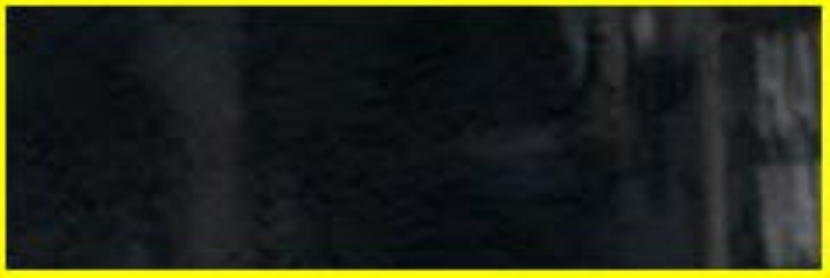}
			\\
			(a) input&\hspace{-4mm} (b) $R_1$ &\hspace{-4mm} (c) $R_2$&\hspace{-4mm} (d) $R_3$%&\hspace{-4mm} (e) $R_4$
		\end{tabular}
	\end{center}
	\caption{The results of different modules from real-world datasets.}
	\label{fig:ablation}
\end{figure*}

In Fig.~\ref{fig:ablation} we provide two deraining examples with the three modules on real-world datasets. We can observe that the self-attention, shown in Fig.~\ref{fig:ablation} (d), plays an important role in the real-world deraining results. This phenomenon further demonstrates the the importance of self-attention that has content adaptation ability to better help the network remove the rain streaks, especially on real-world datasets.

%Tab~\ref{tab: Ablation of the results in syn datasets} is different from the previous experiments, showing SSIM values under various models without PSNR values, because our model is a SSIM-oriented training model. PSNR is greatly affected by $l_1$ loss, and is not a research object for ablation learning.
%$R_4$ achieve the highest numerical result in all experiments.
%
%In addition, we conduct comparative experiments on these four models in real-world datasets, as shown in Fig~\ref{fig:ablation}. As a pyramid model, $R_1$ can remove part of rain streaks, reflecting the effectiveness of the baseline model. As the most complete model, $R_4$ get the cleanest rain-free images. It can be explained that we have better feature expression ability after adding two modules.

\begin{figure*}[!t]
	\begin{center}
		\begin{tabular}{ccccccccc}
			\includegraphics[width = 0.118\linewidth]{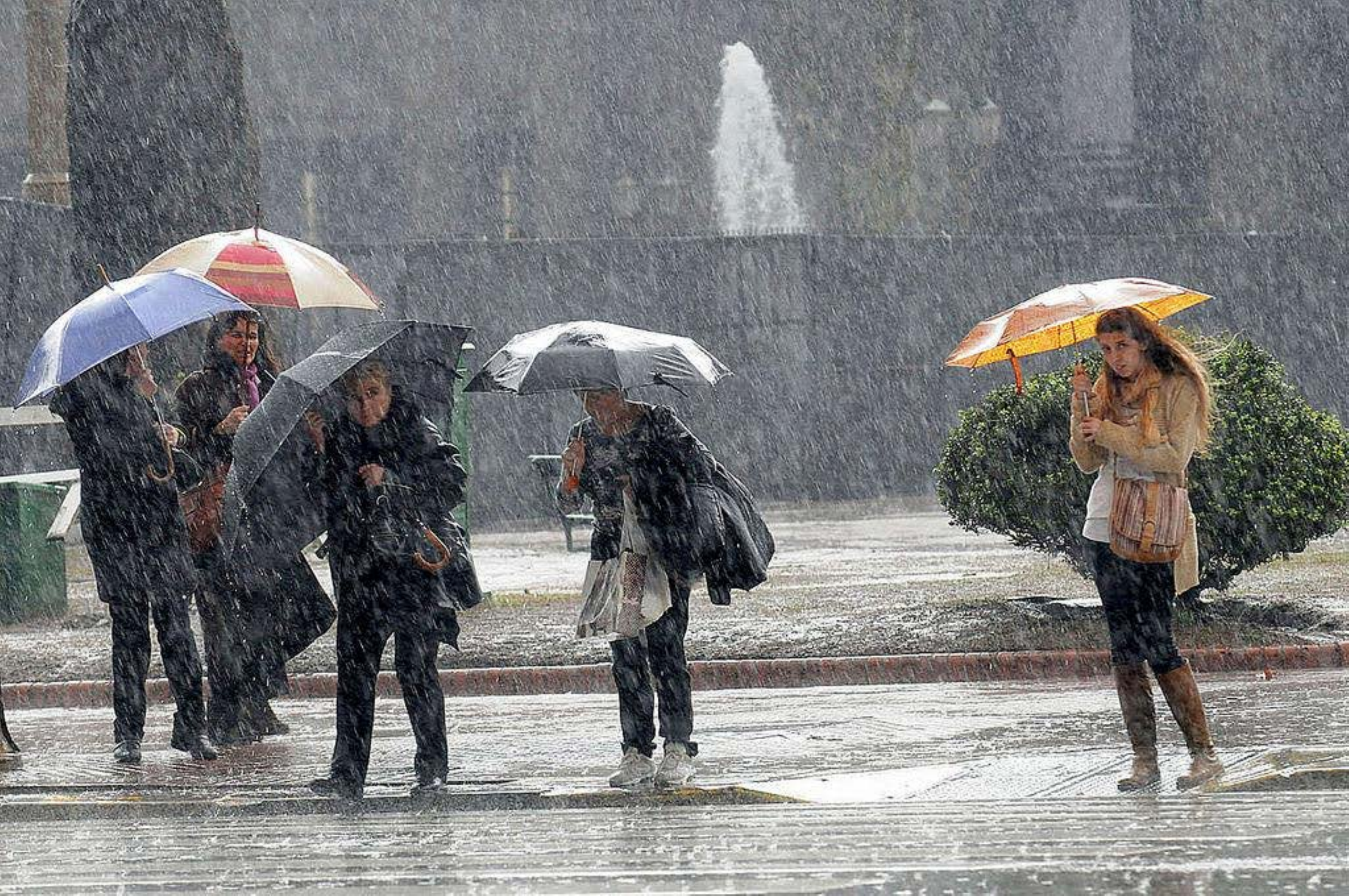}&\hspace{-4mm}
			\includegraphics[width = 0.118\linewidth]{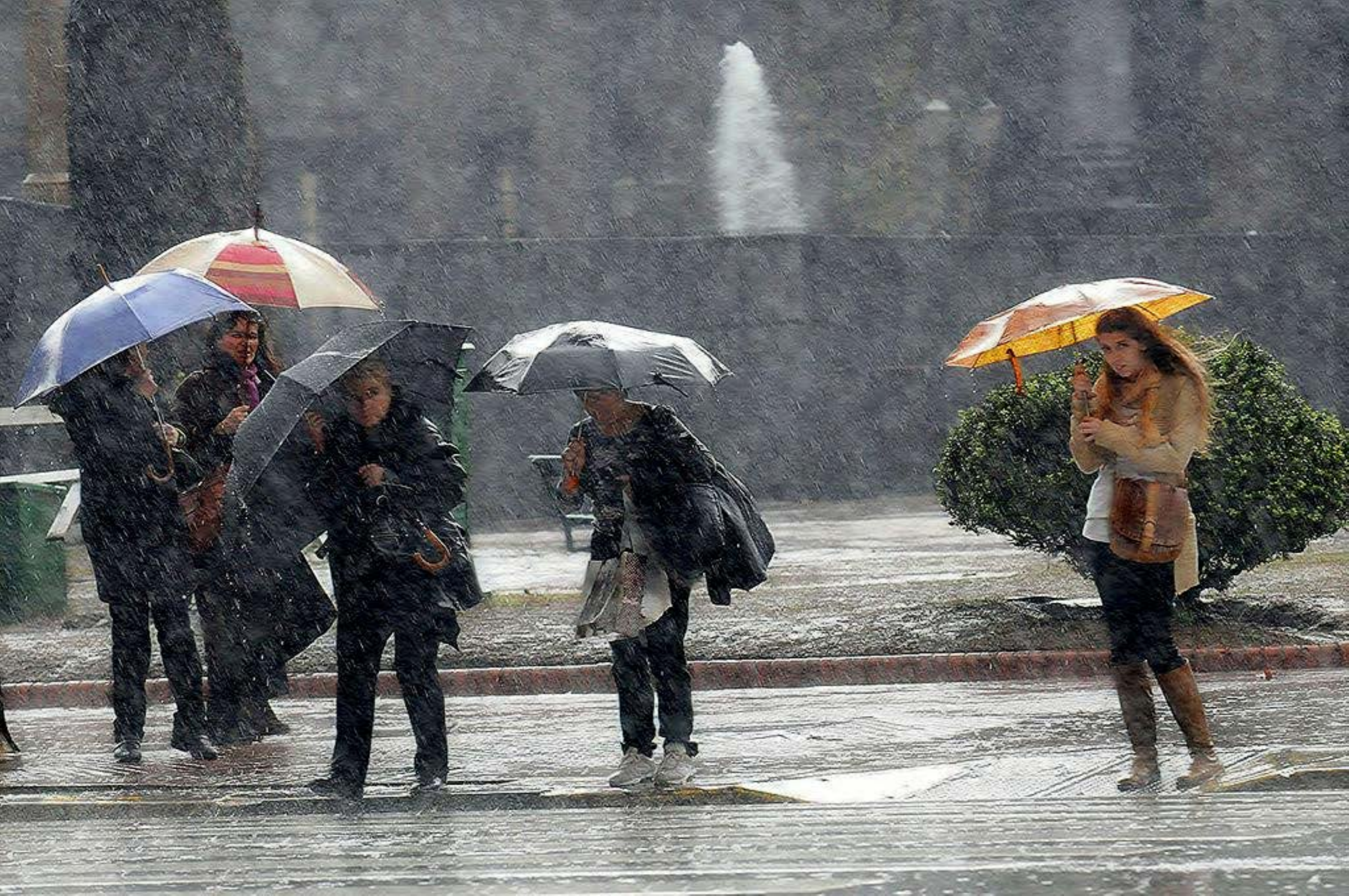}&\hspace{-4mm}
			\includegraphics[width = 0.118\linewidth]{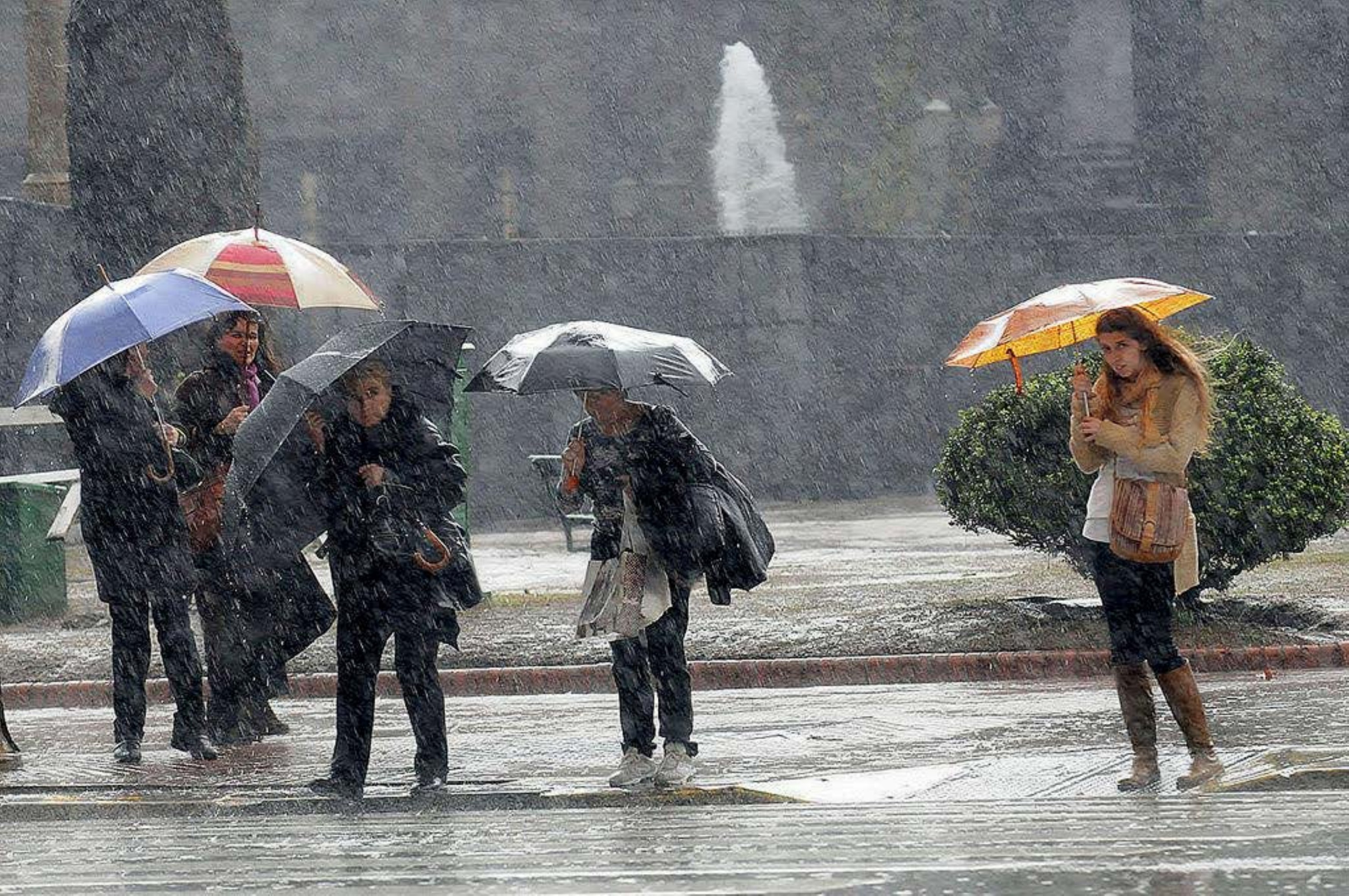}&\hspace{-4mm}
			\includegraphics[width = 0.118\linewidth]{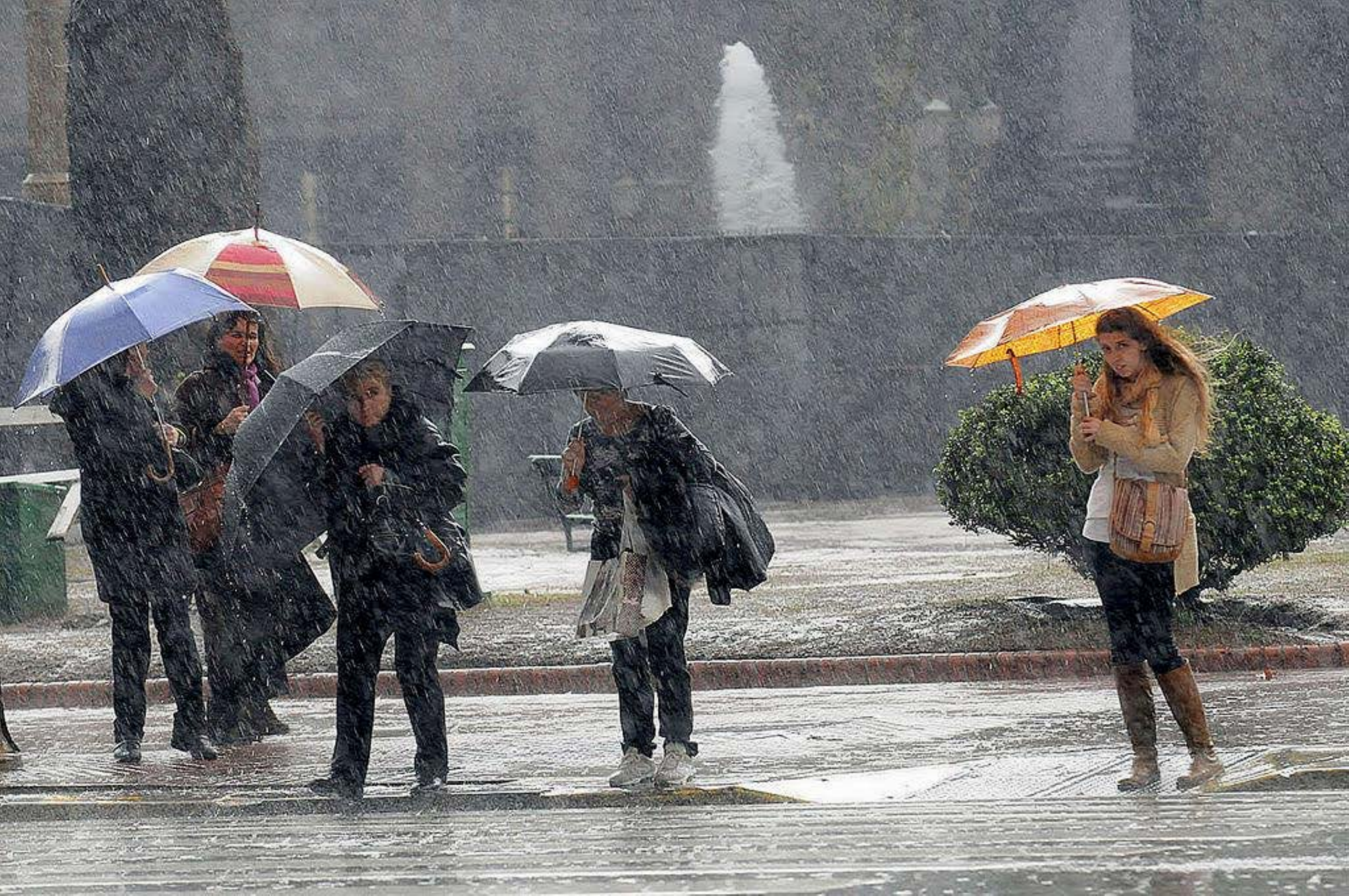}&\hspace{-4mm}
			\includegraphics[width = 0.118\linewidth]{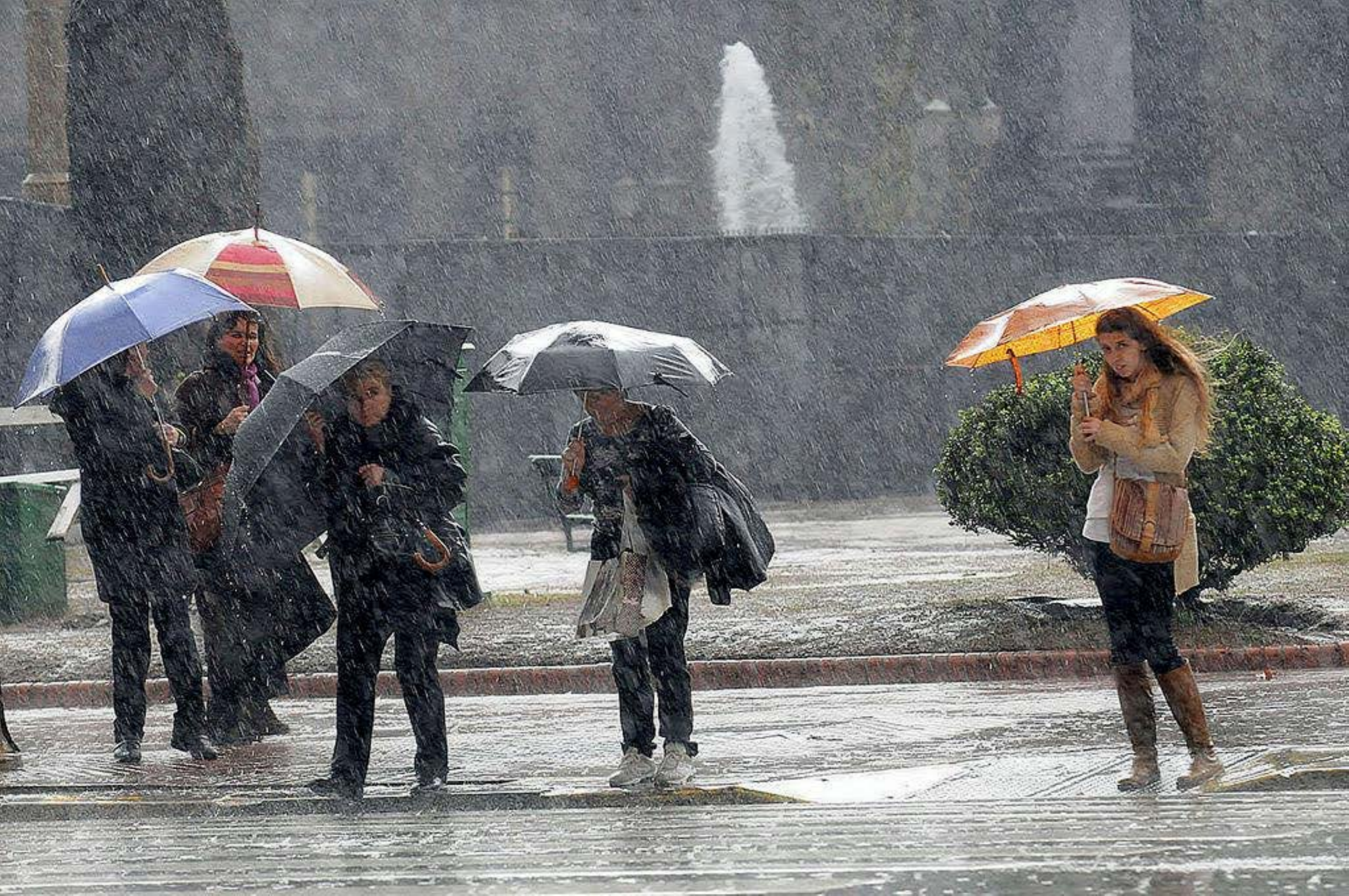}&\hspace{-4mm}
			\includegraphics[width = 0.118\linewidth]{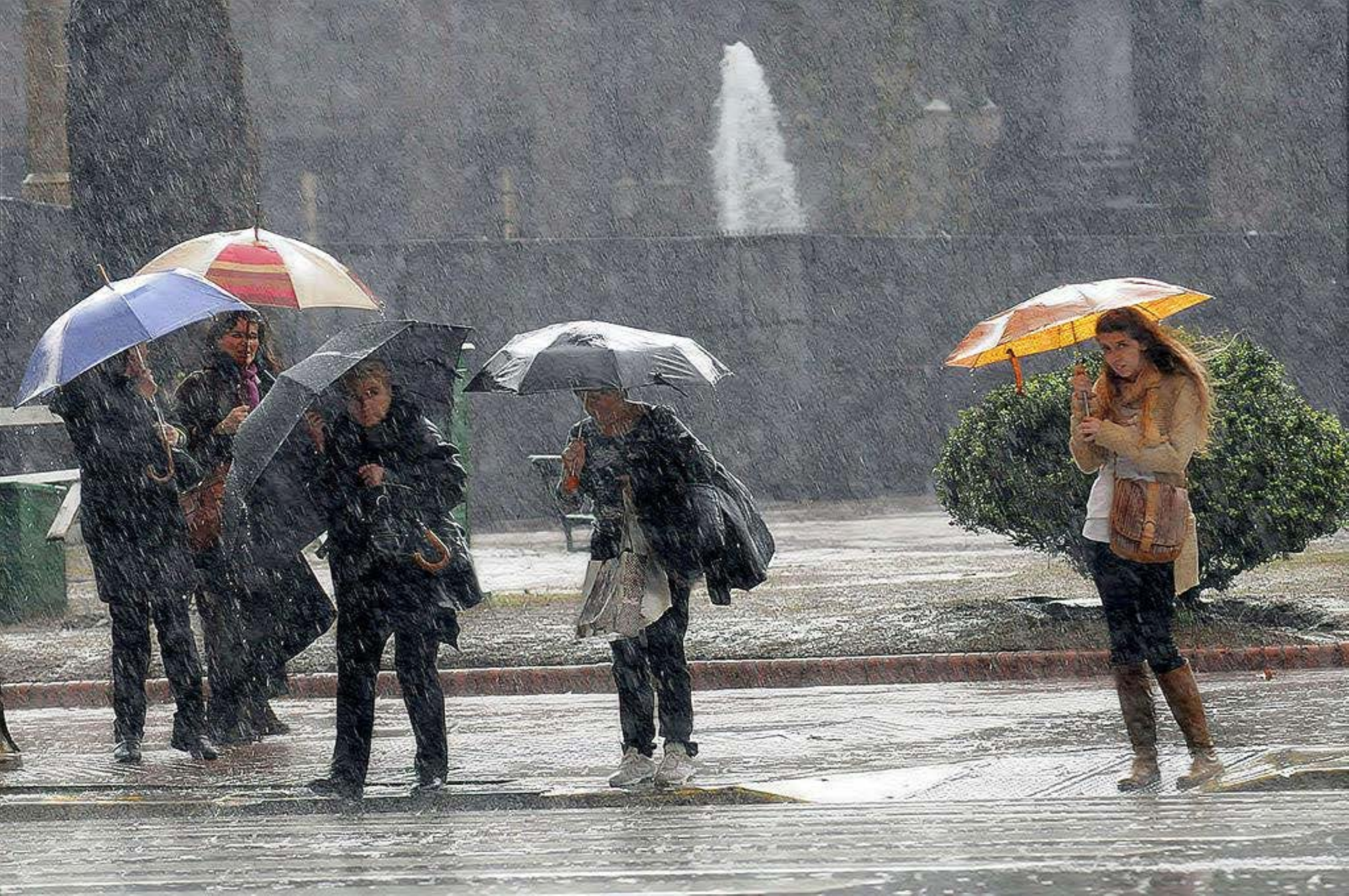}&\hspace{-4mm}
			\includegraphics[width = 0.118\linewidth]{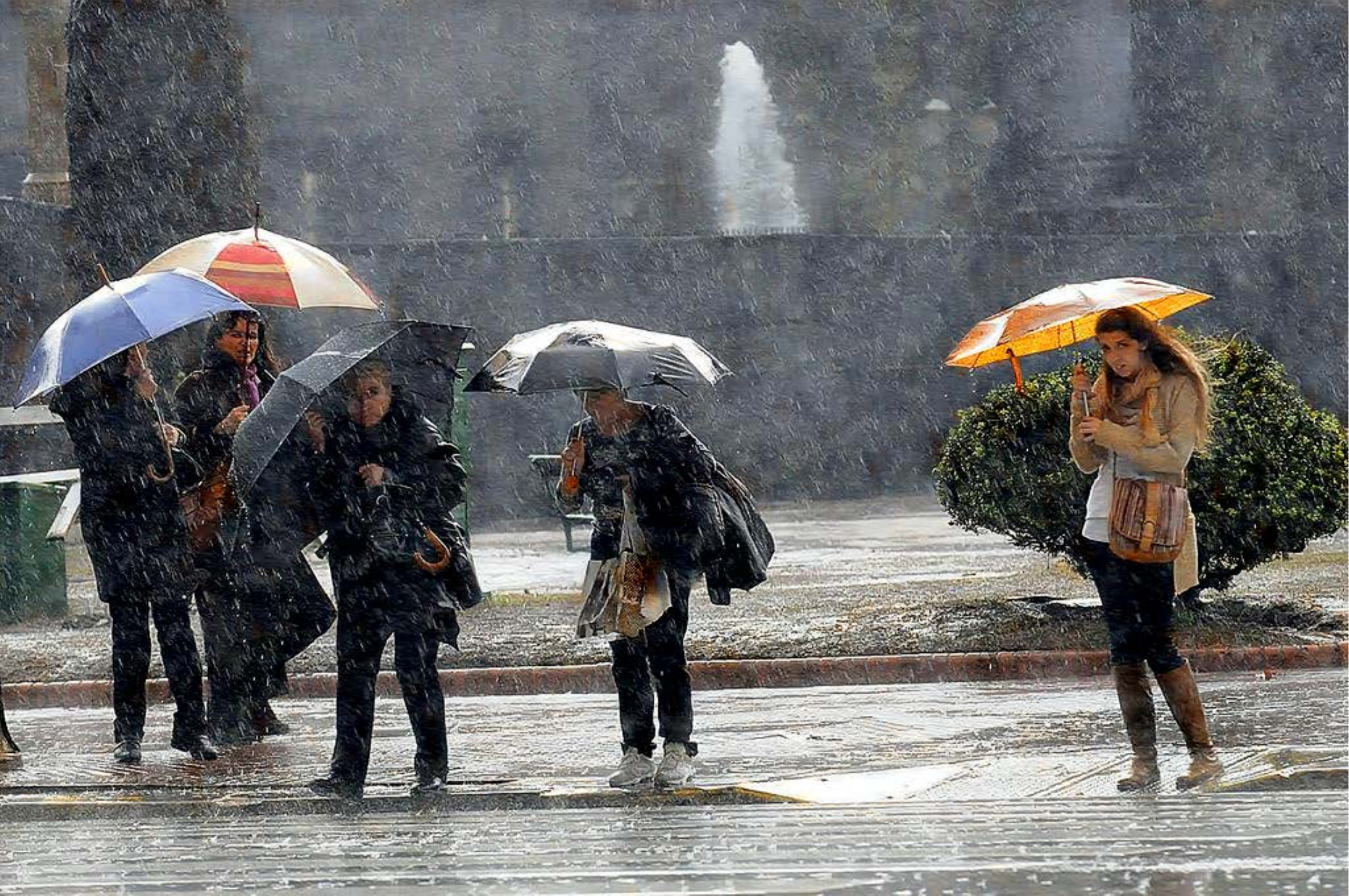}&\hspace{-4mm}
			\includegraphics[width = 0.118\linewidth]{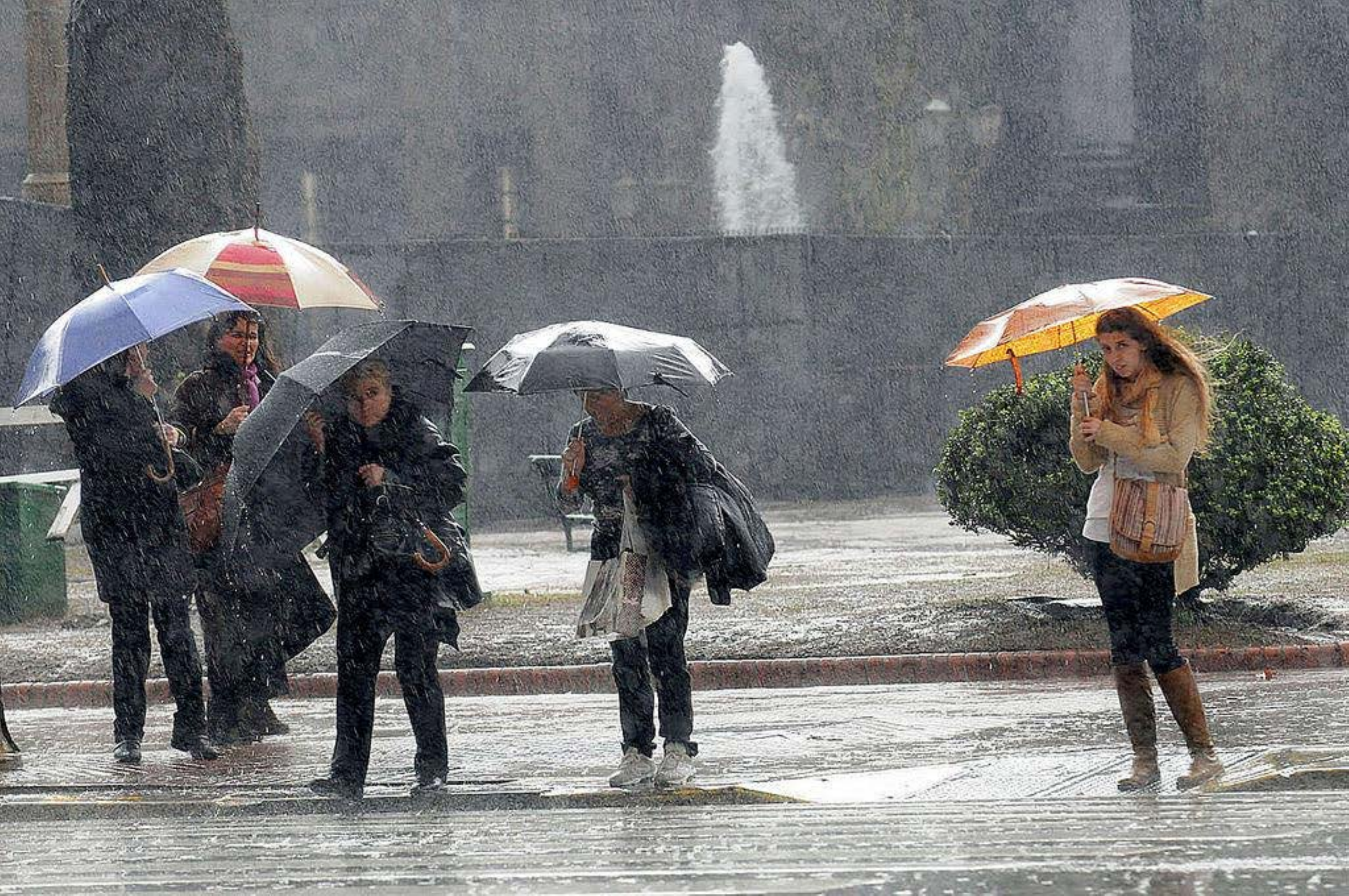}
			\\
			\includegraphics[width = 0.118\linewidth]{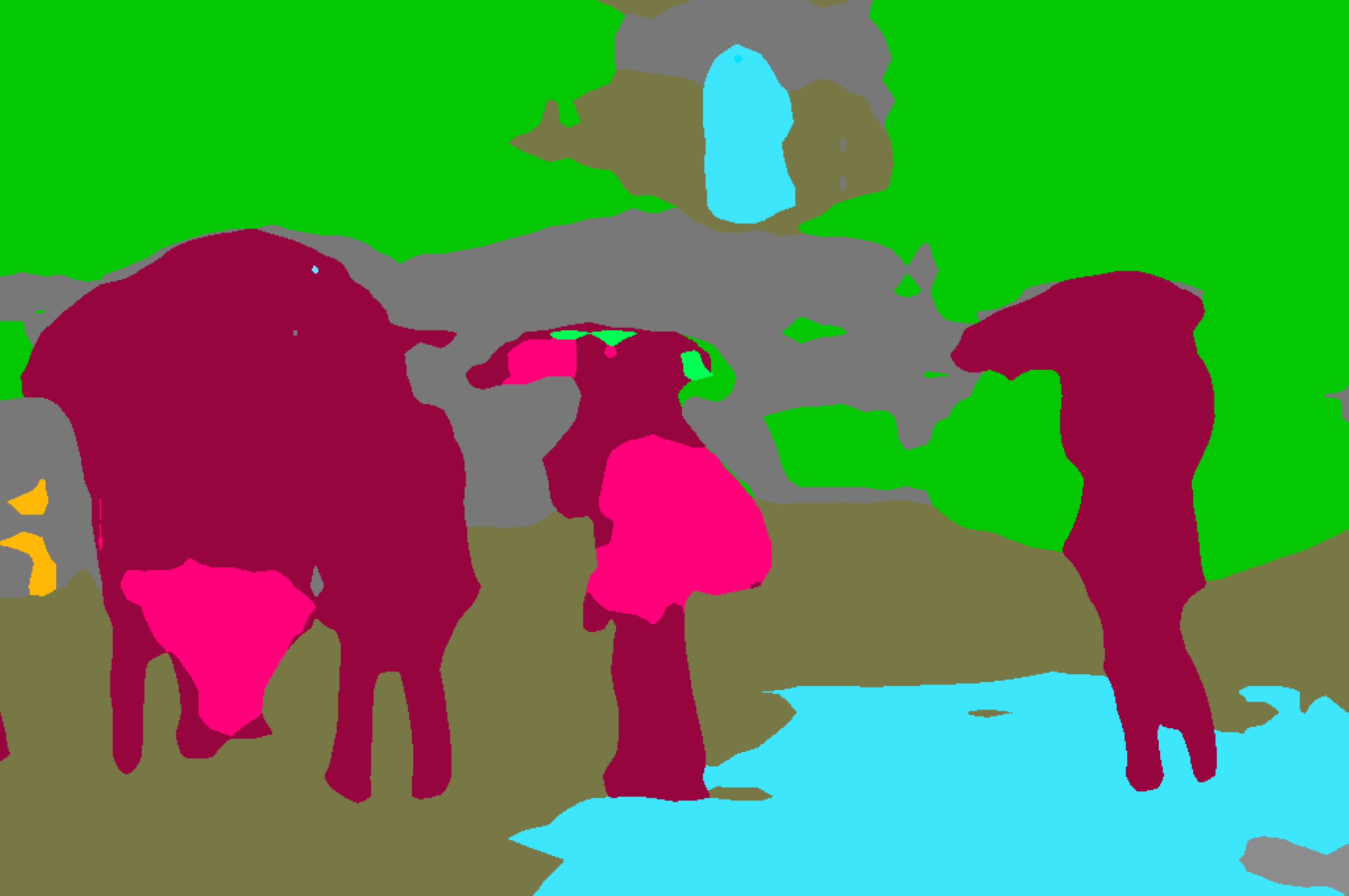}&\hspace{-4mm}
			\includegraphics[width = 0.118\linewidth]{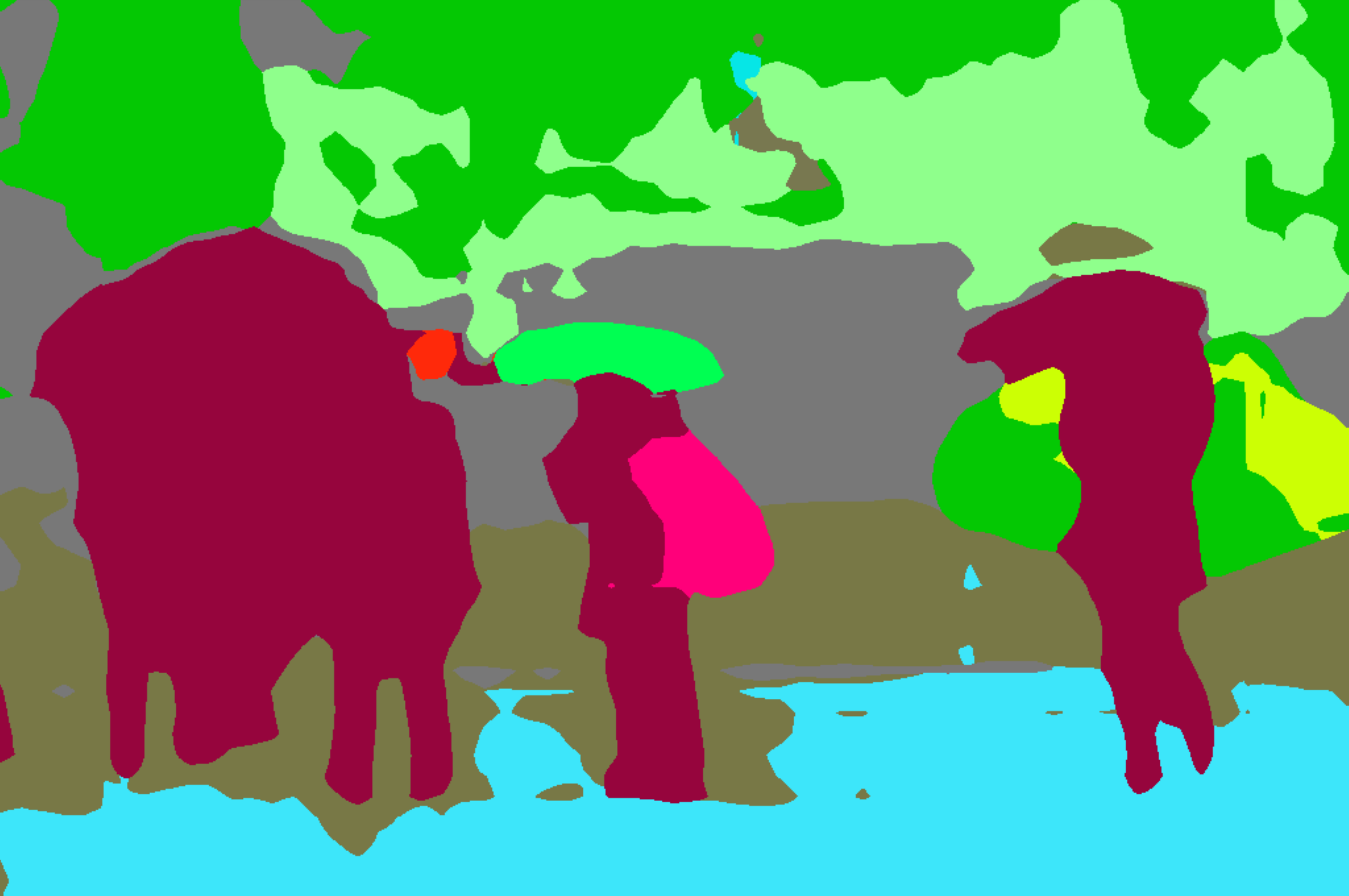}&\hspace{-4mm}
			\includegraphics[width = 0.118\linewidth]{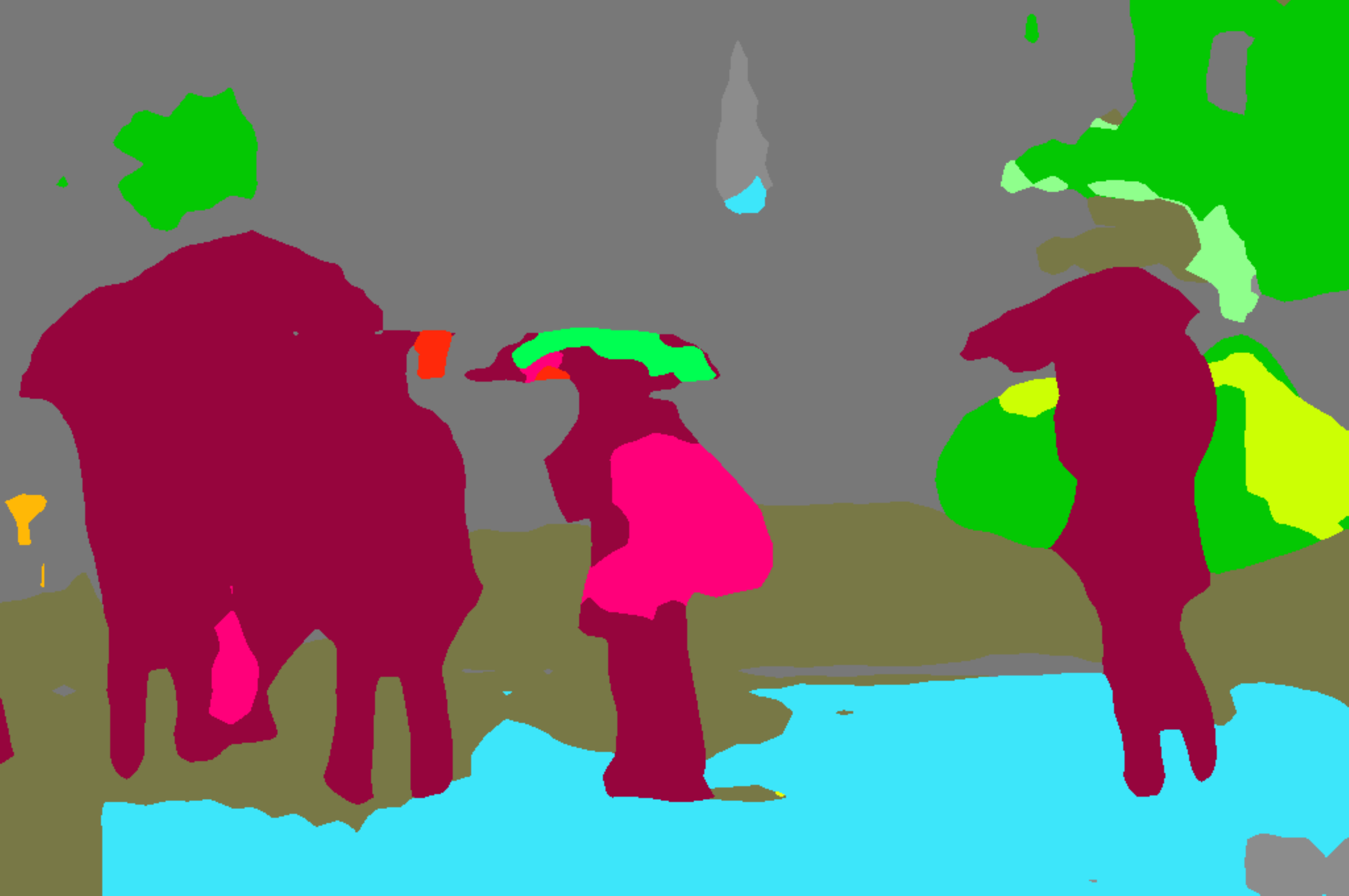}&\hspace{-4mm}
			\includegraphics[width = 0.118\linewidth]{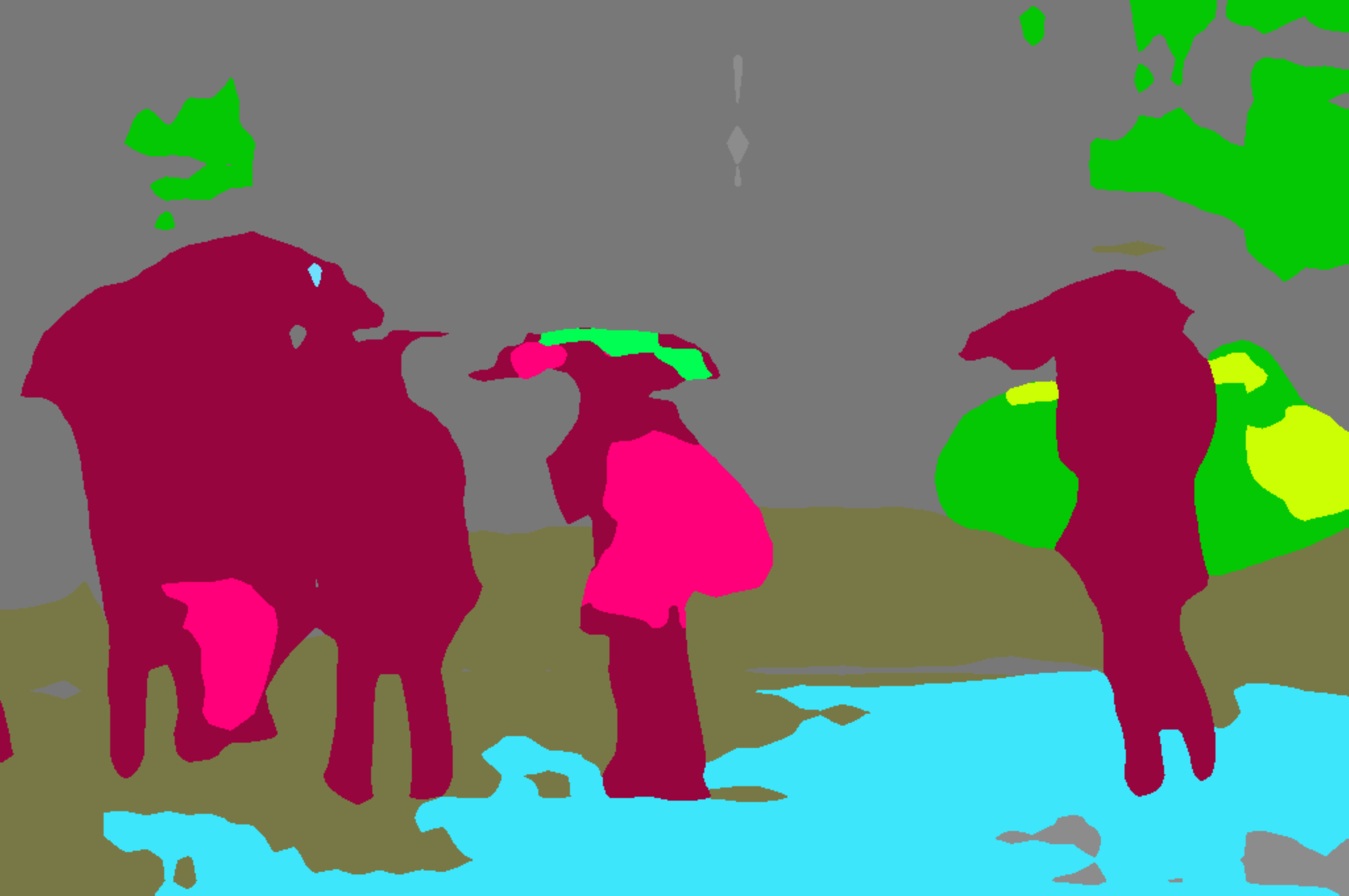}&\hspace{-4mm}
			\includegraphics[width = 0.118\linewidth]{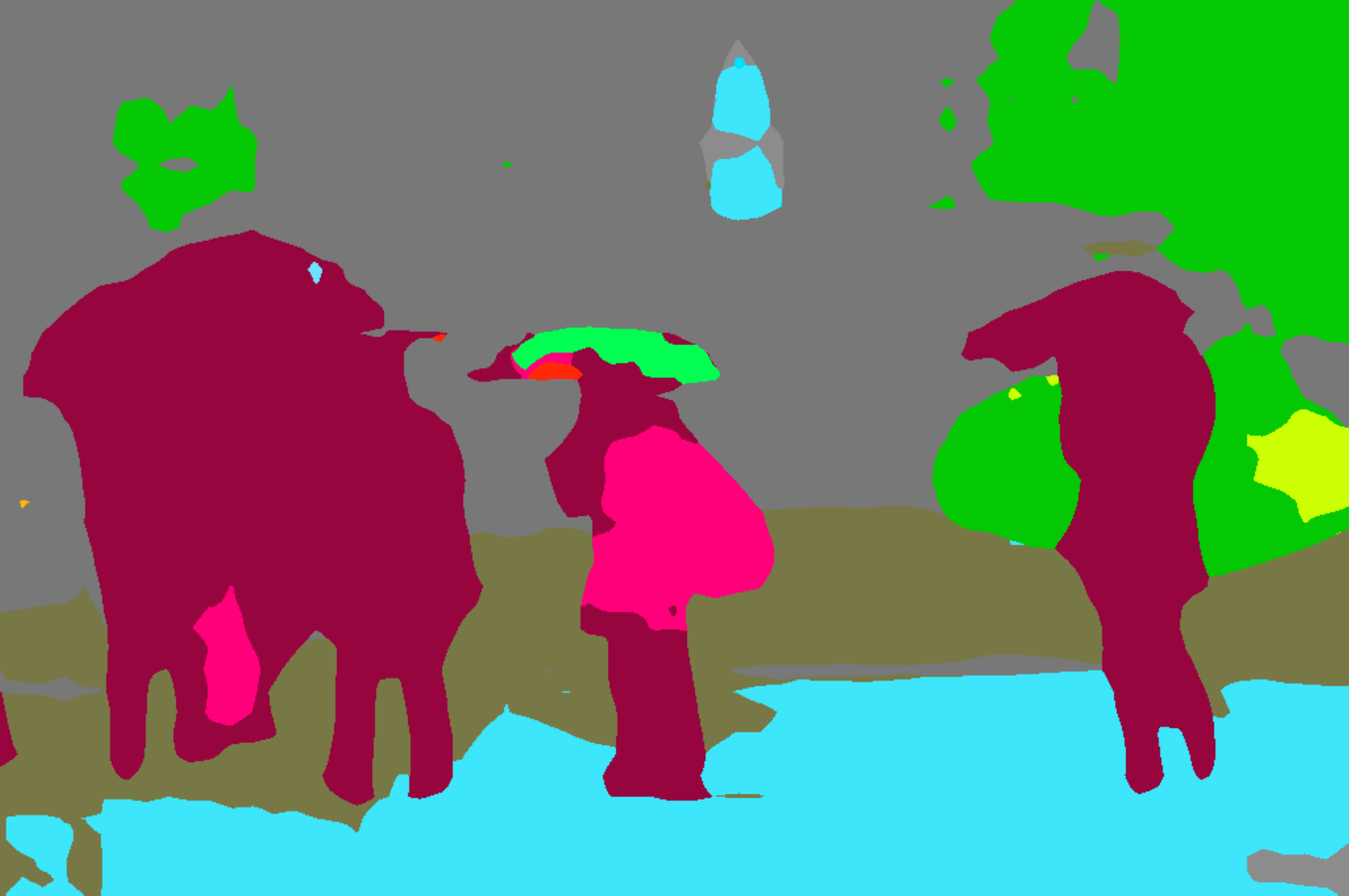}&\hspace{-4mm}
			\includegraphics[width = 0.118\linewidth]{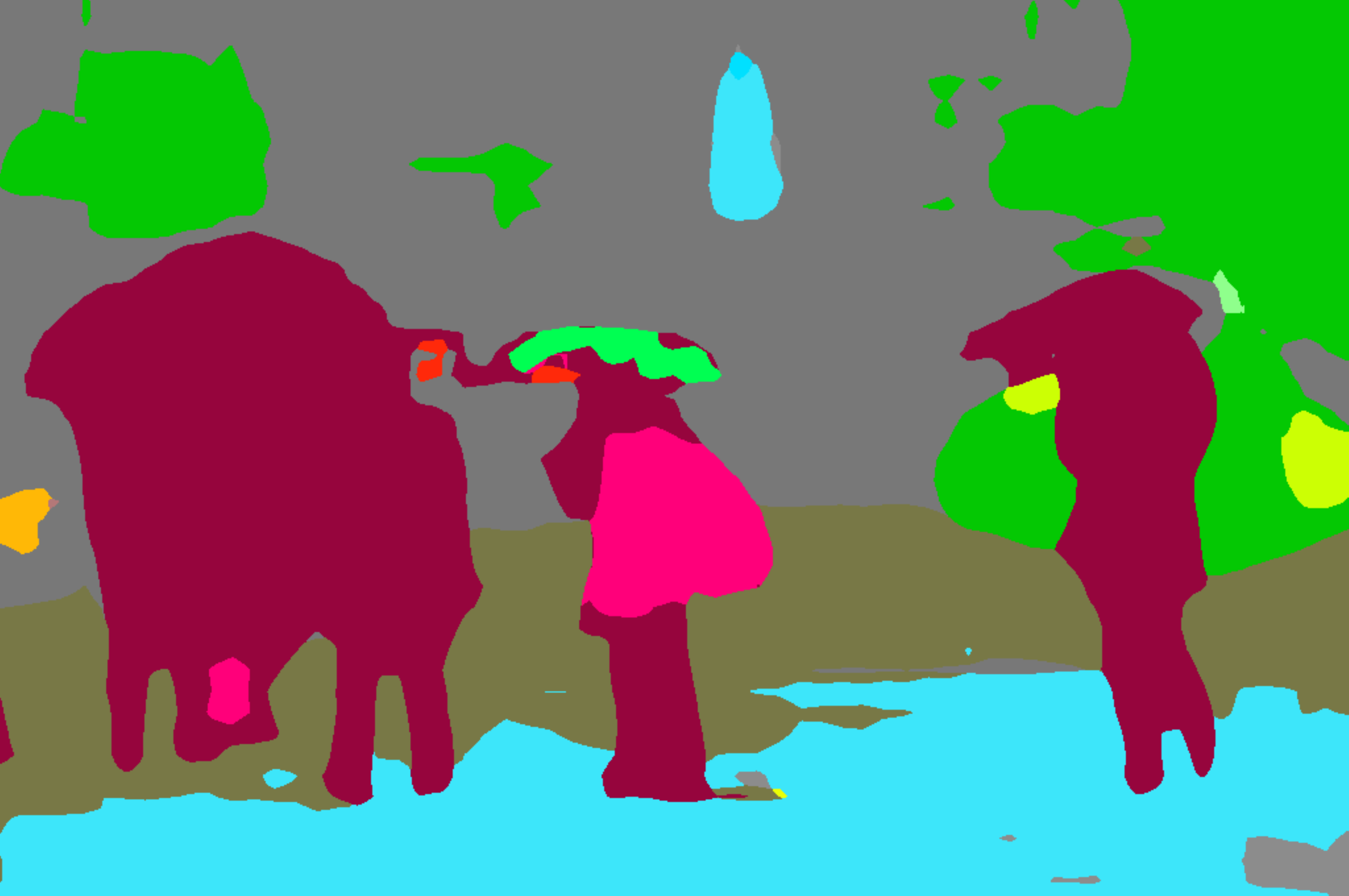}&\hspace{-4mm}
			\includegraphics[width = 0.118\linewidth]{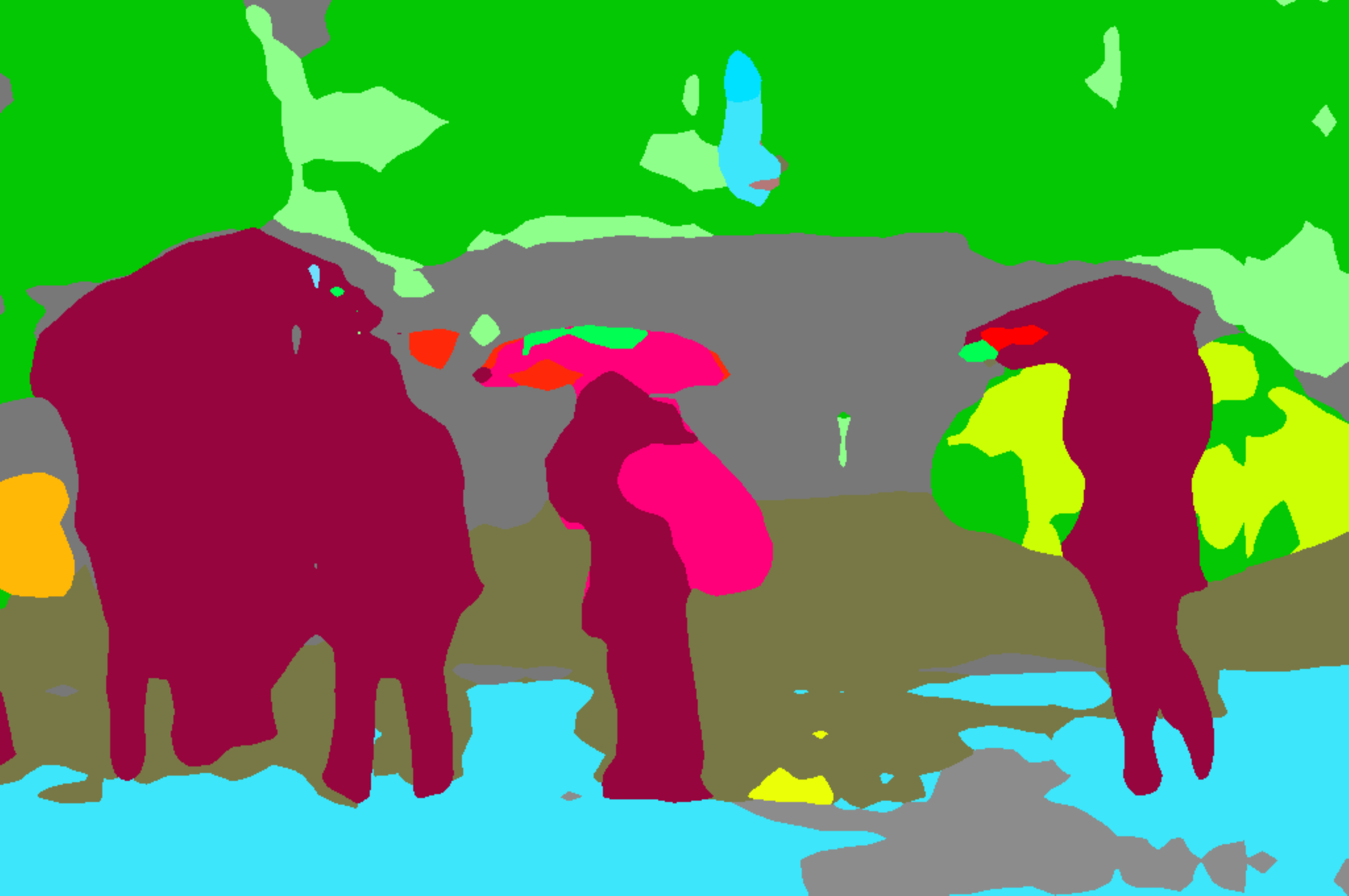}&\hspace{-4mm}
			\includegraphics[width = 0.118\linewidth]{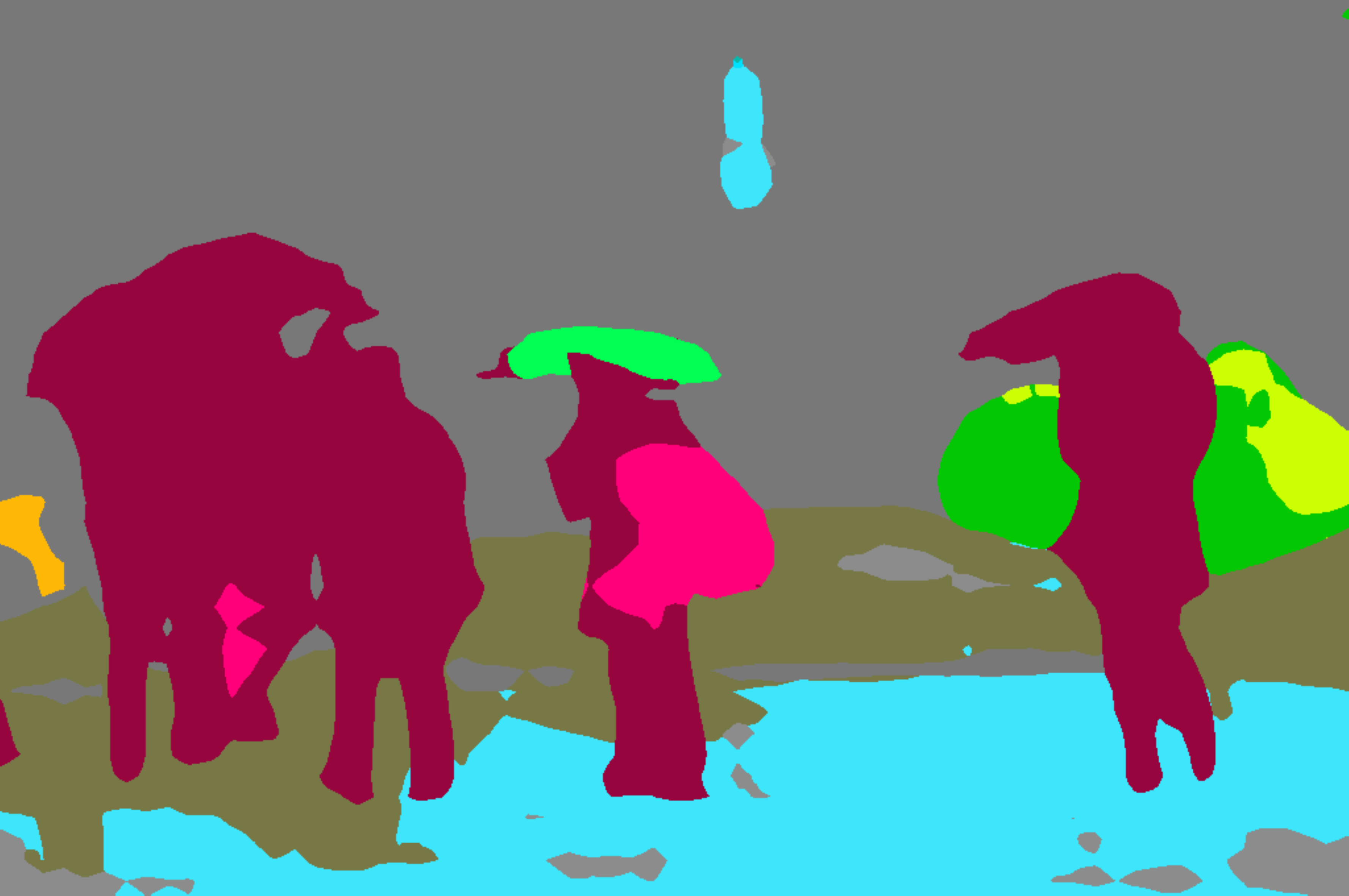}
			\\
			(a) Input&\hspace{-4mm} (b) DDN&\hspace{-4mm} (c) RESCAN&\hspace{-4mm} (d) REHEN&\hspace{-4mm} (e) PreNet&\hspace{-4mm} (f) SpaNet&\hspace{-4mm} (g) SSIR&\hspace{-4mm} (h) JDNet
			\\
		\end{tabular}
	\end{center}
	\caption{Semantic segmentation from real-world datasets.}
	\label{fig:semantic segmentation}
\end{figure*}

\begin{figure*}[!t]
	\begin{center}
		\begin{tabular}{ccccccccc}
			\includegraphics[width = 0.118\linewidth]{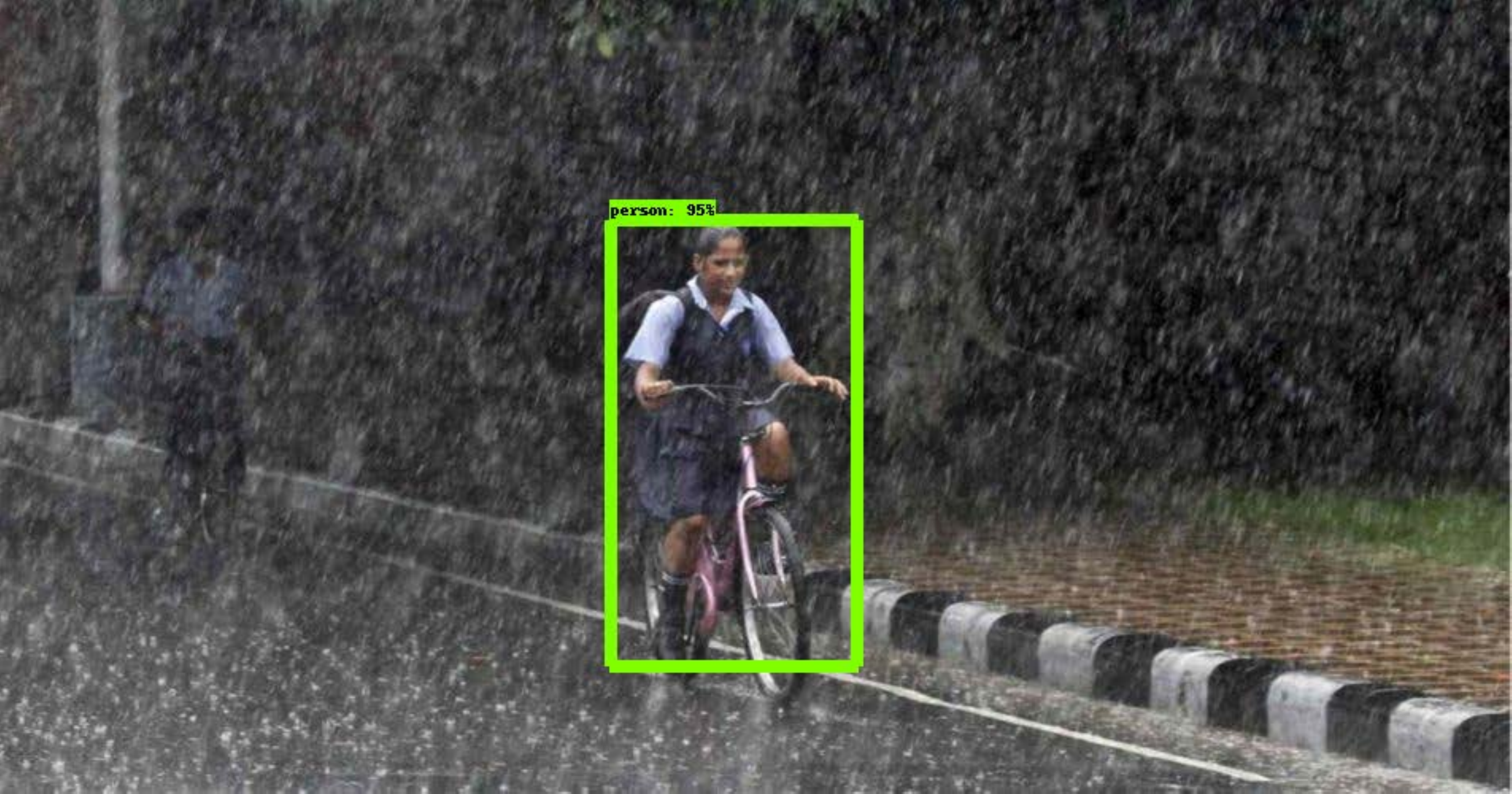}&\hspace{-4mm}
			\includegraphics[width = 0.118\linewidth]{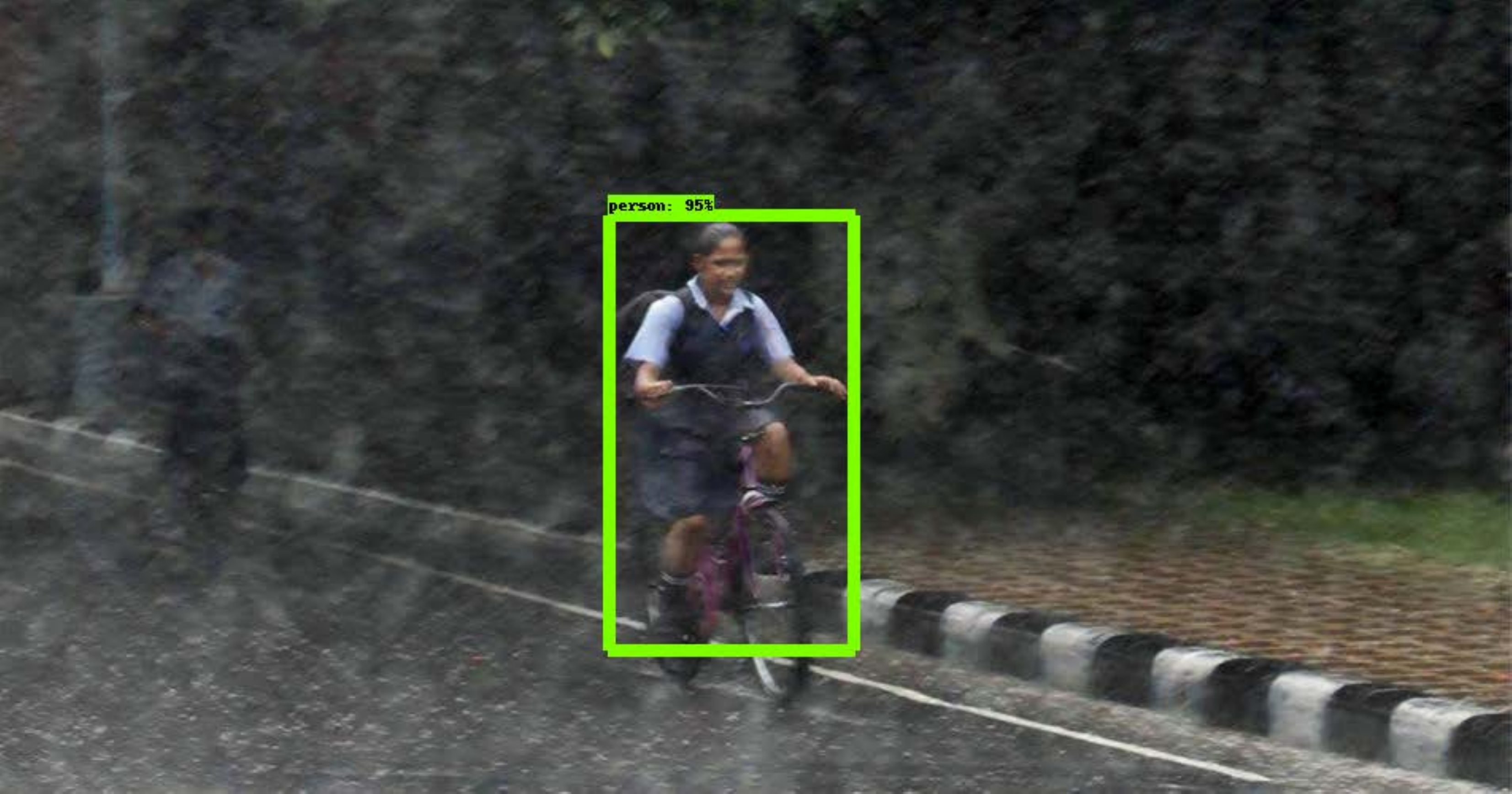}&\hspace{-4mm}
			\includegraphics[width = 0.118\linewidth]{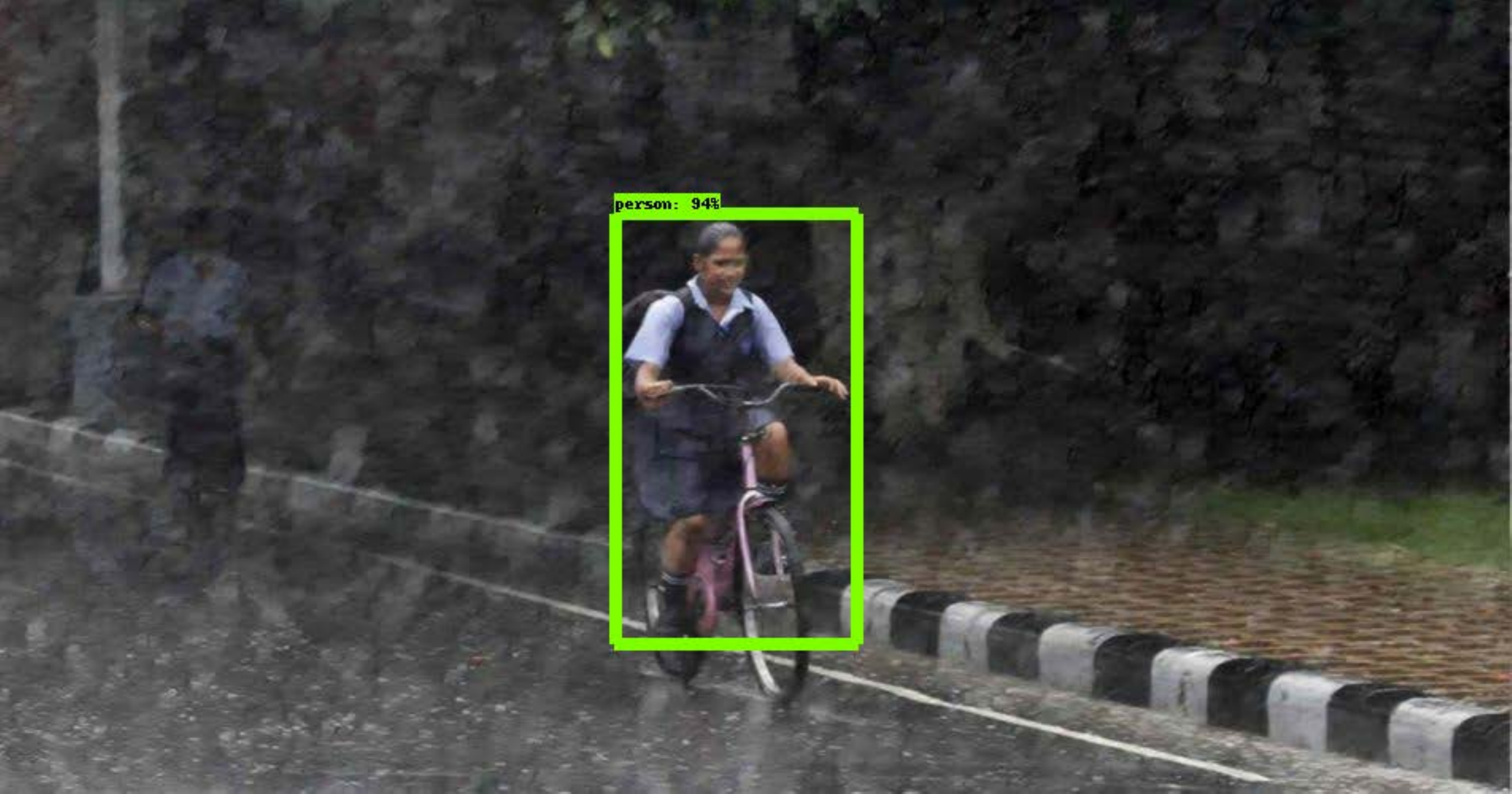}&\hspace{-4mm}
			\includegraphics[width = 0.118\linewidth]{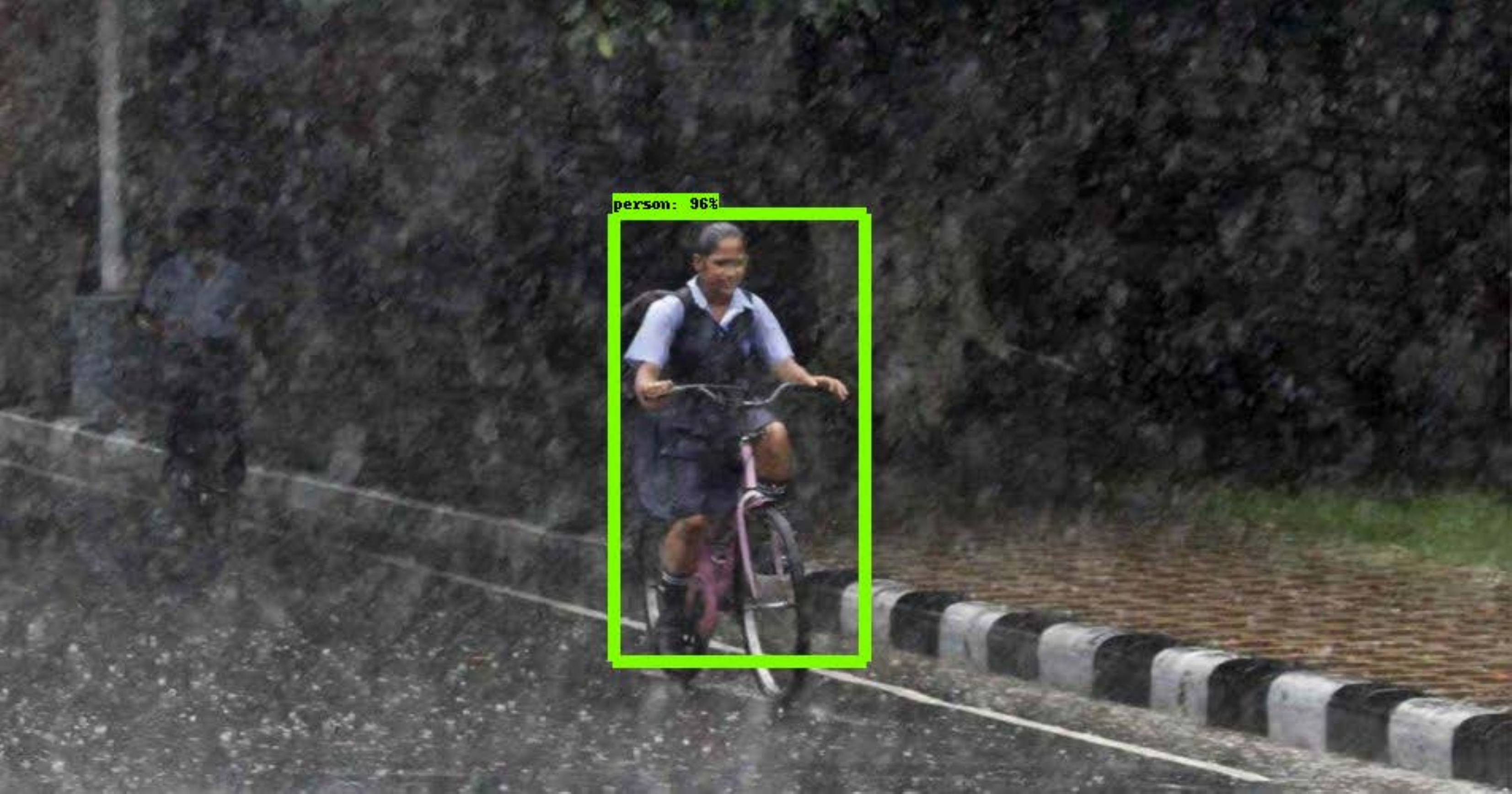}&\hspace{-4mm}
			\includegraphics[width = 0.118\linewidth]{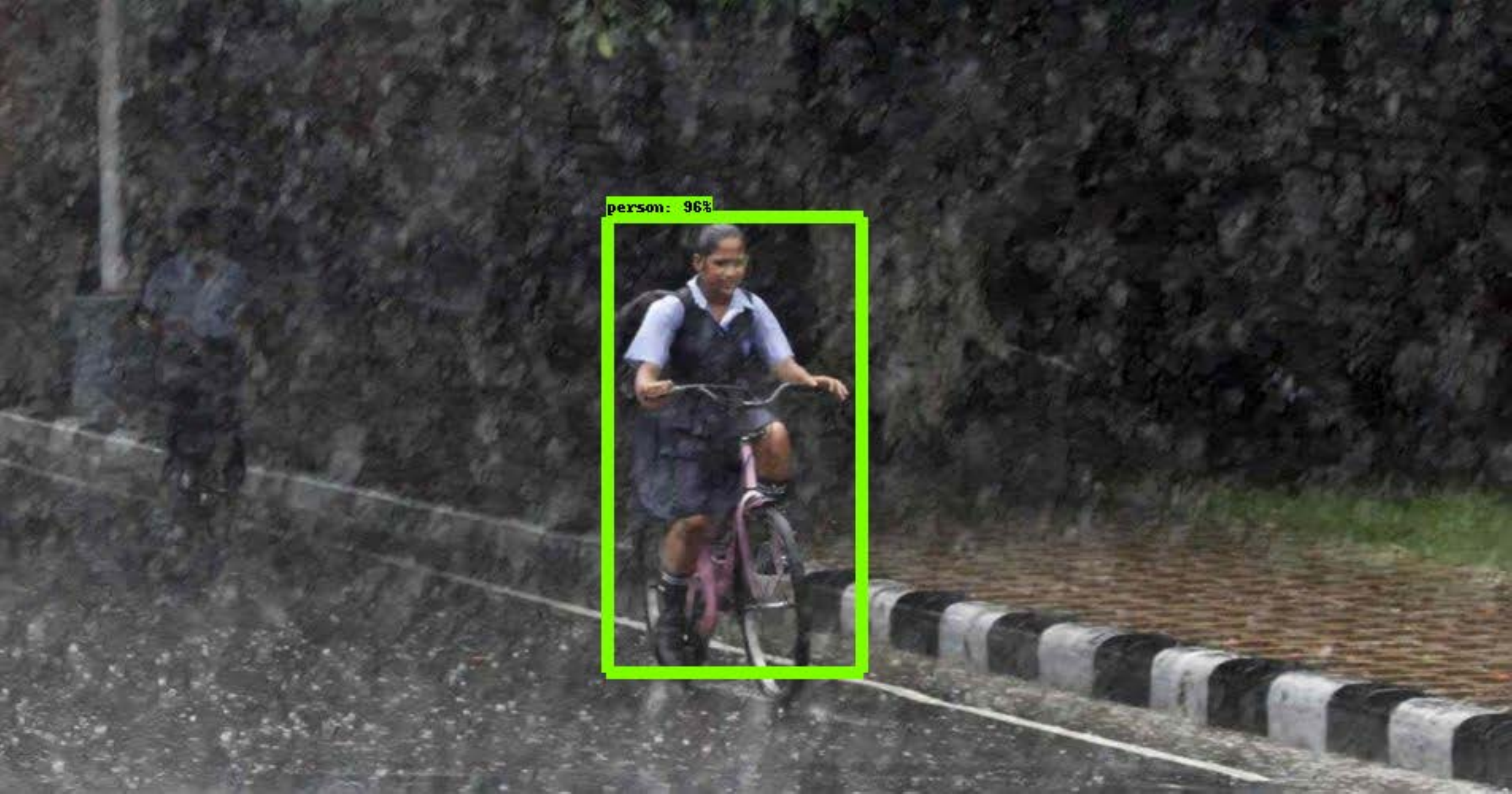}&\hspace{-4mm}
			\includegraphics[width = 0.118\linewidth]{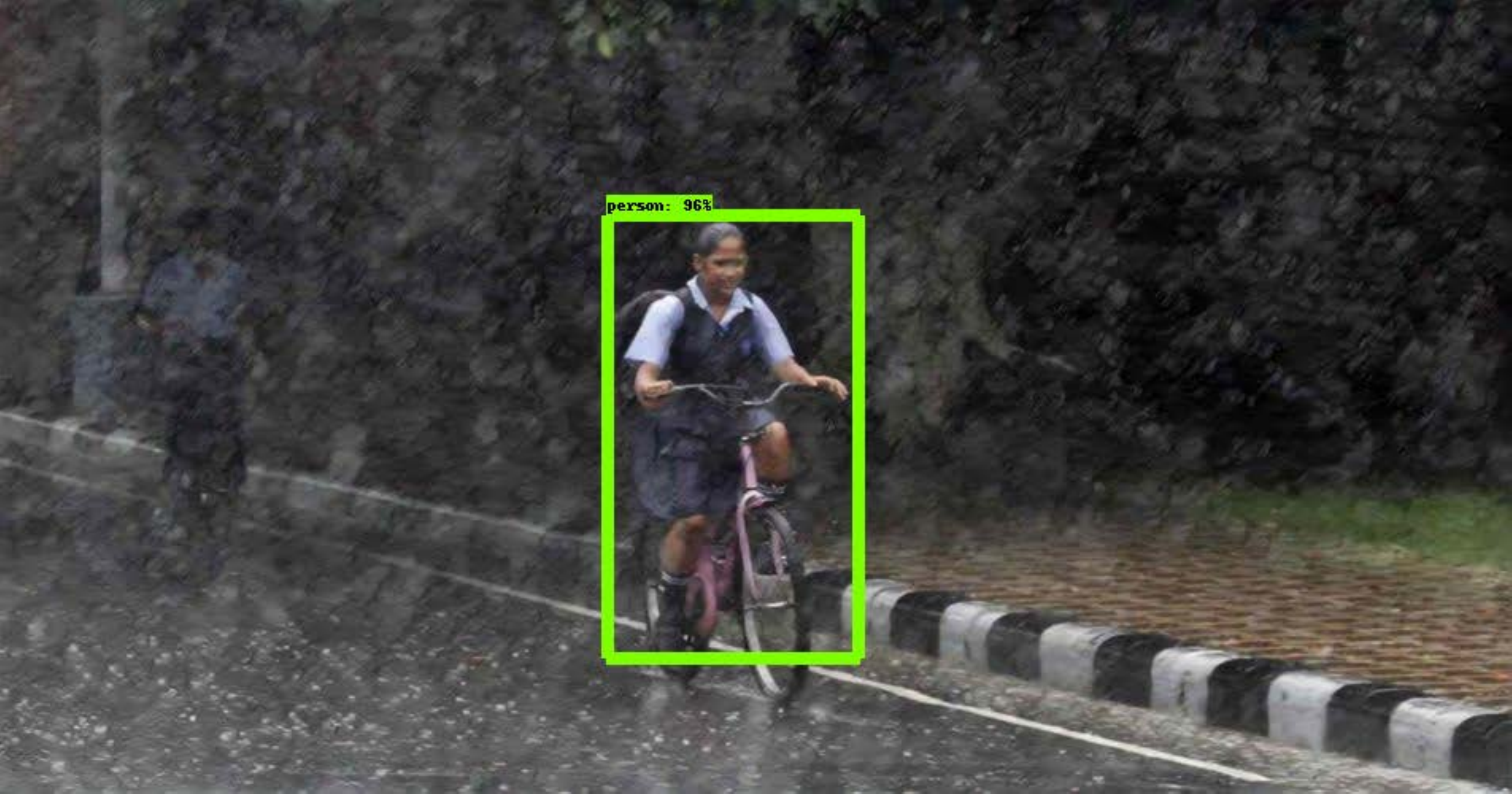}&\hspace{-4mm}
			\includegraphics[width = 0.118\linewidth]{app-image/od-spanet-1.pdf}&\hspace{-4mm}
			\includegraphics[width = 0.118\linewidth]{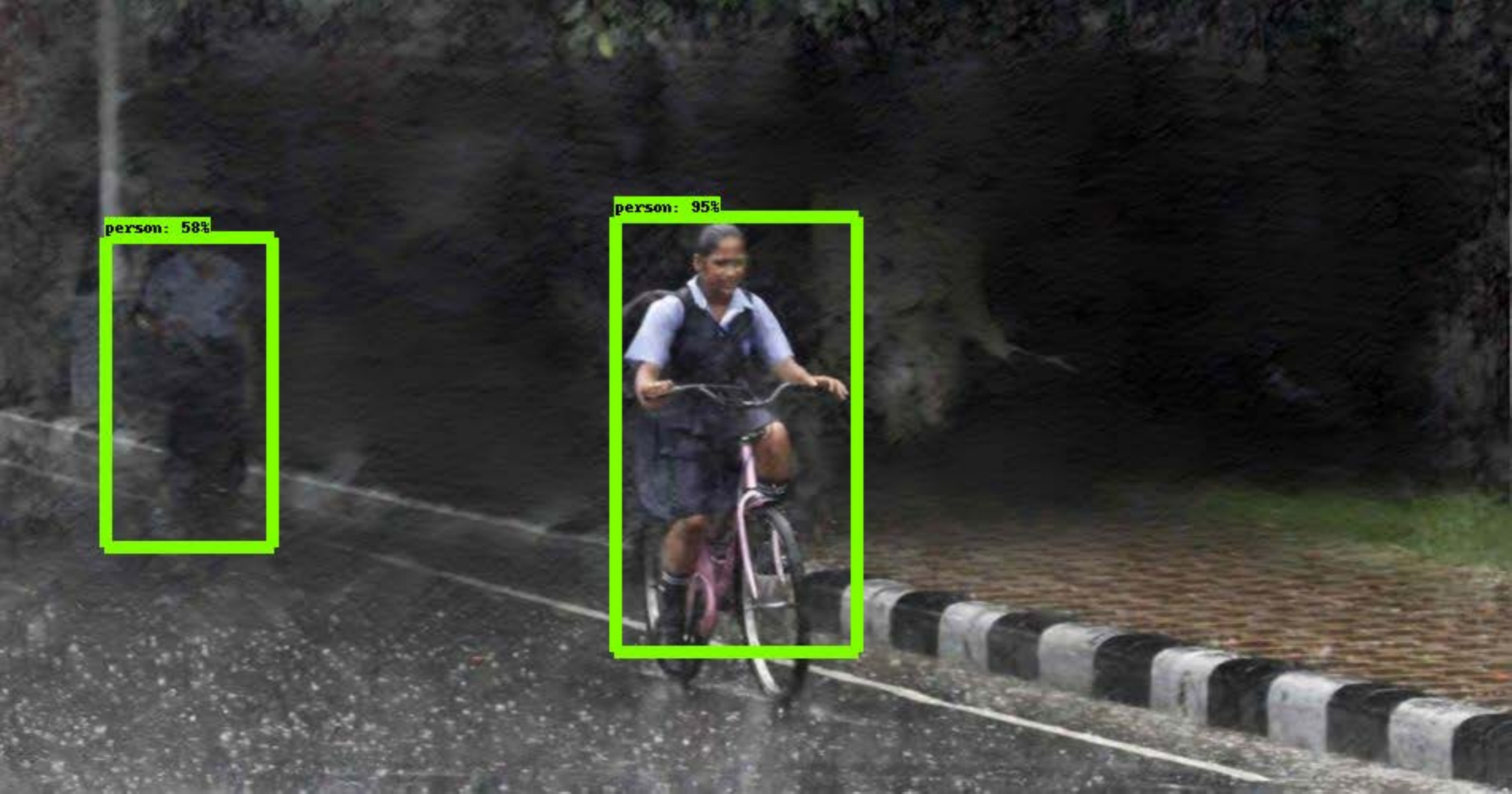}
			\\
			(a) Input&\hspace{-4mm} (b) DDN&\hspace{-4mm} (c) RESCAN&\hspace{-4mm} (d) REHEN  &\hspace{-4mm} (e) PreNet&\hspace{-4mm} (f) SpaNet&\hspace{-4mm} (g) SSIR  & \hspace{-4mm} (h) JDNet
			\\
		\end{tabular}
	\end{center}
	\caption{Object detection from real-world datasets.}
	\label{fig:object detection}
\end{figure*}

\subsection{Discussion on Losses}
In order to show the effectiveness of negative SSIM loss, we conduct experiments with MAE loss and MSE loss under the same conditions. As shown in  Tab.~\ref{tab: The results in syn datasets of Different Losses}, we can see that the negative SSIM loss has the best performance among the three losses. This illustrates that the negative SSIM is the most effective loss for the deraining task.

\begin{table}[!h]
	\caption{The results of different losses on Rain100H.}
	\centering
	\scalebox{1}{
		\begin{tabular}{ccccccccccc}
			\toprule
			Loss      &  SSIM  &PSNR \\
			\midrule		
			MAE              &0.9002 &29.2365     \\
			MSE              &0.8896   &28.7797  \\
			Negative SSIM         &\textbf{0.9221}  &\textbf{30.0160}  \\
			\bottomrule
	\end{tabular}}
	\label{tab: The results in syn datasets of Different Losses}
\end{table}

\subsection{Application for High-level Tasks}
In most conditions, the deraining serves as the preprocessing for some high-level tasks, e.g., segmentation and detection.
In this section, we provide some visual examples for these applications.
Fig.~\ref{fig:semantic segmentation} shows the application of different deraining methods for semantic segmentation on a real-world image, which is directly tested using PSPNet~\cite{semantic_segmentation} under the ADE20K~\cite{ade20k} dataset.
We can see that the proposed method has a better deraining performance and so makes PSPNet have a better segmentation result, while other methods can not generate satisfactory label prediction.

Fig.~\ref{fig:object detection} shows the results of different methods in object detection on real-world datasets, where we use open source called Google Object Detection API on the Tensorflow framework.
We can observe that the proposed method generates more accurate predictions, while other methods can not restore clearer rain-free images that lead to they fail to detect other existing objects.

The above two examples illustrate that our method can provide better preprocessing for other high-level tasks as a better derainer.

\section{Conclusion}
In this paper, we propose an effective deraining approach for single image deraining.
The proposed model consists of three parts, including the Self-Attention module, Scale-Aggregation module and Self-Calibrated convolution.
Each part of these modules can boost to generate clearer and cleaner rain-free images so that the overall network has better deraining performance.
By exploring the inner correlation between different positions of convolution layers, the model is more robust to real-world rainy conditions.
Extensive analysis and discussion illustrate the superiority of the proposed method compared with state-of-the-art approaches both on synthetic and real-world datasets.

\section*{Acknowledgement}
This work was supported by the Natural Science Foundation of China [grant numbers 61976041].

%\begin{acks}
%To Robert, for the bagels and explaining CMYK and color spaces.
%\end{acks}

\bibliographystyle{ACM-Reference-Format}
\bibliography{sample-base}

\appendix



%Lorem ipsum dolor sit amet, consectetur adipiscing elit. Morbi
%malesuada, quam in pulvinar varius, metus nunc fermentum urna, id
%sollicitudin purus odio sit amet enim. Aliquam ullamcorper eu ipsum
%vel mollis. Curabitur quis dictum nisl. Phasellus vel semper risus, et
%lacinia dolor. Integer ultricies commodo sem nec semper.


%Etiam commodo feugiat nisl pulvinar pellentesque. Etiam auctor sodales
%ligula, non varius nibh pulvinar semper. Suspendisse nec lectus non
%ipsum convallis congue hendrerit vitae sapien. Donec at laoreet
%eros. Vivamus non purus placerat, scelerisque diam eu, cursus
%ante. Etiam aliquam tortor auctor efficitur mattis.


%Nam id fermentum dui. Suspendisse sagittis tortor a nulla mollis, in
%pulvinar ex pretium. Sed interdum orci quis metus euismod, et sagittis
%enim maximus. Vestibulum gravida massa ut felis suscipit
%congue. Quisque mattis elit a risus ultrices commodo venenatis eget
%dui. Etiam sagittis eleifend elementum.

%Nam interdum magna at lectus dignissim, ac dignissim lorem
%rhoncus. Maecenas eu arcu ac neque placerat aliquam. Nunc pulvinar
%massa et mattis lacinia.

\end{document}